\documentclass[sigconf]{acmart}
\usepackage{multirow}
\usepackage{array}
\usepackage{makecell}   
\usepackage{graphicx}
\usepackage{amsfonts}   
\usepackage{amsmath}

\usepackage{booktabs}        
\usepackage{bm} 
\usepackage{color}      
\usepackage{algorithm}
\usepackage{algorithmic}
\usepackage{setspace}
\usepackage{multirow}
\usepackage{tikz}
\usepackage{pgfplots}
\usepackage{amssymb}
\usepackage{adjustbox}
\usepackage{amsmath}
\usepackage{float}
\usepackage{cleveref}
\usepackage{subcaption}
\usepackage{float}
\AtBeginDocument{%
  \providecommand\BibTeX{{%
    \normalfont B\kern-0.5em{\scshape i\kern-0.25em b}\kern-0.8em\TeX}}}

\setcopyright{none}
\renewcommand\footnotetextcopyrightpermission[1]{}
\begin{document}

\title{3D Reconstruction and Novel View Synthesis of Indoor Environments based on a Dual Neural Radiance Field}

% \affiliation{%

\author{Zhenyu Bao}
\affiliation{%
  \institution{Peking University}
  % \city{Shenzhen}
  % \country{China}
  \city{}
  \country{}
  }
% \email{zybao@pku.edu.cn}
 
\author{Guibiao Liao}
\affiliation{%
  \institution{Peking University}
  \city{}
  \country{}
  % \city{Shenzhen}
  % \country{China}
  }
% \email{gbliao@stu.pku.edu.cn}
 
\author{Zhongyuan Zhao}
\affiliation{%
  \institution{Peking University}
  \city{}
  \country{}
  % \city{Shenzhen}
  % \country{China}
  }
% \email{zy_efforts@stu.pku.edu.cn}

\author{Kanglin Liu ${}^\dag$}
\affiliation{%
  \institution{Pengcheng Laboratory}
  % \city{Shenzhen}
  % \country{China}
  \city{}
  \country{}
  }
% \email{max.liu.426@gmail.com}

\author{Qing Li ${}^\dag$}
\affiliation{%
  \institution{Pengcheng Laboratory}
  \city{}
  \country{}
  % \city{Shenzhen}
  % \country{China}
  }
% \email{lqing900205@gmail.com}

\author{Guoping Qiu}
\affiliation{%
  \institution{University of Nottingham}
  \city{}
  \country{}
  % \city{Shenzhen}
  % \country{China}
  }
% \email{guoping.qiu@nottingham.ac.uk}

% \author{Anonymous Authors}
% \author{
% Zhenyu Bao\footnotemark[1] \textsuperscript{ 1,2}, 
% Guibiao Liao \textsuperscript{1,2},
% Zhongyuan Zhao \textsuperscript{1,2},
% Kanglin Liu\footnotemark[2] \textsuperscript{ 2}, 
% Qing Li\footnotemark[2] \textsuperscript{ 2},
% Guoping Qiu \textsuperscript{3,4} \and
% $^{1}$Peking University \ \ \ \ 
% $^{2}$Pengcheng Laboratory \\
% $^{3}$University of Nottingham 
% $^{4}$Shenzhen University
% % $^{1}$Peking University 
% % $^{2}$Pengcheng Laboratory 
% % $^{3}$University of Nottingham 
% % $^{4}$Shenzhen University
% }
%% You do not have to enter your paper ID


\begin{abstract}
Simultaneously achieving 3D reconstruction and novel view synthesis for indoor environments has widespread applications but is technically very challenging. State-of-the-art methods based on implicit neural functions can achieve excellent 3D reconstruction results, but their performances on new view synthesis can be unsatisfactory. The exciting development of neural radiance field (NeRF) has revolutionized novel view synthesis, however, NeRF-based models can fail to reconstruct clean geometric surfaces. %In this paper, 
We have developed a dual neural radiance field (Du-NeRF) to simultaneously achieve high-quality geometry reconstruction and view rendering. Du-NeRF contains two geometric fields, one derived from the SDF field to facilitate geometric reconstruction and the other derived from the density field to boost new view synthesis. One of the innovative features of Du-NeRF is that it decouples a view-independent component from the density field and uses it as a label to supervise the learning process of the SDF field. This reduces shape-radiance ambiguity and enables geometry and color to benefit from each other during the learning process. Extensive experiments demonstrate that Du-NeRF can significantly improve the performance of novel view synthesis and 3D reconstruction for indoor environments and it is particularly effective in constructing areas containing fine geometries that do not obey multi-view color consistency. 
Our code is available at: \href{https://github.com/pcl3dv/DuNeRF}{\textcolor{magenta}{https://github.com/pcl3dv/DuNeRF}}.
\end{abstract}







\begin{teaserfigure}
    \includegraphics[width=\textwidth]{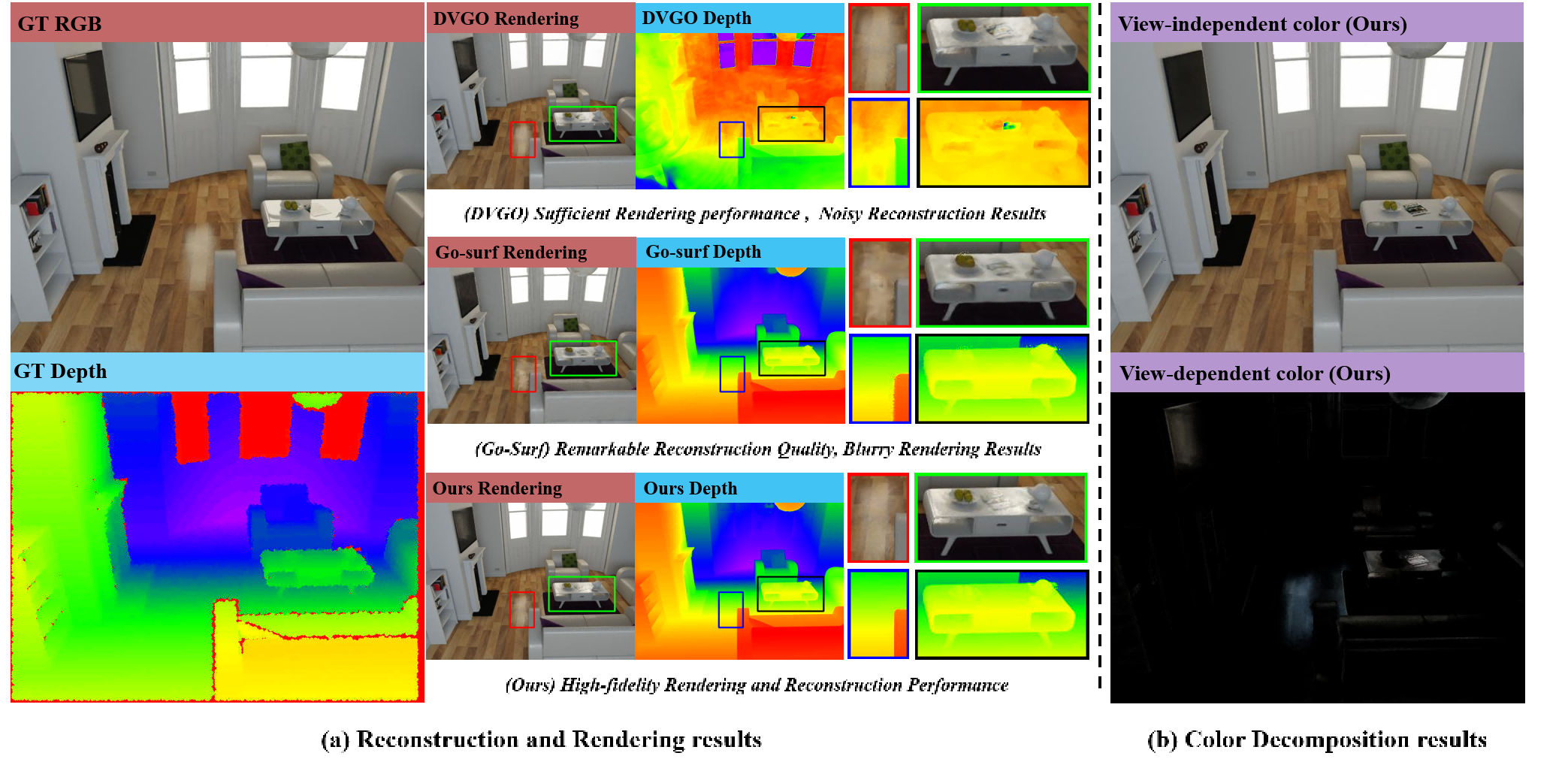}
    \label{fig:teaserfigure}
    \caption{NeRF-based view synthesis methods (eg. \textit{DVGO}) often produce noisy 3D reconstruction results (top right in fig. 1 (a)), while NeRF-based reconstruction approaches (eg. \textit{Go-Surf}) struggle to produce satisfactory rendering image (middle right in fig. 1 (a)). We propose a dual-field architecture to simultaneously obtain high-fidelity view synthesis and reconstruction performance (bottom right in fig. 1 (a)). Fig. 1 (b) illustrates the view-independent and view-dependent color components disentangled by our proposed method. }
    \Description{figure description}
\end{teaserfigure}

%  Meanwhile, our methods could 

\maketitle
% The NeRF-based methods focused on rendering 总是struggle在平滑的几何重建，while methods focused on reconstruction 难以获取高频的图像细节。我们首次使用双场的方式，利用训练过程中免费的与视图无关的信号来关联双场，使得二者在训练过程中螺旋受益，最终得到高质量的几何重建及新视角渲染结果。

\section{Introduction}
Novel view synthesis and 3D reconstruction for indoor environments are of great interest in the computer vision and graphics communities \cite{Ref79, Ref27, Ref10, Ref28, Ref13, Ref0, Ref6, Ref29, Ref30, Ref34}. They provide fundamental support for applications such as robot perception and navigation, virtual reality, and indoor design.
Classical indoor 3D reconstruction methods perform registration and fusion to obtain dense geometry using depth images in an explicit manner where the depth images are obtained either with range sensors like Kinect or inferred from RGB images \cite{Ref31, Ref32, Ref33, Ref35, Ref36, Ref37, Ref38, Ref39}. However, due to noise and holes in the depth images, complete and smooth indoor geometry is difficult to generate. Additionally, such an explicit representation can often fail to preserve sufficient details due to storage limitations thus making it very difficult to synthesize realistic novel views. 

In recent years, coordinate neural networks have been extensively used to describe 3D geometry and appearance due to their powerful implicit and continuous representation capacities \cite{Ref0, Ref40, Ref41, Ref7, Ref8, Ref42, Ref43}. These models take the 3D coordinates as input and output the signed distance value \cite{Ref41, Ref7, Ref8}, density \cite{Ref0, Ref6}, or occupancy of the scene \cite{Ref40, Ref42, Ref43}. 
Although methods such as \textit{Neus} \cite{Ref7} and \textit{VolSDF} \cite{Ref8} can achieve accurate 3D reconstruction of objects, they do not perform well in indoor scenes containing textureless regions or when the observations are sparse. In such cases, depth images are often introduced to provide additional supervision for network training to improve performances \cite{Ref10, Ref13, Ref14}. Furthermore, these methods focus on 3D reconstruction and their performances on novel view synthesis can be unsatisfactory (middle right in fig. 1(a)).

Neural Radiance  Field (NeRF) \cite{Ref0} and a series of its extensions \cite{Ref1, Ref2, Ref3, Ref4, Ref5, Ref6, Ref11, Ref18, Ref23, Ref29, Ref50, Ref51, Ref52, Ref53, Ref54} have achieved exciting results in novel view synthesis. It implicitly represents the density field and color field and performs novel view synthesis via volume rendering. However, these methods can fail to reconstruct clean indoor surfaces (top right in fig. 1(a)), as the density used in NeRF samples the whole space rather than in the vicinity of the surfaces.

In this paper, we propose a dual neural radiance field (Du-NeRF) to simultaneously achieve high-quality geometry reconstruction and view rendering (bottom right in fig. 1(a)).
Specifically, our framework contains two geometric fields, one is derived from the SDF field with clear boundary definitions, and the other is a density field that is more conducive to rendering. They share the same underlying input features, which are interpolated from multi-resolution feature grids and then decoded by different decoders.  We enable the two branches to each play their respective strength, while the former is used to extract geometric features, the latter is used to support the task of new view synthesis. In addition, we decouple a view-independent component from the density field and use it as a label to supervise the learning process of SDF during the network optimization process (the top row in fig. 1(b)). 
In our method, the two geometric fields share the underlying input geometric features to facilitate the optimization of the underlying geometric feature grid. 
Moreover, we use a view-invariant component decoupled from the density branch to replace the view-varying ground truth (GT) images to guide the geometric learning process to reduce shape-radiance ambiguity and allow geometry and color to benefit from each other during the learning process. Experimental results show that this design can effectively construct %or fill in missing 
fine geometries to achieve smooth scene reconstruction, especially in those areas that do not obey multi-view color consistency. %, and make the scene smoother. 
Our contributions are as follows:

\begin{itemize}
    \item We have developed a novel neural radiance field termed Dual Neural Radiance Field (Du-NeRF) for simultaneously improving 3D reconstruction and new view synthesis of indoor environments.
     
    %\item We propose a novel neural radiance field termed Dual neural radiance field (Du-Ne, improving indoor 3D reconstruction and novel view synthesis simultaneously.
      % We propose a color decomposition guided geometry reconstruction network, improving the reconstruction accuracy and contributing to scene reconstruction .

      %  high fidelity novel view synthesis: applying image d
      % 1) Introducing a self-supervised training method for view-independent component extraction 
      
      % high geometry reconstruction accuracy: supervising the geometry reconstruction via view independent color component, improving the reconstruction quality. 

    \item We introduce a novel self-supervised method to extract a view-independent color component for supervising 3D reconstruction, which significantly enhances the smoothness of surfaces and fills in missing parts of indoor objects.

     %We introduce a new self-supervised method to derive a view-independent color component that effectively contributes to making the surface smoother and filling missing parts in 3D reconstruction.  
     
    \item Extensive experiments demonstrate that our method can significantly improve the performance of novel view synthesis and 3D reconstruction for indoor environments.
\end{itemize}

\section{Related Work}
\label{sec:relate}

\textbf{Neural radiance-based novel view synthesis.} 
The introduction of the neural radiance field (NeRF) marks remarkable progress in novel view generation. NeRF models a full-space implicit differentiable and continuous radiance field with neural networks and uses volume rendering to obtain color information. Many variants have been proposed to improve training, inferencing and rendering performances \cite{Ref1, Ref4, Ref5, Ref6, Ref18, Ref50, Ref51, Ref52, Ref53, Ref54}. In particular, \textit{DirectVoxelGO}~\cite{Ref5} and \textit{InstantNGP}~\cite{Ref6} combine explicit and implicit representations and use hybrid grid representations and shallow neural networks for density and color estimation respectively to achieve faster rendering and higher rendering quality. Liu \textit{et al.} \cite{Ref1} introduce a progressive voxel pruning and growing strategy to sample the effective region near the scene surface. Chen \textit{et al.}~\cite{Ref4} use a combination of 2D planes and 1D lines to approximate the grid to achieve faster reconstruction, improved rendering quality, and smaller model sizes. 
To address the ambiguity arising from pixels represented by a single ray, Barron \textit{et al.} \cite{Ref18} introduce the concept of cones instead of points, to increase the receptive field of a single ray. This approach, known as \textit{Mip-nerf}, effectively tackles issues of jaggies and aliasing and enhances rendering quality. To reduce the blurring effect, Lee \textit{et al.} \cite{Ref50} design a rigid blurry kernel module which takes into account both motion blur and defocus blur during the real acquisition process. Kun \textit{et al.} \cite{Ref51} further improve the rendering performance by learning a degradation-driven inter-viewpoint mixer. Additionally, some works attempt to improve the rendering performance by jointly optimizing the poses of the training images \cite{Ref52, Ref53, Ref54}. In contrast, our method adopts geometry-guided sampling, which benefits from the reconstruction result and allows for more accurate sampling of points near the surface, thus improving the performance of view synthesis.

\textbf{Neural implicit 3D reconstruction.}  
Neural implicit functions take a 3D location as input and output occupancy, density, and color~\cite{Ref40, Ref42, Ref9, Ref0, Ref64}. \textit{Scene Representation Networks} employ MLPs to map 3D coordinates to latent features that encode geometry and color information~\cite{Ref65}.  
Yariv \textit{et al.}~\cite{Ref8} and Wang \textit{et al.}~\cite{Ref7} propose two approaches to converting SDF values into density and performing volume rendering to supervise object reconstruction in the Nerf-based framework. \textit{Neuralangelo}~\cite{Ref12} introduces numerical gradients and utilizes multiresolution hash grids to reconstruct detailed scenes. However, while these methods perform well on scenes with rich texture, they struggle with indoor scenes with textureless walls and ceilings.
To address these issues, Yu \textit{et al.}~\cite{Ref10} employ predicted depth and normal maps from a pre-trained network to enhance the reconstruction of indoor scenes. Azinovi{\'c} \textit{et al.}~\cite{Ref13} combine a TSDF representation with the NeRF framework and use a depth representation from an off-the-shelf RGBD sensor to improve the accuracy of indoor geometry reconstruction. Subsequent studies \cite{Ref14, Ref15, Ref16} have further optimized the Neural-RGBD strategy to speed up training. \cite{Ref14} utilizes voxel representations instead of 3D coordinates to achieve faster query, while \cite{Ref15} trains a dynamically adaptive grid that allocates more voxel resources to more complex objects. \cite{Ref16} pre-trains a feature grid to accelerate the training process. We propose to decouple the view-independent colour component for guiding the 3D reconstruction, which effectively enhances the smoothness of surfaces and fills in the missing parts of depth-based reconstruction.

\begin{figure*}[t]
  \centering
  \includegraphics[width=\linewidth]{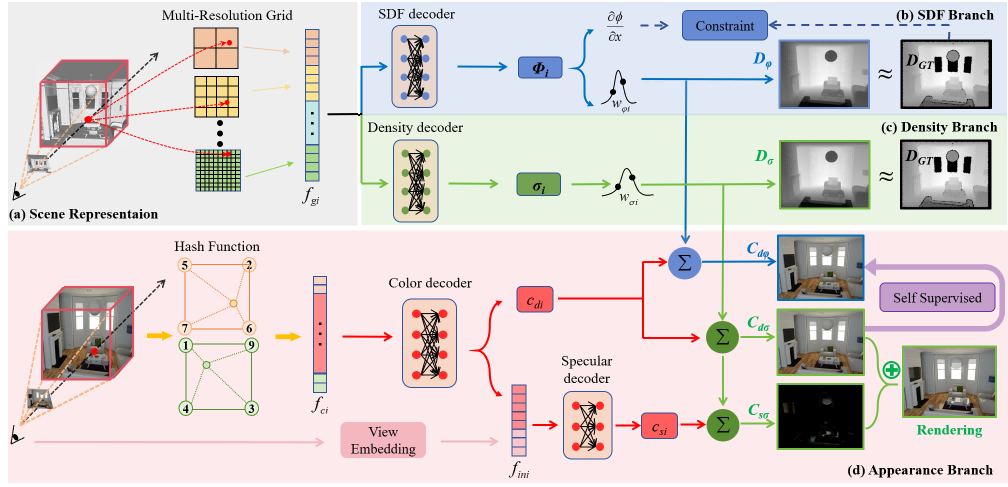}
   \caption{\textbf{Du}al \textbf{Ne}ural \textbf{R}adiance \textbf{F}ield \textbf{(Du-NeRF)}. In (a) Scene Representation, we use a four-layer multi-resolution grid to store geometric features $f_{gi}$, and a hash-based multi-resolution grid for the color features $f_{ci}$ . $f_{gi}$ is decoded to SDF $\phi$ and density $\sigma$ by different MLPs in (b) SDF branch and (c) the Density branch. We provide depth constraints for $\phi$ and introduce additional regularization terms to ensure the stability of its training. We design a depth alignment loss for $\sigma$ to align the two geometry fields. In (d), for color calculation, $\phi$ and $\sigma$ from (b) and (c) are integrated with decoupled view-independent colors to compute the self-supervised loss. The final rendering color sums the view-dependent and view-independent colors to be integrated with the $\sigma$ weights.}
   \label{fig:fig2}
\end{figure*}

\section{Method}
\label{sec:method}

\subsection{Preliminaries}\label{Preliminaries}

\textbf{Neural radiance field.}
Given a collection of posed images, neural radiance field can estimate the color and depth of each pixel by computing the weighted sum of sampling points based on their color and distance from the center of the camera \cite{Ref21}:
% , as illustrated in \cref{eq:eq1}.

\begin{equation}
  C_p=\sum_{i=1}^{N}w_i c_i, D_p=\sum_{i=1}^{N}w_i z_i,
  \label{eq:eq1}
\end{equation}
where the color and distance from sampling points to the camera center, are denoted by $c_i$ and $z_i$ respectively, $w_i$ are the contribution weights of each sampling point to the color and depth and is calculated as $w_i=\sum_{i}^N (\prod_{j}^{i-1}( 1-\alpha_j(p(x))))\alpha_i (p(x_i))$, where $p(x)$ and $\alpha_i$ are the density and opacity of sampling point $x_i$, respectively. The opacity is then computed using ~\cref{eq:eq24}.
\begin{equation}
  \alpha_i = 1-exp(-p(x_i)(z_{i+1} - z_i)),
  \label{eq:eq24}
\end{equation}
the color $c_i$ and density $p(x_i)$ of a given sampling point are predicted by the Multi-layer Perceptron (MLP). The process is illustrated in \cref{eq:eq2}.

\begin{equation}
    \begin{split}
      p(x_i), f &= \Gamma_{\theta}(x_i), \\
      c_i &= \Gamma_{\kappa}(f, d),
      \label{eq:eq2}
    \end{split}
\end{equation}
% \begin{equation}
%   c_i = \Gamma_{\kappa}(f, d)
%   \label{eq:eq3}
% \end{equation}
where $x$ and $d$ represent the location and ray direction, respectively. $f$ is the feature vector related to the location. $\Gamma_\theta$ and $\Gamma_\kappa$ are implicit functions modelled by MLPs.

\textbf{SDF-based neural implicit reconstruction.}
The signed distance function (SDF)  refers to the nearest distance between a point and surfaces and is often used to implicitly represent geometry. One notable application of SDF is neural implicit reconstruction under the volume rendering framework, achieved by a method called \textit{Neus} \cite{Ref7}.

The key to the success of this method is an unbiased transformation between the SDF values and the density, as demonstrated in \cref{eq:eq4}. This transformation enables the creation of high-quality surface geometry with great accuracy and cleanliness.

\begin{equation}
  \alpha_i=\max \left(\frac{\sigma_s\left(\phi\left(\mathbf{x}_i\right)\right)-\sigma_s\left(\phi\left(\mathbf{x}_{i+1}\right)\right)}{\sigma_s\left(\phi\left(\mathbf{x}_i\right)\right)}, 0\right),
  \label{eq:eq4}
\end{equation}
where $\phi(x_i)$ represents the SDF value of a given sample point, and $\sigma_s(x)=(1+e^{-sx})^{-1}$, while the smoothness of the surface is conditioned on a learnable parameter $s$.

\textbf{Multi-resolution feature grid.}
To improve the efficiency of training, a grid representation is used \cite{Ref5, Ref2, Ref4}. However, using only single-resolution grid limits the optimization of density and color to local information, resulting in disruptions to the smoothness and continuity of the scene texture \cite{Ref10, Ref14}. A multi-resolution grid expands the local optimization to nearby continuous fields by varying the receptive field and gradient backpropagation of sample points, enabling higher rendering quality and smoother geometry. The embedding feature is obtained by concatenating the features of each level, as shown in \cref{eq:eq9}.

\begin{equation}
        f = \Omega (V_1(x), V_2(x),..., V_n(x)),
    \label{eq:eq9}
\end{equation}
where $\Omega$ indicates concatenation, and $V_i(x)$ is trilinear interpolation in the $i$-th grid. The final feature vector of the input network is denoted as $f$.

To achieve a higher resolution, hash-based feature grid is employed in \cite{Ref6}. It represents resolutions as:
\begin{equation}
        R_l:= \left \lfloor R_{\min}b^l \right \rfloor, \ b:=\exp(\frac{\ln{R_{\max}}-\ln{R_{\min}}}{L-1}) ,
    \label{eq:eq26}
\end{equation}
where $R_{\min}$, $R_{\max}$ are the coarsest and finest resolution, respectively. $R_l$ represents $l-$th level resolution and $L$ is the total levels. Similarly, we extract the interpolated features at each level and concatenate them together as in \cref{eq:eq9}.

\subsection{Dual Neural Radiance Field}\label{Dual-weight}

\textbf{Scene representation.}
To achieve high-fidelity indoor scene reconstruction and rendering simultaneously, we introduce the dual neural radiance field (\textit{Du-NeRF}). The key idea that enables the dual neural radiance field to achieve high-fidelity reconstruction and rendering is to separately represent the geometry field and the color field. %To speed up the training and inference process, the geometry field and the color field is represented with two mult-resolution grids with different level, considering the different complexity of the reconstruction and rendering tasks. 
Taking into account the fact that the reconstruction task and the rendering task have different complexities, the geometry field and the color field are represented by multi-resolution grids with different levels to speed up training and inference. 

As shown in \cref{fig:fig2}, we use a four-level grid to represent the scene geometry where the grid sizes at each level are respectively 3cm, 6cm, 24cm, and 96cm. At each level, the dimension of the geometry feature is set to 4, and the dimension of the geometry feature $f_{gi}$ is therefore 16.  %The feature dimension is set to 4 for each level and the geometry feature $f_{gi}$ is of dimension 16. 
For the color field, the hash-based multi-resolution grid is utilized. The coarsest resolution $R_{\min}$ of the hash-based multi-resolution grid is set to 16. The level of grid resolution is $L=16$, and the feature vector at each level is $2$-dimensional, we therefore obtain a hash color feature vector $f_{ci}$ with a total of 32 dimensions. The geometry feature $f_{gi}$ and color feature $f_{ci}$ are obtained using tri-linear interpolation.

\textbf{Dual neural radiance networks.} The key idea is to use two different geometric decoders to extract %the geometry information SDF and density, 
SDF and density, which are respectively used for 3D reconstruction and image rendering. %respectively. The obtained SDF and density values are used for 3D reconstruction and image rendering. 
The SDF and density are estimated implicitly from the same interpolated features $f_{gi}$: 
\begin{equation}
    \Gamma_\phi (f_{gi}) = \phi_i, \  \Gamma_\sigma (f_{gi}) = \sigma_i,
    \label{eq:eq23}
\end{equation}
where both of the $\Gamma_\phi$ and $\Gamma_\sigma$ are MLPs, and used to decode $f_{gi}$ into SDF $\phi_i$ and density $\sigma_i$, respectively. %The obtained $\phi$ is converted to occupancy through \cref{eq:eq4} , while $\sigma_i$ converted through \cref{eq:eq24}. 
$\phi_i$ and $\sigma_i$ are both converted to occupancy through \cref{eq:eq4} and \cref{eq:eq24}, respectively. The resulting occupancy of the two is calculated via \cref{eq:eq1} to obtain the pixel depth and color. The whole process is supervised with the image reconstruction loss and depth loss.

Sampling points near object surface have a higher contribution to the rendered color of the ray \cite{Ref0, Ref7}. 
To obtain higher rendering quality, we employ the hierarchical sampling strategy as in \cite{Ref0, Ref7}, which contains coarse sampling and fine sampling near the surface. Specifically, we first uniformly sample 96 points along the ray in the coarse stage and then iteratively add 12 sampling points three times according to the cumulative distribution function(CDF) of previous coarse points weights in the fine stage as in \cite{Ref14}. Finally, we got 132 sampling points for depth and color rendering. Note that we use the weight distribution calculated from the SDF branch for sampling as it provides more accurate surface information. For the volume rendering process, the two branches share the sampling points.

\textbf{Self-supervised color decomposition.}\label{sec:color distanglement}
We disentangle the color into view-independent color and view-dependent color. We use the decoupled view-independent color $c_{di}$ to guide geometry learning in a multi-view consistent self-supervised manner by constraining the weight values. This separation allows the color branch to leverage complete color information for rendering, while the geometry branch benefits from view-consistent supervision. Previous work has shown that decoupling color helps find geometry surfaces in NeRF-based methods~\cite{Ref11,Ref26}. For instance, \cite{Ref26} accomplishes the color decomposition via a simple regularization term, which assumes that  the view-dependent color is close to zero. In contrast, we utilize the view-independent color decoupled from the two branches to constrain each other in a self-supervised manner. This design effectively extracts view-independent color to support accurate geometry reconstruction but also boosts the image rendering results via the mutually beneficial learning process.

%我们利用两个分支各自解耦出来的漫反射进行互相约束以一种自监督的方式。

%mitigate shape-radiance ambiguity~\cite{Ref11,Ref26}. \cite{Ref26} accomplished the color decomposition supervised by the complete color. In contrast, we not only exploit the complete color to supervise the color decomposition but the view-independent color consistency obtained from the SDF branch and density branch in a self-supervised manner. This design effectively extracts view-independent color to support accurate geometry reconstruction but also boosts the image rendering without any additional specular regularity term as in \cite{Ref17,Ref26}. 

To achieve it, as shown in \cref{fig:fig2}, we utilize two color decoders consisting of MLPs to implicitly estimate the view-independent color and view-dependent color. The view-independent decoder takes the interpolated color feature $f_{ci}$ as input and outputs the view-independent color $c_{di}$ and an intermediate feature $f_{ini}$. The specular (view-dependent) decoder takes the intermediate feature $f_{ini}$ and the encoded view-direction vectors as input to obtain the specular color $c_{si}$. The overall color at a sampling point is calculated by adding the two color components $c_i = c_{di}+c_{si}$. The final color $C$ is generated by summing the weighted color of each sampling point with \cref{eq:eq1}. In our framework, we integrate $\omega_{\sigma_i}$ with the full color $c_i$ as in \textit{NeRF}, %while
however, we only weight the view-independent color $c_{di}$ with $\omega_{\phi_i}$ to obtain the view-independent color of the corresponding ray. The calculation of color in \cref{eq:eq1} becomes:
\begin{equation}
  C_{d\phi}=\sum_{i=1}^{N}\omega_{\phi i} c_{di},
  \label{eq:eq22}
\end{equation}
where $C_{d\phi}$ is the view-independent color computed by $\omega_{\phi i}$. %Otherwise, 
We can  obtain the view-independent color $C_{d\sigma}$ computed by $c_{di}$ and $\omega_{\sigma_i}$, which will be used as the ground truth to constrain the learning process of $C_{d\phi}$:

\begin{equation}
    \mathcal{L}_{d}=\sum_{i}^{N}  \lambda_{d}\left \| C_{d\phi}-C_{d\sigma} \right \| 
    \label{eq:eq25}.
\end{equation}

\subsection{Pose Refinement}\label{Pose Refinement}
Noisy poses lead to undesirable mesh protrusions of 3D reconstruction and artifact in novel view synthesis. Thus, we adopt a pose optimization strategy in the training process, which treats the poses as learnable parameters as in \cite{Ref13}. Specifically, we convert the transformation matrix to Euler angles and translation vectors ($\mathbb{R}^{3+3}$) and initialize them with results of the Bundlefusion\cite{Ref27}. We optimize them along with the dual fields and are supervised by depth loss and photometric loss.

After pose optimization, poses of training images are transformed into a new coordinate space. A calibration process is required for testing images to transform them into the same coordinate space as the training images. To achieve it, we introduce the proximity frame alignment (PFA) strategy, which uses the pose change of the adjacent training images to help correct the noisy poses of the testing images.
Specially, given a sequence of consecutive images  $\left \{ {I_i\in\mathbb{R}^{H\times W\times 3} \mid i\in \left \{ 1, ... N\right\}} \right \} $,  and their corresponding poses $\left \{ {P_i\in\mathbb{R}^{4\times 4} \mid i\in \left \{ 1, ... N\right\}} \right \}$,we suppose that the $k-$th image  is testing image and the $k+1$-th is the training image. $P_k$ and $P_{k+1}$ are noisy poses, and $P_{k+1}'$ is the optimised pose of the training image, respectively. $P_k'$ is the correct pose of the testing image, which can be calculated by \cref{eq:eq28}.
% To correct the noisy poses of the testing images, we introduce the proximity frame alignment (PFA) strategy. Since we can not obtain the correct poses of testing images in the training process as for training images, we propose to use the pose change of the adjacent training images to help correct the noisy poses of the testing images.
% Specially, given a sequence of consecutive images  $\left \{ {I_i\in\mathbb{R}^{H\times W\times 3} \mid i\in \left \{ 1, ... N\right\}} \right \} $,  and their corresponding poses $\left \{ {P_i\in\mathbb{R}^{4\times 4} \mid i\in \left \{ 1, ... N\right\}} \right \}$,we suppose that the $k-$th image  is testing image and the $k+1$-th is the training images. $P_k$ and $P_{k+1}$ are noisy poses, and $P_{k+1}'$ is the optimised pose of the training, respectively. $p_k'$ is the correct pose of the testing image, which can be calculated by \cref{eq:eq28}.
\begin{equation}
    P_k' = \mathfrak{A} \times P_k,
    \label{eq:eq28}
\end{equation}
where $\mathfrak{A}$ is the transformation matrix of the adjacent training image $k+1$, obtained by:
\begin{equation}
    \mathfrak{A} = P_{k+1}'  \times P_{k+1}^{-1}.
    \label{eq:eq29}
\end{equation}
In this way, we can calibrate the testing images into the coordinate system of the training images.

% 当进行训练位姿优化后，用于训练的位姿与用于测试的位姿不再统一，这是因为测试的位姿没有在训练过程中进行优化。为了对渲染结果进行定量的评估，我们提出一种临近帧位姿对齐评估策略。具体的，给定一组输入的连续帧序列的位姿 \left \{ {P_i\in\mathbb{R}^{4\times 4} \mid i\in \left \{ 1, ... N\right\}} \right \} 及对应的图像$\left \{ {I_i\in\mathbb{R}^{H\times W\times 3} \mid i\in \left \{ 1, ... N\right\}} \right \} $，我们假定第k张图像及其对应的位姿被选为测试集，第k+1张为训练集。Pk与Pk+1为优化前的位姿，and pk+1'为优化后的训练位姿，pk'为需要被对齐的测试位姿。
% $P_k' = \mathfrak{A} \times P_k$
% 其中 $\mathfrak{A}$ 为位姿对齐算子，as follows：
% $\mathfrak{A} = P_{k+1}'  \times P_{k+1}^{-1}$
% 我们通过这样的方式来最小化评估误差。

\subsection{Network training}
\label{sec:Network training}
To optimize the dual neural radiance field, we randomly sample $M$ rays during training. Our loss is divided into two components including a $\mathcal{L}_\phi$  loss, and a $\mathcal{L}_\sigma$ loss in  \cref{eq:eq15}.
\begin{equation}
    \mathcal{L}{(\text {P})}= \mathcal{L}_{\phi} +\mathcal{L}_{\sigma}.
    \label{eq:eq15}
\end{equation}

The $\mathcal{L}_\phi$  contains the three components as shown in  \cref{eq:eq16}: view-independent color loss, depth loss and SDF regularization loss, as the following:
\begin{equation}
    \mathcal{L}_\phi=\lambda_d \mathcal{L}_d + \lambda_{\text{depth}} \mathcal{L}_{\text{depth}} + \mathcal{L}_{\text {SDF}}.
    \label{eq:eq16}
\end{equation}

The view-independent loss is calculated as the distance between $C_{d\phi}$ and $C_{d\sigma}$ as shown in \cref{eq:eq25}, and the depth loss is the $L_1$ loss:
\begin{equation}
    \mathcal{L}_{\text{depth}} = \sum_{i}^{N} \left | D_\phi-D_{\text{gt}} \right |.
    \label{eq:eq17}
\end{equation}

In order to improve the robustness of the learning process of the SDF $\phi$, we imposed a series of SDF losses $\mathcal{L}_{\text {SDF}}$ to regularize the $\phi$ value as \cite{Ref14}:
\begin{equation}
    \mathcal{L}_{\text {SDF}}=\lambda_{\text {eik }} \mathcal{L}_{\text {eik }}+\lambda_{\text {fs}} \mathcal{L}_{\text {fs}}+\lambda_{\text {sdf }} \mathcal{L}_{\text {sdf }}+\lambda_{\text {smooth }} \mathcal{L}_{\text {smooth }}.
    \label{eq:eq18}
\end{equation}

The regularization term $\mathcal{L}_{\text{eik}}(x)$ encourages valid SDF predictions in the unsupervised regions, while $\mathcal{L}_{\text{smooth}}(x)$ is an explicit smoothness term to realize smooth surfaces.
\begin{equation}
        \mathcal{L}_{\text{eik}}(x)=(1-\|\nabla \phi(x)\|)^2,
    \label{eq:eq5}
\end{equation}
\begin{equation}
        \mathcal{L}_{\text{smooth}}(x)=\|\nabla \phi(x)-\nabla \phi(x+\epsilon)\|^2.
    \label{eq:eq6}
\end{equation}
$\mathcal{L}_{\text{sdf}}(x)$ and $\mathcal{L}_{\text{fs}}(x)$ are used to constrain the truncation distance $b(x)$.
\begin{equation}
        \mathcal{L}_{\text{sdf}}(x)=|\phi(x)-b(x)|,
    \label{eq:eq7}
\end{equation}
\begin{equation}
        \mathcal{L}_{\text{fs}}(x)= \max \left(0, e^{-\alpha \phi(x)}-1, \phi(x)-b(x)\right).
    \label{eq:eq8}
\end{equation}
A detailed explanation of these regularity constraints can be found in \cite{Ref14}.

The $\mathcal{L}_{\sigma}$ loss contains two parts as shown in \cref{eq:eq19}:
\begin{equation}
    \mathcal{L}_{\sigma}= \lambda_{\text {rgb}} \sum_{i}^{N} \left \| C_\sigma-C_{\text{gt}}\right \| + \lambda_{\text {align}} \sum_{i}^{N} \left | D_\sigma-D_{\text{gt}} \right |,
    \label{eq:eq19}
\end{equation}
% \begin{equation}
%     \mathcal{L}_{\text {rgb}}=\sum_{i}^{N} \left \| C_\sigma-C_{gt}\right \| 
%     \label{eq:eq20}
% \end{equation}
% \begin{equation}
%     \mathcal{L}_{\text {align}}=\sum_{i}^{N} \left | D_\sigma-D_{gt} \right |  
%     \label{eq:eq21}
% \end{equation}
where $C_\sigma$ is the final rendering color and $D_\sigma$ is the depth calculated from the color branch. $\lambda_{\text{align}}$ is used to align two geometric fields. We experimentally found that $\mathcal{L}_{\text {align}}$ is important for decoupling consistent view-independent color.
For each of the geometry coefficients, we follow the practice of \cite{Ref14} and set them as follows: $\lambda_d=5$, $\lambda_{\text{depth}}=1$, $\lambda_{\text{eik}}=1.0$, $\lambda_{\text{fs}}=1.0$, $\lambda_{\text{sdf}}=10.0$, $\lambda_{\text{smooth}}=1.0$. 
For color coefficients we experimentally set them as $\lambda_{\text{rgb}}=50$ and $\lambda_{\text{align}}=1$.

\section{Experiment}
\label{sec:Results}

\subsection{Setup}

\textbf{Datasets.} We utilize the \textit{NeuralRGBD} (10 scenes), \textit{Replica} (8 scenes), and \textit{Scannet} (5 scenes) datasets to evaluate the proposed method. The \textit{NeuralRGBD} dataset and \textit{Replica} dataset are synthetic datasets, and images are of high quality.  The \textit{Scannet} dataset is real world dataset, captured with a handheld device such as an iPad and suffers from severe motion blur. 
% The \textit{NeuralRGBD} dataset comprises ten synthetic scenes, and the \textit{Replica} dataset contains eight photo-realistic rooms and offices. There are five \textcolor{red}{real-world} scenes in the \textit{Scannet} dataset. 
Each scene in the three datasets includes RGB images, depth images, and corresponding poses. The poses are obtained via the \textit{BundleFusion} algorithm. For each scene, about 10 percent of images are used as validation sets, and the rest of the images are used as training sets. Specifically, starting with the $10$-th image, we choose one image in every ten consecutive images as the validation set.

% We tested our method on 5 Scannet scenes collected by attaching the sensor to a handheld device such as an iPad in real-world conditions. The collected data suffer from severe motion blur in RGB images.

\textbf{Implementation details.}
The SDF decoder $\Gamma_\phi$, density decoder $\Gamma_\sigma$, color decoder $\Gamma_d$ and specular decoder $\Gamma_s$ all use two-layers MLPs with the hidden dimension of 32. We sample $M=6144$ rays for each iteration, and each ray contains $N_c=96$ coarse samples and $N_f=36$ fine samples. Our method is implemented in Pytorch and trained with the ADAM optimizer with a learning rate of $1 \times 10^{-3}$, and $1 \times 10^{-2}$ for MLP decoders, multi-resolution grids features, respectively. We run 20K iterations in all scenes with the learning rate decay at iteration 10000 and 15000, and the decay rate is $1/3$.
% we run all experiments on a desktop PC with a 3.60GHz Intel i7-12700K CPU and an NVIDIA RTX-3090 GPU.  

\textbf{Baselines.}
% 分为3*2的体系，方法包括经典算法和最新算法，类别分为几何、渲染、几何+渲染。
To validate the effectiveness of the proposed method, we compare it with approaches for indoor 3D  reconstruction and view rendering, respectively. For indoor 3D reconstruction, we compare with registration-based method including \textit{BundleFusion}~\cite{Ref27}, \textit{Colmap}~\cite{Ref66, Ref67, Ref68}, \textit{Convolutional Occupancy Networks}~\cite{Ref42}, \textit{SIREN}~\cite{Ref70}, and recent volumetric rendering-based geometry reconstruction methods such as \textit{Neus}~\cite{Ref7}, \textit{VolSDF}~\cite{Ref8}, \textit{Neuralangelo}~\cite{Ref12}, \textit{Neural RGBD}~\cite{Ref13}, and \textit{Go-Surf}~\cite{Ref14}. 
We run marching cubes at the resolution of 1cm to extract meshes. We cull the points and faces in the areas that are not observed in any camera views as in ~\cite{Ref13,Ref14}. For view rendering, we compare some representative approaches based on the neural radiance field, including \textit{DVGO}~\cite{Ref5}, and \textit{InstantNGP}~\cite{Ref6}. Their performance on 3D reconstruction is also compared. 

% \textbf{Evaluation Metrics.}
% we use Accuracy (Acc), Completeness (Com), Chamfer-$l_1$ Distance (C-$l_1$), Normal Consistency (NC), and F-score to report reconstruction performance, as~\cite{Ref13,Ref14}, and Peak Signal-to-Noise Ratio (PSNR), SSIM, and L-PIPS to report view synthesis performance as~\cite{Ref5,Ref6}.

% For $COLMAP$, $ConvOccNets$ and $SIREN$, we directly report the results from in \textit{Go-surf} without re-experimentation, since in the original method they exhibit a significant decrease compared to \textit{Go-surf} or \textit{NeuralRGBD}, on the same dataset. For \textit{NeuralRGBD} and \textit{Go-surf}, we directly implemented the source code from their github, except for changing the original training set to 90\% to be consistent with all our experiments. In addition, \textit{sdfstudio} and \textit{nerfstudio} are used to implement the \textit{Neus}, \textit{VolSDF},  \textit{InstantNGP} and \textit{Neuralangelo} the officially provided execution instructions, with $pipeline.model.sdf-field.inside-outside$ set to True. The \textit{DVGO} we implemented the version of v2, which uses cuda codes for faster convergence.

% \textbf{Metrics.}
% We used evaluation metrics for geometry reconstruction and color rendering, respectively. The geometry evaluation metrics include accuracy, completion, chamfer $l_1$ distance, normal consistency and F-score, while color evaluation metrics include PSNR, SSIM and LPIPS(Alex). 

\subsection{Comparison}\label{Comparison}

\begin{figure}[ht!]
	\centering
        
	\begin{subfigure}{\linewidth}
            \rotatebox[origin=c]{90}{\footnotesize{N.-RGBD}\hspace{-1.1cm}}
            \begin{minipage}[t]{0.235\linewidth}
                \centering
                \includegraphics[width=1\linewidth]{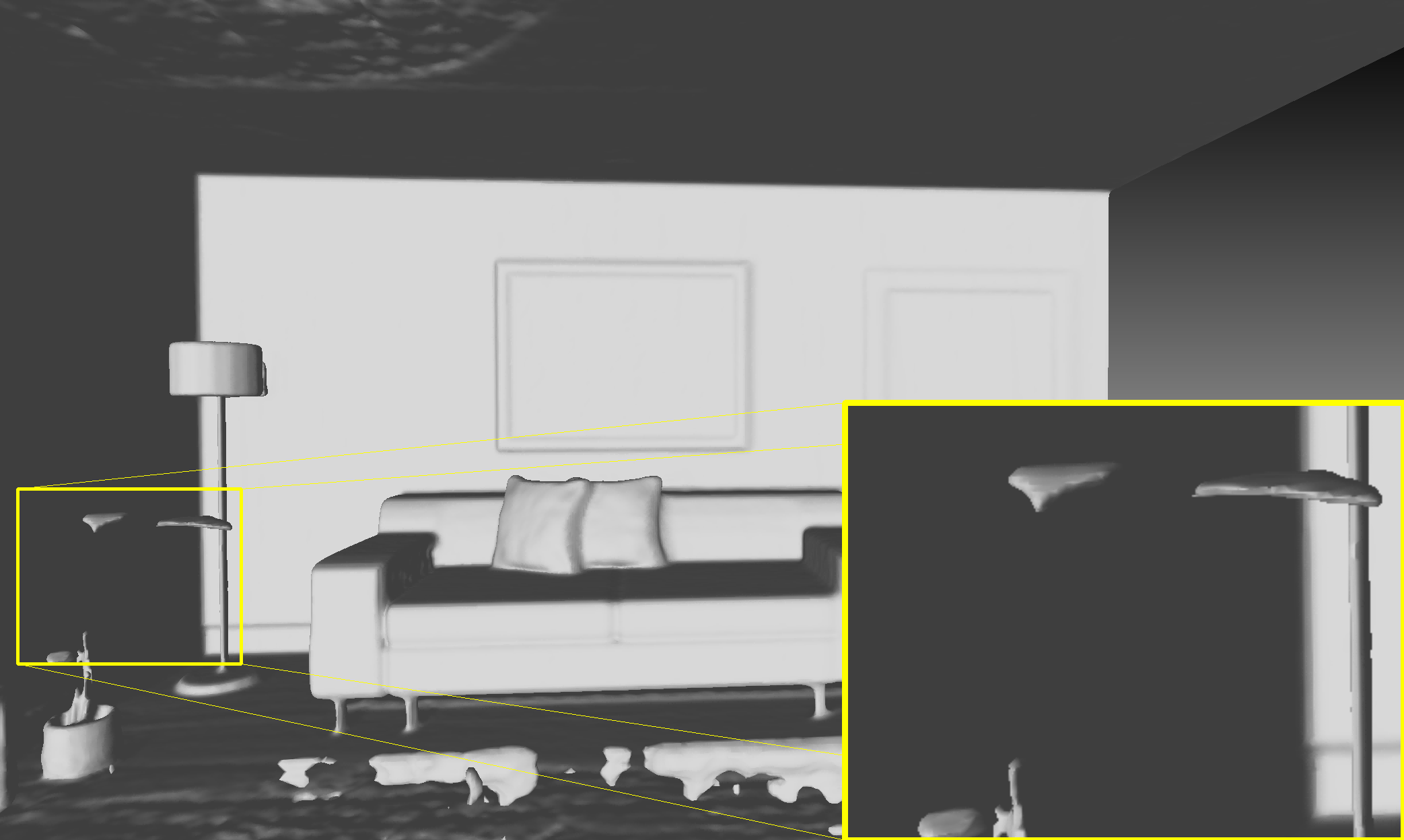}
            \end{minipage}
            \begin{minipage}[t]{0.235\linewidth}
                \centering
                \includegraphics[width=1\linewidth]{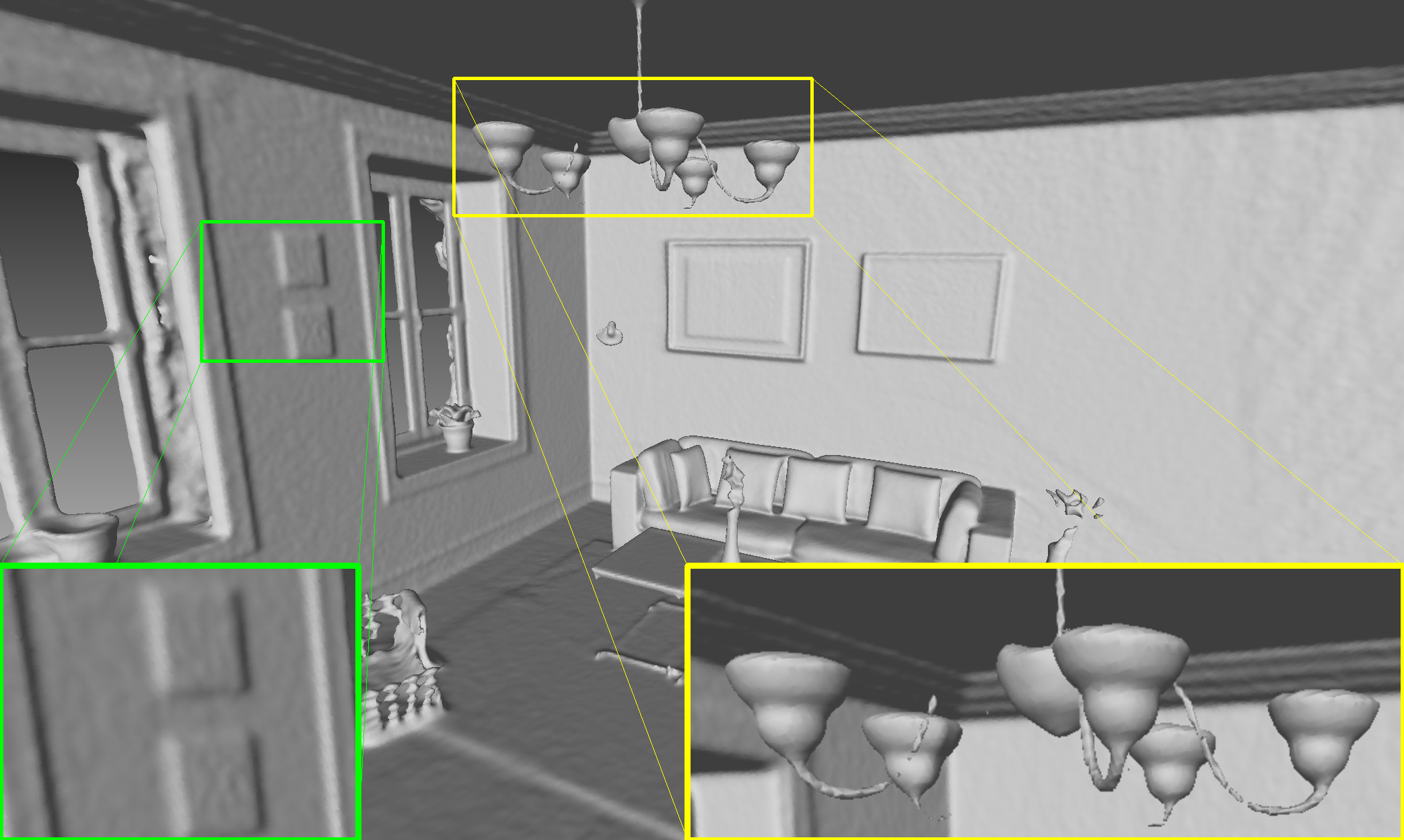}
            \end{minipage}
            \begin{minipage}[t]{0.235\linewidth}
                \centering
                \includegraphics[width=1\linewidth]{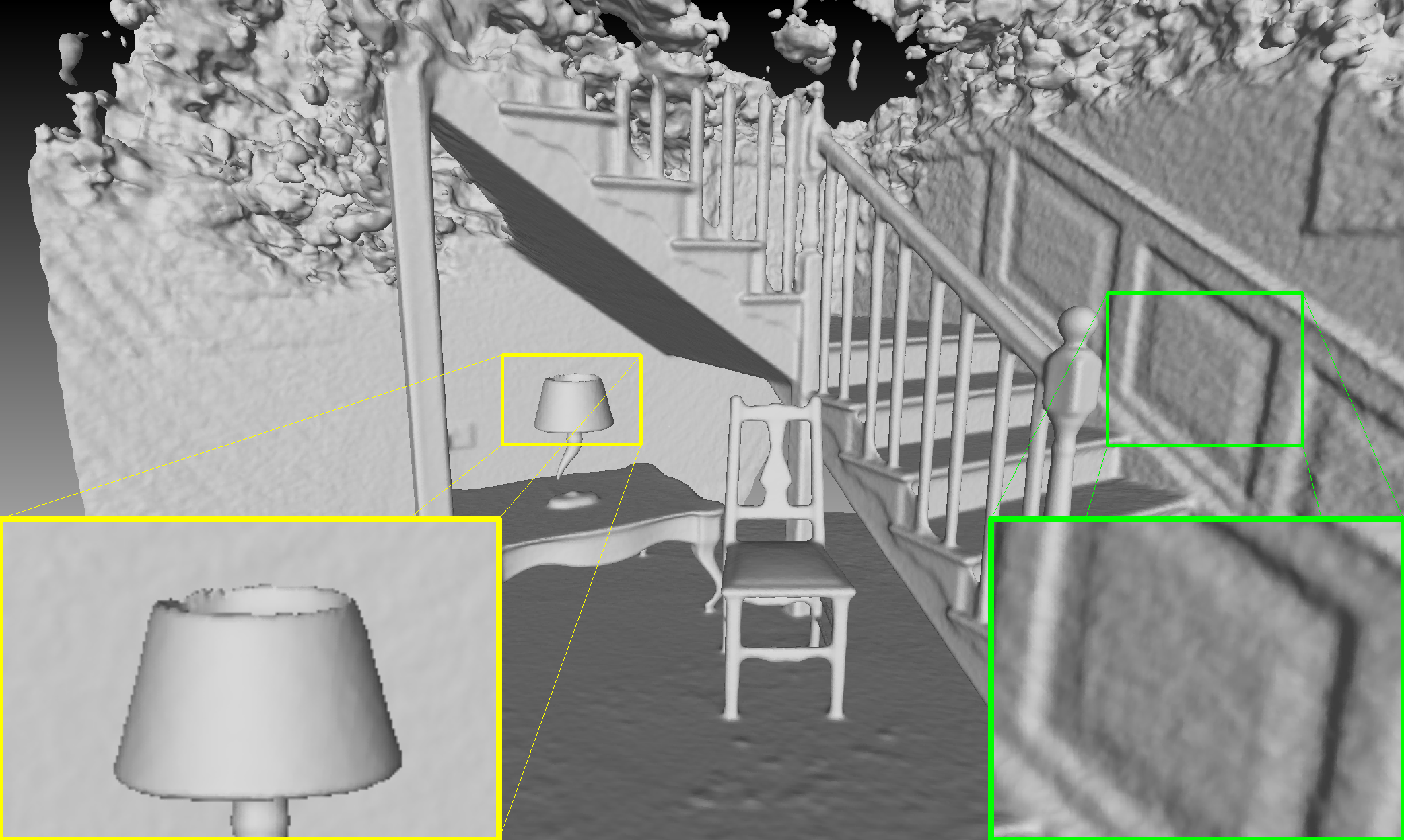}
            \end{minipage}
            \begin{minipage}[t]{0.235\linewidth}
                \centering
                \includegraphics[width=1\linewidth]{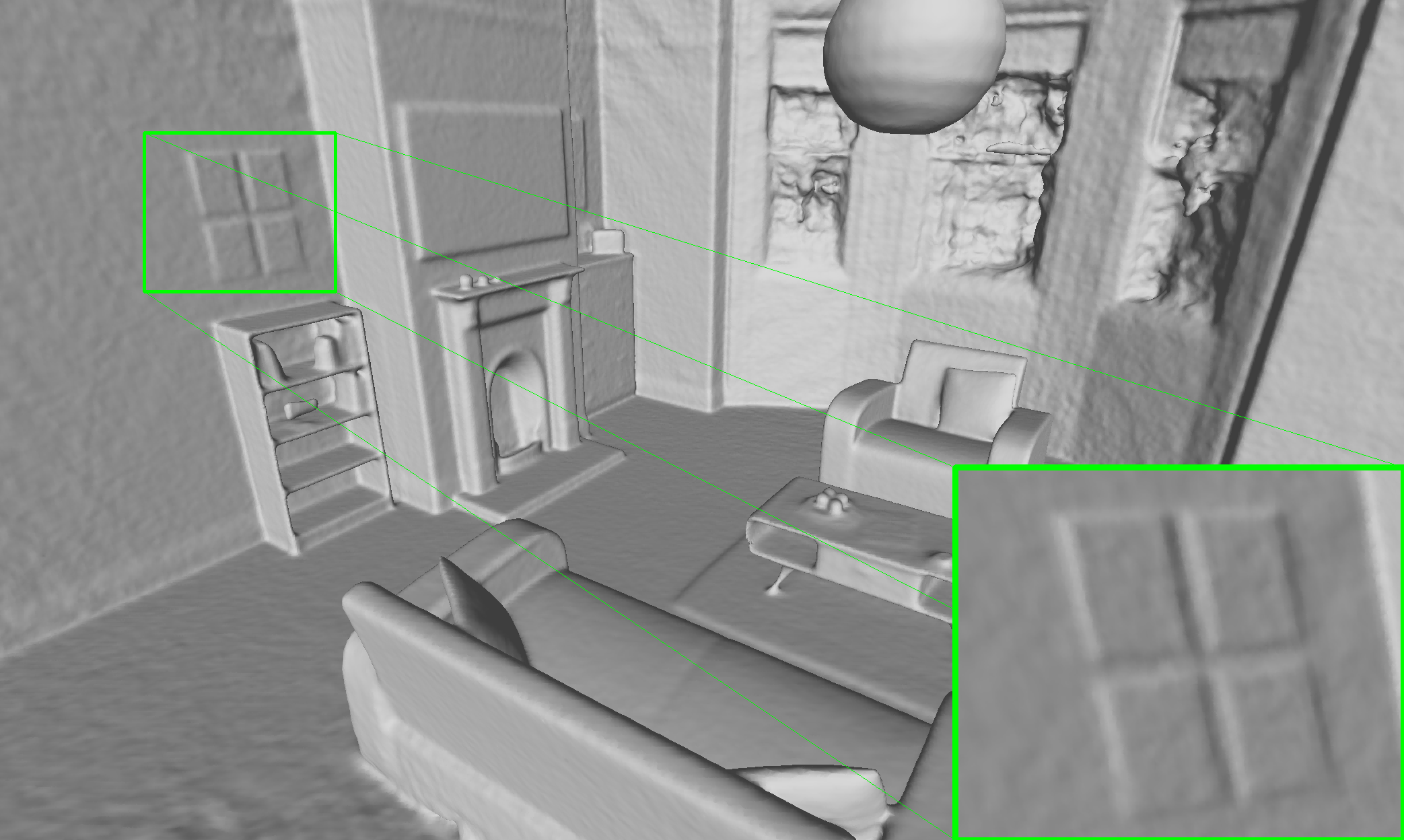}
            \end{minipage}
        \end{subfigure}

	\begin{subfigure}{\linewidth}
            \rotatebox[origin=c]{90}{\footnotesize{Go-Surf}\hspace{-1.1cm}}
            \begin{minipage}[t]{0.235\linewidth}
                \centering
                \includegraphics[width=1\linewidth]{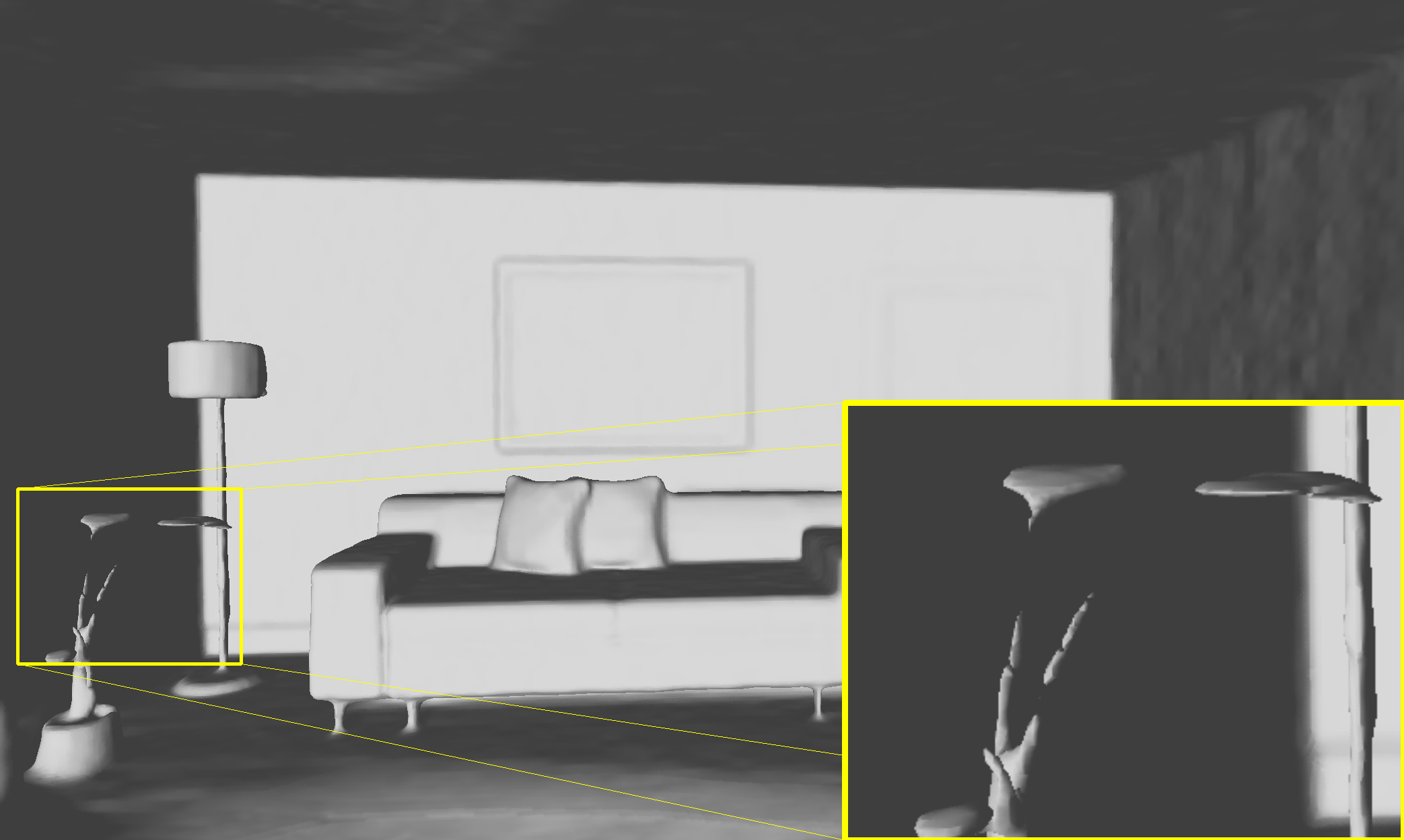}
            \end{minipage}
            \begin{minipage}[t]{0.235\linewidth}
                \centering
                \includegraphics[width=1\linewidth]{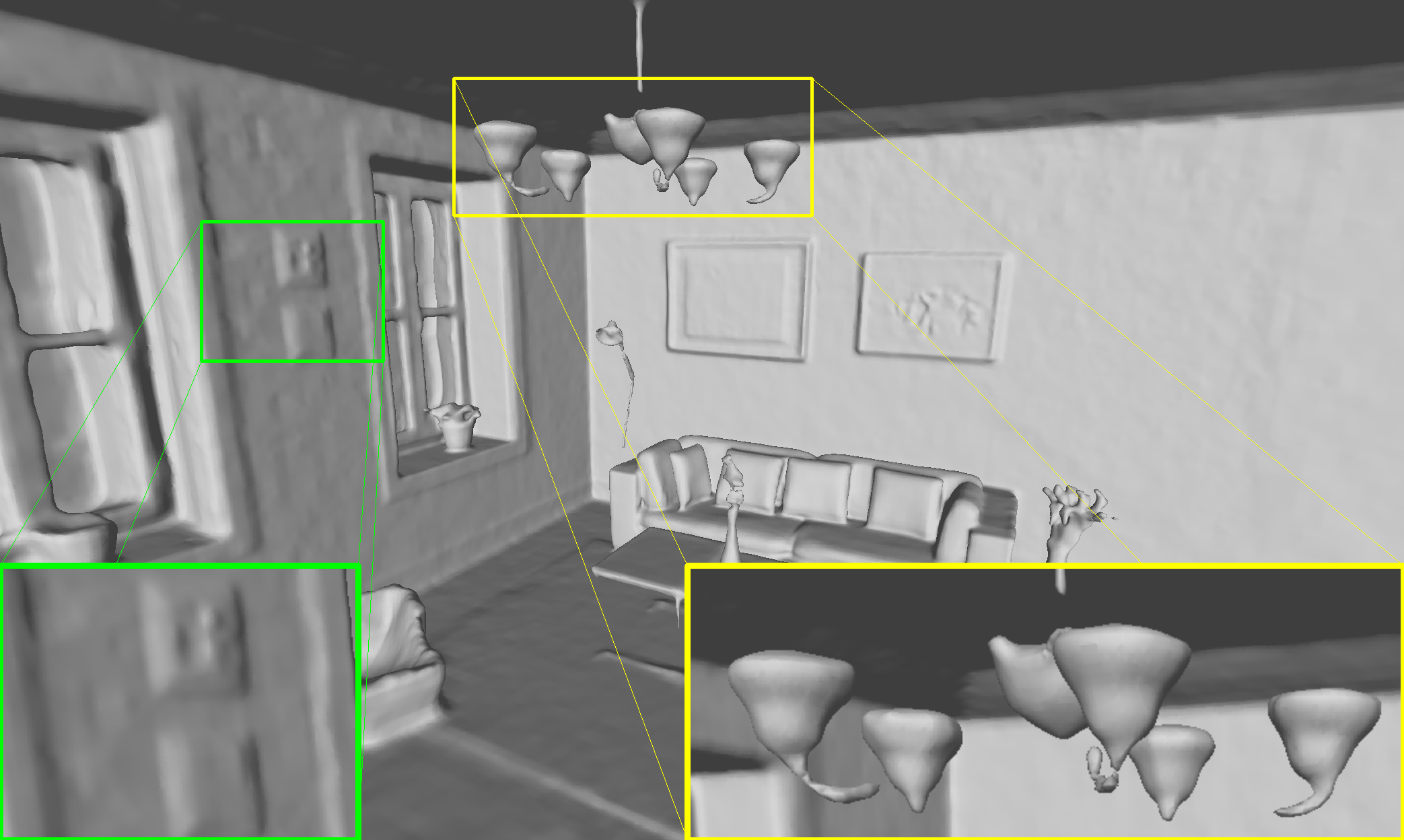}
            \end{minipage}
            \begin{minipage}[t]{0.235\linewidth}
                \centering
                \includegraphics[width=1\linewidth]{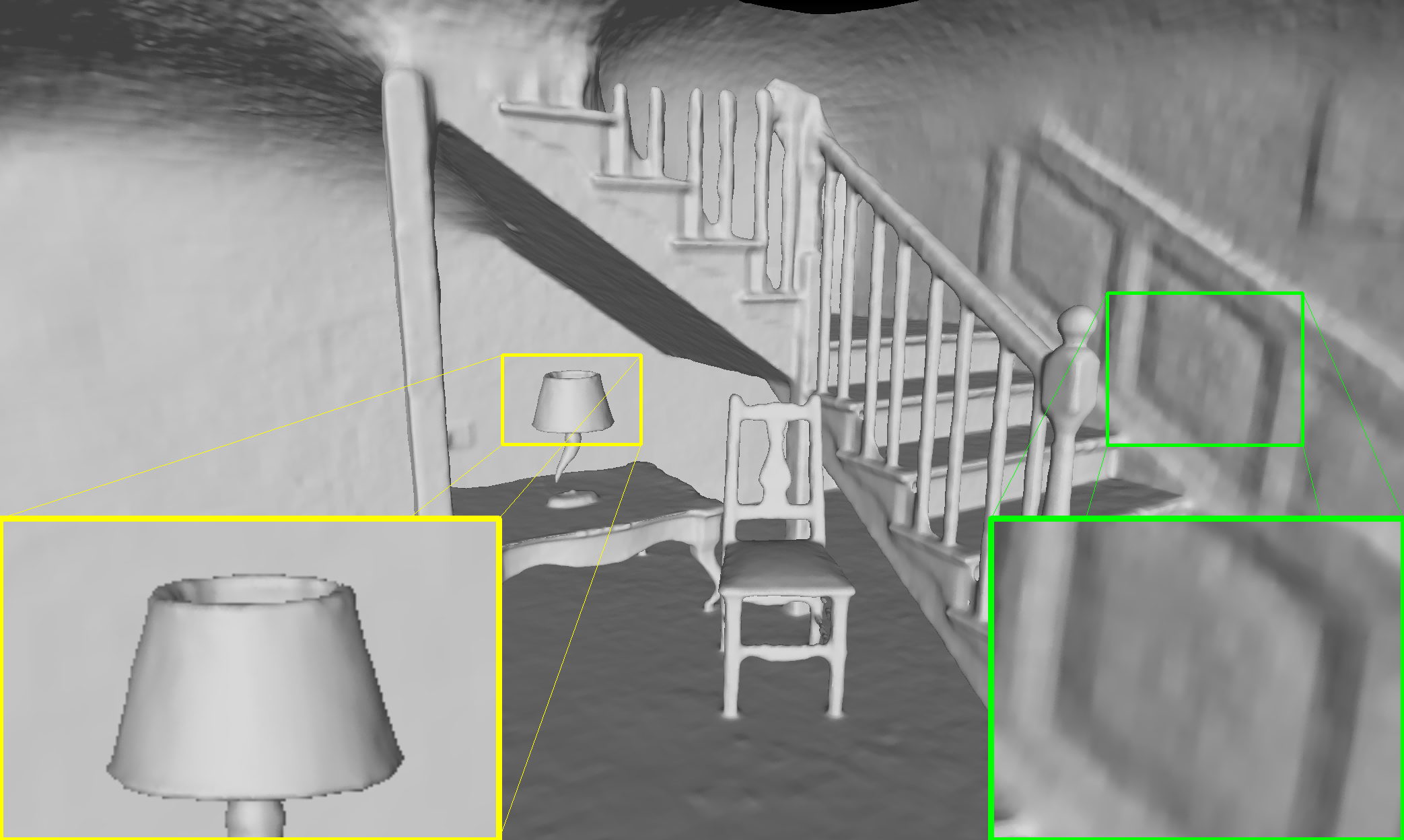}
            \end{minipage}
            \begin{minipage}[t]{0.235\linewidth}
                \centering
                \includegraphics[width=1\linewidth]{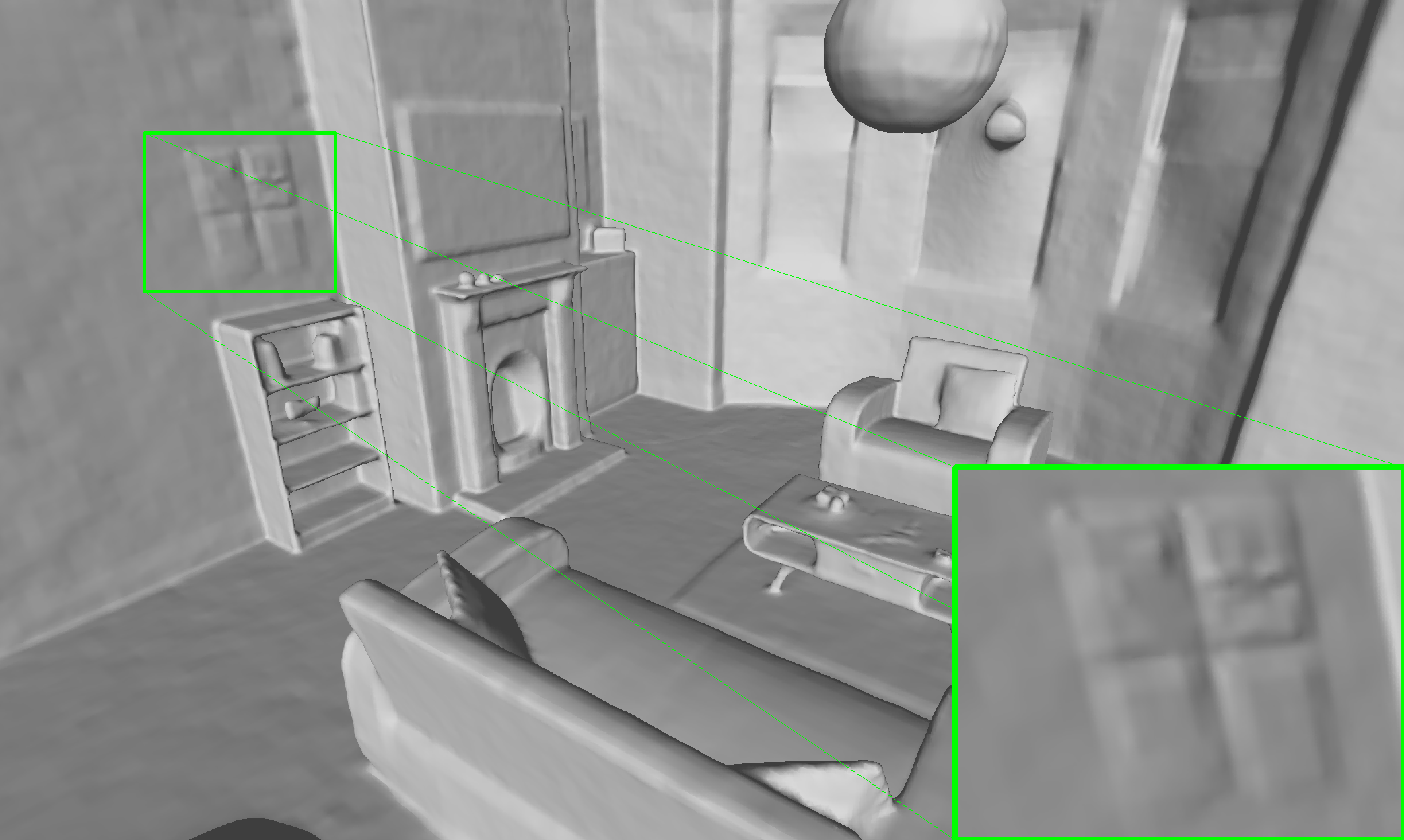}
            \end{minipage}
        \end{subfigure}

	\begin{subfigure}{\linewidth}
            \rotatebox[origin=c]{90}{\footnotesize{Ours}\hspace{-1.1cm}}
            \begin{minipage}[t]{0.235\linewidth}
                \centering
                \includegraphics[width=1\linewidth]{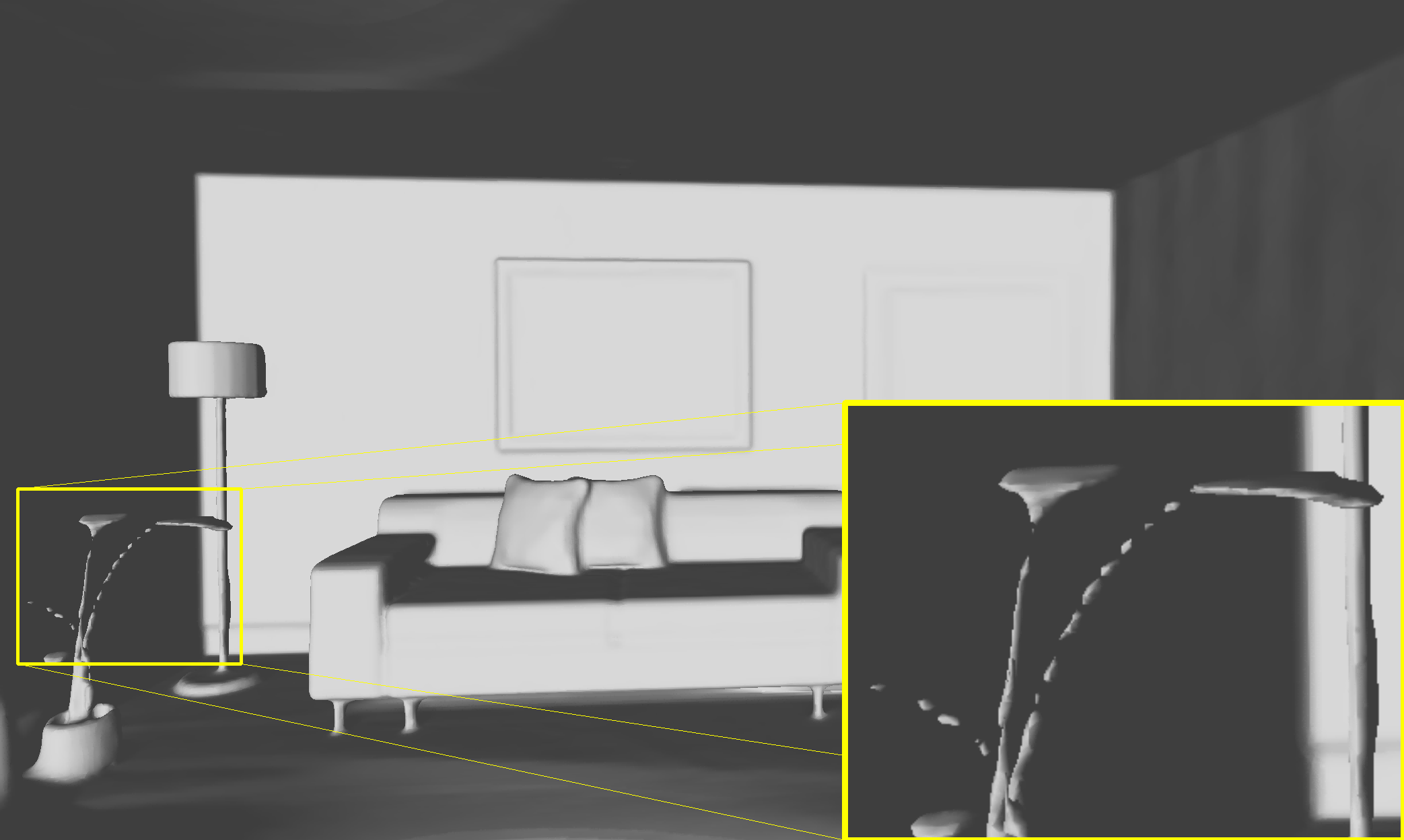}
            \end{minipage}
            \begin{minipage}[t]{0.235\linewidth}
                \centering
                \includegraphics[width=1\linewidth]{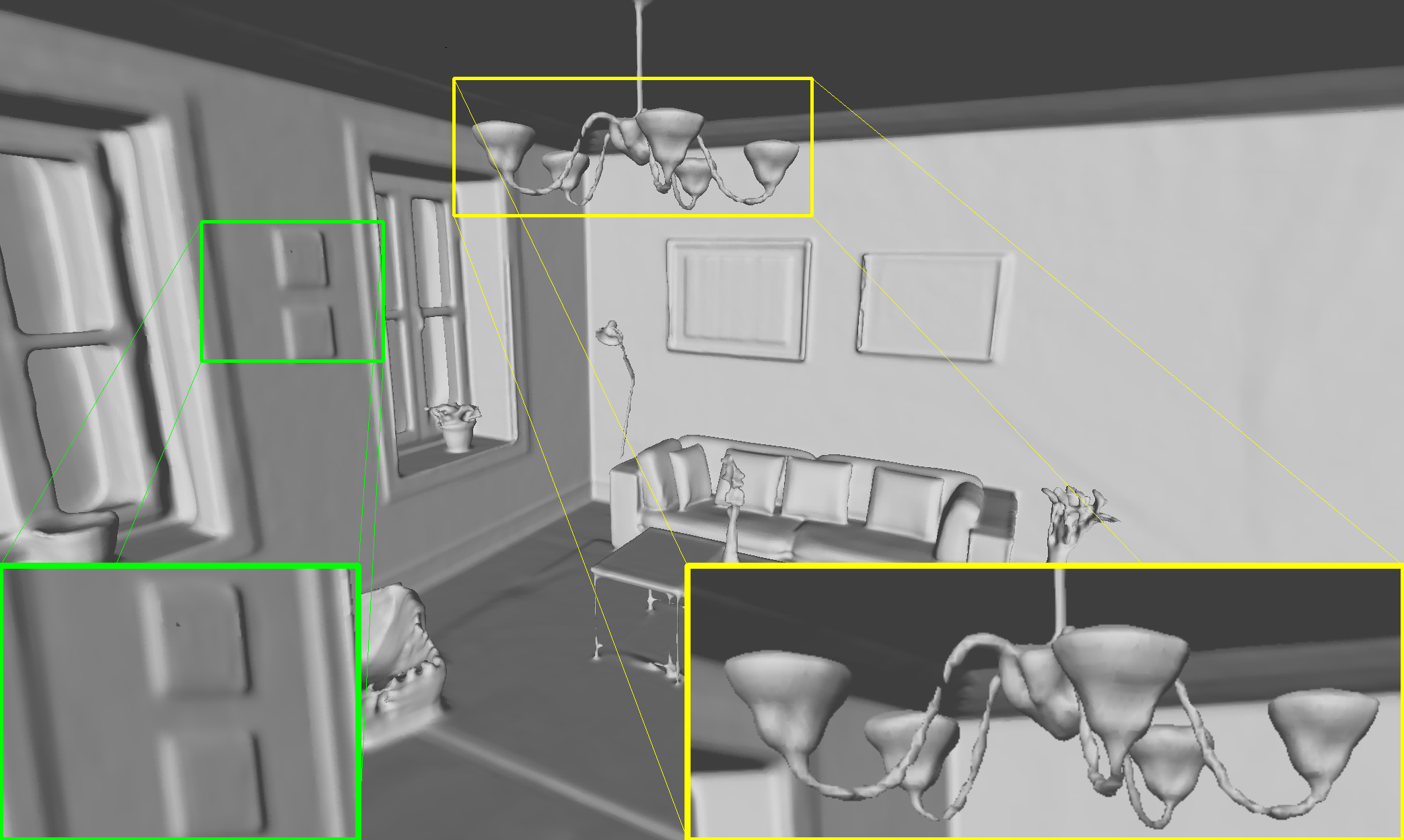}
            \end{minipage}
            \begin{minipage}[t]{0.235\linewidth}
                \centering
                \includegraphics[width=1\linewidth]{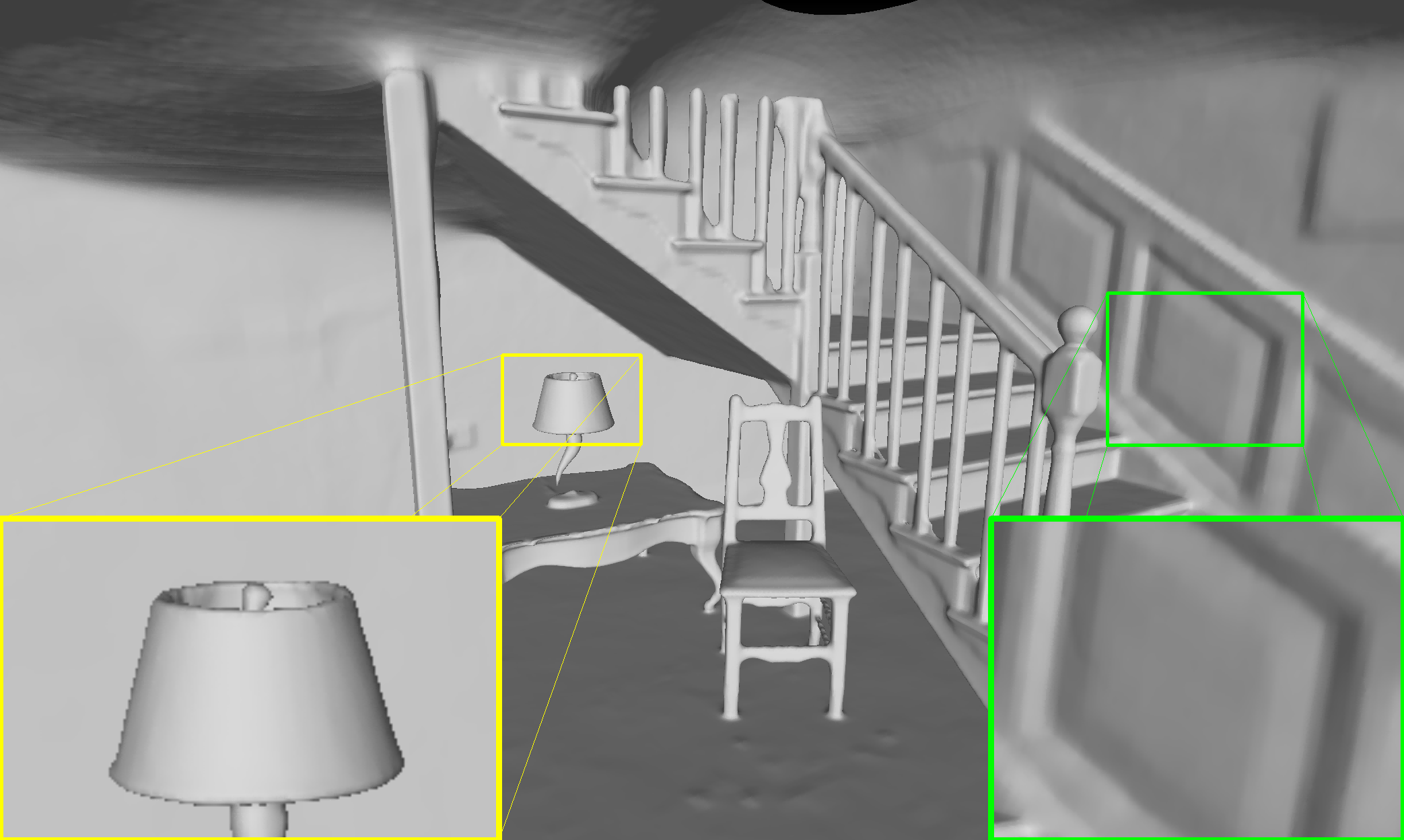}
            \end{minipage}
            \begin{minipage}[t]{0.235\linewidth}
                \centering
                \includegraphics[width=1\linewidth]{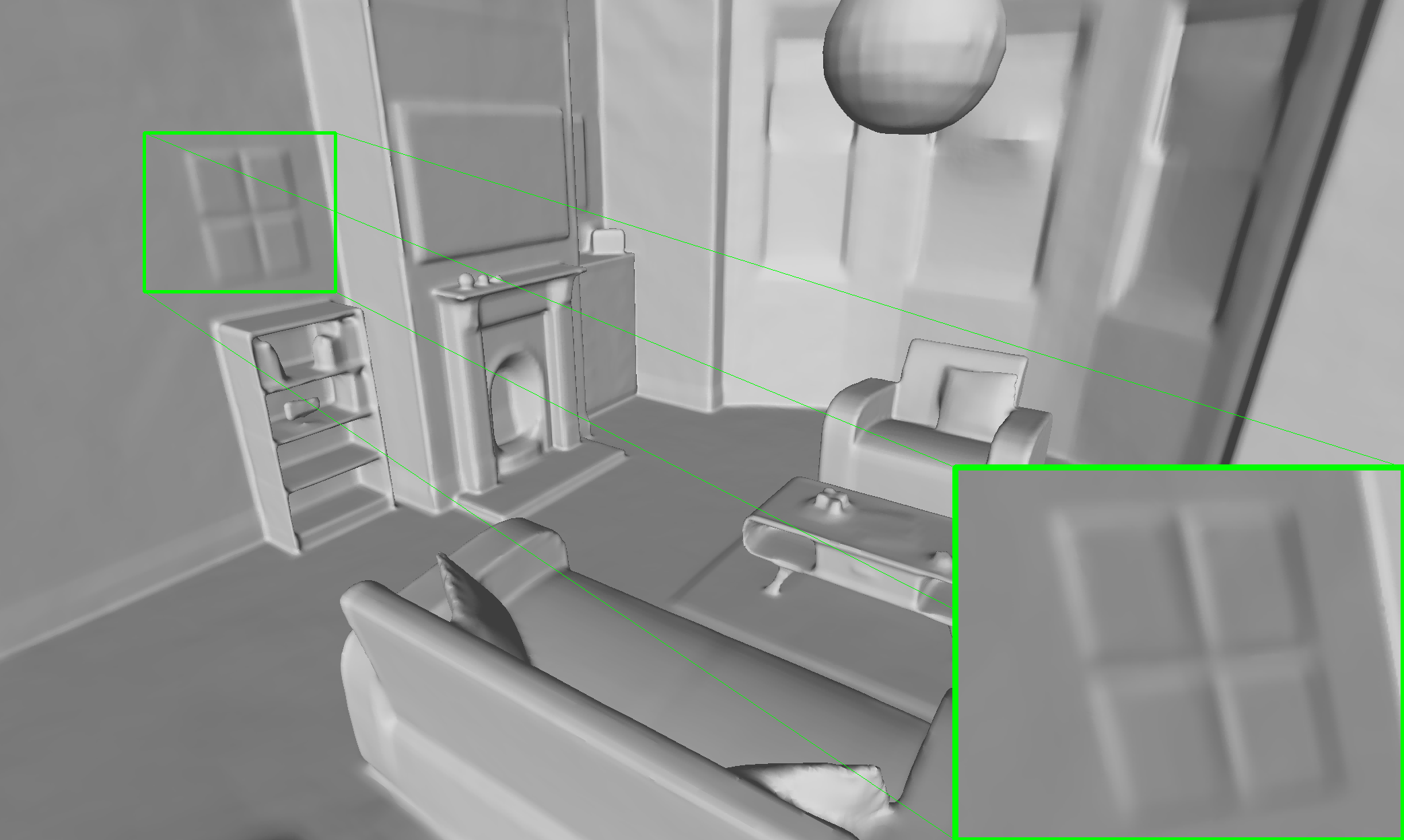}
            \end{minipage}
        \end{subfigure}

	\begin{subfigure}{\linewidth}
            \rotatebox[origin=c]{90}{\footnotesize{GT}\hspace{-1.1cm}}
            \begin{minipage}[t]{0.235\linewidth}
                \centering
                \includegraphics[width=1\linewidth]{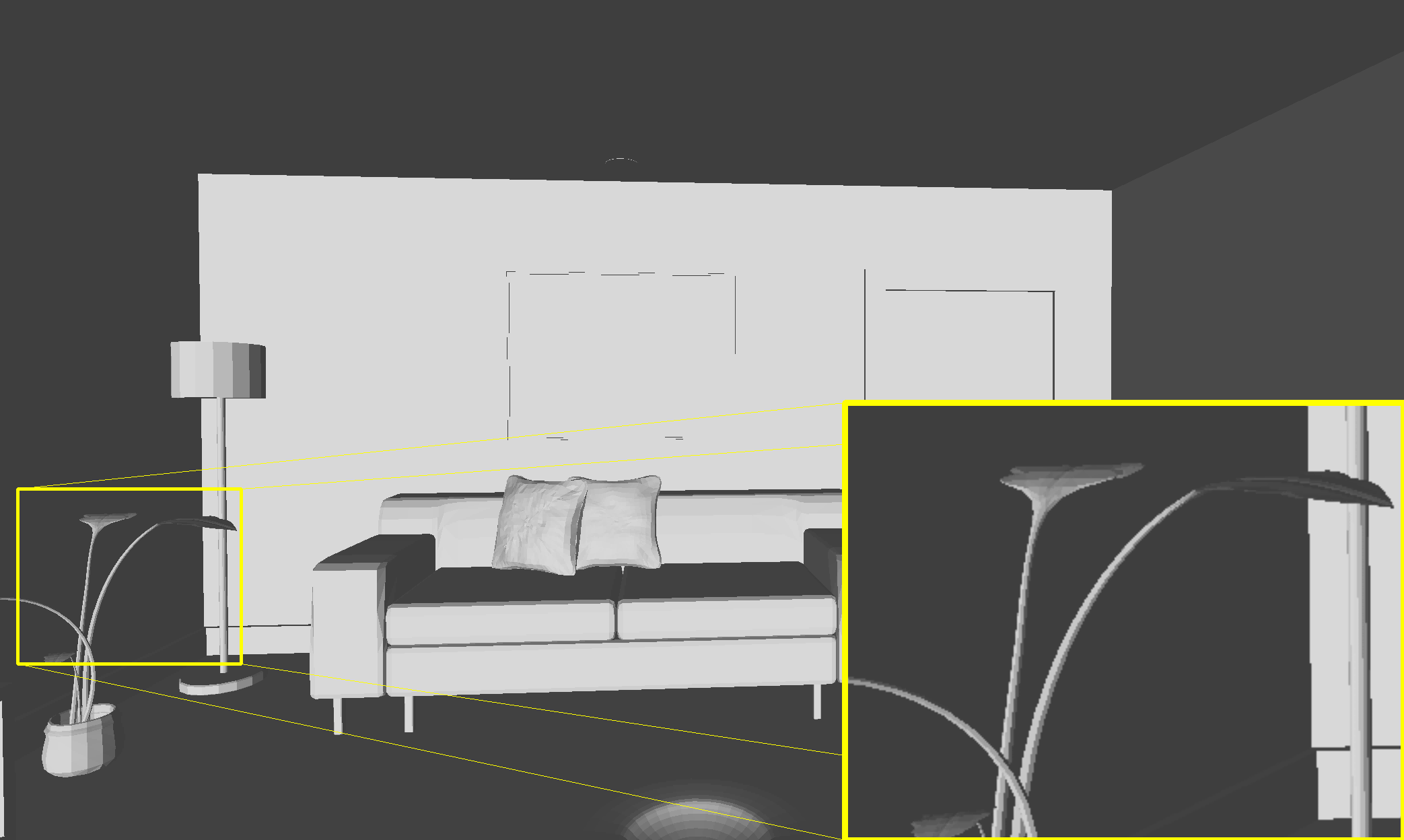}
            \end{minipage}
            \begin{minipage}[t]{0.235\linewidth}
                \centering
                \includegraphics[width=1\linewidth]{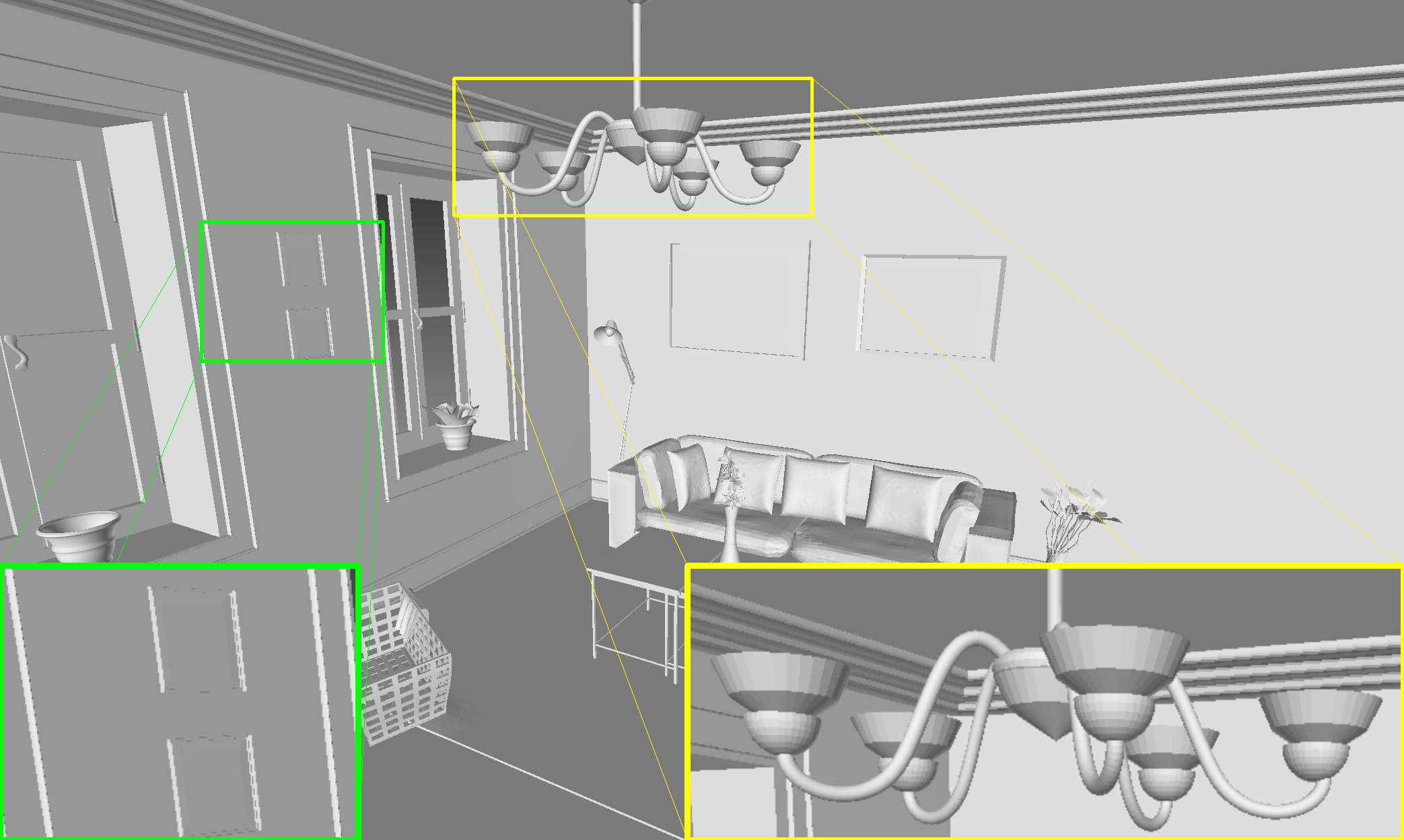}
            \end{minipage}
            \begin{minipage}[t]{0.235\linewidth}
                \centering
                \includegraphics[width=1\linewidth]{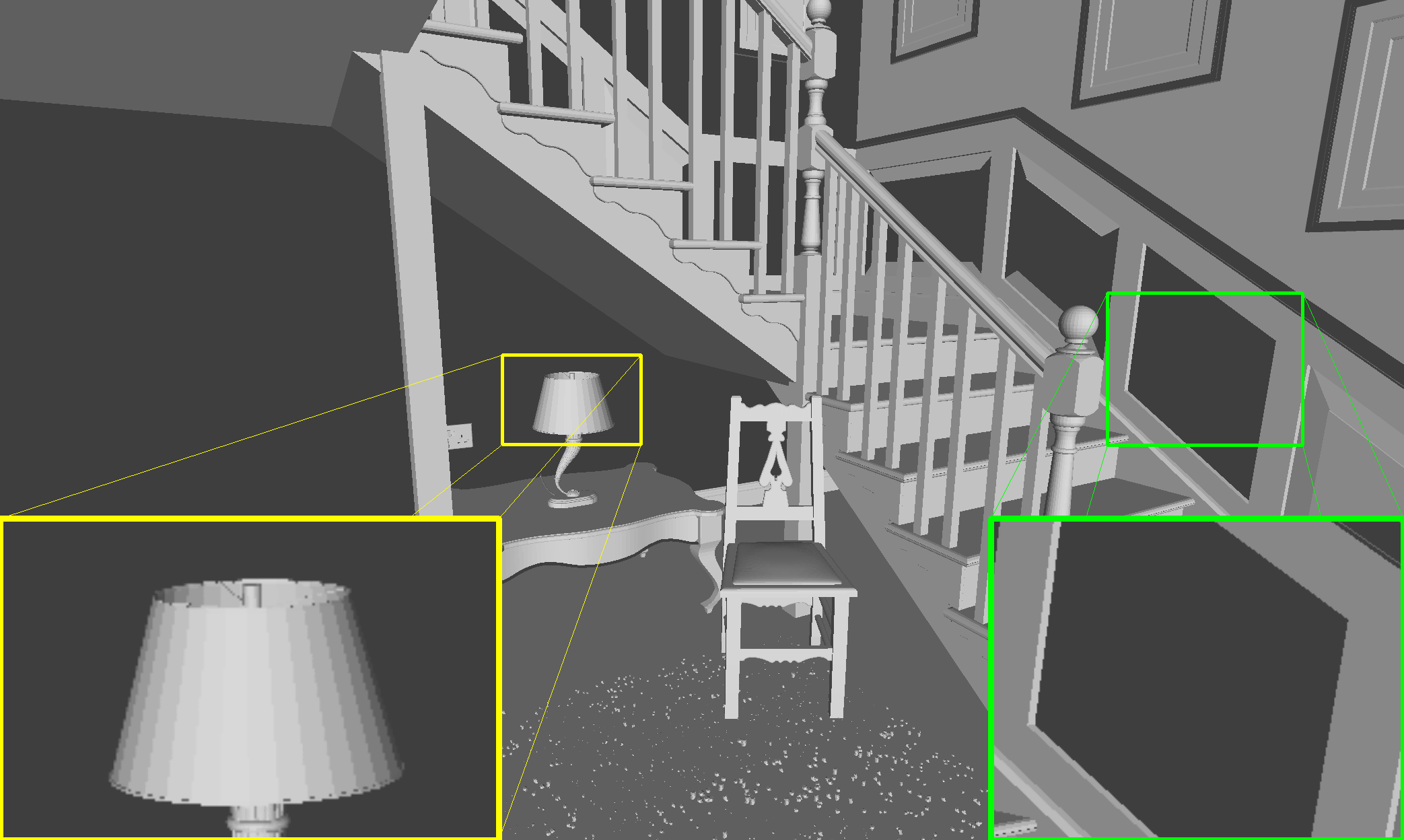}
            \end{minipage}
            \begin{minipage}[t]{0.235\linewidth}
                \centering
                \includegraphics[width=1\linewidth]{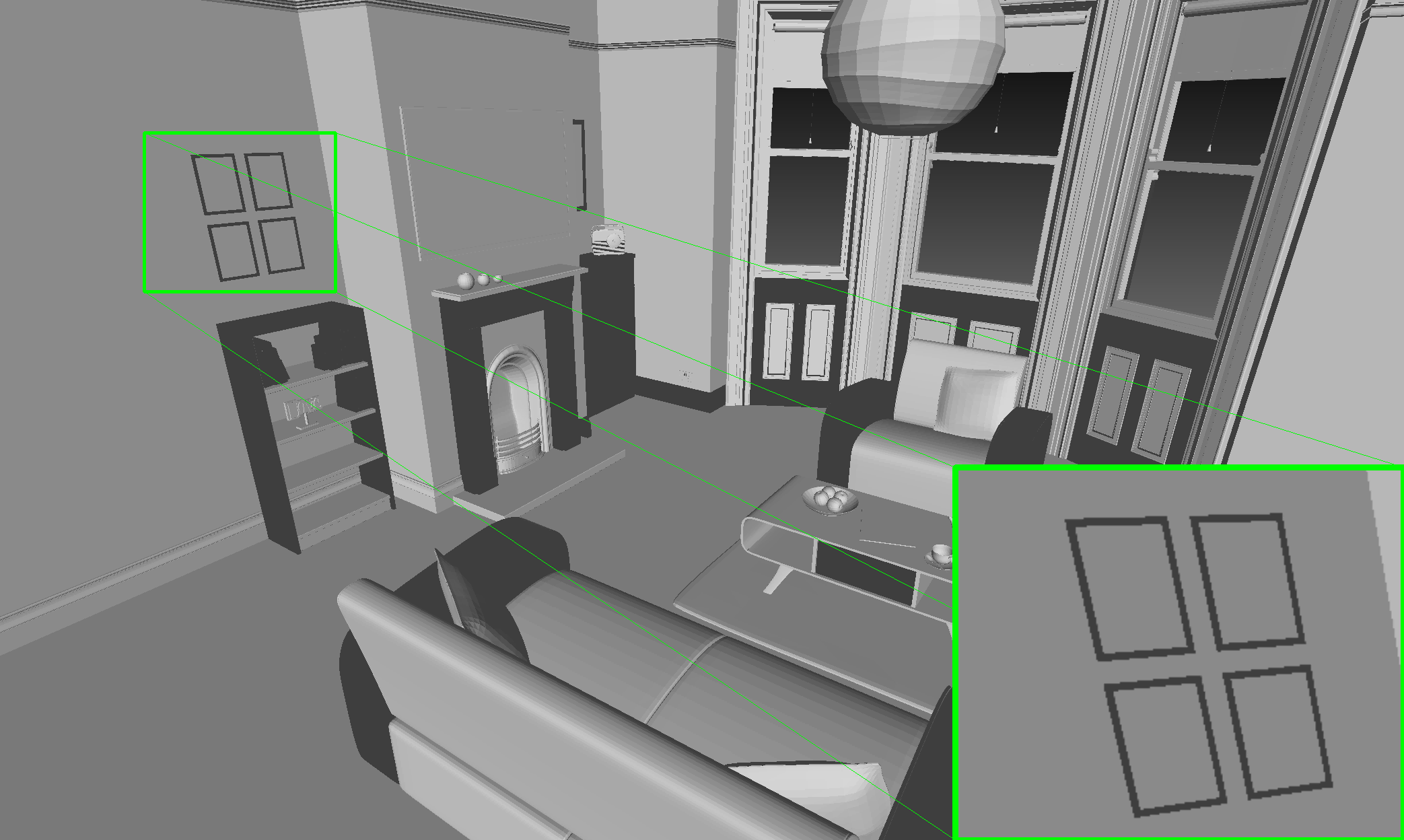}
            \end{minipage}
        \end{subfigure}

	\caption{The qualitative reconstruction results on NeuralRGBD datasets. The proposed method can fill in the missing part (highlighted in yellow boxes) and produce  smoother planes and clear edges (highlighted in green boxes) }
	\label{fig:fig3}
\end{figure}

 \begin{figure}[ht!]
	\centering
	\begin{subfigure}{\linewidth}
            \begin{minipage}[t]{0.242\linewidth}
                \centering
                \includegraphics[width=1\linewidth]{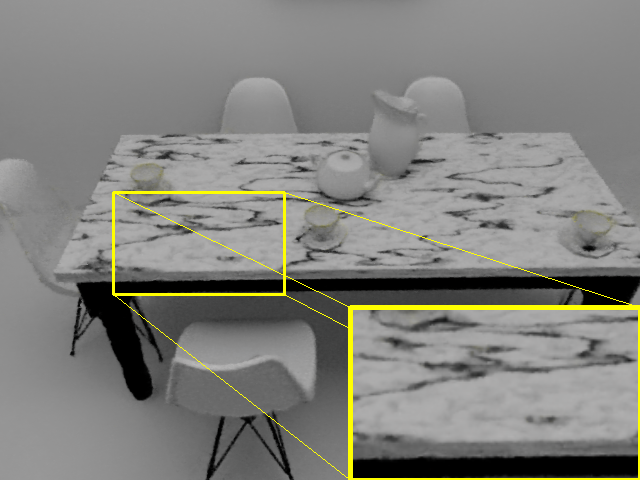}
            \end{minipage}
            \begin{minipage}[t]{0.242\linewidth}
                \centering
                \includegraphics[width=1\linewidth]{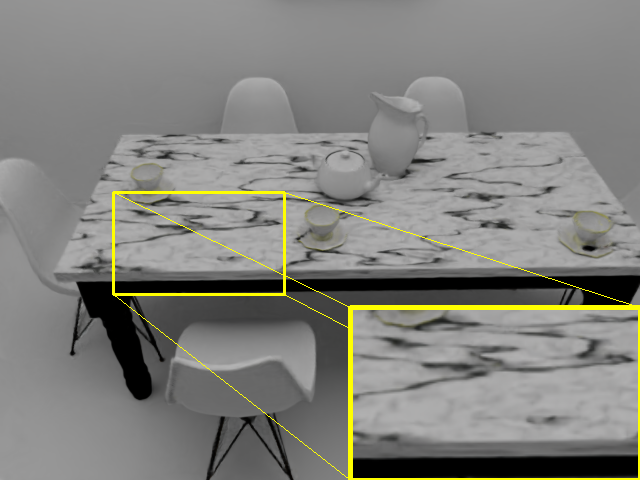}
            \end{minipage}
            \begin{minipage}[t]{0.242\linewidth}
                \centering
                \includegraphics[width=1\linewidth]{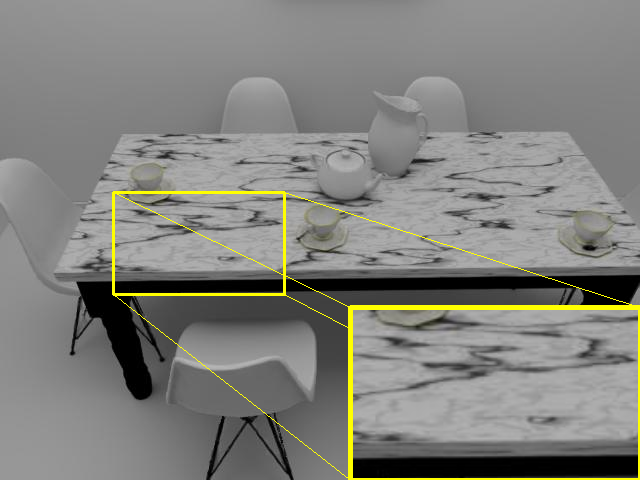}
            \end{minipage}
            \begin{minipage}[t]{0.242\linewidth}
                \centering
                \includegraphics[width=1\linewidth]{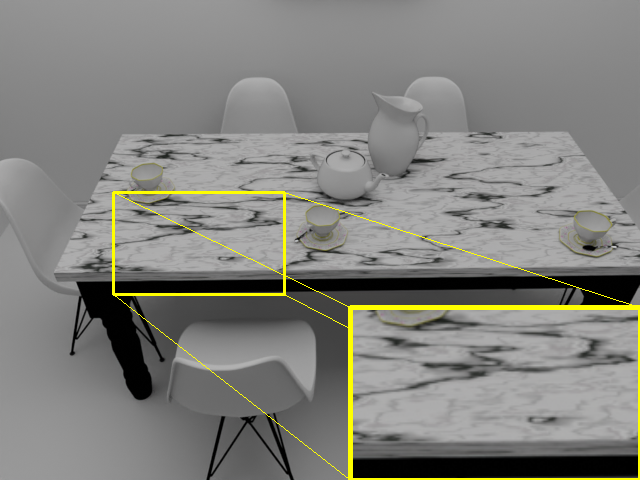}
            \end{minipage}
        \end{subfigure}
        
	\begin{subfigure}{\linewidth}
            \begin{minipage}[t]{0.242\linewidth}
                \centering
                \includegraphics[width=1\linewidth]{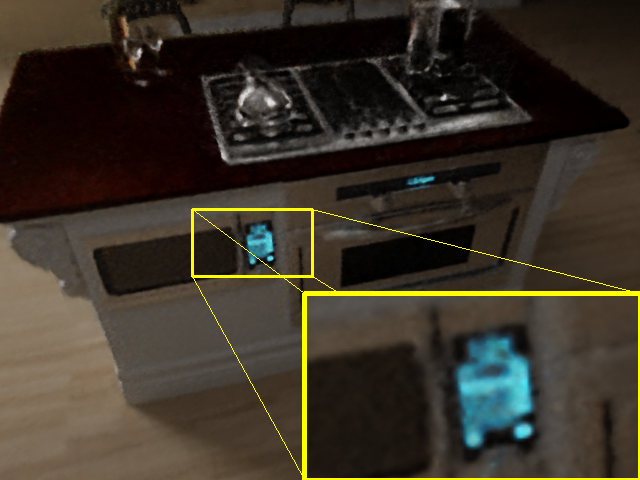}
            \end{minipage}
            \begin{minipage}[t]{0.242\linewidth}
                \centering
                \includegraphics[width=1\linewidth]{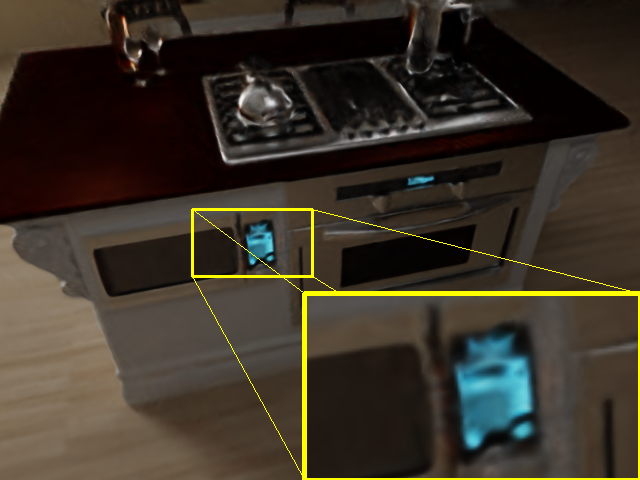}
            \end{minipage}
            \begin{minipage}[t]{0.242\linewidth}
                \centering
                \includegraphics[width=1\linewidth]{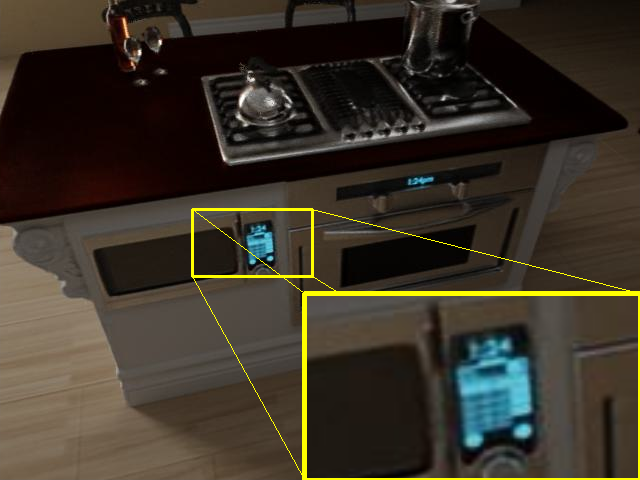}
            \end{minipage}
            \begin{minipage}[t]{0.242\linewidth}
                \centering
                \includegraphics[width=1\linewidth]{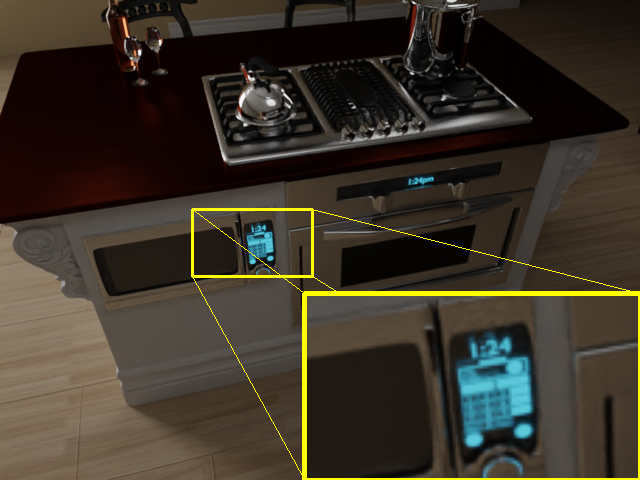}
            \end{minipage}
        \end{subfigure}

	\begin{subfigure}{\linewidth}
            \begin{minipage}[t]{0.242\linewidth}
                \centering
                \includegraphics[width=1\linewidth]{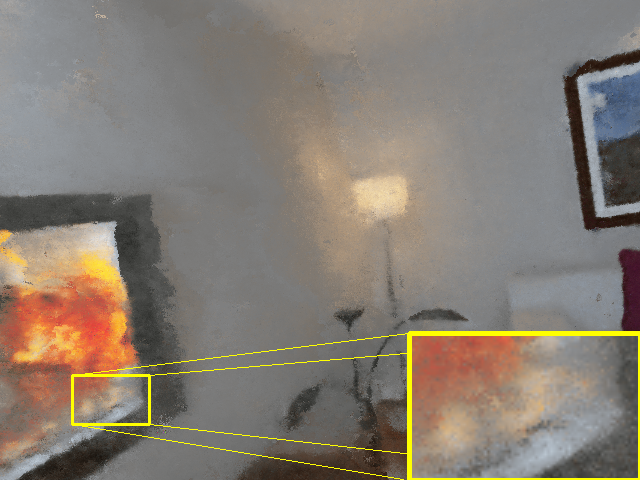}
            \end{minipage}
            \begin{minipage}[t]{0.242\linewidth}
                \centering
                \includegraphics[width=1\linewidth]{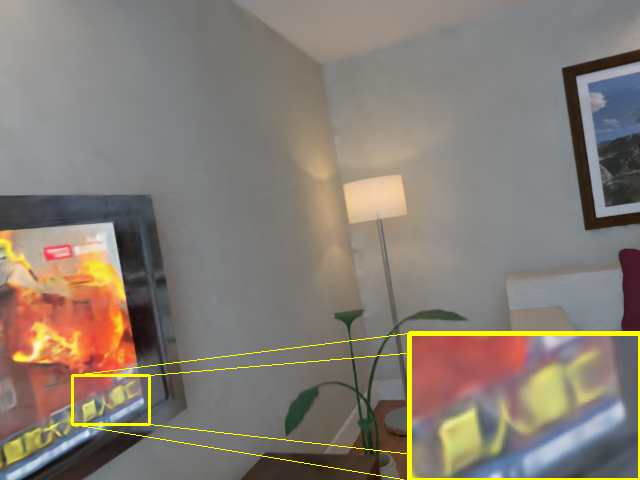}
            \end{minipage}
            \begin{minipage}[t]{0.242\linewidth}
                \centering
                \includegraphics[width=1\linewidth]{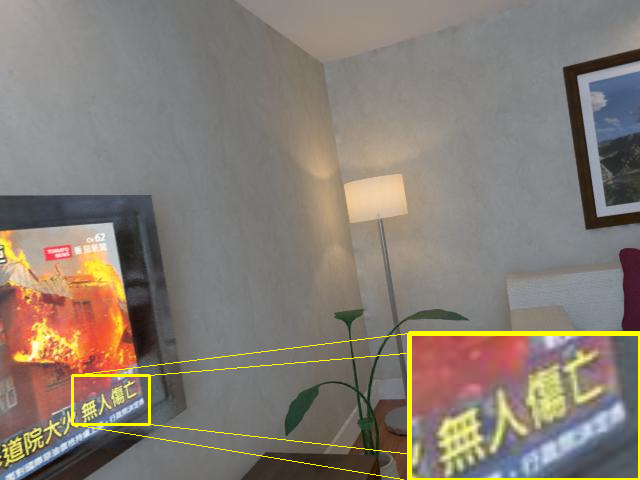}
            \end{minipage}
            \begin{minipage}[t]{0.242\linewidth}
                \centering
                \includegraphics[width=1\linewidth]{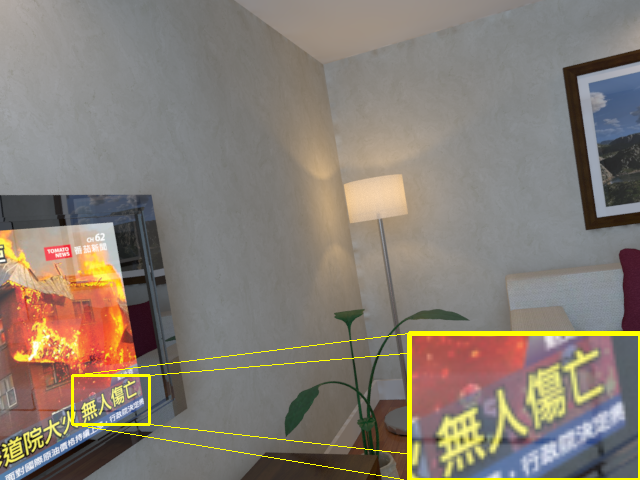}
            \end{minipage}
        \end{subfigure}

	\begin{subfigure}{\linewidth}
            \begin{minipage}[t]{0.242\linewidth}
                \centering
                \includegraphics[width=1\linewidth]{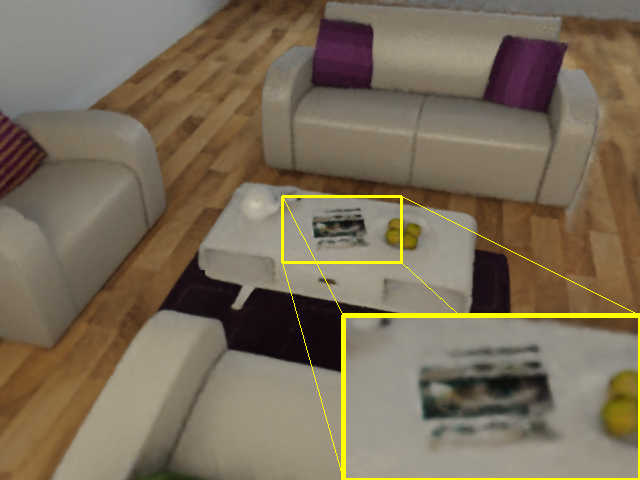}
                \caption{InstantNGP}
            \end{minipage}
            \begin{minipage}[t]{0.242\linewidth}
                \centering
                \includegraphics[width=1\linewidth]{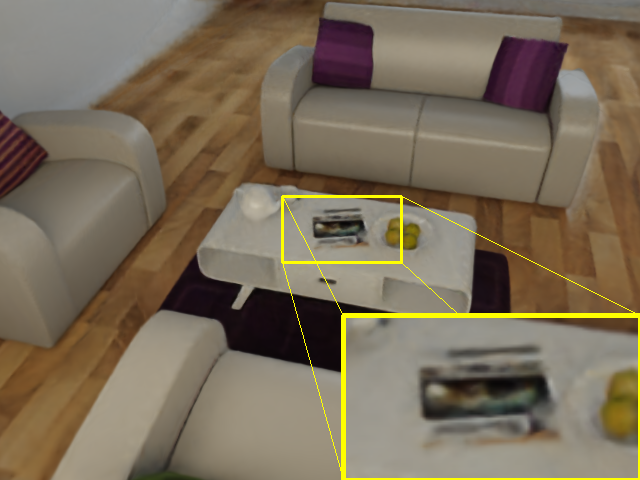}
                \caption{DVGO}
            \end{minipage}
            \begin{minipage}[t]{0.242\linewidth}
                \centering
                \includegraphics[width=1\linewidth]{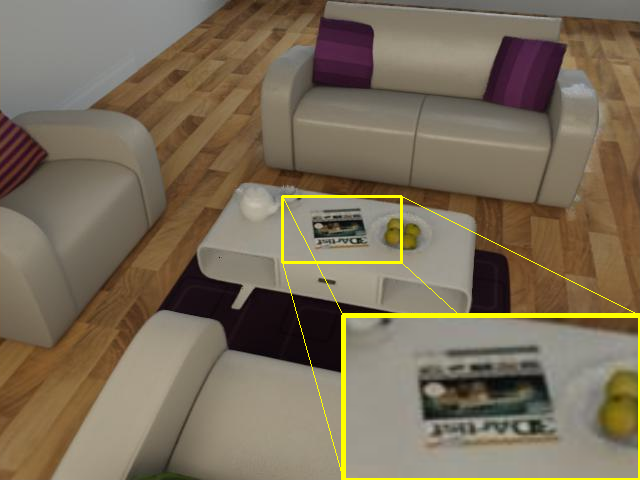}
                \caption{Ours}
            \end{minipage}
            \begin{minipage}[t]{0.242\linewidth}
                \centering
                \includegraphics[width=1\linewidth]{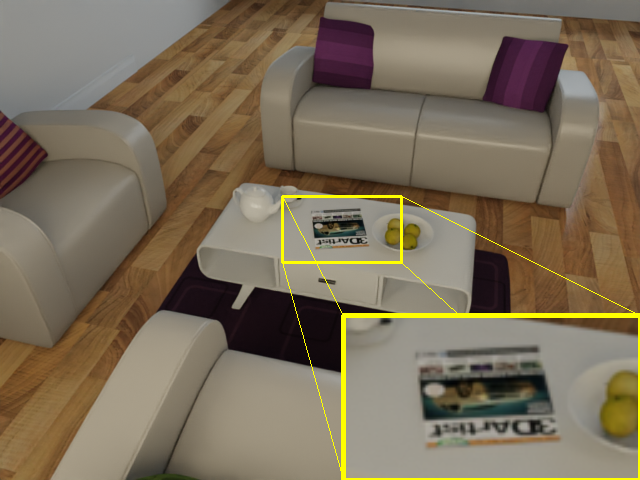}
                \caption{GT}
            \end{minipage}
        \end{subfigure}

	\caption{Qualitative comparison of novel view synthesis results of NeuralRGBD dataset. It can be seen from the results that our method has better visual rendering effects, whether they are striped structures or text on books.}
	\label{fig:fig4}
\end{figure}

\textbf{Evaluation on NeuralRGBD Dataset.}We evaluated our method on 10 synthesis data used in \textit{Neural RGB-D} and \textit{Go-Surf}. Qualitative and quantitative results are shown in \cref{fig:fig4,fig:fig3} and \cref{tab:tab1}.
It can be seen from the \cref{tab:tab1} that our method achieves the best performance on both view rendering and 3D reconstruction. Detailed rendering and reconstruction results can be found in the Supplementary.

For the 3D reconstruction task, the proposed method shows superior results on all metrics except the \textit{NeuralRGBD} approach in terms of accuracy (\cref{tab:tab1}). \cref{fig:fig3} shows that our method generate smoother geometry ( green boxes), and can fill in the missing stripes of the indoor objects (yellow boxes). For the view synthesis task, our method shows an improvement of $3-10$ db over the previous representative approaches (\cref{tab:tab1}), and the rendered images have richer textures (\cref{fig:fig3}), \textit{e.g.}
the subtitles on the TV and the magazine on the table.

% For the 3D reconstruction task, it can be seen that our method outperforms the traditional registration-based methods in all evaluation metrics of 3D reconstruction. Compared to the volume-rendering-based 3D reconstruction methods, the proposed method also shows superior results on all metrics except the neural RGBD approach in terms of accuracy. As depicted in the \cref{fig:fig3}, our method contains smoother geometry for indoor scenes, including walls and ceilings, and is capable of filling in the missing strip parts of the indoor objects.
% Furthermore, we obtain better view rendering performance by a large margin as can be seen from \cref{tab:tab1}, compared to the volume-rendering-based reconstruction methods. 

% For the view synthesis task, we outperform two representative approaches, which have similar scene representations to our method. Additionally, our 3D reconstruction result is evidently better than these two approaches. Specifically, our method shows an improvement of $3-10$ dB over \textit{DVGO} and \textit{Instant-NGP}, while achieving approximately $3$ times higher than them in F-score.
% The \cref{fig:fig4} clearly illustrates the superiority in view rendering tasks, where we can distinctly discern the subtitles on the TV and the magazine on the table, while \textit{DVGO} and \textit{Instant-NGP} appear blurred.

\begin{table}[ht!]
  \caption{Reconstruction and view synthesis results of NeuralRGBD dataset. The best performances are highlighted in bold.}
  \label{tab:tab1}
  \centering
  \Huge
  \resizebox{0.5\textwidth}{!}{
  \begin{tabular}{@{}lcccccccc@{}}
    \toprule
    \textbf{Method} & \textbf{Acc $\downarrow$} & \textbf{Com $\downarrow$} & \textbf{C-$l_1 \downarrow$} & \textbf{NC $\uparrow$} & \textbf{F-score $\uparrow$} & \textbf{PSNR $\uparrow$} & \textbf{SSIM $\uparrow$} & \textbf{LPIPS $\downarrow$} \\
    
    \midrule
    BundleFusion & 0.0178 & 0.4577 & 0.2378 & 0.851 & 0.680 & - & - & - \\

    COLMAP & 0.0271 & 0.0364 & 0.0293 & 0.888 & 0.874  & - & - & -\\ 
    ConvOccNets & 0.0498 & 0.0524 & 0.0511 & 0.861 & 0.682  & - & - & -\\
    SIREN & 0.0229 & 0.0412 & 0.0320 & 0.905 & 0.852  & - & - & -\\

    \midrule
    Neus & 0.3174 & 0.6911 & 0.5043 & 0.628 & 0.103 & 27.465 & 0.849 & 0.192 \\
    VolSDF & 0.1627 & 0.4815 & 0.3222 & 0.681 & 0.262 & 28.717 & 0.882 & 0.174 \\
    Neuralangelo & 0.3212 & 0.6938 & 0.5075 & 0.531 & 0.101 & 29.786 & 0.892 & 0.121 \\
    Neural RGB-D & \textbf{0.0145} & 0.0508 & 0.0327 & 0.920 & 0.936 & 31.994 & 0.901 & 0.183 \\
    Go-Surf & 0.0164 & 0.0213 & 0.0189 & 0.932 & 0.949 & 29.586 & 0.889 & 0.183 \\
    
    \midrule
    DVGO & 0.2389 & 0.5558 & 0.3973 & 0.564 & 0.317 & 33.633 & 0.940 & 0.125 \\
    Instant-NGP & 0.2641 & 0.7318 & 0.4976 & 0.555 & 0.208 & 27.976 & 0.799 & 0.255 \\

    \midrule
    Ours & 0.0156 & \textbf{0.0197} & \textbf{0.0177} & \textbf{0.933} & \textbf{0.960} & \textbf{36.503} & \textbf{0.966} & \textbf{0.048} \\

    \bottomrule
  \end{tabular}
  }

\end{table}

\textbf{Evaluation on Replica dataset.}
As shown in \cref{tab:tab2}, our method also demonstrates the superiority on the Replica dataset. Specifically, in terms of rendering metrics, our method achieves a 5 dB higher rendering quality than methods like \textit{Instant-NGP} and \textit{DVGO}. Compared to the volume rendering-based reconstruction methods, our method achieves better view rendering performance. 
Furthermore, our method outperforms reconstruction-focused methods in these metrics. These further show the robustness of the proposed method on synthetic datasets. The qualitative results can be found in the Supplementary.

\begin{table}[ht!]
  \caption{Reconstruction and view synthesis results of Replica dataset. The best performances are highlighted in bold.}
  \label{tab:tab2}
  \centering
  \Huge
  \resizebox{0.5\textwidth}{!}{
  \begin{tabular}{@{}lcccccccc@{}}
    \toprule
    \textbf{Method} & \textbf{Acc $\downarrow$} & \textbf{Com $\downarrow$} & \textbf{C-$l_1 \downarrow$} & \textbf{NC $\uparrow$} & \textbf{F-score $\uparrow$} & \textbf{PSNR $\uparrow$} & \textbf{SSIM $\uparrow$} & \textbf{LPIPS $\downarrow$} \\
    
    \midrule
    BundleFusion & 0.0145 & 0.0453 & 0.0299 & 0.961 & 0.936  & - & - & -\\

    \midrule
    Neus & 0.1623 & 0.2956 & 0.2288 & 0.754 & 0.194 & 28.939 & 0.855 & 0.181 \\
    VolSDF & 0.1348 & 0.3009 & 0.2180 & 0.747 & 0.339 & 30.375 & 0.866 & 0.175 \\
    Neuralangelo & 0.4747 & 0.6288 & 0.5519 & 0.649 & 0.112 & 30.213 & 0.868 & 0.178 \\
    Neural RGB-D & \textbf{0.0096} & 0.2447 & 0.1271 & 0.934 & 0.847 & 32.668 & 0.893 & 0.198 \\
    Go-Surf & 0.0120 & 0.0122 & 0.0121 & 0.9718 & 0.9896 & 30.967 & 0.884 & 0.217 \\
    
    \midrule
    DVGO & 0.2399 & 0.3511 & 0.2955 & 0.6040 & 0.239 & 31.962 & 0.893 & 0.223 \\
    Instant-NGP & 0.3332 & 1.1151 & 0.7288 & 0.5470 & 0.1583 & 32.352 & 0.884 & 0.150 \\

    \midrule
    Ours & 0.0112 & \textbf{0.0111} & \textbf{0.0112} & \textbf{0.9748} & \textbf{0.9911} & \textbf{37.104} & \textbf{0.955} & \textbf{0.074} \\

    \bottomrule
  \end{tabular}
  }

\end{table}

\begin{figure}[ht!]
	\centering
 
	\begin{subfigure}{\linewidth}
            \rotatebox[origin=c]{90}{\footnotesize{InstantNGP}\hspace{-1.4cm}}
            \begin{minipage}[t]{0.235\linewidth}
                \centering
                \includegraphics[width=1\linewidth]{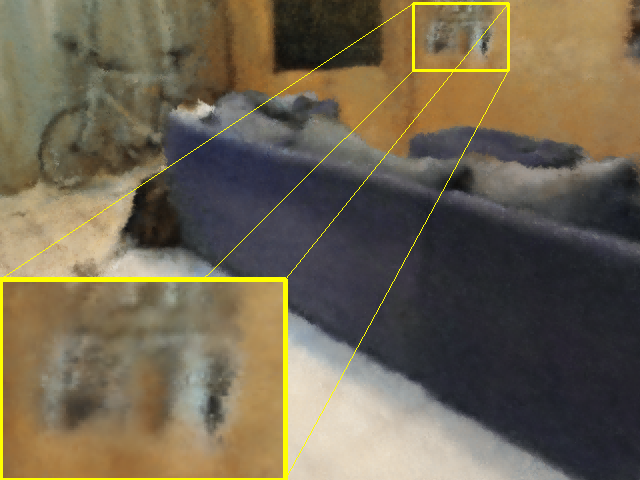}
            \end{minipage}
            \begin{minipage}[t]{0.235\linewidth}
                \centering
                \includegraphics[width=1\linewidth]{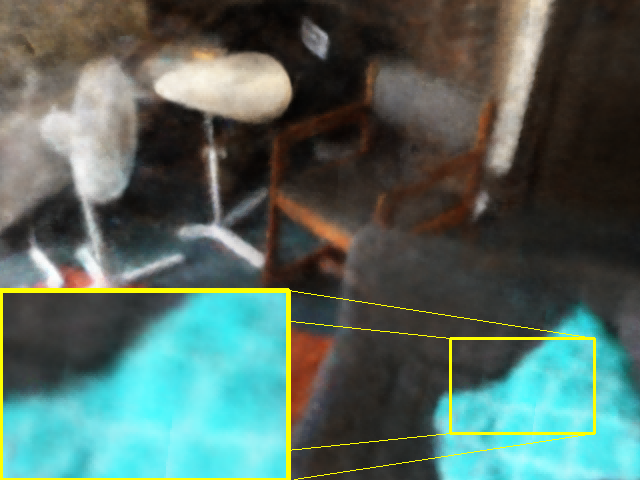}
            \end{minipage}
            \begin{minipage}[t]{0.235\linewidth}
                \centering
                \includegraphics[width=1\linewidth]{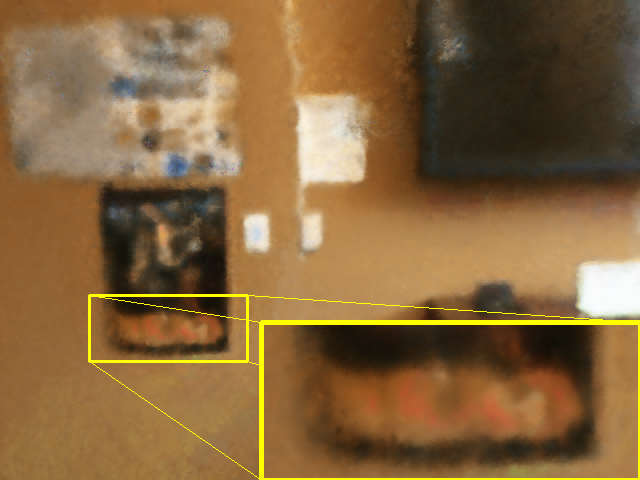}
            \end{minipage}
            \begin{minipage}[t]{0.235\linewidth}
                \centering
                \includegraphics[width=1\linewidth]{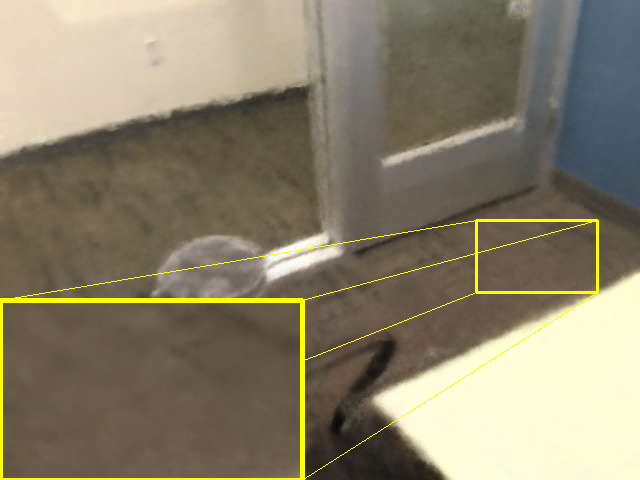}
            \end{minipage}
        \end{subfigure}

	\begin{subfigure}{\linewidth}
            \rotatebox[origin=c]{90}{\footnotesize{DVGO}\hspace{-1.4cm}}
            \begin{minipage}[t]{0.235\linewidth}
                \centering
                \includegraphics[width=1\linewidth]{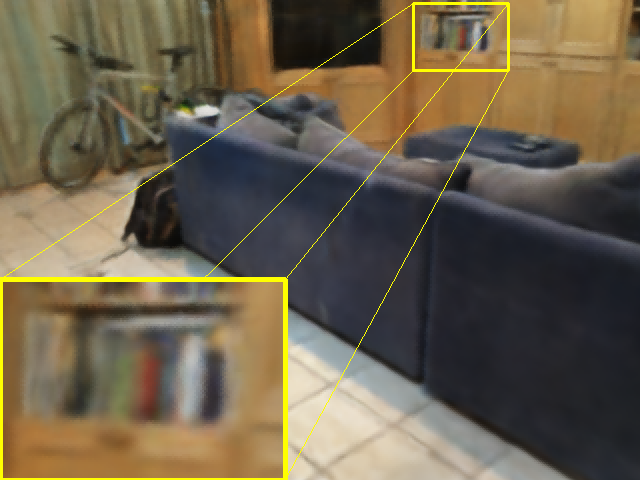}
            \end{minipage}
            \begin{minipage}[t]{0.235\linewidth}
                \centering
                \includegraphics[width=1\linewidth]{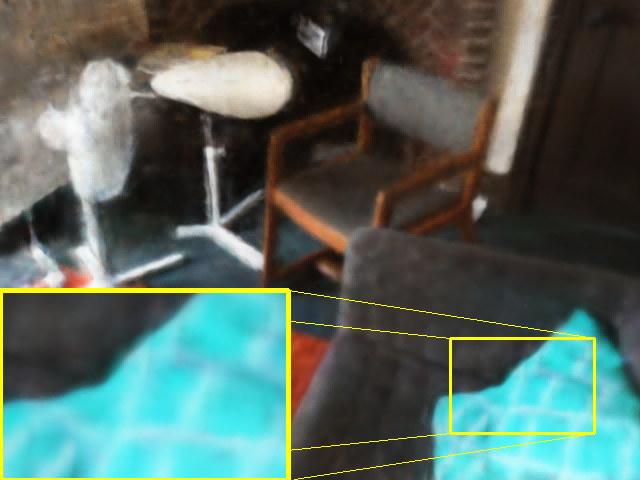}
            \end{minipage}
            \begin{minipage}[t]{0.235\linewidth}
                \centering
                \includegraphics[width=1\linewidth]{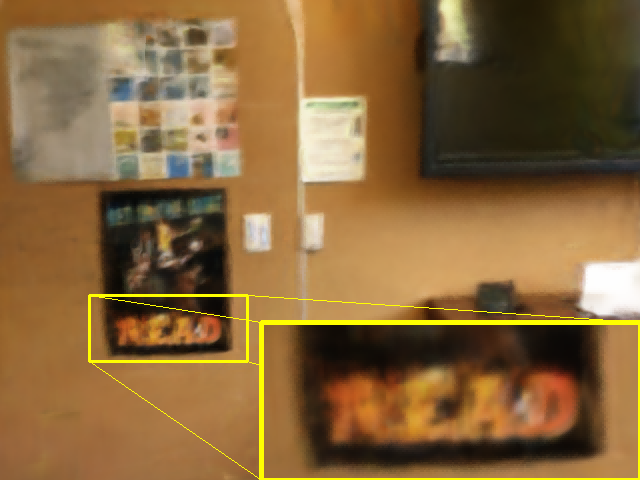}
            \end{minipage}
            \begin{minipage}[t]{0.235\linewidth}
                \centering
                \includegraphics[width=1\linewidth]{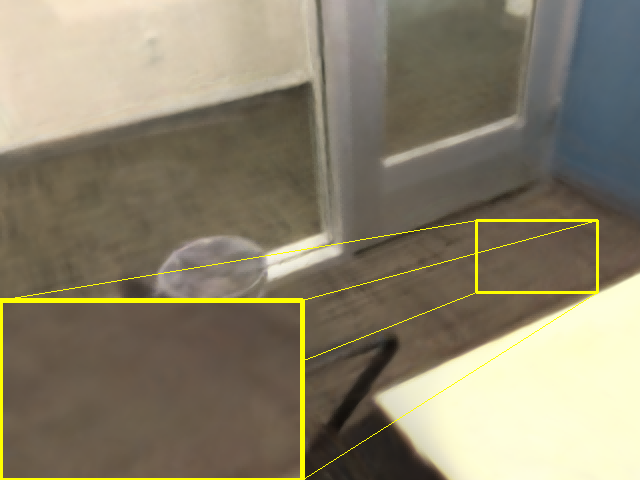}
            \end{minipage}
        \end{subfigure}

	\begin{subfigure}{\linewidth}
            \rotatebox[origin=c]{90}{\footnotesize{Ours}\hspace{-1.4cm}}
            \begin{minipage}[t]{0.235\linewidth}
                \centering
                \includegraphics[width=1\linewidth]{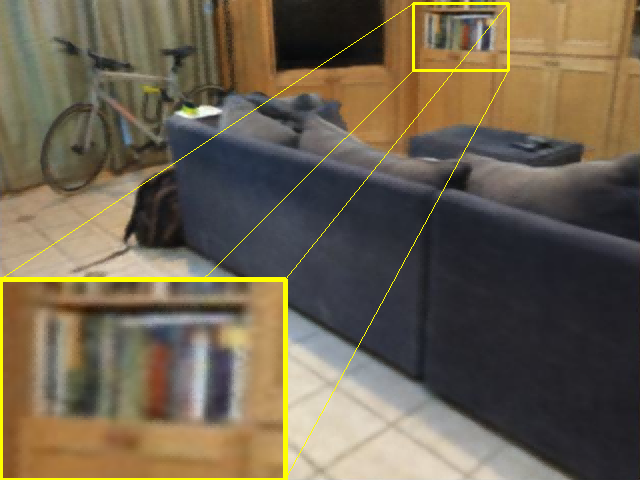}
            \end{minipage}
            \begin{minipage}[t]{0.235\linewidth}
                \centering
                \includegraphics[width=1\linewidth]{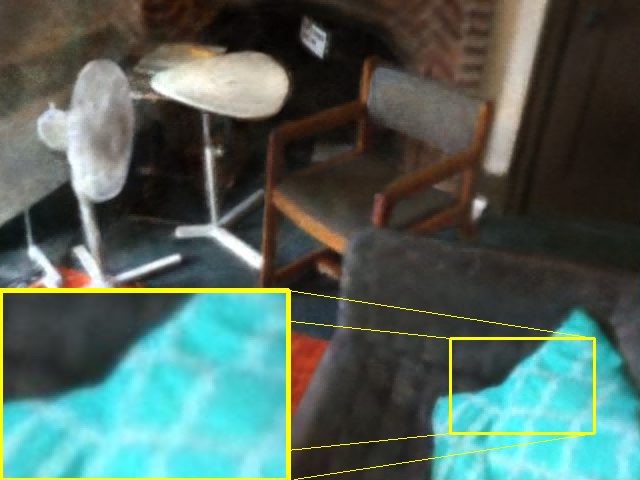}
            \end{minipage}
            \begin{minipage}[t]{0.235\linewidth}
                \centering
                \includegraphics[width=1\linewidth]{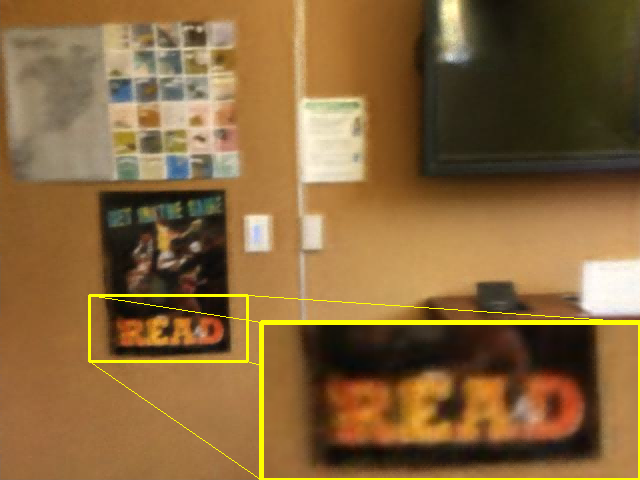}
            \end{minipage}
            \begin{minipage}[t]{0.235\linewidth}
                \centering
                \includegraphics[width=1\linewidth]{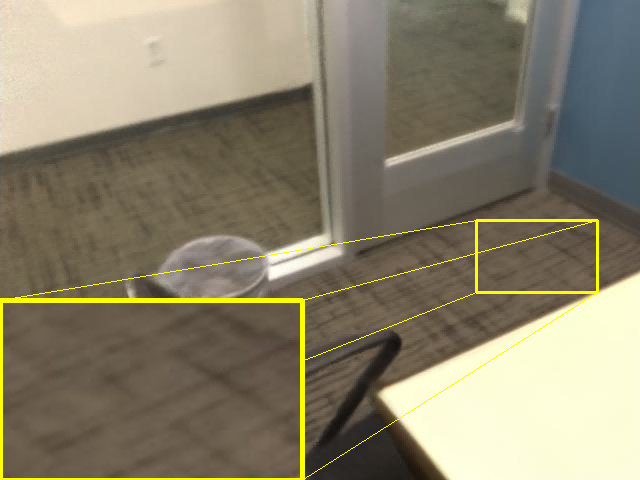}
            \end{minipage}
        \end{subfigure}

	\begin{subfigure}{\linewidth}
            \rotatebox[origin=c]{90}{\footnotesize{GT}\hspace{-1.4cm}}
            \begin{minipage}[t]{0.235\linewidth}
                \centering
                \includegraphics[width=1\linewidth]{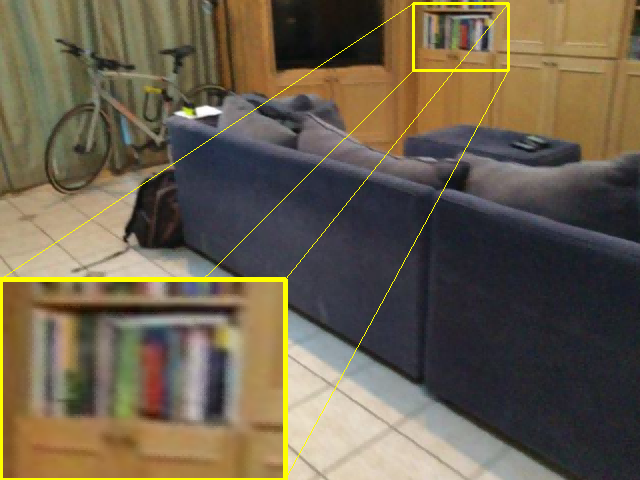}
                % \caption{Whiteroom}
            \end{minipage}
            \begin{minipage}[t]{0.235\linewidth}
                \centering
                \includegraphics[width=1\linewidth]{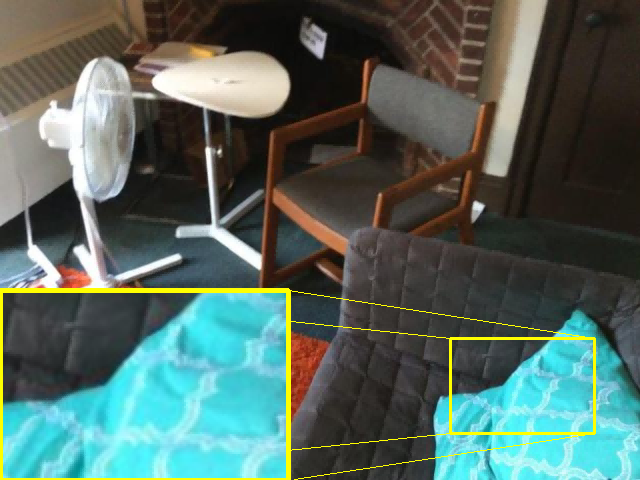}
                % \caption{Office3}
            \end{minipage}
            \begin{minipage}[t]{0.235\linewidth}
                \centering
                \includegraphics[width=1\linewidth]{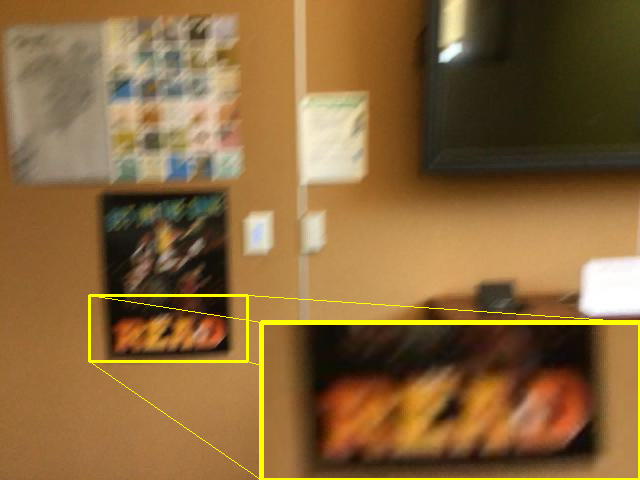}
                % \caption{Room0}
            \end{minipage}
            \begin{minipage}[t]{0.235\linewidth}
                \centering
                \includegraphics[width=1\linewidth]{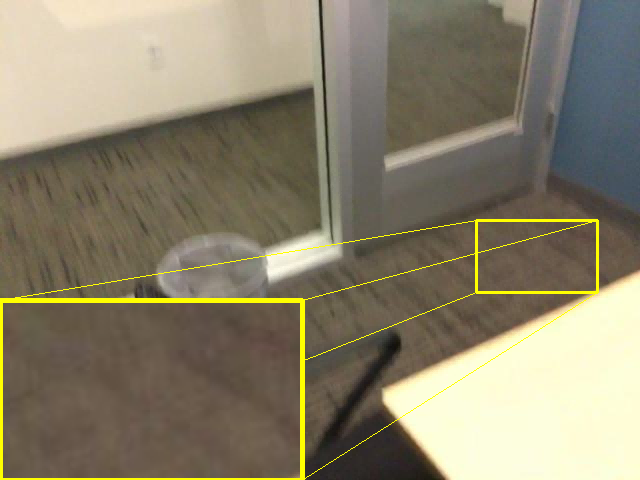}
                % \caption{Room1}
            \end{minipage}
        \end{subfigure}

	\caption{Qualitative results on ScanNet scenes demonstrate the superior rendering quality of our approach compared to previous NeRF-based methods, especially for images exhibiting severe motion blur, such as the text on a poster (column 3) and the stripes on the floor (column 4).}
	\label{fig:scannet_color}
\end{figure}

\begin{figure}[ht!]
	\centering
	\begin{subfigure}{\linewidth}
            \rotatebox[origin=c]{90}{\footnotesize{N.-RGBD}\hspace{-1.1cm}}
            \begin{minipage}[t]{0.235\linewidth}
                \centering
                \includegraphics[width=1\linewidth]{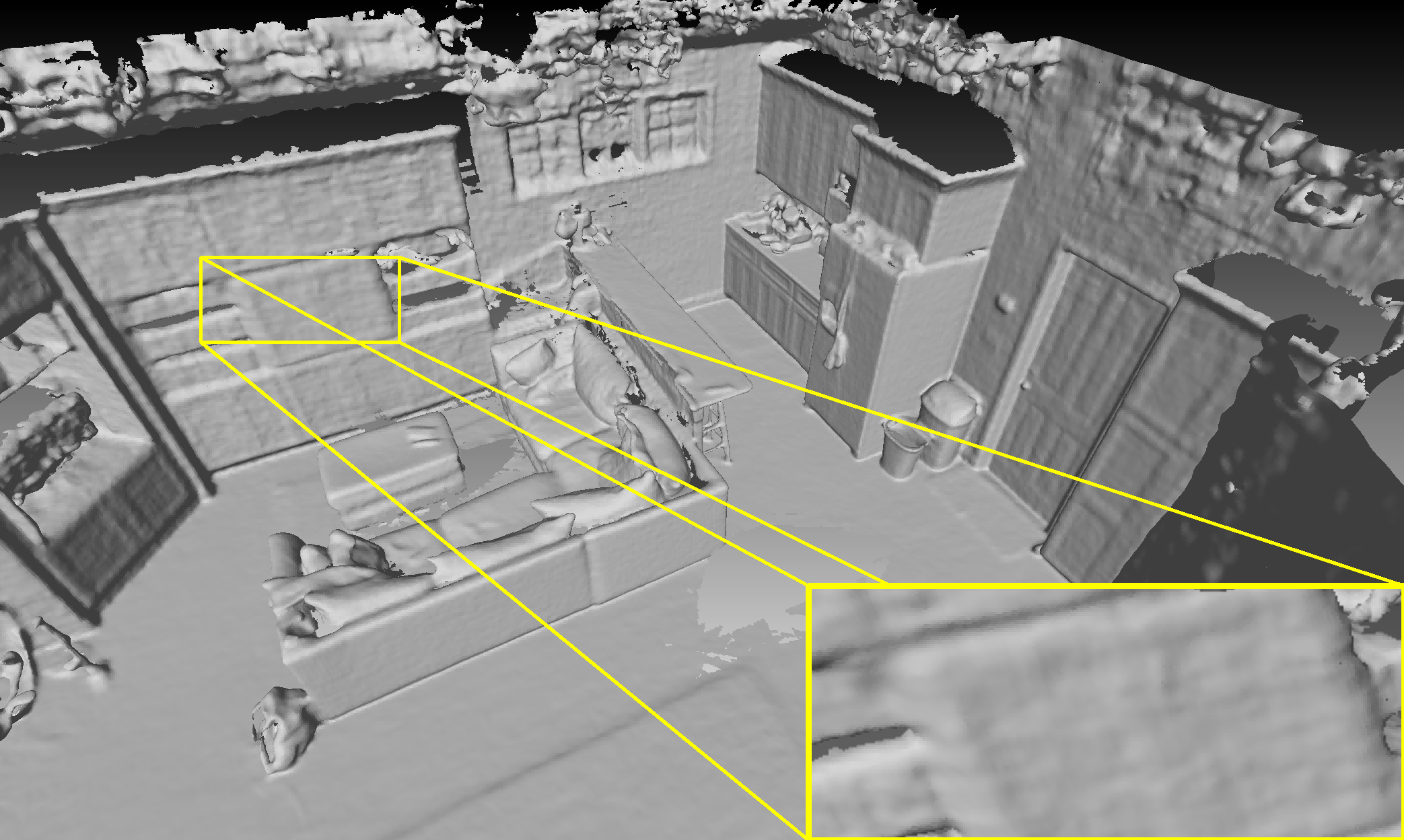}
            \end{minipage}
            \begin{minipage}[t]{0.235\linewidth}
                \centering
                \includegraphics[width=1\linewidth]{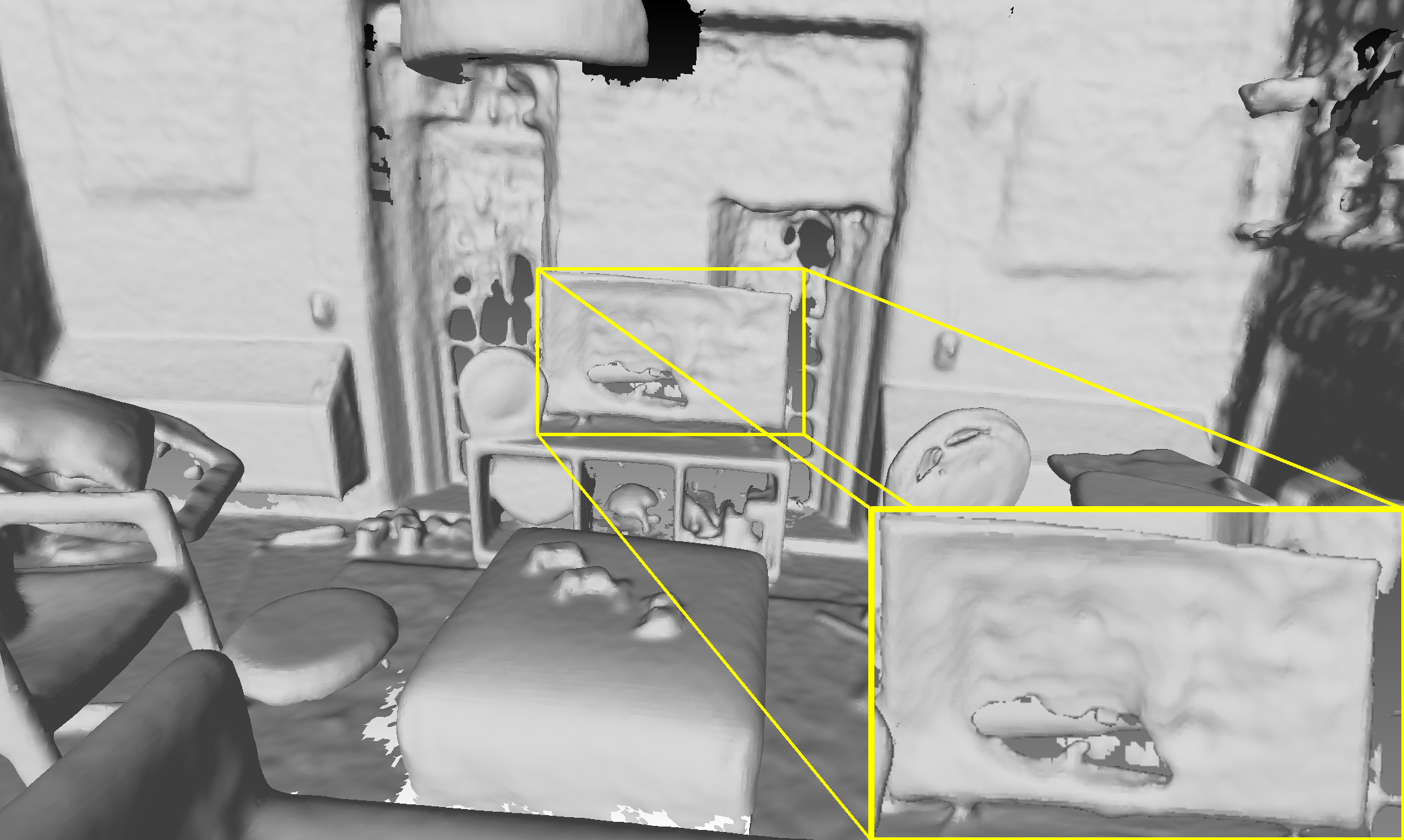}
            \end{minipage}
            \begin{minipage}[t]{0.235\linewidth}
                \centering
                \includegraphics[width=1\linewidth]{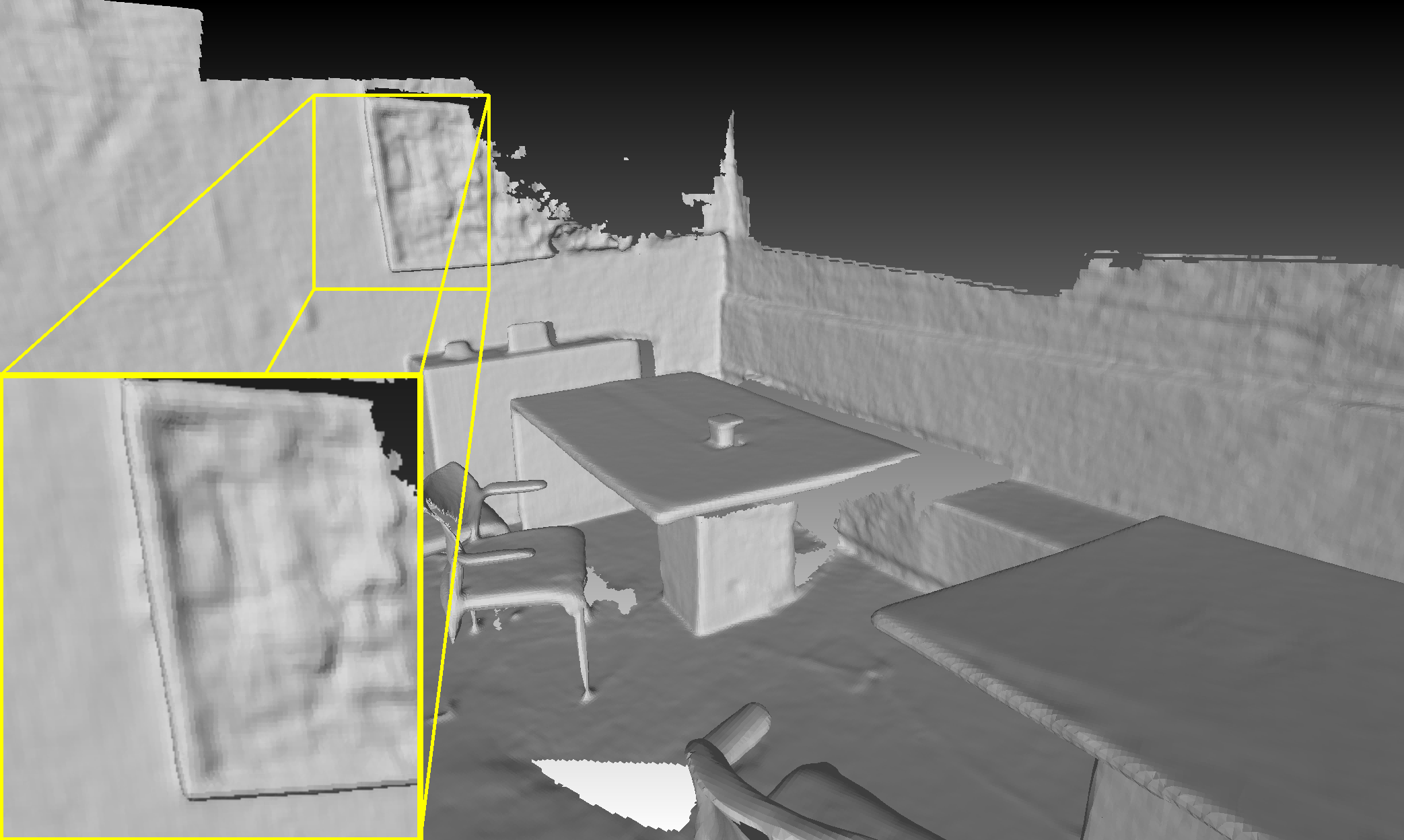}
            \end{minipage}
            \begin{minipage}[t]{0.235\linewidth}
                \centering
                \includegraphics[width=1\linewidth]{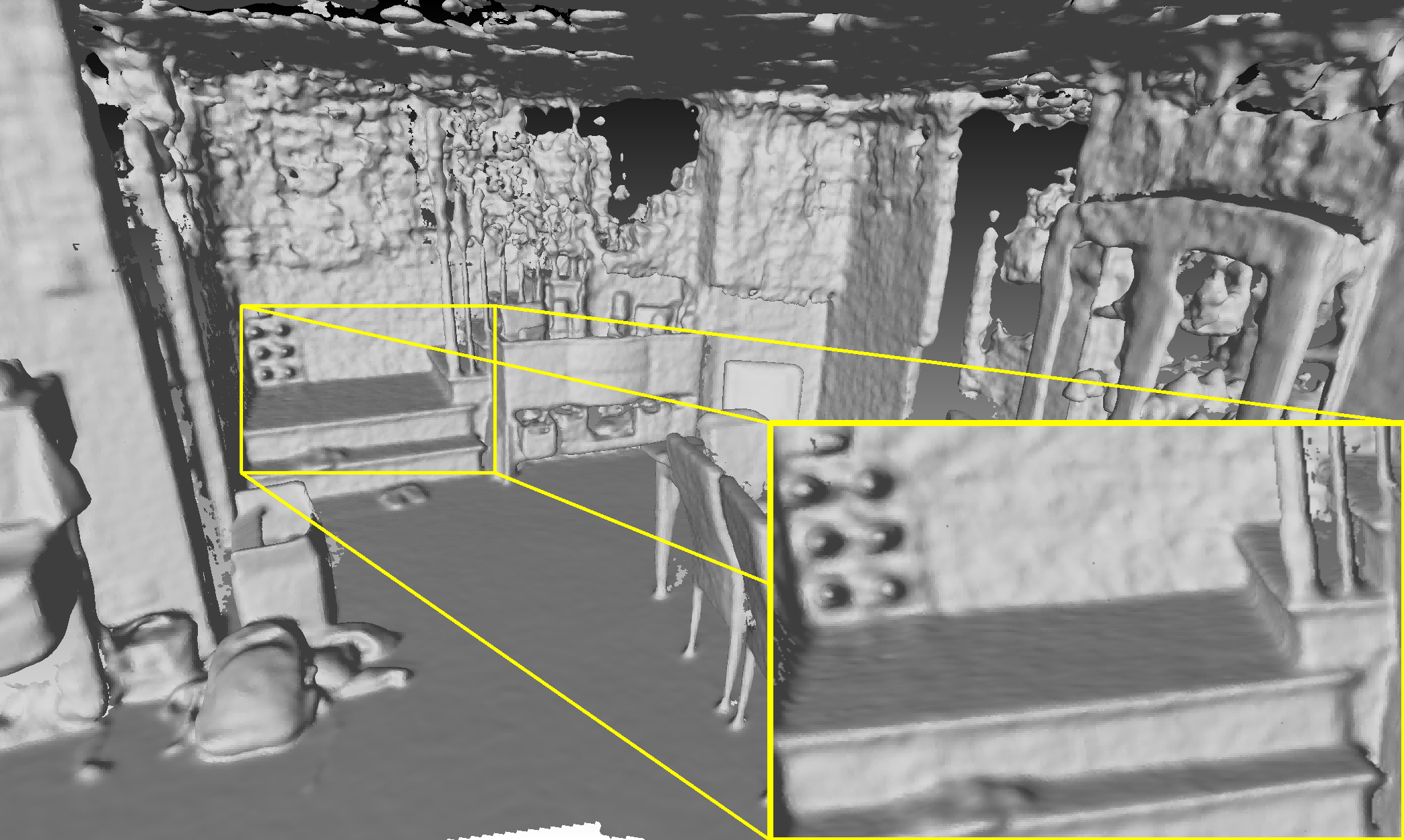}
            \end{minipage}
        \end{subfigure}

	\begin{subfigure}{\linewidth}
            \rotatebox[origin=c]{90}{\footnotesize{Go-Surf}\hspace{-1.1cm}}
            \begin{minipage}[t]{0.235\linewidth}
                \centering
                \includegraphics[width=1\linewidth]{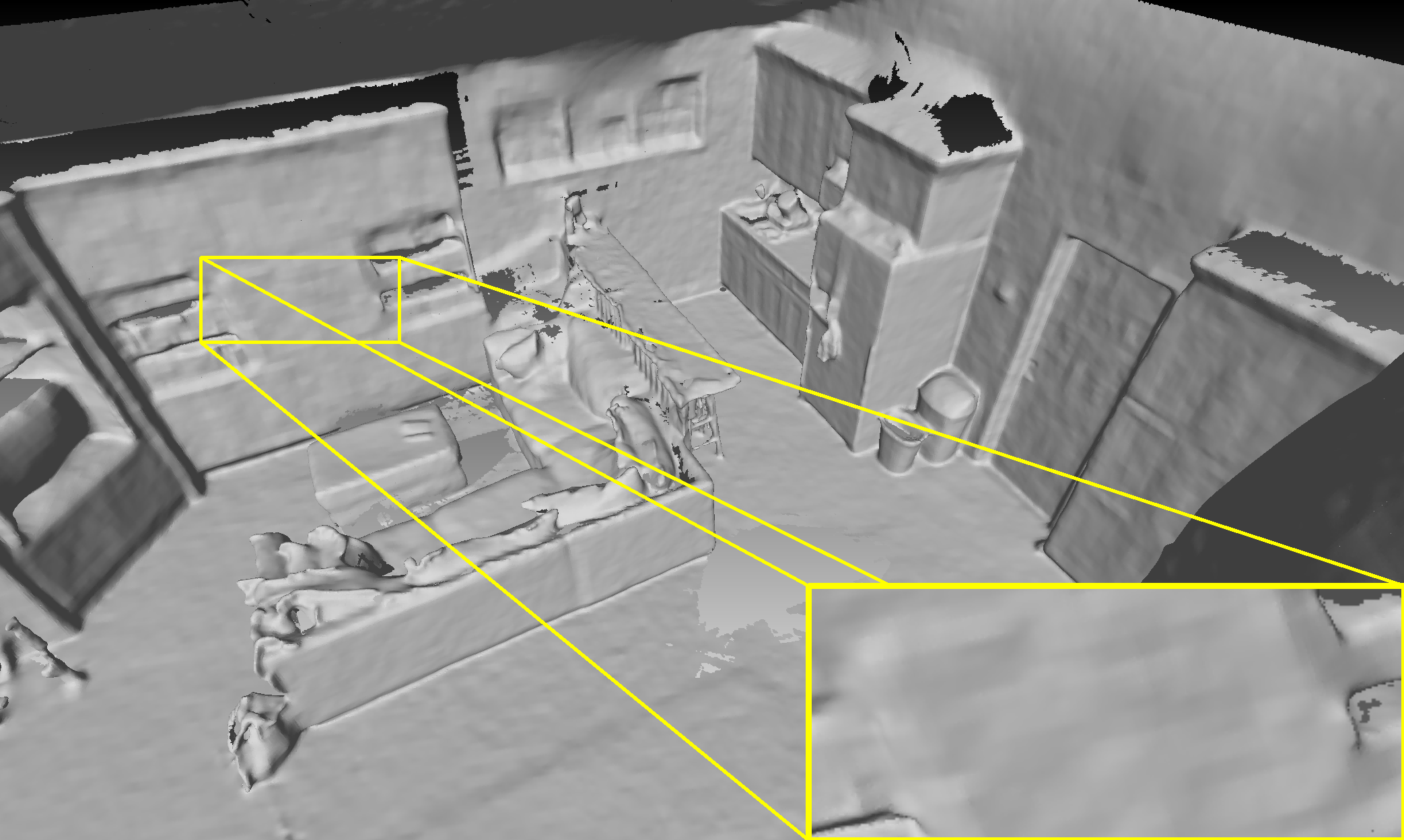}
            \end{minipage}
            \begin{minipage}[t]{0.235\linewidth}
                \centering
                \includegraphics[width=1\linewidth]{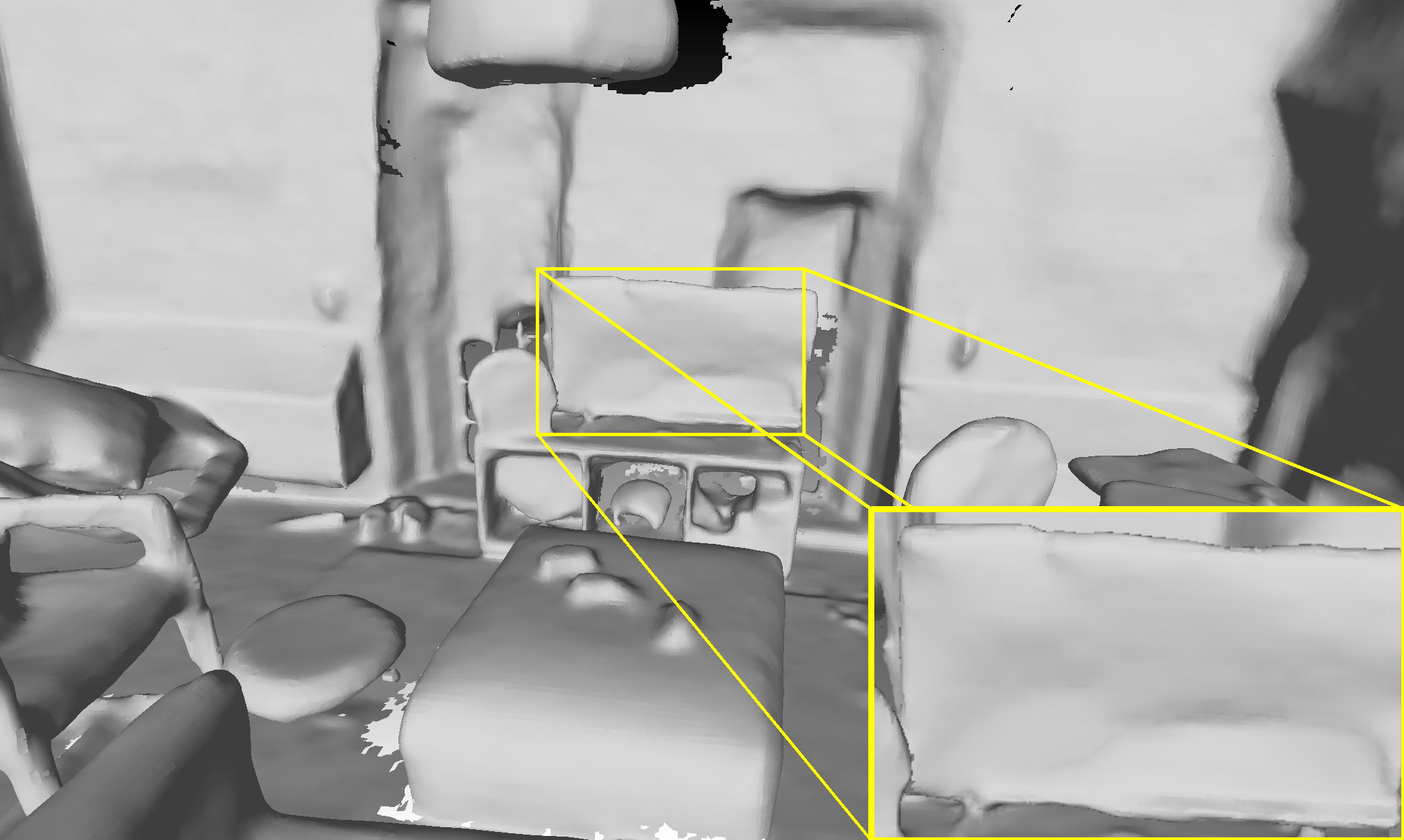}
            \end{minipage}
            \begin{minipage}[t]{0.235\linewidth}
                \centering
                \includegraphics[width=1\linewidth]{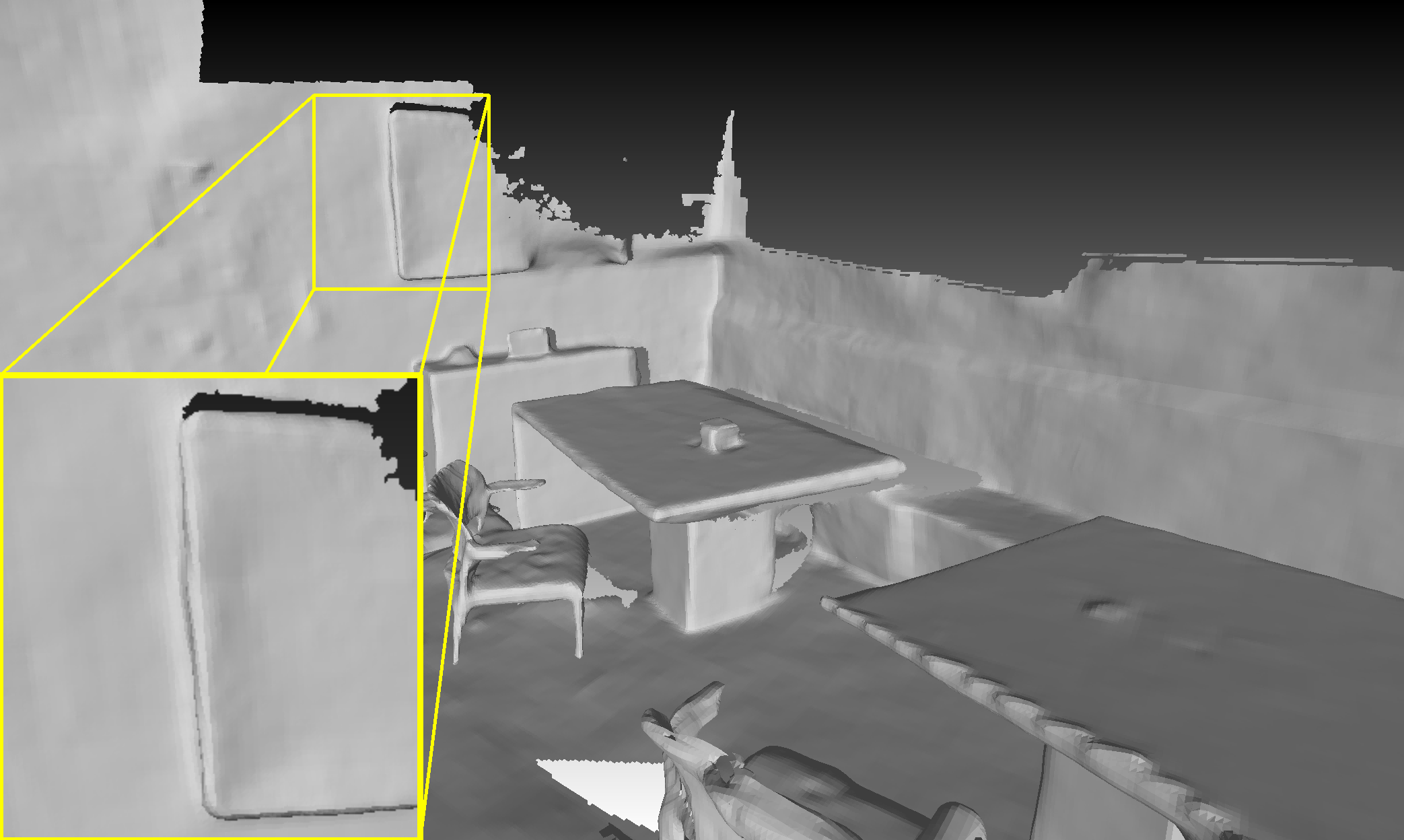}
            \end{minipage}
            \begin{minipage}[t]{0.235\linewidth}
                \centering
                \includegraphics[width=1\linewidth]{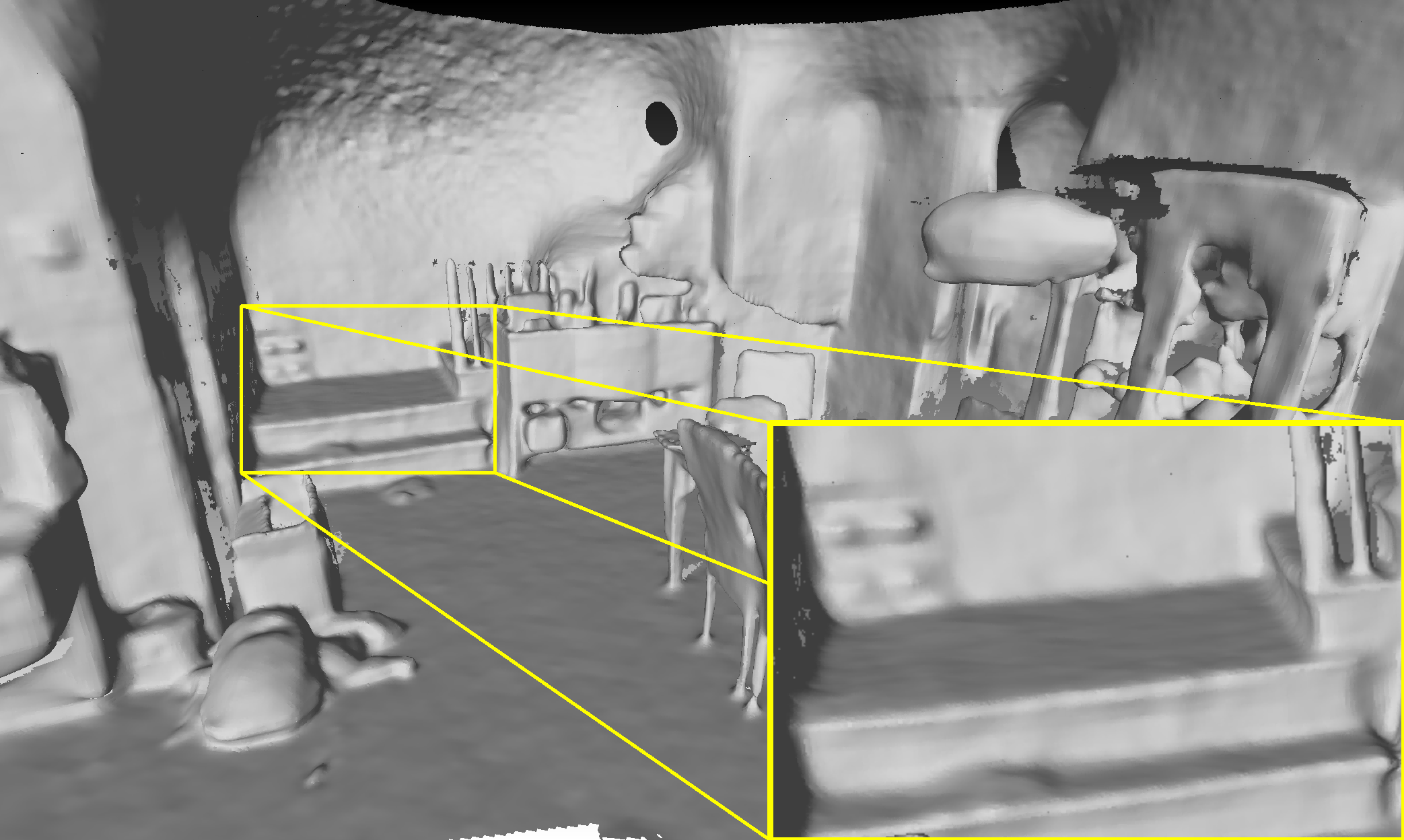}
            \end{minipage}
        \end{subfigure}

	\begin{subfigure}{\linewidth}
            \rotatebox[origin=c]{90}{\footnotesize{Ours}\hspace{-1.1cm}}
            \begin{minipage}[t]{0.235\linewidth}
                \centering
                \includegraphics[width=1\linewidth]{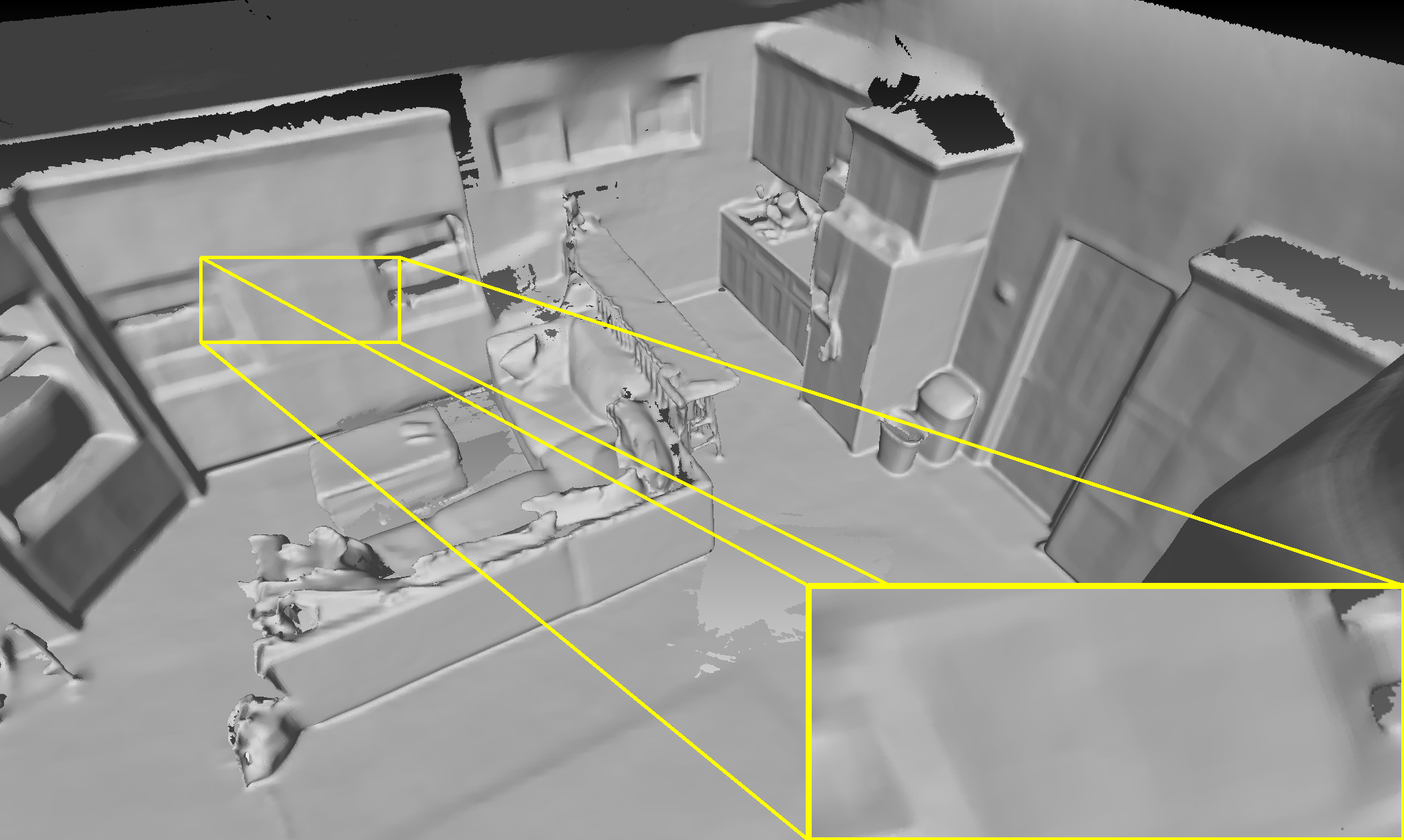}
            \end{minipage}
            \begin{minipage}[t]{0.235\linewidth}
                \centering
                \includegraphics[width=1\linewidth]{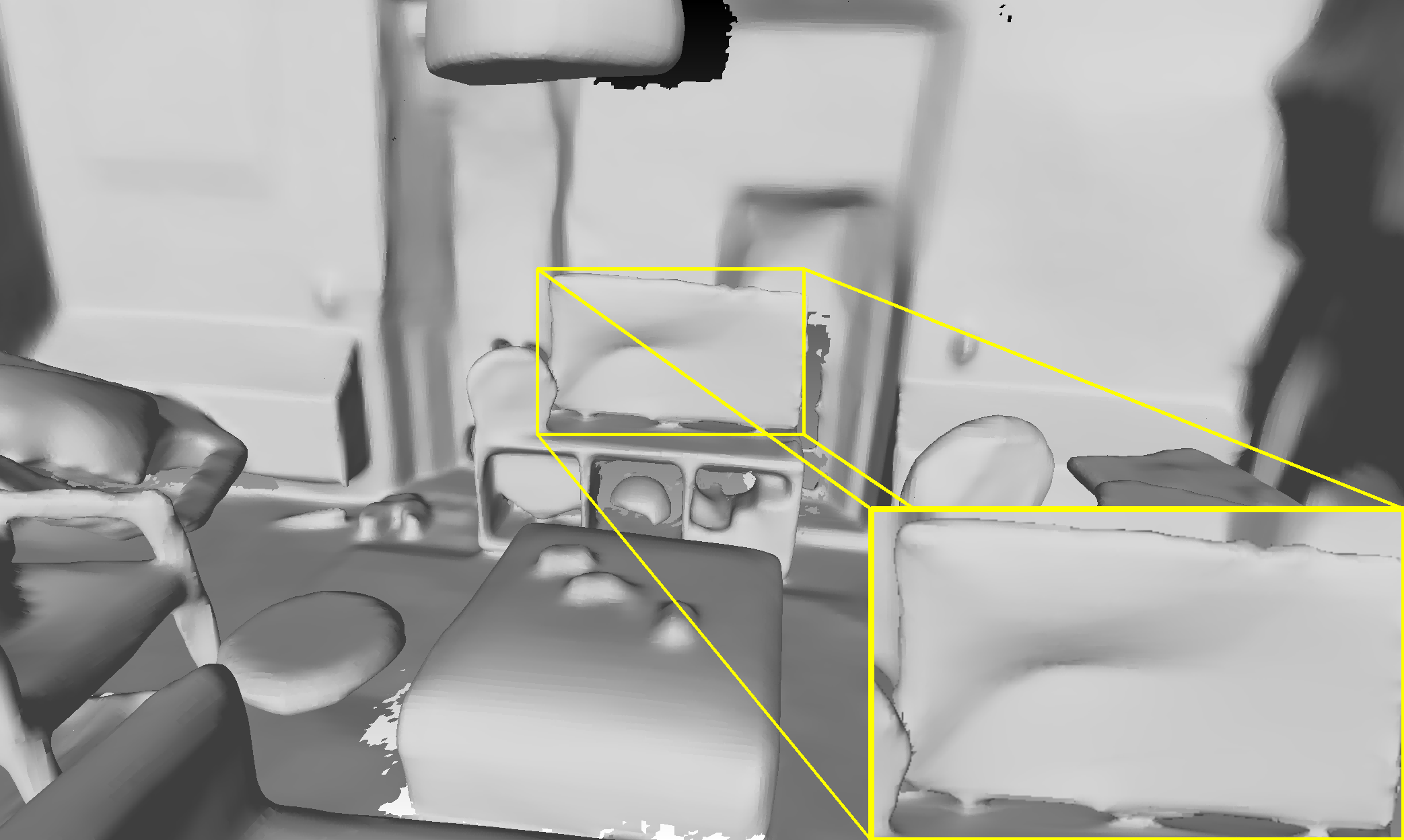}
            \end{minipage}
            \begin{minipage}[t]{0.235\linewidth}
                \centering
                \includegraphics[width=1\linewidth]{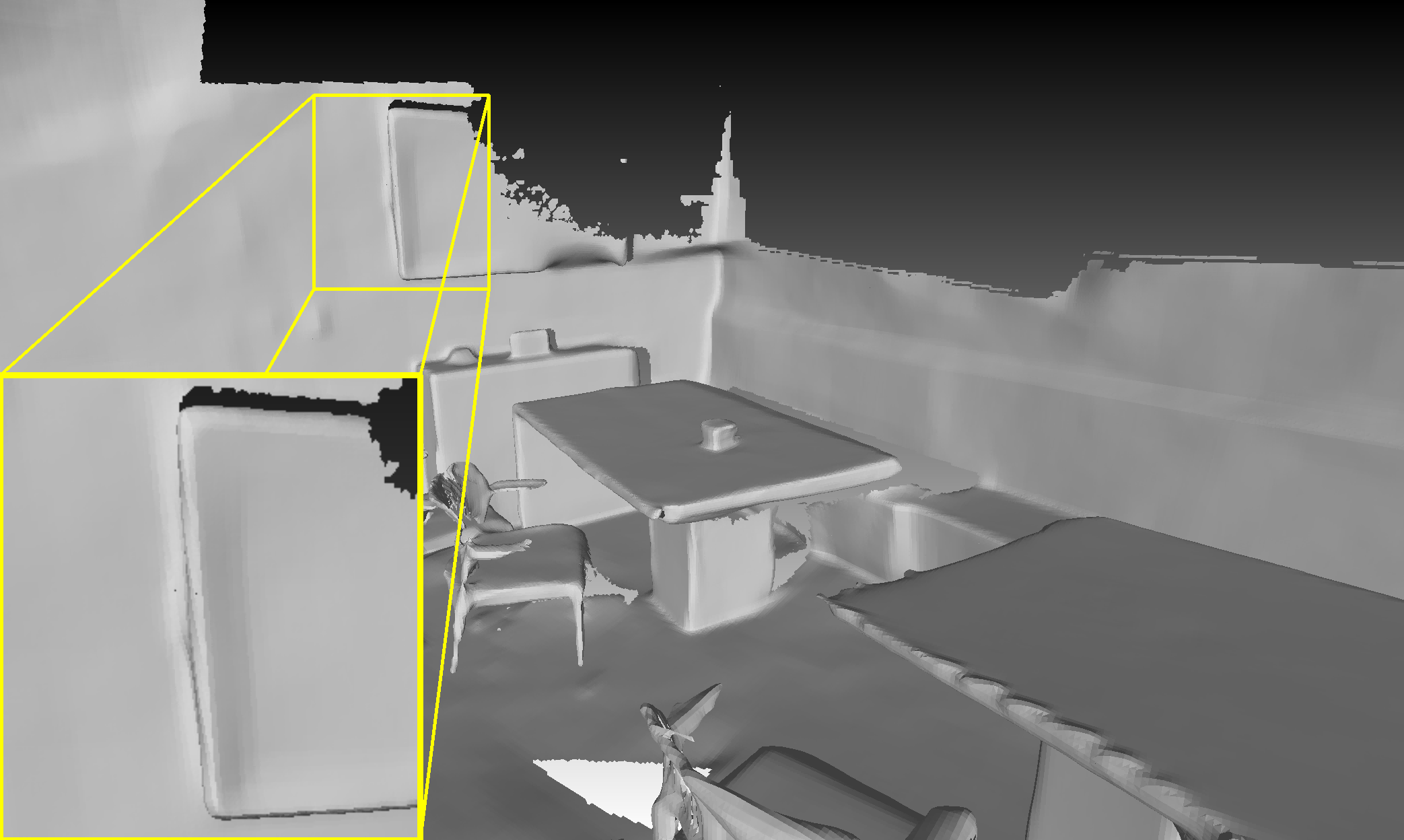}
            \end{minipage}
            \begin{minipage}[t]{0.235\linewidth}
                \centering
                \includegraphics[width=1\linewidth]{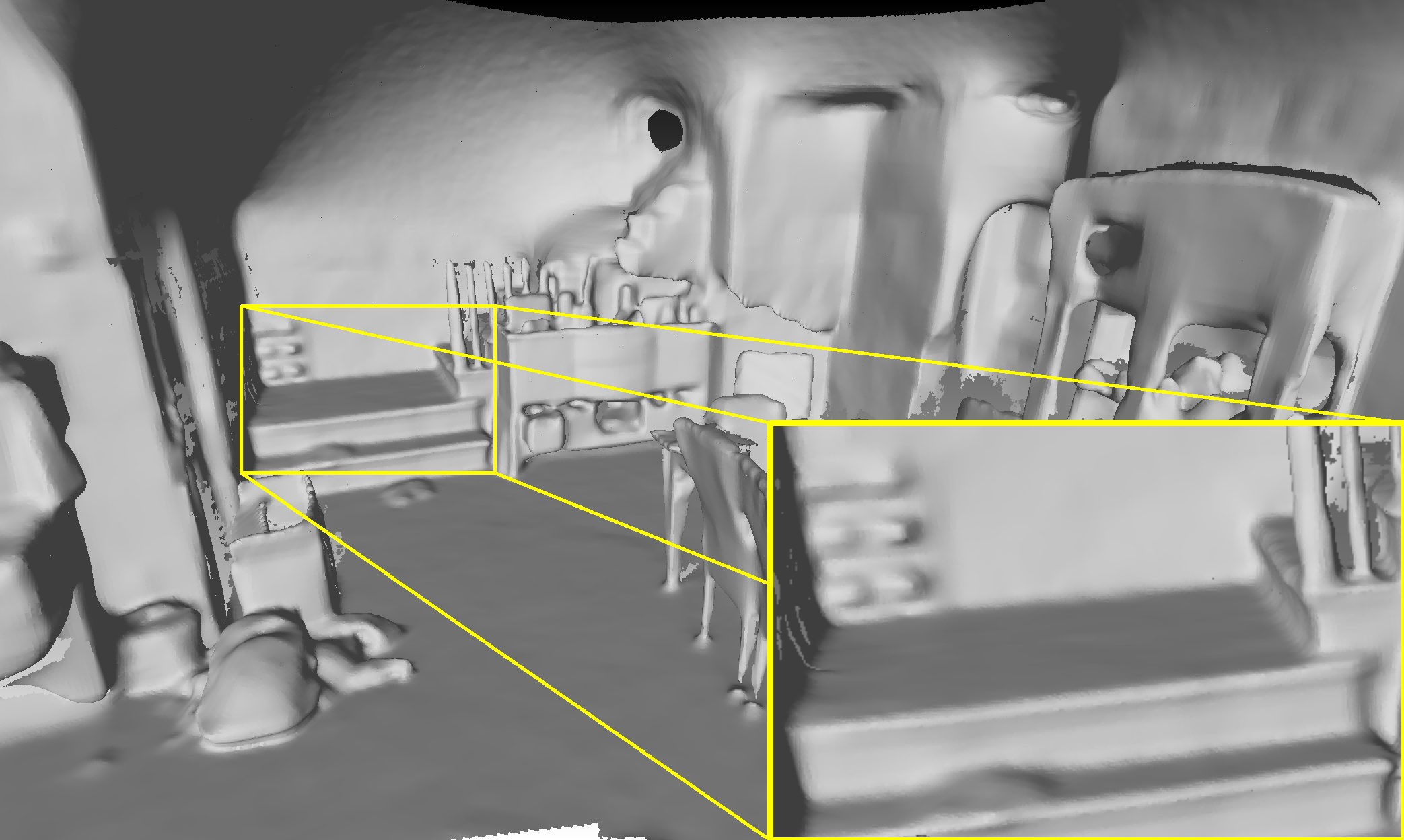}
            \end{minipage}
        \end{subfigure}

	\begin{subfigure}{\linewidth}
            \rotatebox[origin=c]{90}{\footnotesize{GT}\hspace{-1.1cm}}
            \begin{minipage}[t]{0.235\linewidth}
                \centering
                \includegraphics[width=1\linewidth]{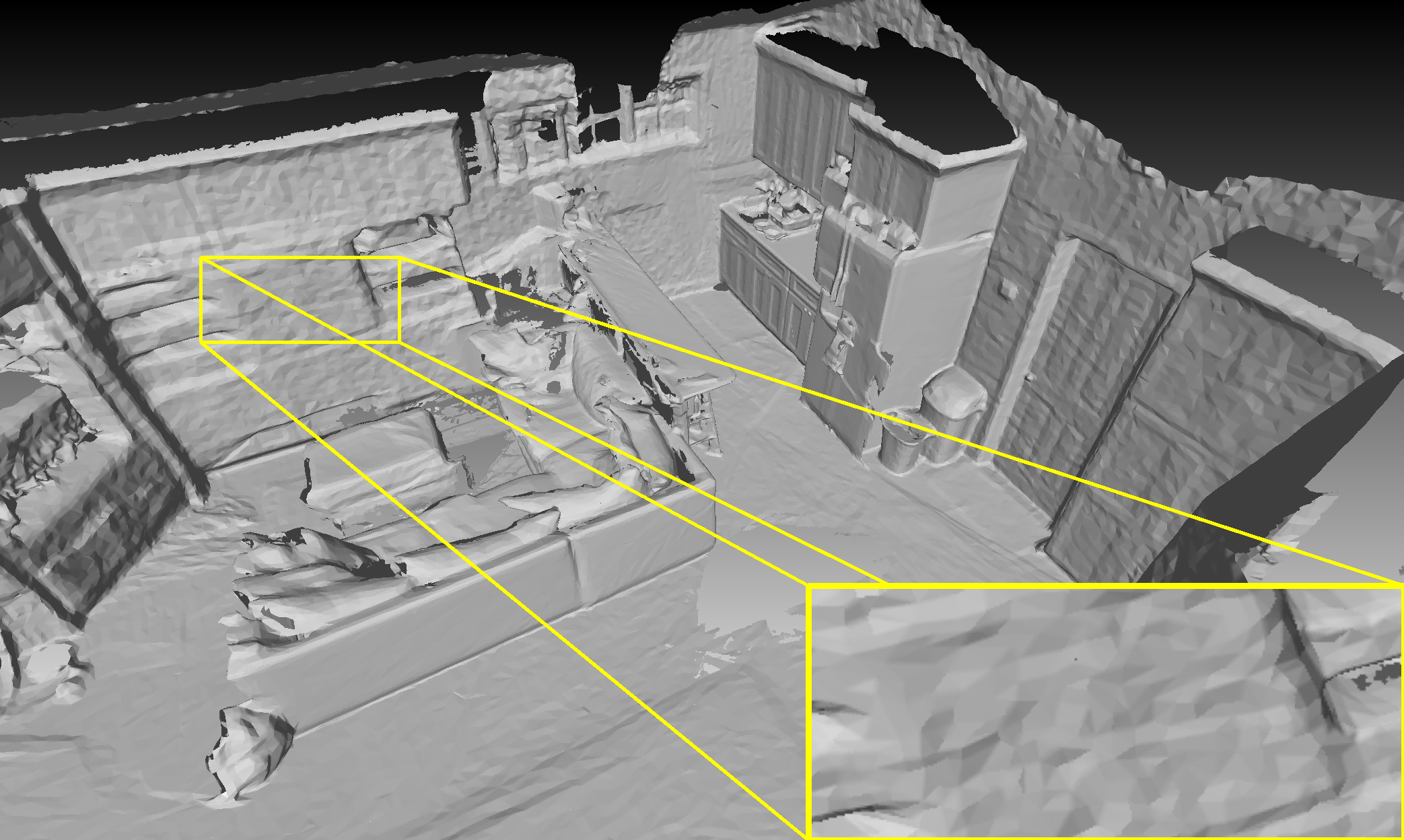}
            \end{minipage}
            \begin{minipage}[t]{0.235\linewidth}
                \centering
                \includegraphics[width=1\linewidth]{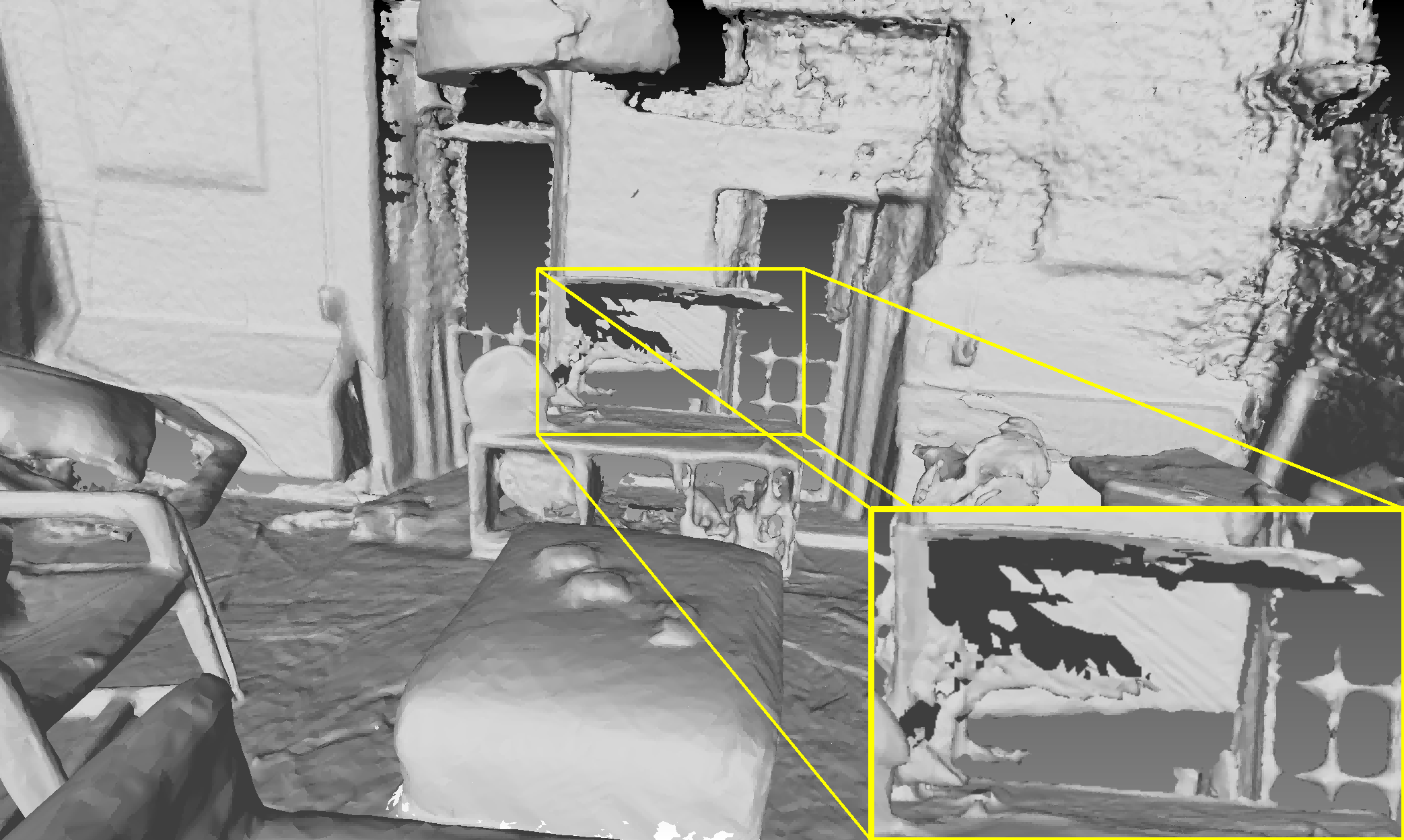}
            \end{minipage}
            \begin{minipage}[t]{0.235\linewidth}
                \centering
                \includegraphics[width=1\linewidth]{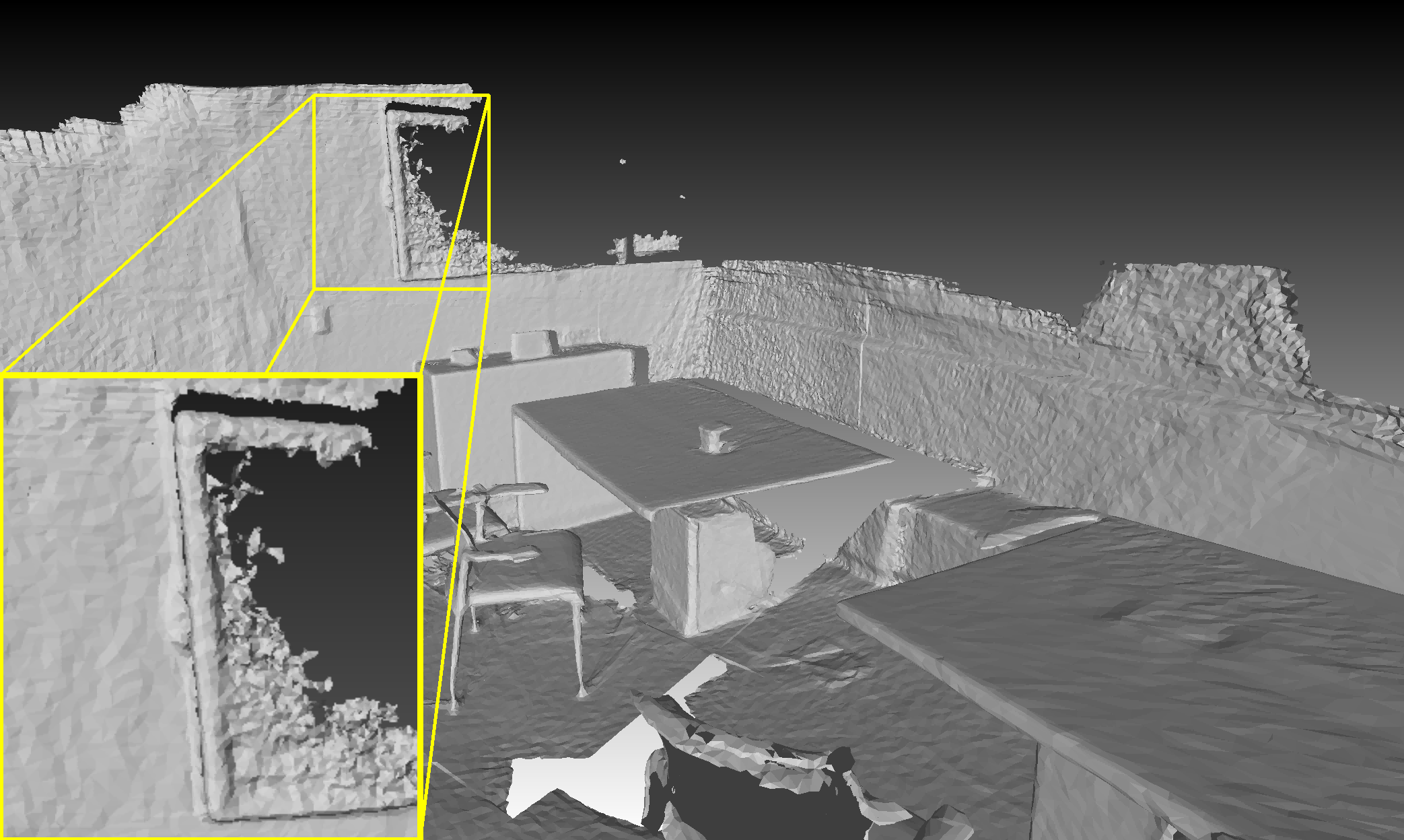}
            \end{minipage}
            \begin{minipage}[t]{0.235\linewidth}
                \centering
                \includegraphics[width=1\linewidth]{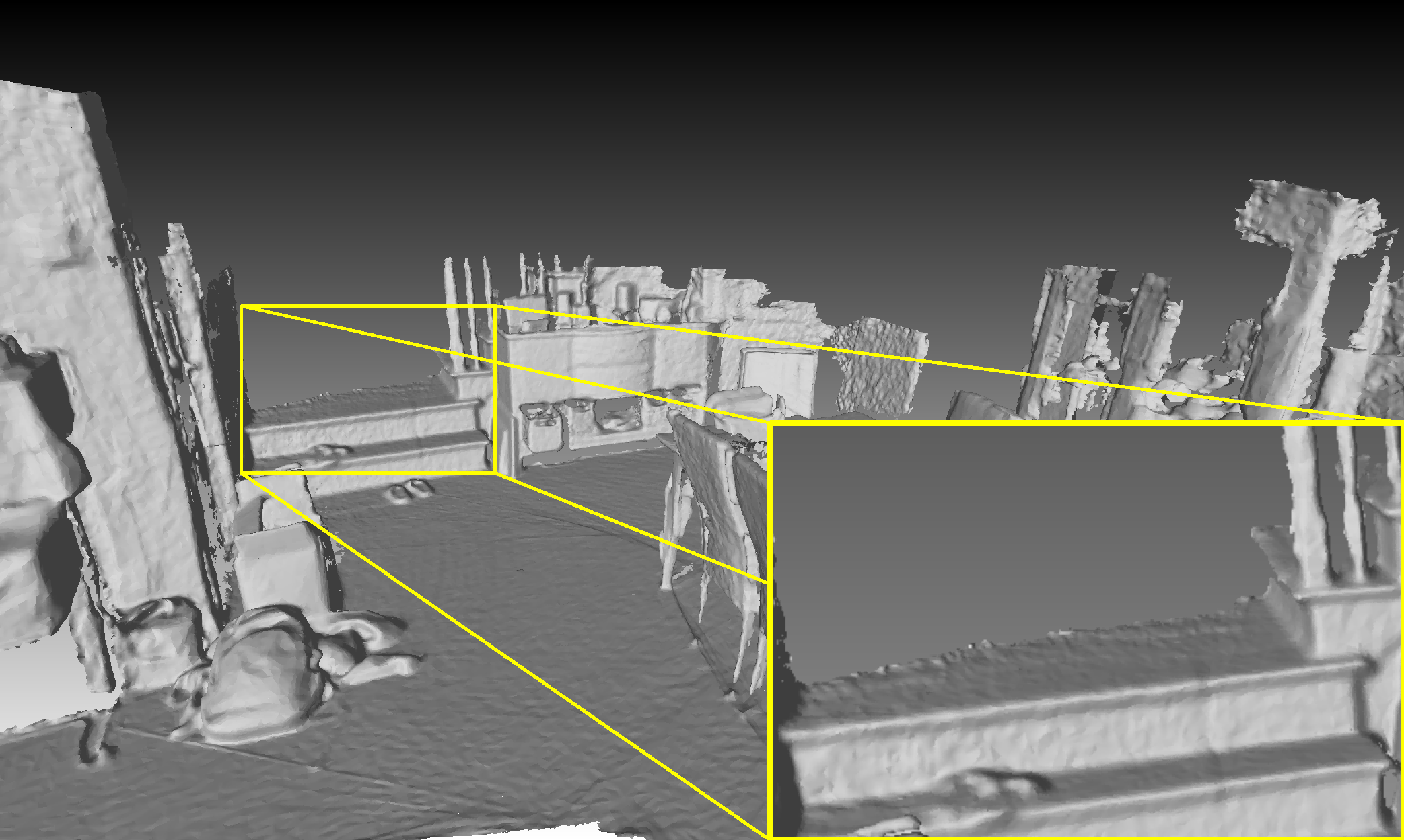}
            \end{minipage}
        \end{subfigure}

	\caption{Qualitative mesh reconstruction on Scannet scenes. Our method produces visually smoother and cleaner meshes compared to previous methods, as demonstrated by the zoomed-in details provided for comparison.}
	\label{fig:scannet_mesh}
\end{figure}

\textbf{Evaluation on Scannet Dataset.} We present the quantitative results (\cref{tab:scannet comparison}) and the qualitative results (\cref{fig:scannet_color} and \cref{fig:scannet_mesh}) evaluated on five real-world scenes. It can be seen from \cref{tab:scannet comparison} that our method achieves the best metrics of PSNR and Chamfer-$l_1$.

For the 3D reconstruction task, our approach yields smoother surface and more complete objects, \textit{e.g.} whiteboard and TV (\cref{fig:scannet_mesh}). Remarkably, for the view synthesis task, our method successfully generates clear  floor stripes and poster text, which appear blurred in other results (\cref{fig:scannet_color}).

% For the 3D reconstruction task, the performance of our model is comparable with the volume-rendering-based methods, achieving the lowest metrics of Chamfer-$l_1$. Compared to the \textit{DVGO} and \textit{InstantNGP}, our method has an exponential superiority over them in term of the Chamfer-$l_1$. As can be seen in \cref{fig:scannet_mesh}, the mesh derived from our method performs smoother and more complete than other methods, such as the whiteboard in the third column and the TV in the second column.
% For the view synthesis task, our method consistently demonstrates superior performance across all scenes. Specifically, our method achieves an average gain of $1.5-5.5$ dB compared to the representative nerf-based methods, as shown in \cref{tab:scannet comparison}. Furthermore, the \cref{fig:scannet_color} shows that our method clearly generates the stripe on the ground and the words in the poster, while \textit{DVGO} and \textit{Go-Surf} both show a blurred rendering.
% \begin{table}[ht!]
%   \centering
%   \Huge
%   \resizebox{0.5\textwidth}{!}{

\begin{table}[ht!]
\renewcommand{\arraystretch}{1.4}
\caption{
Performance comparison with other nerf-based methods on rendering and reconstruction results of the Scannet dataset. Best results are highlighted as \textbf{bold}. 
}
\label{tab:scannet comparison}
\centering
% \begin{adjustbox} {width=1.05\linewidth}
% \Huge
\Huge
\resizebox{0.5\textwidth}{!}{
\begin{tabular}{ l | c c | c c | c c | c c | c c | c c }
\Xhline{4\arrayrulewidth}
\multirow{2}{*}{Method}  
& \multicolumn{2}{c|}{scene0000}  & \multicolumn{2}{c|}{scene0002} & \multicolumn{2}{c|}{scene0005} & \multicolumn{2}{c|}{scene0024}
& \multicolumn{2}{c|}{scene0494} & \multicolumn{2}{c}{Average} \\
& C-$l_1 \downarrow$ & PSNR$ \uparrow$ & C-$l_1 \downarrow$ & PSNR$ \uparrow$ & C-$l_1 \downarrow$ & PSNR$ \uparrow$ & C-$l_1 \downarrow$ & PSNR$ \uparrow$ & C-$l_1 \downarrow$ & PSNR$ \uparrow$ & C-$l_1 \downarrow$ & PSNR$ \uparrow$ \\ 
\hline\hline
Neus  & 0.410   & 24.056   & 0.364   & 21.397   & 0.444   & 25.985   & 0.568   & 21.888   & 0.543   & 28.770 & 0.466   & 24.419 \\
VolSDF  & 0.652   & 24.849   & 0.606   & 21.284   & 0.573   & 25.456   & 0.793   & 22.560   & 0.541   &  27.831  & 0.633   & 24.396 \\
Neuralangelo  & 0.457   & 22.847   & 0.350   & 22.227   & 0.339   & 28.075   & 0.462   & 20.496   & 0.336   & 28.865 & 0.389   & 24.502 \\
Neural RGB-D  & 0.068   & 21.779   & 0.060   & 17.330   & 0.075   & 23.635   & 0.185   & 16.266   & \textbf{0.044}   & 28.274 & 0.086   & 21.457 \\
Go-surf  & 0.030   & 24.033   & 0.024   & 20.246   & 0.054   & 24.977   & \textbf{0.106}   & 21.538   & 0.103   & 27.572 & 0.063   & 23.673 \\

\hline
DVGO  & 0.133   & 26.315   & 0.143   & 21.206   & 0.153   & 28.126   & 0.142   & 22.689   & 0.155   & 29.655 & 0.145   & 25.598 \\
Instant-NGP  & 0.476   & 23.145  &  0.249   & 22.854   & 0.349   & 23.988   & 0.958   & 15.285   & 0.305   & 28.786   & 0.467   & 22.812 \\

\hline
\textbf{Ours} 
& \textbf{0.029} & \textbf{26.712} & \textbf{0.022} & \textbf{23.032} 
& \textbf{0.053} & \textbf{29.146} & \textbf{0.106} & \textbf{24.240}  
& 0.075 & \textbf{31.675} & \textbf{0.058} & \textbf{26.961} \\
\Xhline{4\arrayrulewidth}
\end{tabular}
}
% \end{adjustbox}
\end{table}

    \begin{figure}[t]
    	\centering
            
    	\begin{subfigure}{\linewidth}
                \begin{minipage}[t]{0.32\linewidth}
                    \centering
                    \includegraphics[width=1\linewidth]{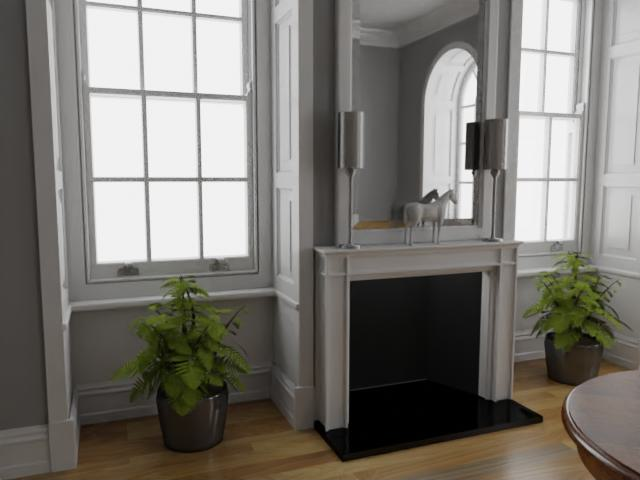}
                \end{minipage}
                \begin{minipage}[t]{0.32\linewidth}
                    \centering
                    \includegraphics[width=1\linewidth]{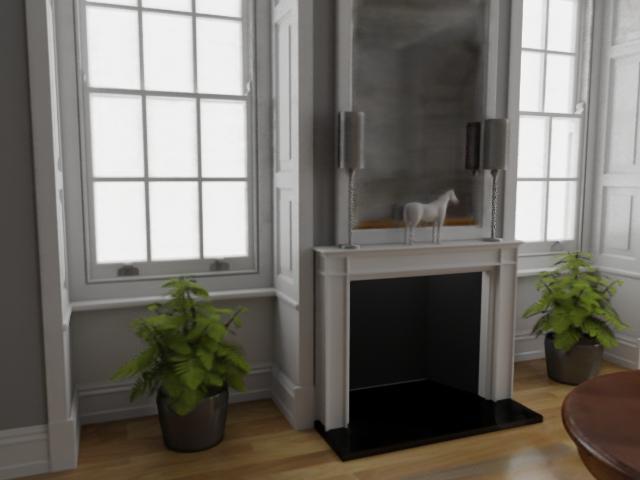}
                \end{minipage}
                \begin{minipage}[t]{0.32\linewidth}
                    \centering
                    \includegraphics[width=1\linewidth]{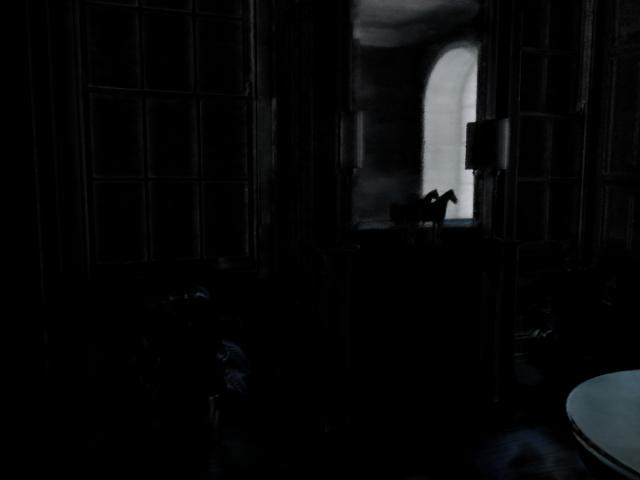}
                \end{minipage}
            \end{subfigure}
    
    	\begin{subfigure}{\linewidth}
                \begin{minipage}[t]{0.32\linewidth}
                    \centering
                    \includegraphics[width=1\linewidth]{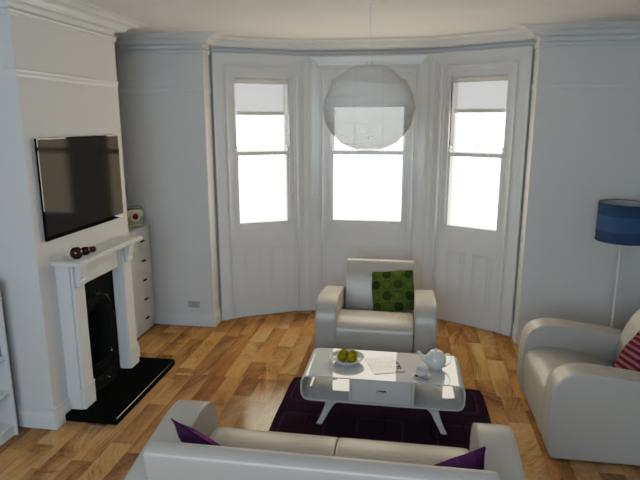}
                    \caption{Complete color}
                \end{minipage}
                \begin{minipage}[t]{0.32\linewidth}
                    \centering
                    \includegraphics[width=1\linewidth]{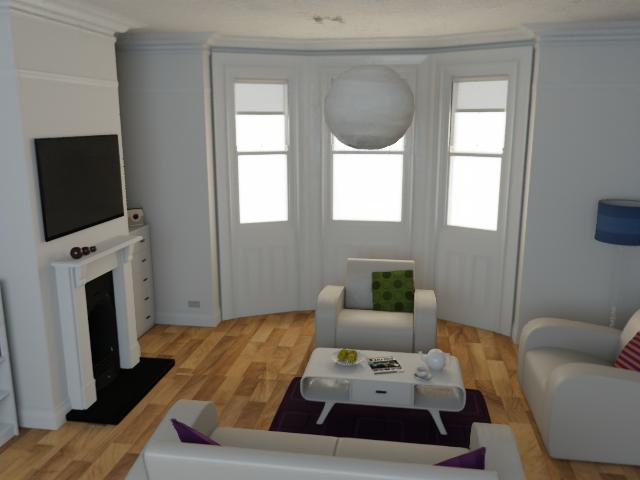}
                    \caption{Vi color}
                \end{minipage}
                \begin{minipage}[t]{0.32\linewidth}
                    \centering
                    \includegraphics[width=1\linewidth]{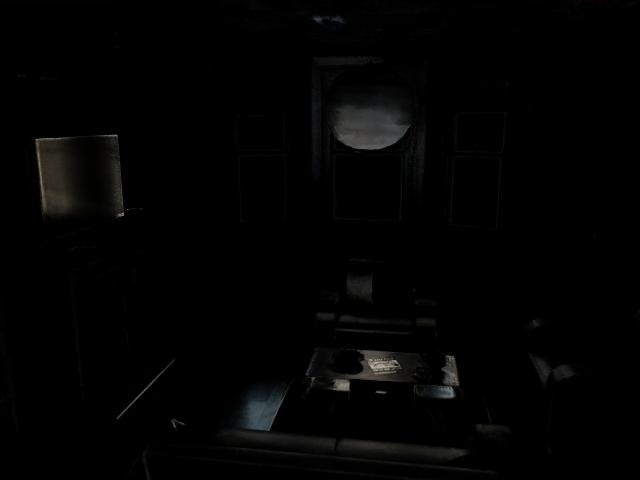}
                    \caption{Vd color}
                \end{minipage}
            \end{subfigure}
    
    	 \caption{Color decoupling results of the proposed method on NeuralRGBD dataset. The Vi color represents the view-independent color, while Vd color is the view-dependent color. It can be seen that our method can effectively decouple the complete color into the view-dependent (specular reflective surfaces) and view-independent (diffuse surfaces) colors.} 
    	\label{fig:fig5}
    \end{figure}

\subsection{Ablations}\label{Ablations}
% We conduct ablation studies to demonstrate the effectiveness of designing the blocks and choice of scene representation. The quantitative result is shown in \cref{tab:tab3,tab:tab4,tab:tab5,tab:tab7}, where we evaluate these methods in the RGBD synthetic dataset.
We conduct ablation studies to demonstrate the effectiveness of designing the blocks and choice of scene representation. The quantitative result is shown in \cref{tab:tab3,tab:tab4,tab:tab5,tab:pose_refine}, where we evaluate these methods in the RGBD synthetic dataset. Other ablation studies can be found in the Supplementary.

\textbf{Effect of the Du-NeRF architecture.}
To analyze the network architecture of our dual-field, we experimented with various architectural configurations, detailed in \cref{tab:tab3}. 
The SDF-only represents that the method contains only the SDF branch and the Density-only involves only the density branch. Dual is the proposed method which has the density branch and  SDF branch. The experimental results show that the two branches achieve optimal rendering and reconstruction performance. The SDF branch exhibits superior reconstruction capabilities, while the Density branch excels in rendering, as evidenced by the fact that the SDF-only branch outperforms the Density-only branch in terms of F-score, and the latter achieves higher PSNR values.

% Results indicate that the SDF-only outperforms the Density-only in F-score, though the latter scores higher in PSNR. These findings corroborate that while the SDF branch excels in reconstruction, the density branch is superior in rendering. Importantly, integrating both branches enhances both view rendering and reconstruction performance.

% To investigate the effectiveness of our dual-field, we conduct experiments with different architecture settings. The results are shown in \cref{tab:tab3}. Specifically, the SDF-only denotes that only the SDF branch is used for neural volume rendering without the density branch, and the Density-only represents the structure that only contains the density branch. Dual is the proposed method, in which the final color comes from the density branch, and the geometry is extracted from the SDF branch. Experimental results show that SDF-only obtains a better F-score value than Density-only, while the latter has a higher PSNR value than the former. The experimental results further confirm our observation that a single SDF branch excels in reconstruction, while a single density branch excels in rendering. Furthermore, The dual field significantly improves the view rendering performance, and simultaneously increases reconstruction performance.

\begin{table}[ht!]
  \caption{Ablation study of the Du-NeRF architecture. }
  \label{tab:tab3}
  \centering
  \footnotesize
  \resizebox{0.5\textwidth}{!}{
  \begin{tabular*}{\linewidth}{@{\extracolsep{\fill}}lcccc@{}}
    \toprule
    \textbf{Method}  & \textbf{C-$l_1 \downarrow$} & \textbf{F-score $\uparrow$} & \textbf{PSNR $\uparrow$}  & \textbf{LPIPS $\downarrow$} \\
    
    \midrule
    SDF-only   & 0.0133 & 0.974 & 31.741 & 0.110 \\
    Density-only  & 0.0168  & 0.934 & 32.744 & 0.090 \\
    Dual(SDF+Density) & \textbf{0.0115}  & \textbf{0.977} & \textbf{34.775} & \textbf{0.074} \\
    
    \bottomrule
  \end{tabular*}
  }

\end{table}

% As shown in \cref{tab:tab3}, the dual field significantly improves the view rendering performance, and simultaneously increases reconstruction performance. The SDF-only in \cref{tab:tab3} means only the SDF branch used for neural volume rendering without the density branch, while the Density-only represents only the density branch without that of SDF. Dual field means that two branches are used for volume rendering at the same time, in which the final color comes from the density branch, and the geometry is extracted from the SDF branch. 
% Experimental results show that SDF-only is higher than Density-only, both in F-score and PSNR. We suppose that SDF-only provides more accurate sampling positions compared to the latter. The dual-field architecture could take advantage of both the sampling of the former and the rendering of the latter, achieving the best performance in geometry reconstruction and novel view synthesis.

% This demonstrates the effectiveness of the proposed dual-field structure in enhancing the view rendering and indoor 3D reconstruction.
% We consider that the distribution of weights has more freedom in the newly added branch to find a set of weight parameters more suitable for the final color, thus improving the rendering quality. Otherwise, shared geometry feature help original SDF geometry learn more information.

\textbf{Effect of color disentanglement.} 
To explore the impact of color decoupling on view rendering and 3D reconstruction, we conducted several tests summarized in \cref{tab:tab4}. FC indicates supervision with color loss without decoupling, while VdC+ViC involves color loss from both View-dependent Color (VdC) and View-independent Color (ViC), calculated separately in VdC and ViC conditions. The results presented in \cref{tab:tab4} show that ViC supervision provides optimal performance for 3D reconstruction and volume rendering tasks. Furthermore, the comparable performance observed in the first two rows indicates that color decoupling does not improve reconstruction or rendering quality. In particular, the use of VdC as the supervision signal leads to a significant degradation in reconstruction and rendering performance compared to FC and VdC+ViC. \cref{fig:fig5} demonstrates that our method can accurately decompose the ViC and VdC color. For more experimental results on color disentanglement, please refer to the Supplementary.

\begin{table}[ht!]
  % \caption{Ablation study of color disentanglement (C-D) on reconstruction and novel view synthesis. \cref{tab:tab4} shows the effect of different color for SDF supervision on the reconstruction and novel view synthesis results after color decoupling. ViC (Ours) denotes the view-independent component, VdC is the view-dependent component, and FC represents the complete color. Experimental results demonstrate that supervised by the view-independent color component gives the best performance on reconstruction and rendering.}
  \caption{Ablation study of color disentanglement (C-D), where ViC (Ours) denotes the view-independent component, VdC is the view-dependent component, and FC represents the complete color, respectively.}
  \label{tab:tab4}
  \centering
  \footnotesize
  \resizebox{0.5\textwidth}{!}{
  \begin{tabular*}{\linewidth}{@{\extracolsep{\fill}}cccccc@{}}
    \toprule
    \textbf{C-D} & \textbf{Supervised} & \textbf{C-$l_1 \downarrow$} & \textbf{F-score $\uparrow$} & \textbf{PSNR $\uparrow$}  & \textbf{LPIPS $\downarrow$} \\
    
    \midrule
    w/o & FC & 0.0182 & 0.9549 & 34.494 & 0.090 \\
    w/ & VdC+ViC & 0.0185 & 0.9526 & 34.710 & 0.080 \\
    w/ & VdC & 0.1658  & 0.6503 & 32.780 & 0.122 \\
    w/ & ViC & \textbf{0.0177}  & \textbf{0.9597} & \textbf{35.225} & \textbf{0.072} \\
    
    \bottomrule
  \end{tabular*}
  }

\end{table}

% The \cref{tab:tab4} shows the ablation experiments of the color disentanglement (C-D). We tested the impact of different components of the color decoupling process on the final result, including the view-independent component (ViC), view-dependent component (VdC), and complete color (FC) is the sum of VdC and ViC. For the first row we use Fc instead of VdC+ViC due to color decoupling is not used in this experiment. 
% It can be seen from the results that pure color decoupling cannot improve the quality of reconstruction and rendering from the first two rows. In addition, using ViC supervision to guide the learning of the SDF branch, the performance on F-score and PSNR are higher than other ways of supervision and the method without C-D, which confirms the effectiveness of our approach discussed in \cref{Dual-weight}. The color decomposition results are shown in \cref{fig:fig5}. It can be seen that the proposed method successfully decomposes the view-dependent color and view-independent color (see regions in the mirror, the reflective books, and the desk). 

\textbf{Effect of different grid representation.}
We examined how different scene representations affect reconstruction and rendering using three strategies, detailed in \cref{tab:tab5}. $\text{G}{\text{M}}-\text{C}{\text{M}}$ uses four-level resolution grids for both geometry and color, $\text{G}{\text{H}}-\text{C}{\text{H}}$ employs hash-based grids for both, and $\text{G}{\text{M}}-\text{C}{\text{H}}$ combines a four-level grid for geometry with a hash-based grid for color.
Findings in \cref{tab:tab5} indicate that while hash-based color grids improve rendering outcomes, hash-based geometry grids reduce reconstruction quality. The mixed representation of $\text{G}{\text{M}}-\text{C}{\text{H}}$ yields the best results in both F-score and PSNR, suggesting that color benefits from a complex representation, whereas geometry performs better with a simpler way.

\begin{table}[ht!]
  % \caption{Evaluation of different scene representation, where $\text{G}_{\text{M}}$, $\text{C}_{\text{M}}$, $\text{G}_{\text{H}}$, $\text{C}_{\text{H}}$, denote geometric multi-resolution grid, color multi-resolution grid, geometric hash grid, and color hash grid, respectively. It can be seen that a four-level multi-resolution grid has better geometry representation and a hash-based multi-resolution grid gives better image rendering performance.}
  \caption{Ablation study of different scene representation, where $\text{G}_{\text{M}}$, $\text{C}_{\text{M}}$, $\text{G}_{\text{H}}$, $\text{C}_{\text{H}}$, denote geometric multi-resolution grid, color multi-resolution grid, geometric hash grid, and color hash grid, respectively. }
  \label{tab:tab5}
  \centering
  \footnotesize
  \begin{tabular*}{\linewidth}{@{\extracolsep{\fill}}ccccc@{}}
    \toprule
    \textbf{Method} & \textbf{C-$l_1 \downarrow$} & \textbf{F-score $\uparrow$} & \textbf{PSNR $\uparrow$} & \textbf{LPIPS $\downarrow$} \\
    
    \midrule
    $\text{G}_{\text{M}}-\text{C}_{\text{M}}$  & \textbf{0.0177} & 0.9597 & 35.225 & 0.072 \\
    $\text{G}_{\text{H}}-\text{C}_{\text{H}}$ & 0.0294 & 0.8976 & 36.087 & 0.108 \\
    \textbf{$\text{G}_{\text{M}}-\text{C}_{\text{H}}$} & \textbf{0.0177} & \textbf{0.9600} & \textbf{36.503} & \textbf{0.048} \\
    \bottomrule
  \end{tabular*}

\end{table}

\begin{figure}[ht!]
	\centering

	\begin{subfigure}{\linewidth}
            \begin{minipage}[t]{0.48\linewidth}
                \centering
                \includegraphics[width=1\linewidth]{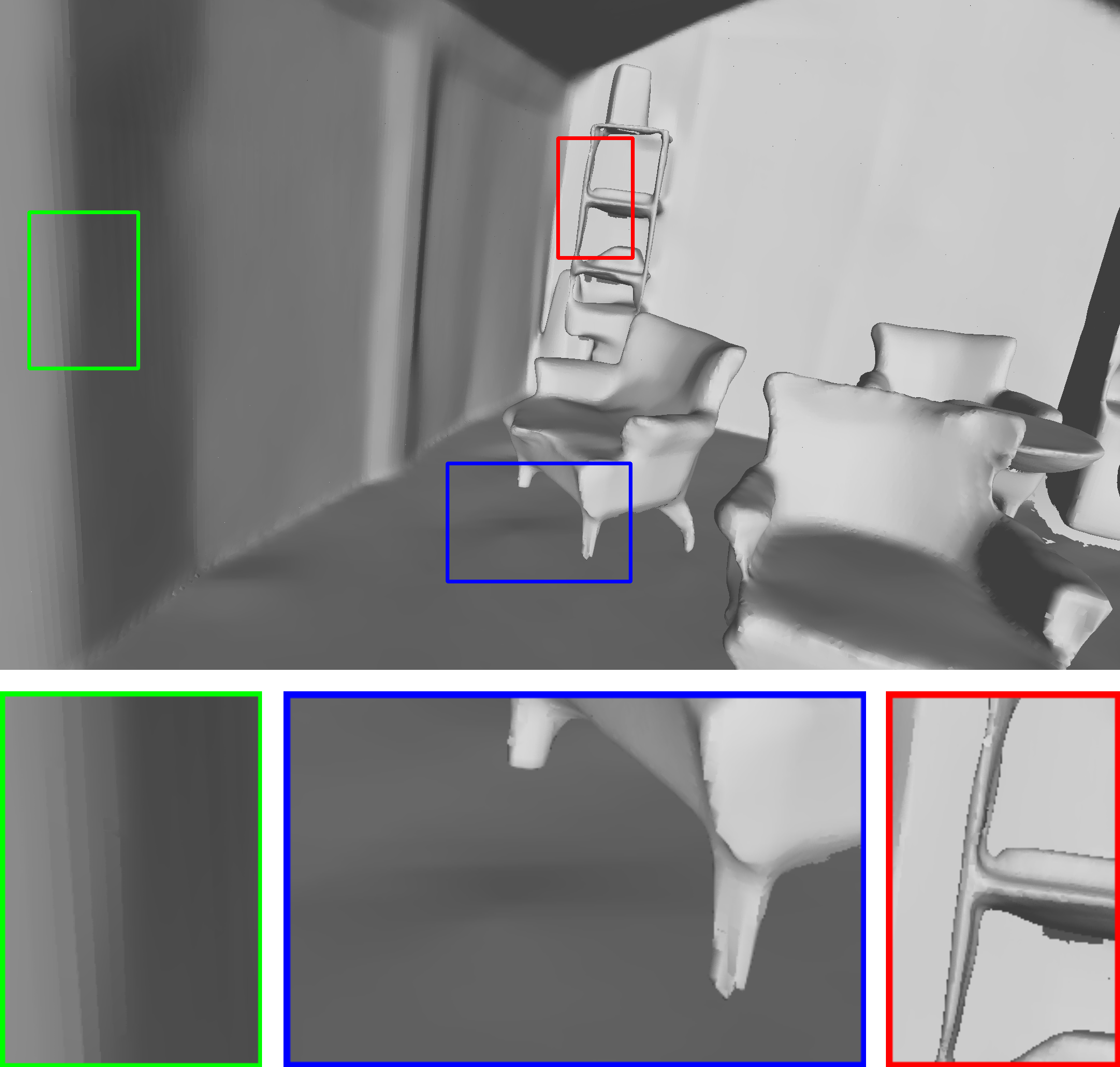}
                \caption{Ours}
            \end{minipage}
            \begin{minipage}[t]{0.48\linewidth}
                \centering
                \includegraphics[width=1\linewidth]{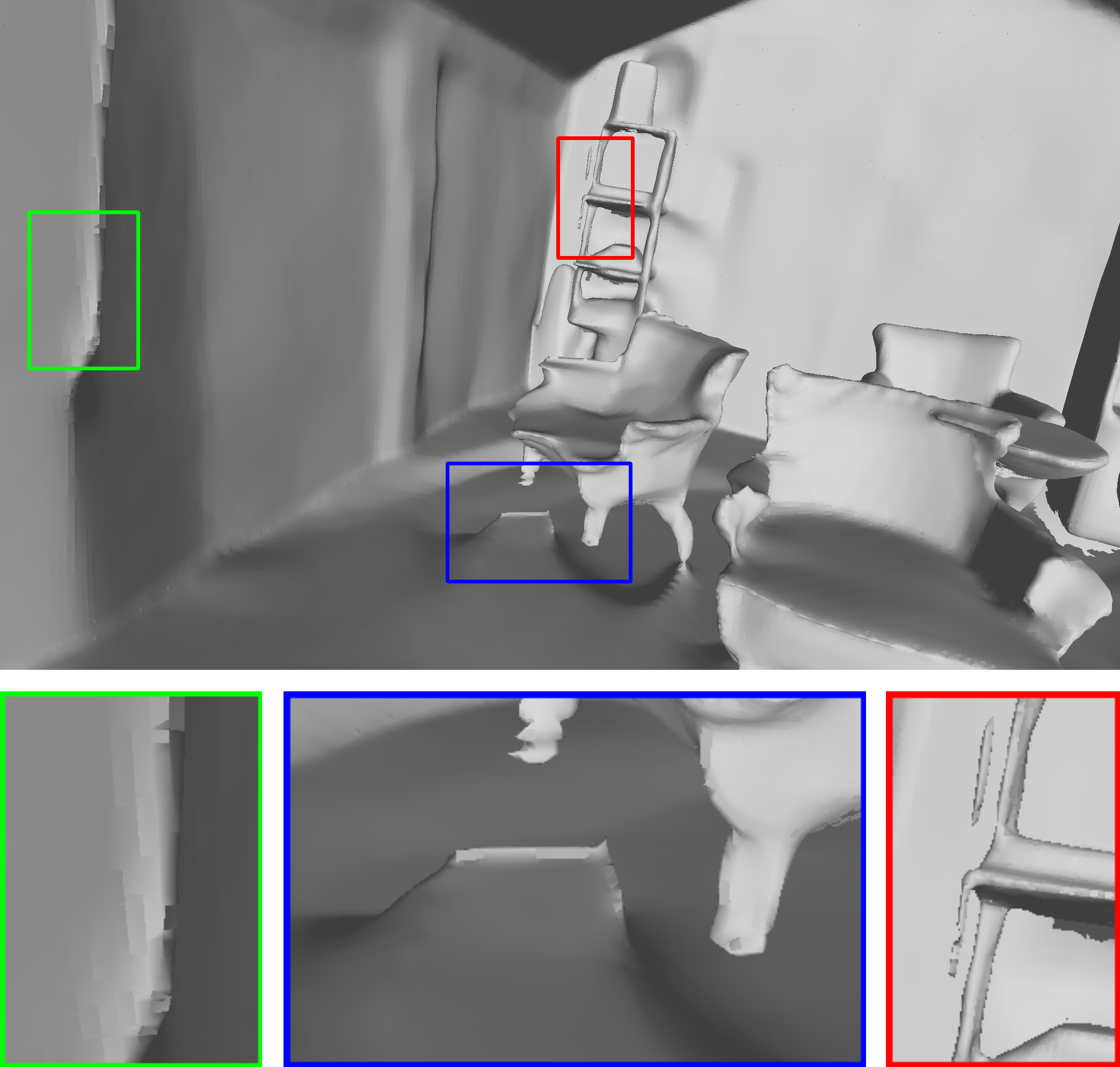}
                \caption{W/o pose optimization}
            \end{minipage}

        \end{subfigure}

	 \caption{The comparison experiments about pose optimization. Floating points and artifacts are present in the scene when pose optimization is turned off. } 
	\label{fig:pose_refine}
\end{figure}

\textbf{Effect of pose refinement and PFA evaluation strategy.} 
To assess the impact of pose optimization on 3D reconstruction, we tested two scenarios: with and without pose optimization, as illustrated in \cref{fig:pose_refine}. Results indicate that lacking pose optimization leads to defective meshes with unwanted floaters and compromised geometry.

Further, to ascertain the correct poses for testing images, we explored three configurations. "W/o Alignment" uses uncorrected provided poses, "IA" (Interpolated Alignment) derives poses from averaging those of two adjacent training images, and "PFA" (Proximity Frame Alignment) is our method. Results in \cref{tab:pose_refine} show that without proper pose alignment, view rendering quality significantly declines. Notably, our PFA method surpasses the IA strategy, which relies on simple interpolation, in rendering quality.

% To verify the effect of pose optimization, we conduct experiments to compare the 3D reconstruction performance in two conditions: with and without pose optimization. The results are shown in \cref{fig:pose_refine}. 
% It can be seen that the mesh without pose optimization tends to produce unwanted floaters and the geometry is corrupted. 

% To obtain the correct poses of the testing images, we conduct experiments in three settings, where \textit{W/o Alginment} denotes that the testing images use the provided poses without correction.
% \textit{IA} represents that the poses of the testing images are obtained with the average poses of the two adjacent training images, and PFA is the proposed method. The results are shown in \cref{tab:pose_refine}. It can be observed that without an appropriate pose alignment strategy, the view rendering quality is severely degraded. Furthermore, our proposed pose alignment strategy outperforms the IA method, which employs a simple interpolation approach.

\begin{table}[ht!]
  \caption{Evaluation of pose alignment strategy, included W/o Alignment, Interpolation Alignment (IA) and Proximity Frame Alignment (PFA). It can be seen that our PFA strategy could achieve the best performance.}
  % \caption{Ablation study of pose alignment strategy. For more details refer to \cref{Ablations}.}
  \label{tab:pose_refine}
  \centering
  \footnotesize
  \begin{tabular*}{\linewidth}{@{\extracolsep{\fill}}cccc@{}}
    \toprule
    \textbf{Method} & \textbf{PSNR $\uparrow$} & \textbf{SSIM $\uparrow$} & \textbf{LPIPS $\downarrow$} \\
    
    \midrule
    W/o Alignment  & 18.357 & 0.679 & 0.398 \\
    IA & 21.802 & 0.674 & 0.342 \\
    PFA (ours) & \textbf{26.712} & \textbf{0.796} & \textbf{0.296} \\
    \bottomrule
  \end{tabular*}

\end{table}

\section{Concluding remarks}
\label{sec:Discussion}

We have presented a dual neural radiance field (Du-NeRF) for high-fidelity 3D indoor scene reconstruction and rendering simultaneously. This method incorporates two branches, one for reconstruction and another for rendering, allowing mutual enhancement. Additionally, we developed an effective framework that facilitates the on-the-fly extraction of view-independent color information from the model, which can be leveraged for supervising 3D reconstruction, potentially yielding smoother and more complete mesh representations.

% We have presented a dual neural radiance field for high-fidelity 3D indoor scene reconstruction and rendering simultaneously. By designing two branches for reconstruction and view rendering respectively, the proposed method allows the two tasks to benefit from each other. To further improve the reconstruction and rendering performances, we designed a self-supervised color decomposition method. Experimental results have shown that our self-supervised method simultaneously improves the geometric reconstruction and rendering capabilities. Moreover, our method achieves state-of-the-art performances compared to previous methods.

Besides, the effectiveness of the proposed method remains to be tested under a few-shot scenario where a small number of RGBD images are provided. In addition, trying new feature representations such as replacing hash representation with Gaussian splitting may improve performances. We will address these in future works.

% \textbf{Limitation and future work.}
% Although Du-NeRF has been experimentally proven to be effective, our method may perform poorly if the observed images are blurred because it relies on multi-view color consistency to supervise geometry reconstruction and blur images do not satisfy such requirements. Besides, the effectiveness of the proposed method remains to be tested under a few-shot scenario where a small number of RGBD images are provided. In addition, trying new feature representations such as replacing hash representation with Gaussian splitting may improve performances. We will address these in future works. 

% \begin{acks}
% \end{acks}

\bibliographystyle{ACM-Reference-Format}
\bibliography{sample-authordraft}

\end{document}

\title{3D Reconstruction and Novel View Synthesis of Indoor Environments based on a Dual Neural Radiance Field}

\author{Supplementary Material}

\maketitle
% \maketitlesupplementary

\section{Additional Ablation Study}\label{Additional Ablation Study}
\begin{figure*}[ht!]
	\centering
	\begin{subfigure}{\linewidth}
            \rotatebox[origin=c]{90}{\small{w/o $\lambda_{\text{align}}$}\hspace{-1.2cm}}
            \begin{minipage}[t]{0.232\linewidth}
                \centering
                \includegraphics[width=1\linewidth]{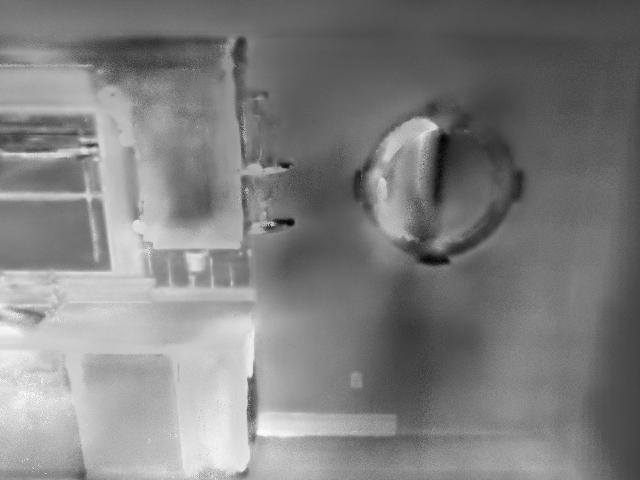}
            \end{minipage}
            \vspace{0.3pt}
            \begin{minipage}[t]{0.232\linewidth}
                \centering
                \includegraphics[width=1\linewidth]{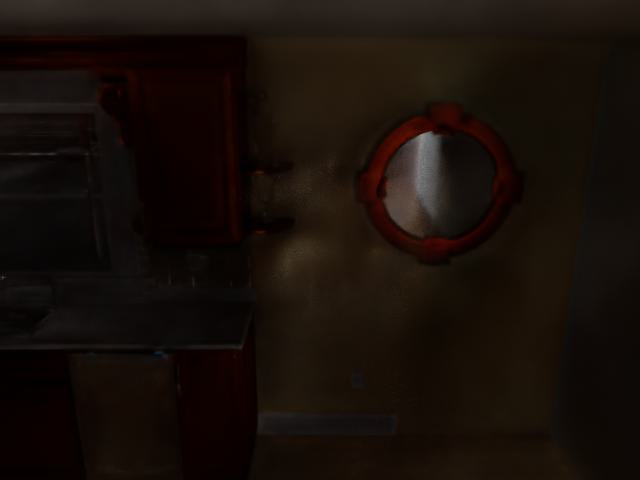}
            \end{minipage}
            \vspace{0.3pt}
            \begin{minipage}[t]{0.232\linewidth}
                \centering
                \includegraphics[width=1\linewidth]{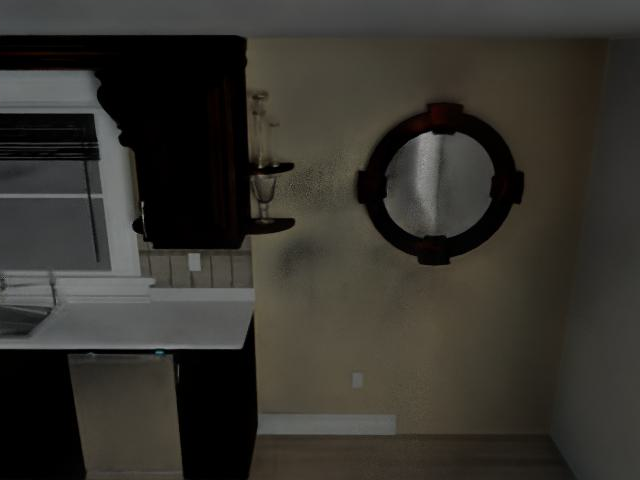}
            \end{minipage}
            \vspace{0.3pt}
            \begin{minipage}[t]{0.232\linewidth}
                \centering
                \includegraphics[width=1\linewidth]{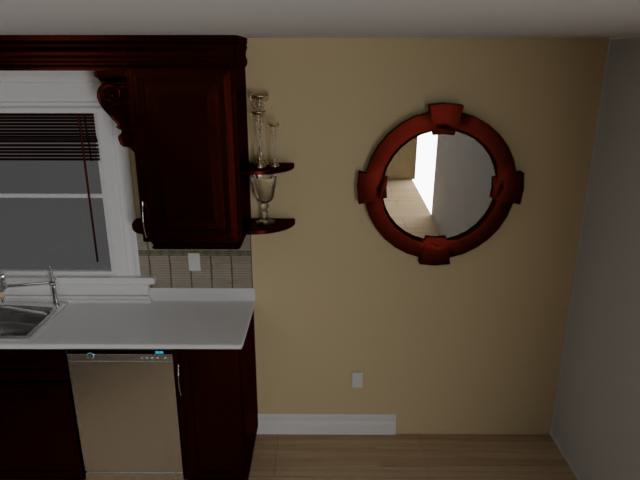}
            \end{minipage}

        \end{subfigure}

	\begin{subfigure}{\linewidth}
            \rotatebox[origin=c]{90}{\small{w/ $\lambda_{\text{align}}$}\hspace{-1.2cm}}
            \begin{minipage}[t]{0.232\linewidth}
                \centering
                \includegraphics[width=1\linewidth]{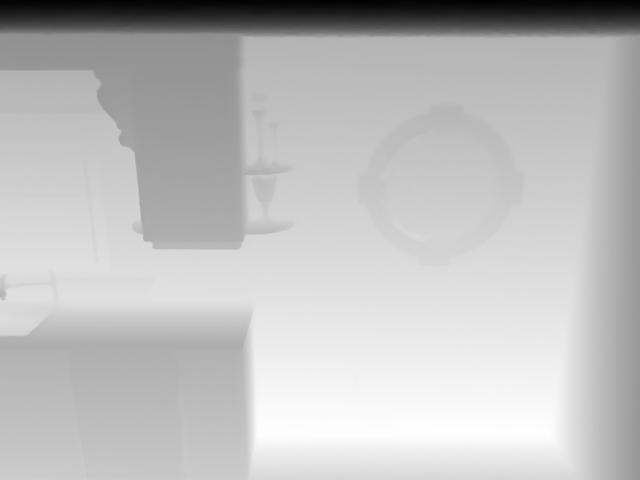}
                \caption{Depth}
                \label{6a}
            \end{minipage}
            \begin{minipage}[t]{0.232\linewidth}
                \centering
                \includegraphics[width=1\linewidth]{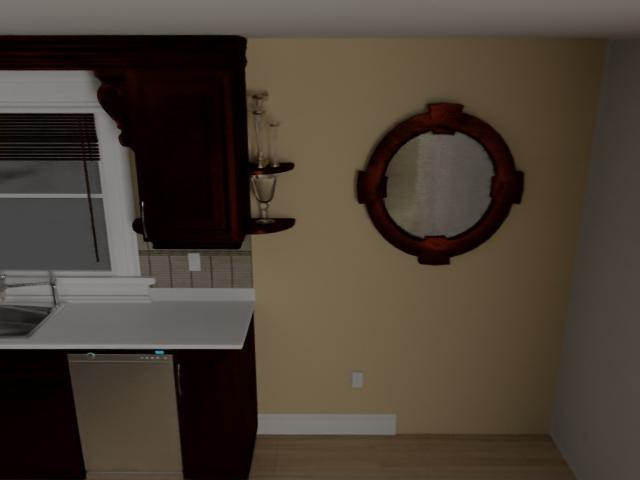}
                \caption{V$_\text{i}$ color}
                \label{6b}
            \end{minipage}
            \begin{minipage}[t]{0.232\linewidth}
                \centering
                \includegraphics[width=1\linewidth]{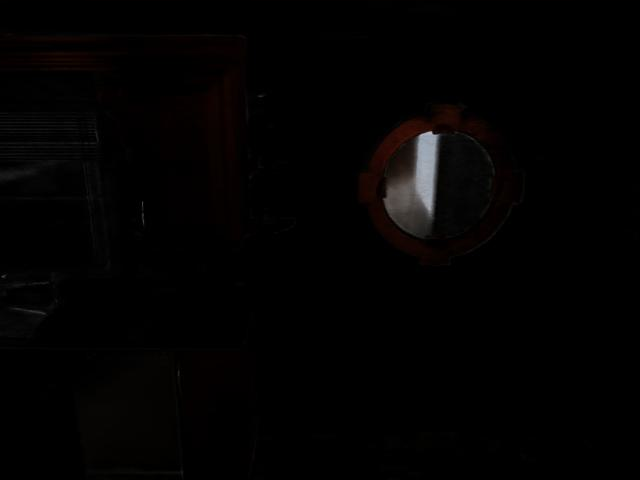}
                \caption{V$_\text{d}$ color}
                \label{6c}
            \end{minipage}
            \begin{minipage}[t]{0.232\linewidth}
                \centering
                \includegraphics[width=1\linewidth]{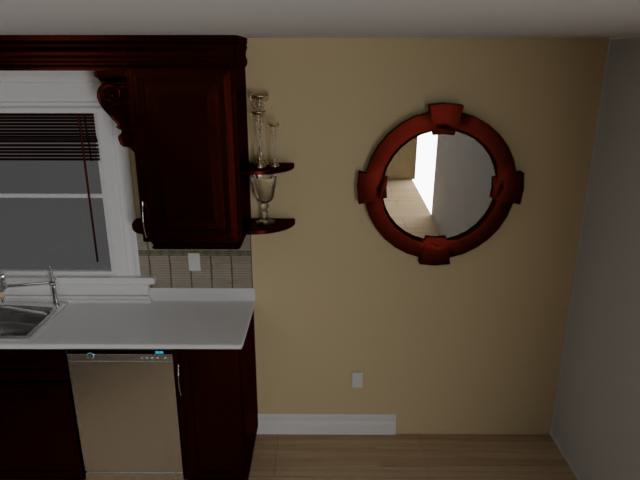}
                \caption{GT}
                \label{6d}
            \end{minipage}

        \end{subfigure}

	\caption{We evaluates the effect of different $\lambda_{\text{align}}$ on the generated view-independent color. The first row shows the results w/o $\lambda_\text{align}$, while second for w/ $\lambda_\text{align}$. The (a), (b), (c) are the rendering depth map, view-independent color, and view-dependent color, respectively, while (d) is the complete images. The experimental results indicate that we could not get an accurate diffuse component without $\lambda_\text{align}$.} 
	\label{fig:depth_align}
\end{figure*}

\textbf{Effect of the depth alignment loss.}
We conduct experiments to verify the importance of $\lambda_\text{align}$ to color decomposition, as shown in \cref{fig:depth_align}. It can be seen that without $\lambda_\text{align}$, the generated depth image is very coarse and the wall and mirror are not smooth while the method with $\lambda_\text{align}$ gives the accurate depth image. Besides, without $\lambda_\text{align}$ the method seems to fail to decompose color into the view-independent component and the view-dependent component, especially in mirror region and wall regions.

\textbf{Effect of self-supervised loss.} We conduct experiments to verify the impact of different $\lambda_d$ on reconstruction performance and rendering quality. Experimental results show that an appropriate $\lambda_d$ value is crucial to balance the effects of color and depth supervision. Specifically, smaller $\lambda_d$ results in enhancing the effect of the depth loss and reducing the effect of the view-independent color supervision, while larger  $\lambda_d$ increases the effect of color supervision and decreases the effect of depth supervision.
We carry out experiments with the different values of $\lambda_d$ ranging from 0.5 to 50. The results in \cref{tab:tab7} show that $\lambda_d=5$ gives the best performance. In addition, the performance drops whenever $\lambda_d$ is larger or smaller. 

\begin{table}[ht!]
  \caption{Ablation study of self-supervised loss. We evaluate the effect of different values of $\lambda_d$ on the view synthesis and reconstruction. It can be seen that best performance is obtained when $\lambda_d$ is 5. Worse results are achieved whenever $\lambda_d$ increases or decreases.}
  \label{tab:tab7}
  \centering
  \footnotesize
  \resizebox{0.5\textwidth}{!}{
  \begin{tabular*}{\linewidth}{@{\extracolsep{\fill}}lcccc@{}}
    \toprule
    \textbf{$\lambda_d$}  & \textbf{C-$l_1 \downarrow$} & \textbf{F-score $\uparrow$} & \textbf{PSNR $\uparrow$}  & \textbf{LPIPS $\downarrow$} \\
    
    \midrule
    $0.5$   & 0.0132 & \textbf{0.985} & 36.926 & 0.0344 \\
    $2.0$  & 0.0134  & \textbf{0.985} & 37.076 & 0.0353 \\
    $5.0$ (Ours) & \textbf{0.0128}  & \textbf{0.985} & \textbf{37.167} & \textbf{0.0324} \\
    $10.0$  & 0.0132  & \textbf{0.985} & 37.146 & 0.0332 \\
    $50.0$ & 0.0131  & \textbf{0.985} & 36.713 & 0.0359 \\
    
    \bottomrule
  \end{tabular*}
  }

\end{table}

\section{Additional Experimental Results}\label{Additional Experimental Results}

\subsection{More detailed experiment results}
Additional qualitative results are shown in \cref{fig:fig7,fig:fig8}. 

We compare the proposed method with \textit{BundleFusion}, \textit{Neus}, \textit{VolSDF}, \textit{NeuralRGBD} and \textit{Go-Surf} in Scene \textit{Grey-white room}, \textit{Office2}, \textit{Office3}, \textit{Room0} for geometry reconstruction, as shown in \cref{fig:fig7}. 
Our method achieves the best visual result on all indoor scenes. It can be seen from \cref{fig:fig7} that our method has richer object details and smoother planes of indoor scenes. Specifically, Our method fills in the office chair legs completely and achieves a smoother result on the background walls and floors in the second and third columns. In the last column, our method is able to reconstruct the complete table compared to \textit{BundleFusion}, \textit{Neus} and \textit{NeuralRGBD}, and refine the shape of the bottle compared to \textit{Go-Surf}.

In addition, view synthesis performance on  \textit{Whiteroom}, \textit{Office3}, \textit{Room0}, and \textit{Room1} scenes of the  \textit{Neus}, \textit{InstantNGP}, \textit{DVGO}, \textit{Go-Surf} and \textit{NeuralRGBD}, are shown in \cref{fig:fig8}. Our approach achieves the best image rendering results, in both the texture-less areas and rich texture areas. For instance, we accurately restore the appearance of texture-less regions such as ceiling corners in the first column and walls in the second and third columns. Additionally, we reproduce the details of complex texture regions, such as the shutter in the third column and the stripes of the quilt in the last column. Methods that focus on view rendering such as \textit{DVGO} and \textit{InstantNGP} suffer in texture-less or sparsely observed regions, as shown in the first and third columns of \cref{fig:fig8}. Approaches focus on surface reconstruction, such as \textit{NeuralRGBD} and \textit{Go-Surf}, often struggle to get good view rendering results on areas of complex texture such as the windows in the third column and the quilt in the fourth column. 

The detailed quantitative results on Replica dataset and NeuralRGBD dataset can be seen in \cref{tab:tab11} and in \cref{tab:tab9,tab:tab10}, respectively.  Our method obtains the SOTA rendering performance in all scenes and gives the highest performance in most of the scenes for geometry reconstruction.

\subsection{More color decomposition results}
\begin{figure*}[ht!]
	\centering
 	\begin{subfigure}{\linewidth}
            \begin{minipage}[t]{0.24\linewidth}
                \centering
                \includegraphics[width=1\linewidth]{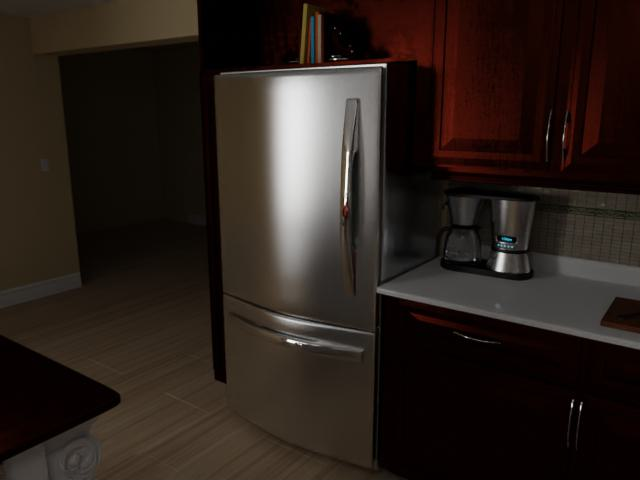}
            \end{minipage}
            \begin{minipage}[t]{0.24\linewidth}
                \centering
                \includegraphics[width=1\linewidth]{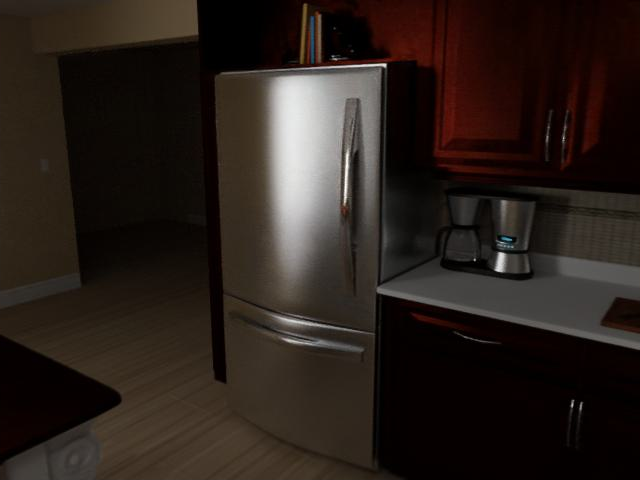}
            \end{minipage}
            \begin{minipage}[t]{0.24\linewidth}
                \centering
                \includegraphics[width=1\linewidth]{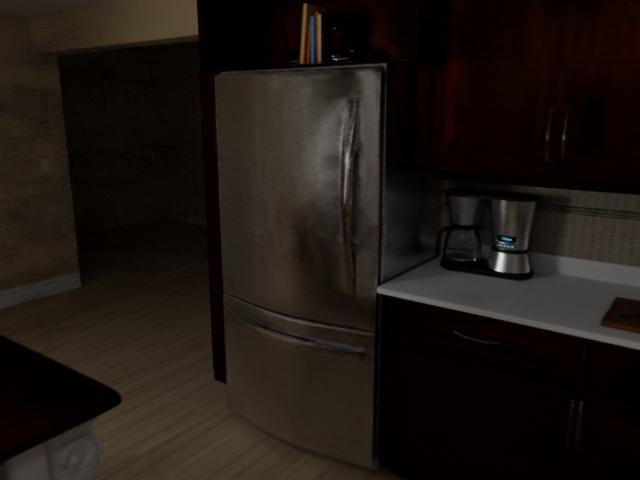}
            \end{minipage}
            \begin{minipage}[t]{0.24\linewidth}
                \centering
                \includegraphics[width=1\linewidth]{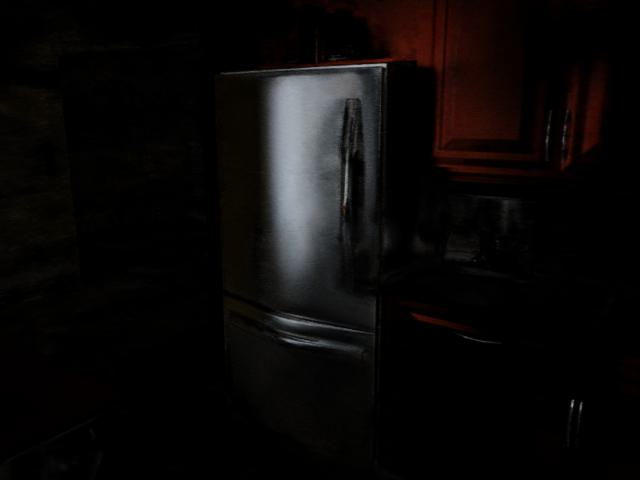}
            \end{minipage}
        \end{subfigure}       
	\begin{subfigure}{\linewidth}
            \begin{minipage}[t]{0.24\linewidth}
                \centering
                \includegraphics[width=1\linewidth]{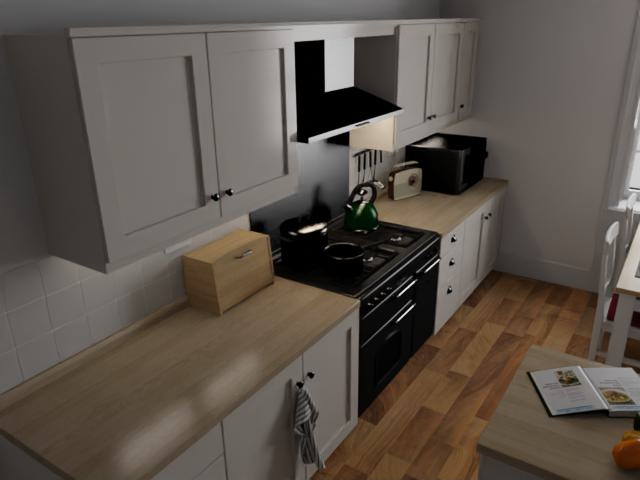}
            \end{minipage}
            \begin{minipage}[t]{0.24\linewidth}
                \centering
                \includegraphics[width=1\linewidth]{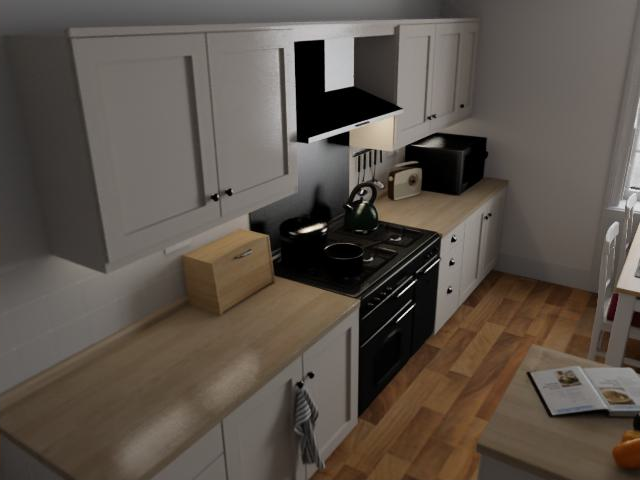}
            \end{minipage}
            \begin{minipage}[t]{0.24\linewidth}
                \centering
                \includegraphics[width=1\linewidth]{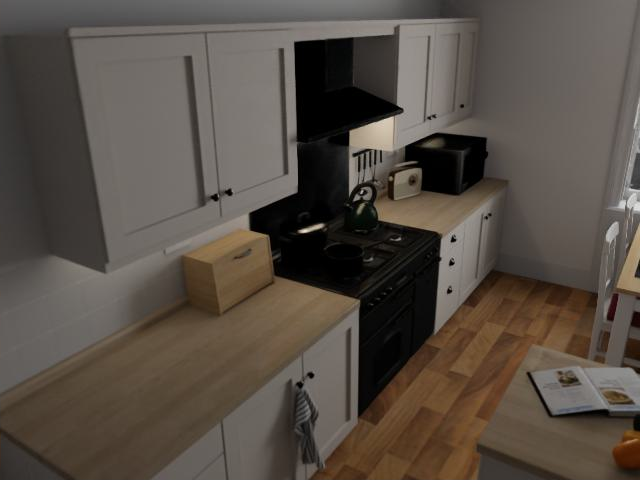}
            \end{minipage}
            \begin{minipage}[t]{0.24\linewidth}
                \centering
                \includegraphics[width=1\linewidth]{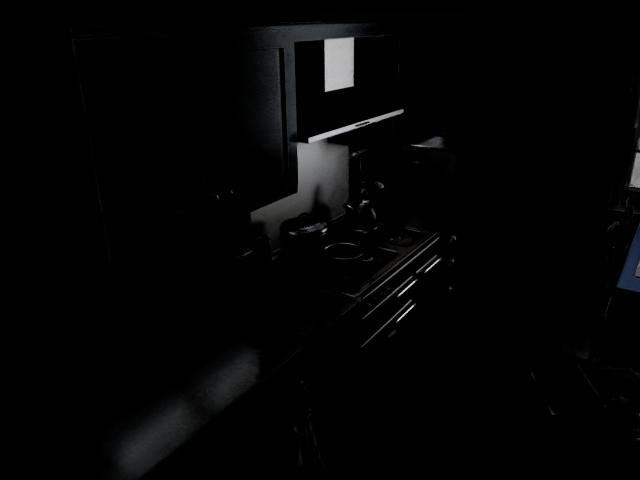}
            \end{minipage}
        \end{subfigure}

	\begin{subfigure}{\linewidth}
            \begin{minipage}[t]{0.24\linewidth}
                \centering
                \includegraphics[width=1\linewidth]{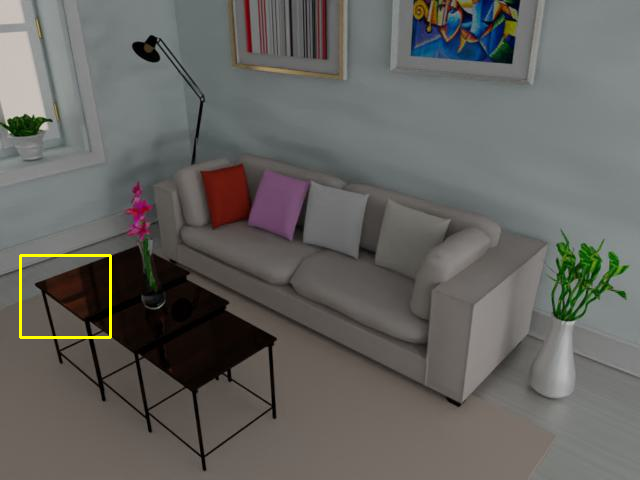}
            \end{minipage}
            \begin{minipage}[t]{0.24\linewidth}
                \centering
                \includegraphics[width=1\linewidth]{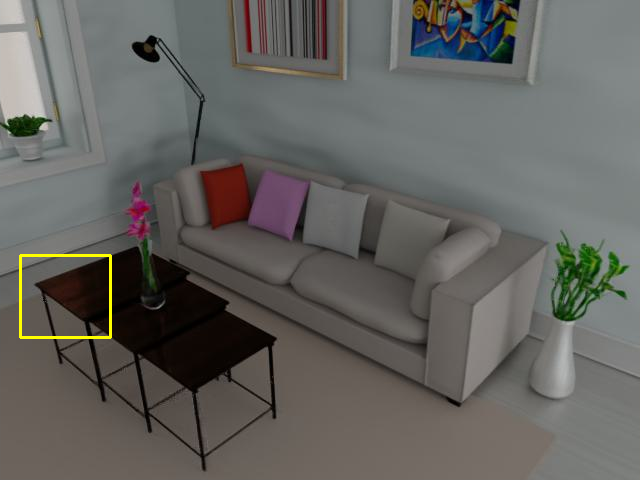}
            \end{minipage}
            \begin{minipage}[t]{0.24\linewidth}
                \centering
                \includegraphics[width=1\linewidth]{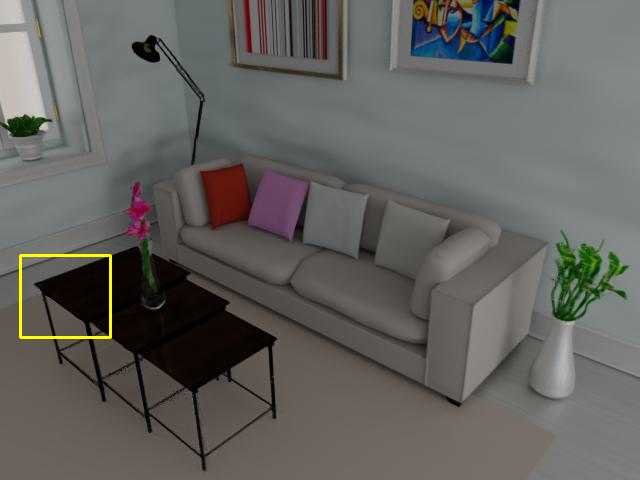}
            \end{minipage}
            \begin{minipage}[t]{0.24\linewidth}
                \centering
                \includegraphics[width=1\linewidth]{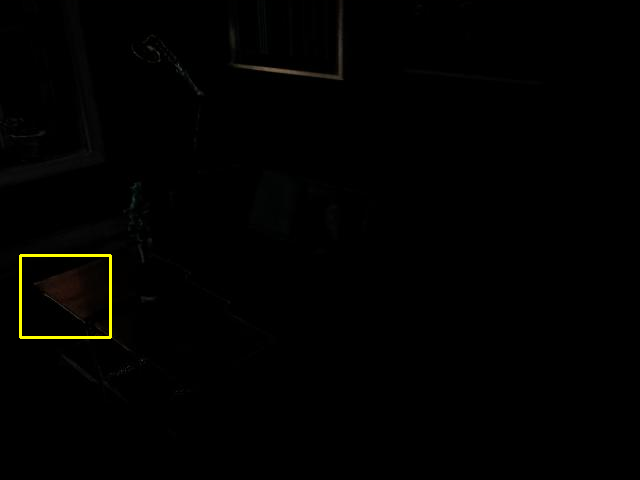}
            \end{minipage}
        \end{subfigure}

	\begin{subfigure}{\linewidth}
            \begin{minipage}[t]{0.24\linewidth}
                \centering
                \includegraphics[width=1\linewidth]{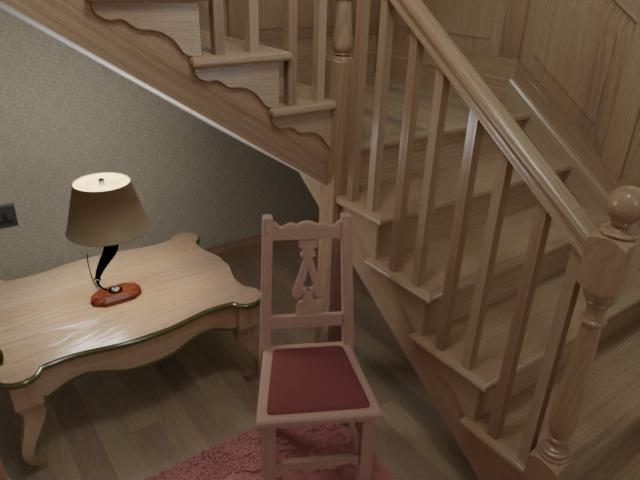}
            \end{minipage}
            \begin{minipage}[t]{0.24\linewidth}
                \centering
                \includegraphics[width=1\linewidth]{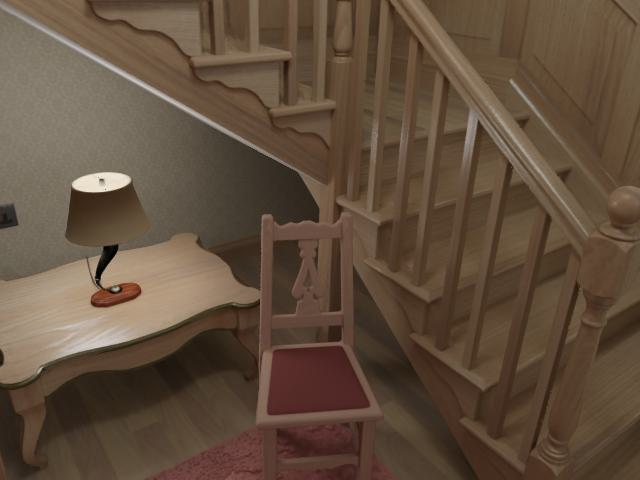}
            \end{minipage}
            \begin{minipage}[t]{0.24\linewidth}
                \centering
                \includegraphics[width=1\linewidth]{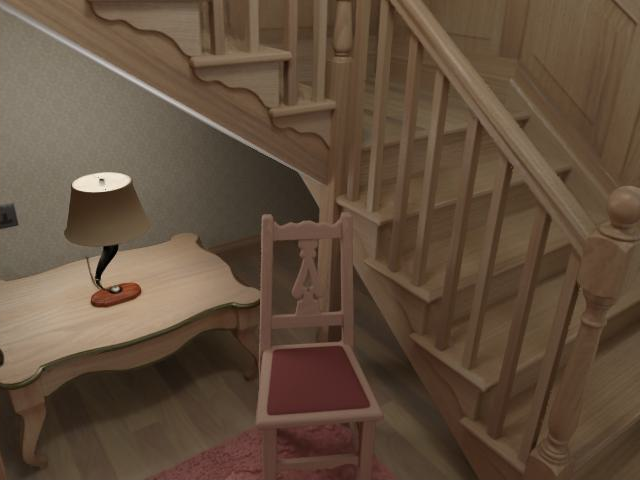}
            \end{minipage}
            \begin{minipage}[t]{0.24\linewidth}
                \centering
                \includegraphics[width=1\linewidth]{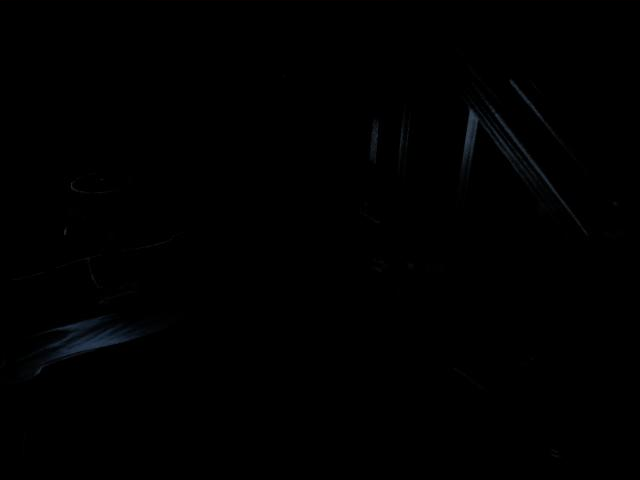}
            \end{minipage}
        \end{subfigure}

	\begin{subfigure}{\linewidth}
            \begin{minipage}[t]{0.24\linewidth}
                \centering
                \includegraphics[width=1\linewidth]{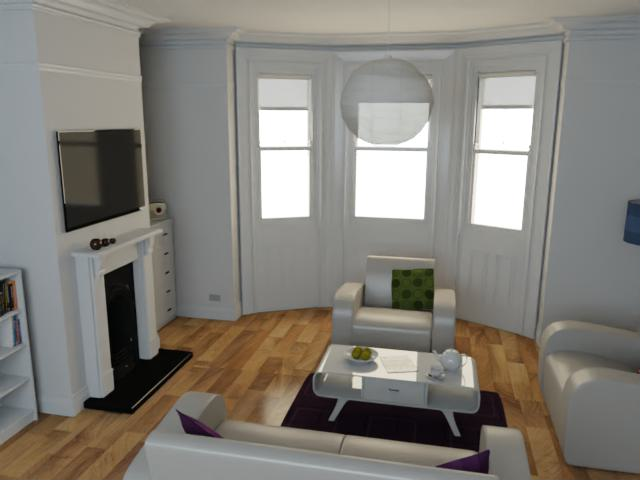}
                \caption{GT color}
            \end{minipage}
            \begin{minipage}[t]{0.24\linewidth}
                \centering
                \includegraphics[width=1\linewidth]{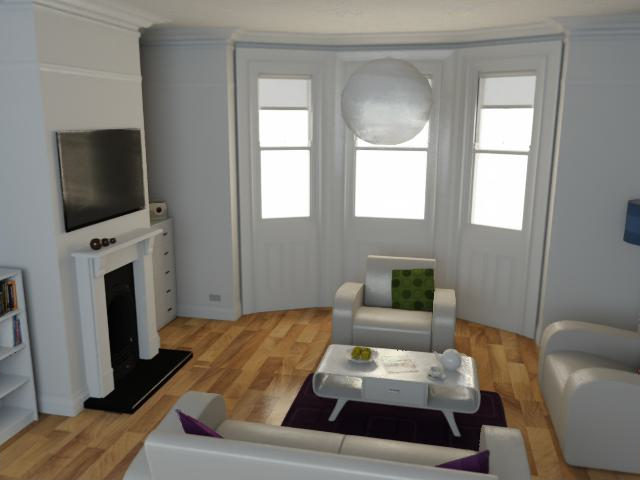}
                \caption{Rendering Color}
            \end{minipage}
            \begin{minipage}[t]{0.24\linewidth}
                \centering
                \includegraphics[width=1\linewidth]{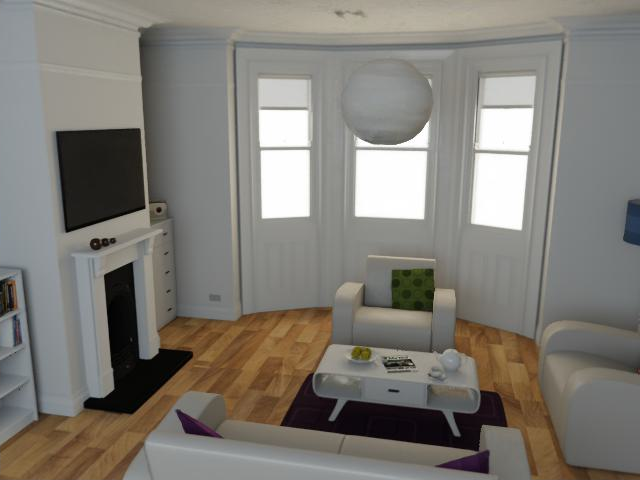}
                \caption{V$_\text{i}$ color}
            \end{minipage}
            \begin{minipage}[t]{0.24\linewidth}
                \centering
                \includegraphics[width=1\linewidth]{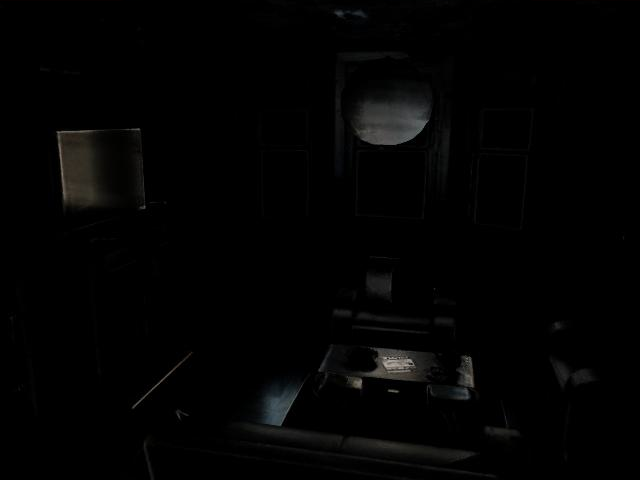}
                \caption{V$_\text{d}$ color}
            \end{minipage}
        \end{subfigure}

	 \caption{More color decoupling results. The V$_\text{i}$ color represents the view-independent color, while V$_\text{d}$ color is the view-dependent color. It can be seen that our method can effectively decouple the complete color into the view-independent (diffusion surfaces) and view-dependent (specular reflective surfaces) colors.} %, such as the mirrors on the wall and the books on the table.}
 % { Intuitively, when we use books with consistent view-independent color to learn the geometry of books, it will be better than using reflective books to learn geometry.}
	\label{fig:fig10}
\end{figure*}

\begin{figure*}[ht!]
	\centering
 
	\begin{subfigure}{\linewidth}
            \rotatebox[origin=c]{90}{\normalsize{BundleFusion}\hspace{-2.2cm}}
            \begin{minipage}[t]{0.24\linewidth}
                \centering
                \includegraphics[width=1\linewidth]{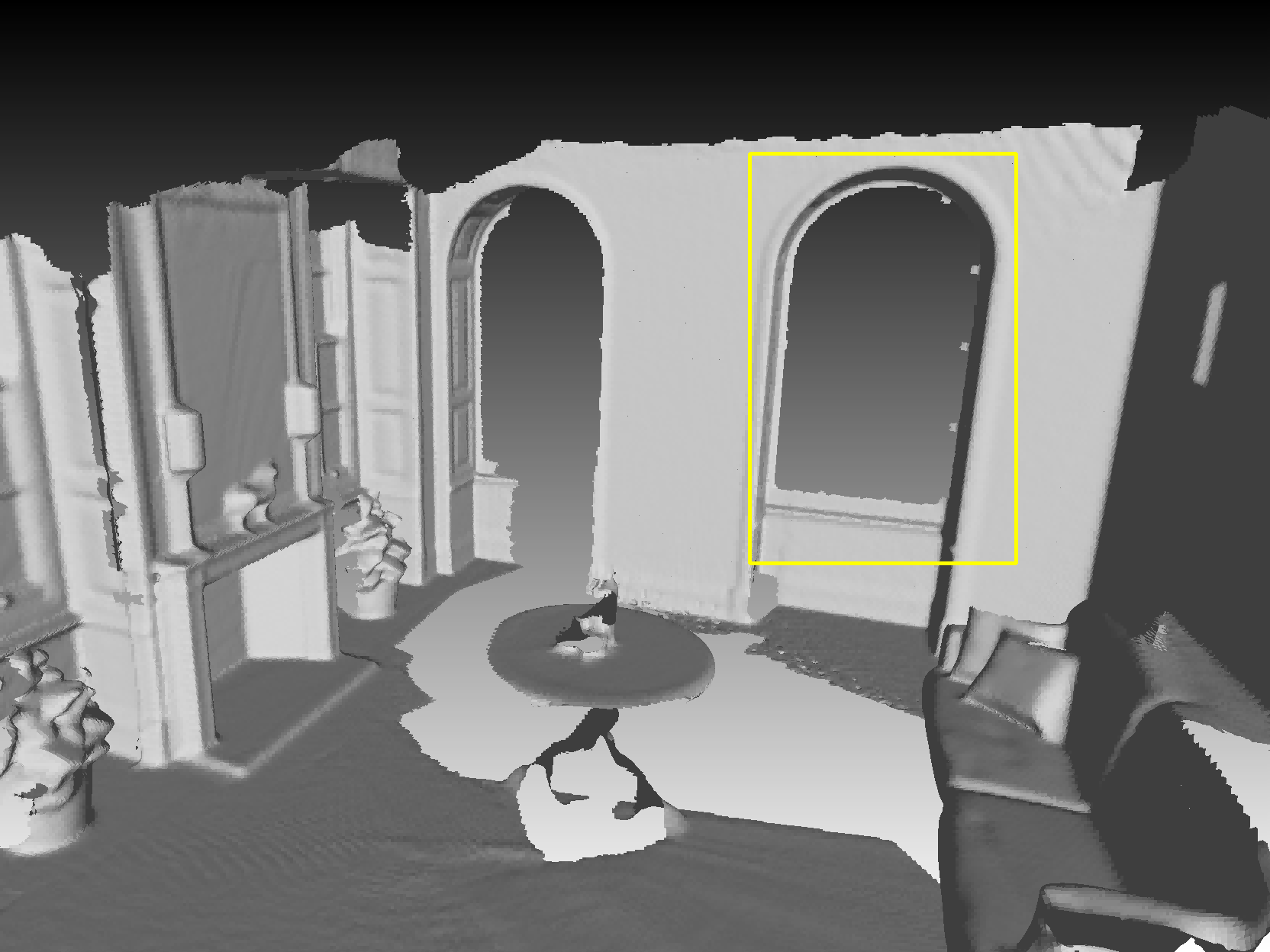}
            \end{minipage}
            \begin{minipage}[t]{0.24\linewidth}
                \centering
                \includegraphics[width=1\linewidth]{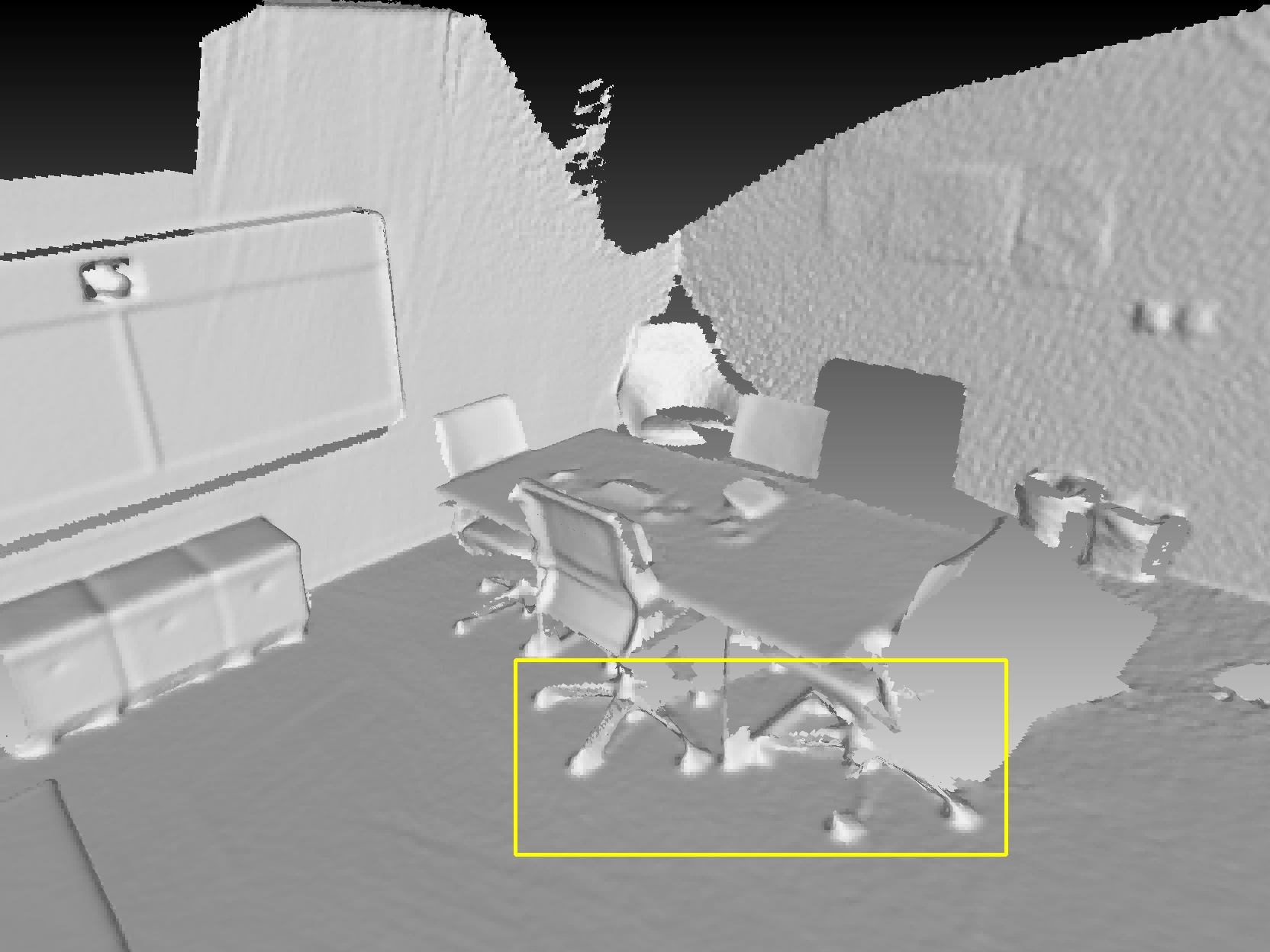}
            \end{minipage}
            \begin{minipage}[t]{0.24\linewidth}
                \centering
                \includegraphics[width=1\linewidth]{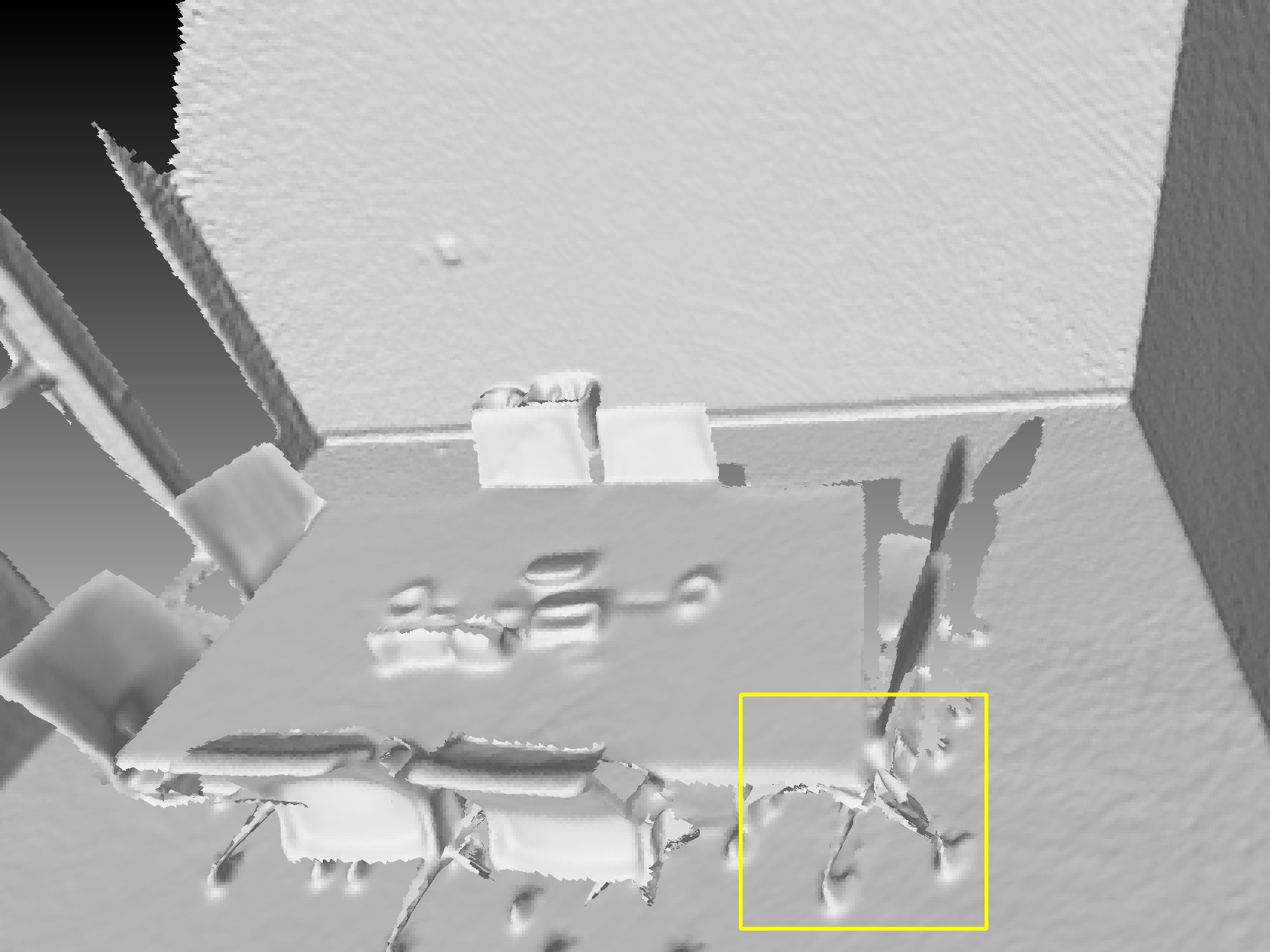}
            \end{minipage}
            \begin{minipage}[t]{0.24\linewidth}
                \centering
                \includegraphics[width=1\linewidth]{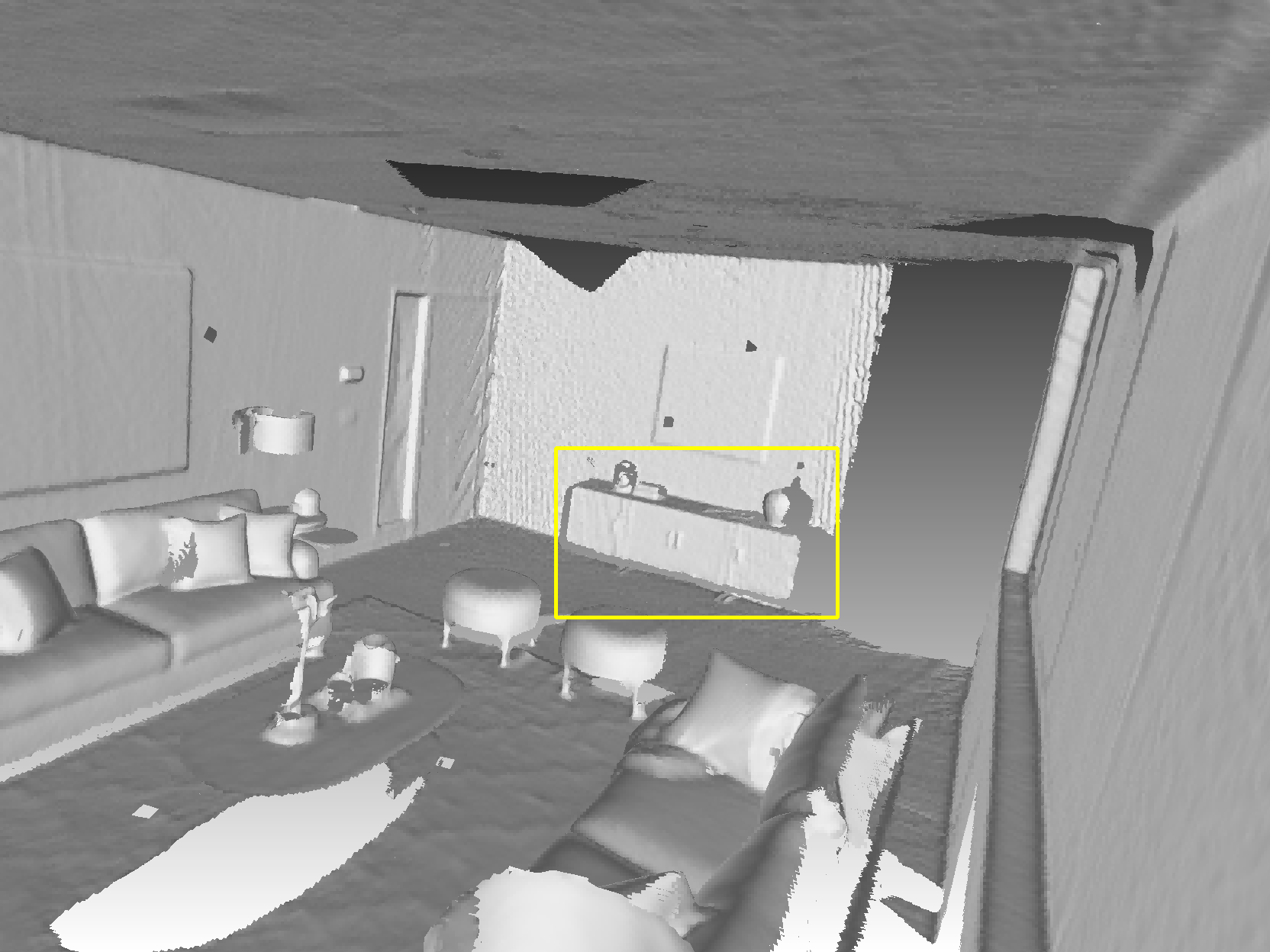}
            \end{minipage}
        \end{subfigure}
        
	\begin{subfigure}{\linewidth}
            \rotatebox[origin=c]{90}{\normalsize{Neus}\hspace{-2.2cm}}
            \begin{minipage}[t]{0.24\linewidth}
                \centering
                \includegraphics[width=1\linewidth]{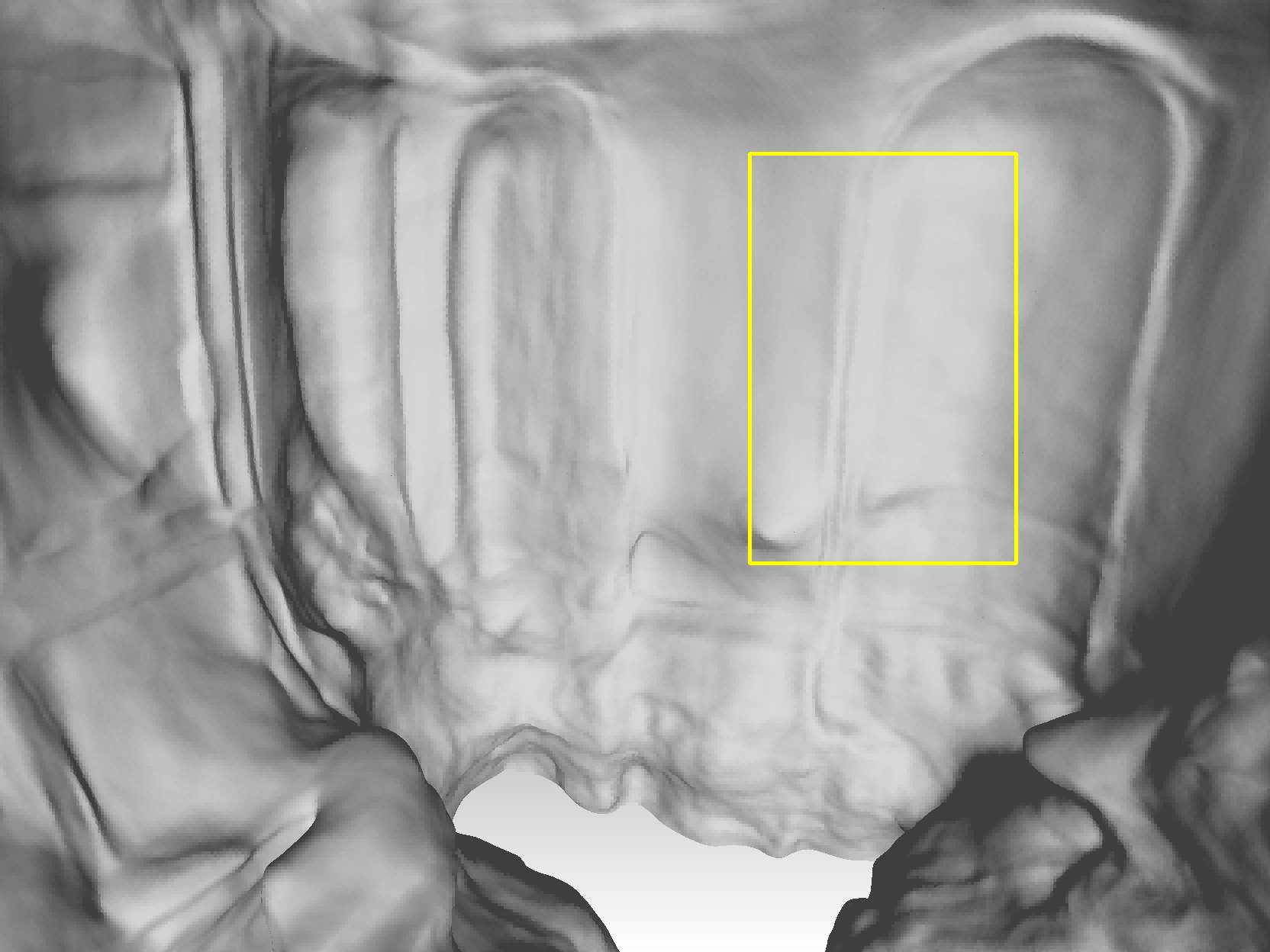}
            \end{minipage}
            \begin{minipage}[t]{0.24\linewidth}
                \centering
                \includegraphics[width=1\linewidth]{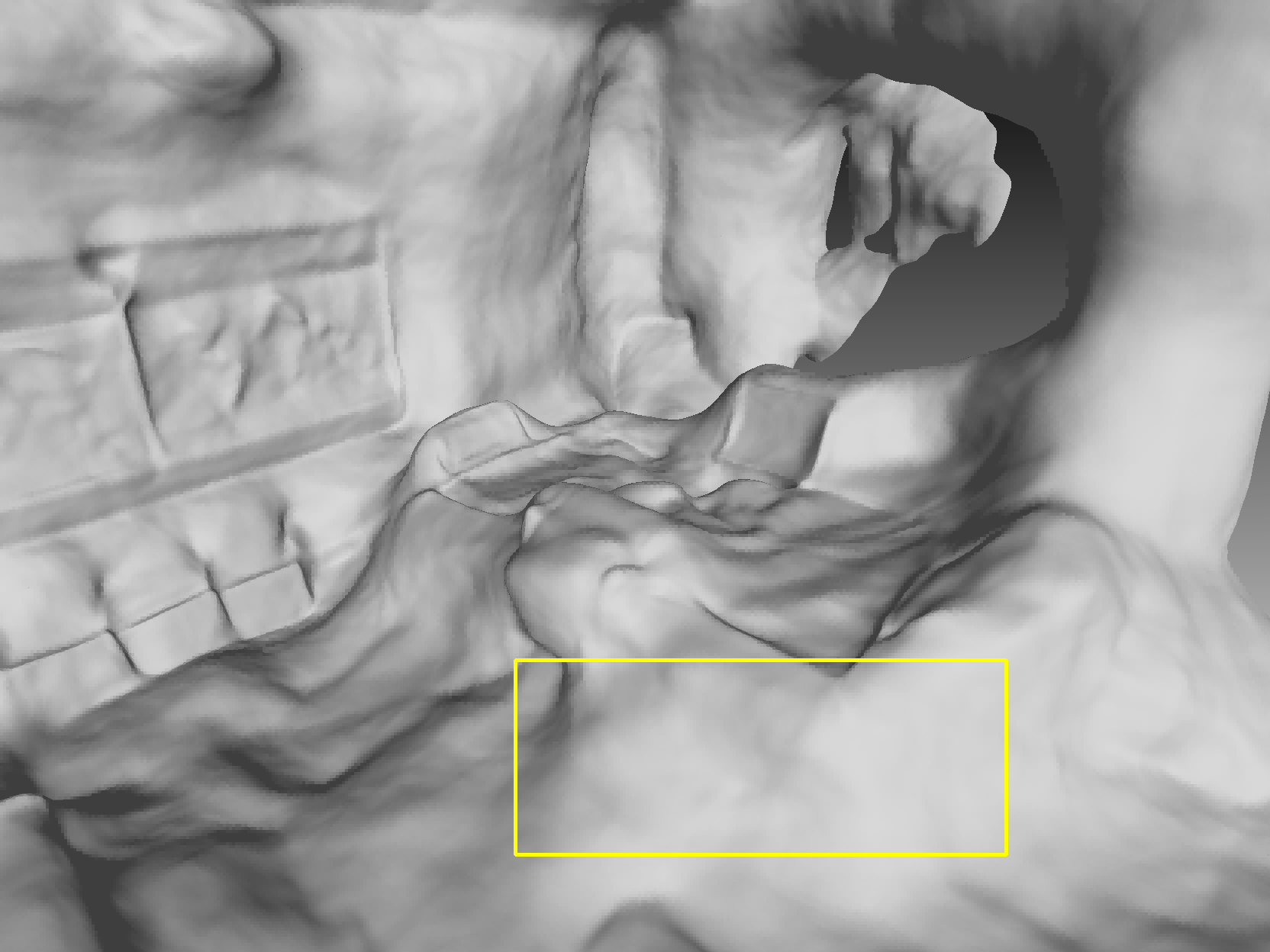}
            \end{minipage}
            \begin{minipage}[t]{0.24\linewidth}
                \centering
                \includegraphics[width=1\linewidth]{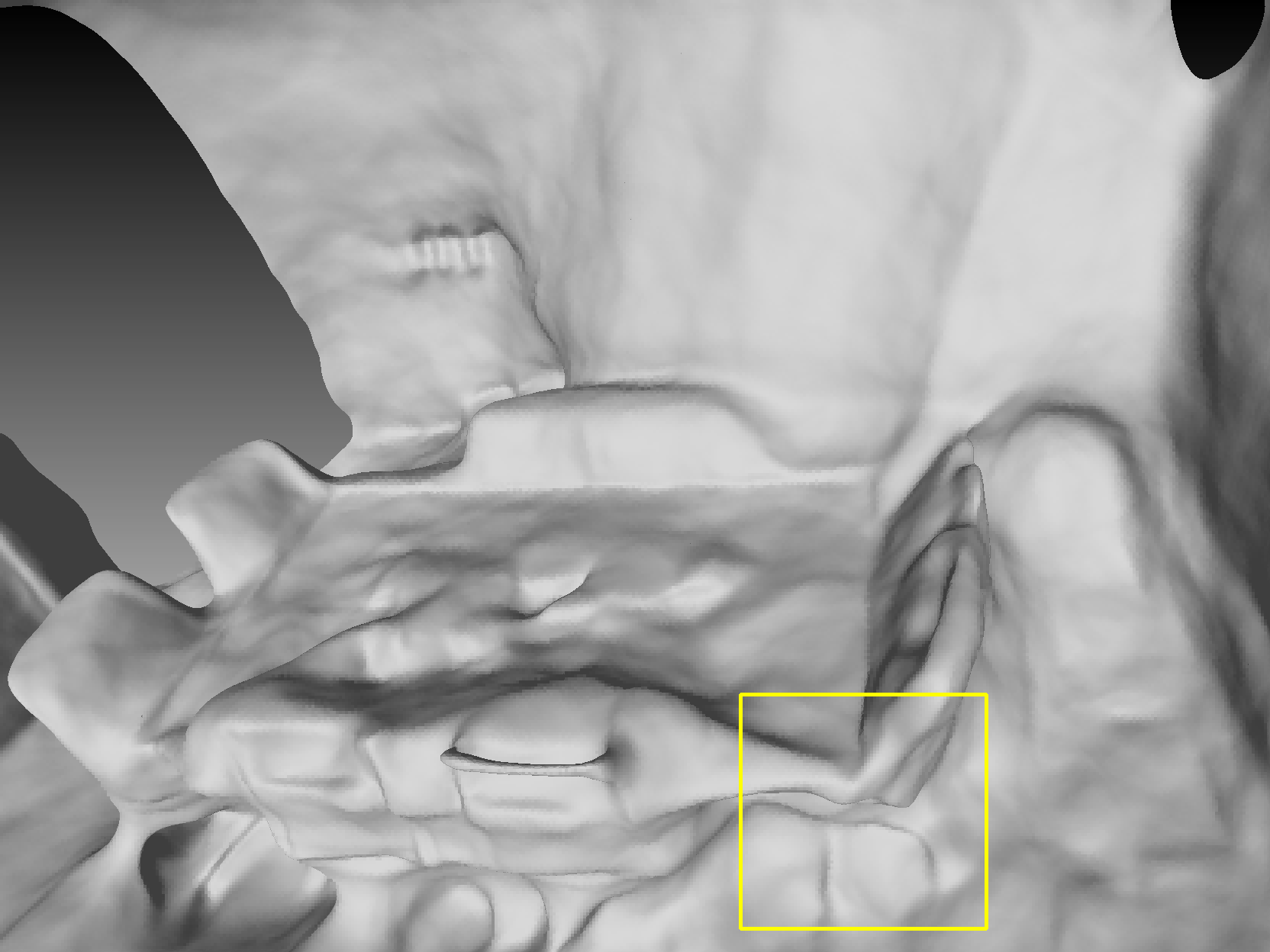}
            \end{minipage}
            \begin{minipage}[t]{0.24\linewidth}
                \centering
                \includegraphics[width=1\linewidth]{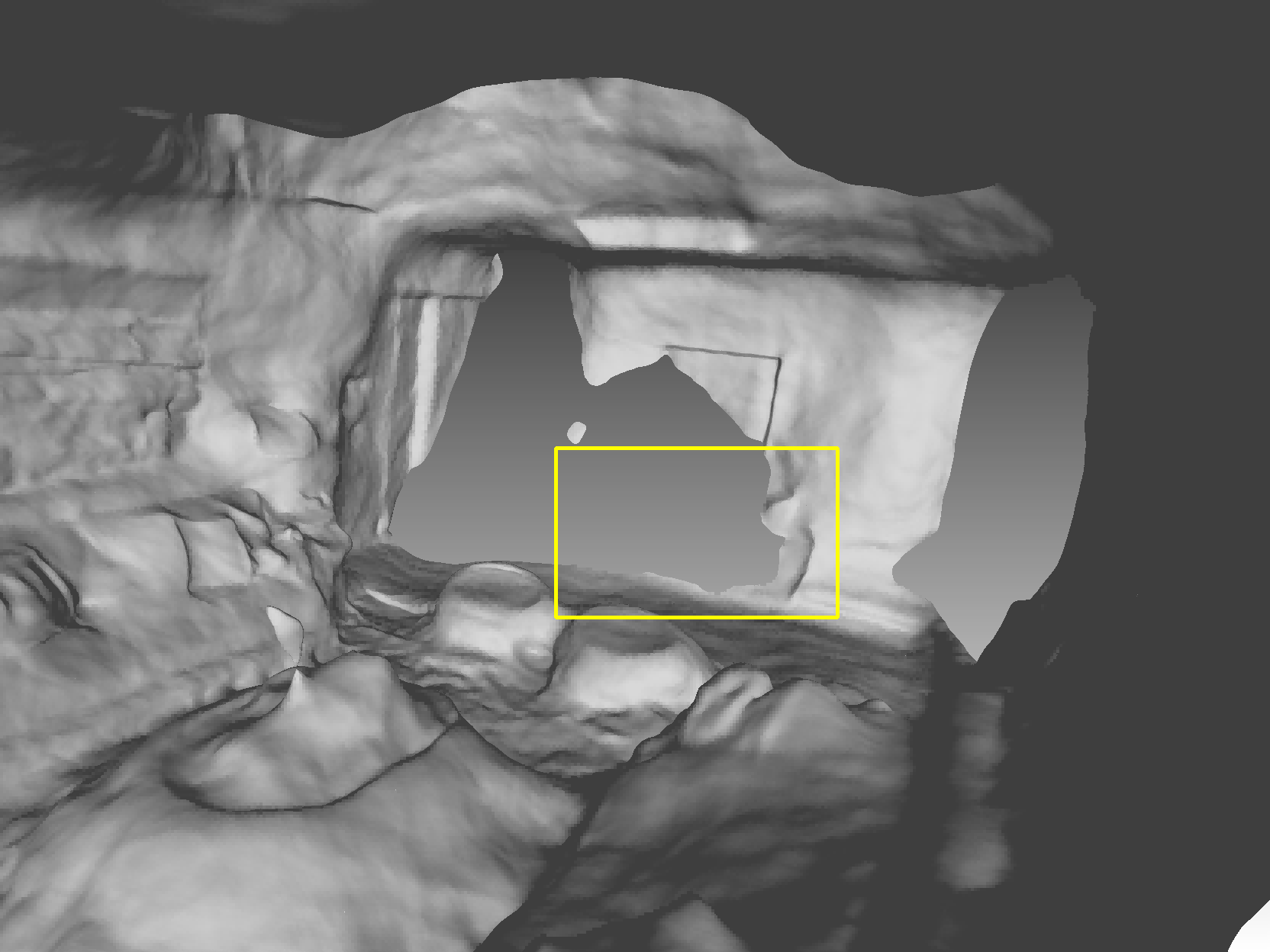}
            \end{minipage}
        \end{subfigure}

	\begin{subfigure}{\linewidth}
            \rotatebox[origin=c]{90}{\normalsize{VolSDF}\hspace{-2.2cm}}
            \begin{minipage}[t]{0.24\linewidth}
                \centering
                \includegraphics[width=1\linewidth]{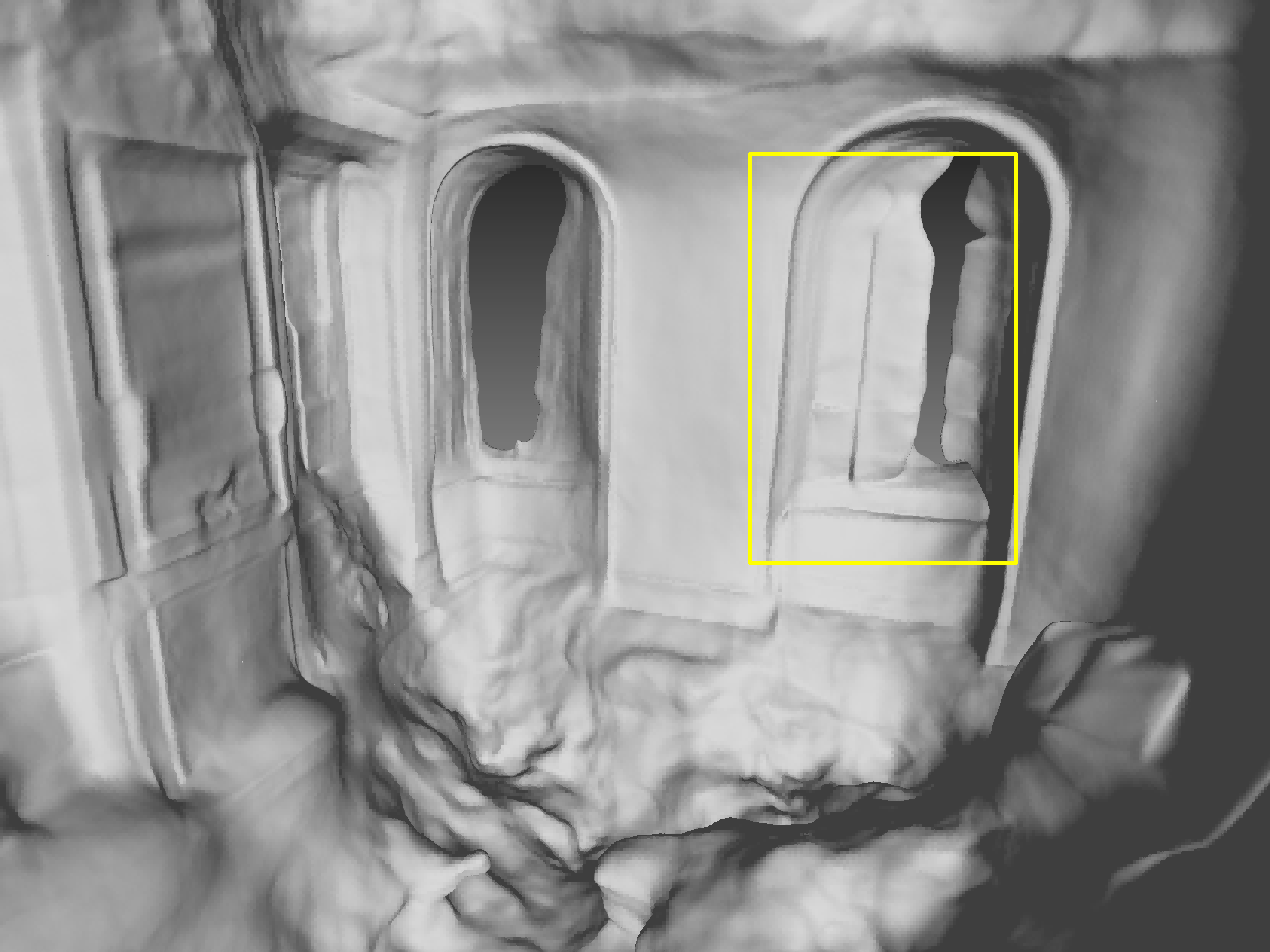}
            \end{minipage}
            \begin{minipage}[t]{0.24\linewidth}
                \centering
                \includegraphics[width=1\linewidth]{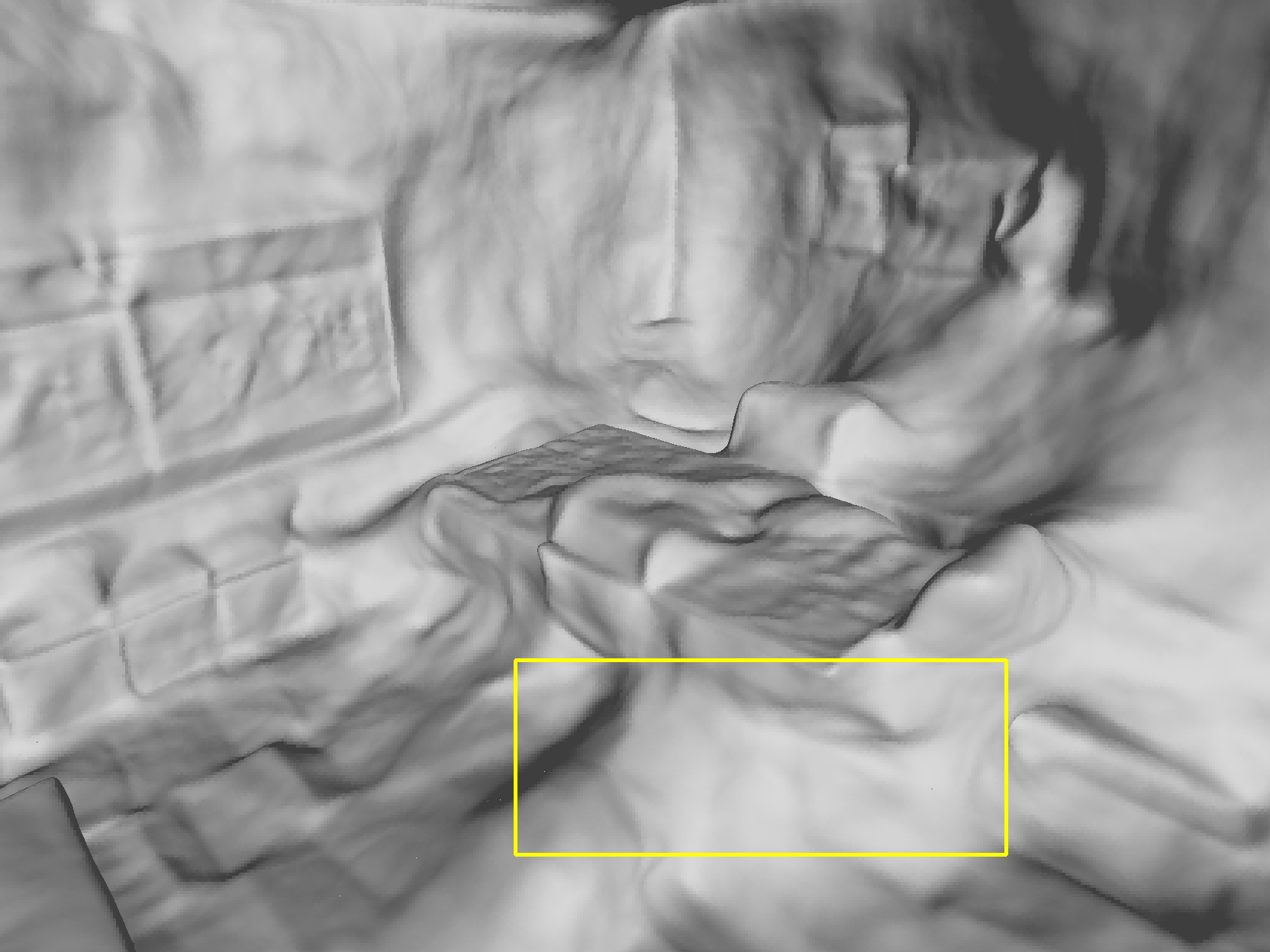}
            \end{minipage}
            \begin{minipage}[t]{0.24\linewidth}
                \centering
                \includegraphics[width=1\linewidth]{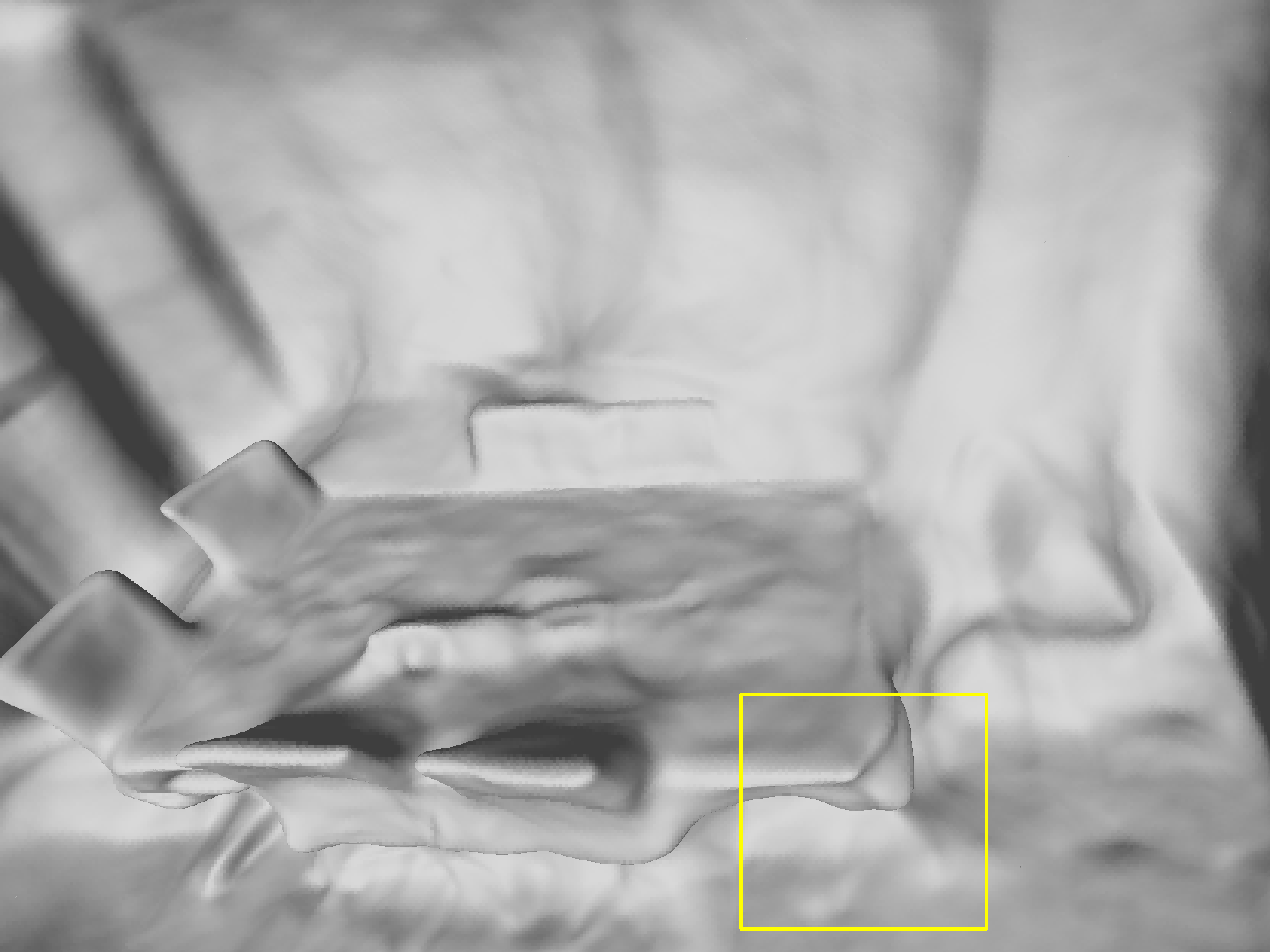}
            \end{minipage}
            \begin{minipage}[t]{0.24\linewidth}
                \centering
                \includegraphics[width=1\linewidth]{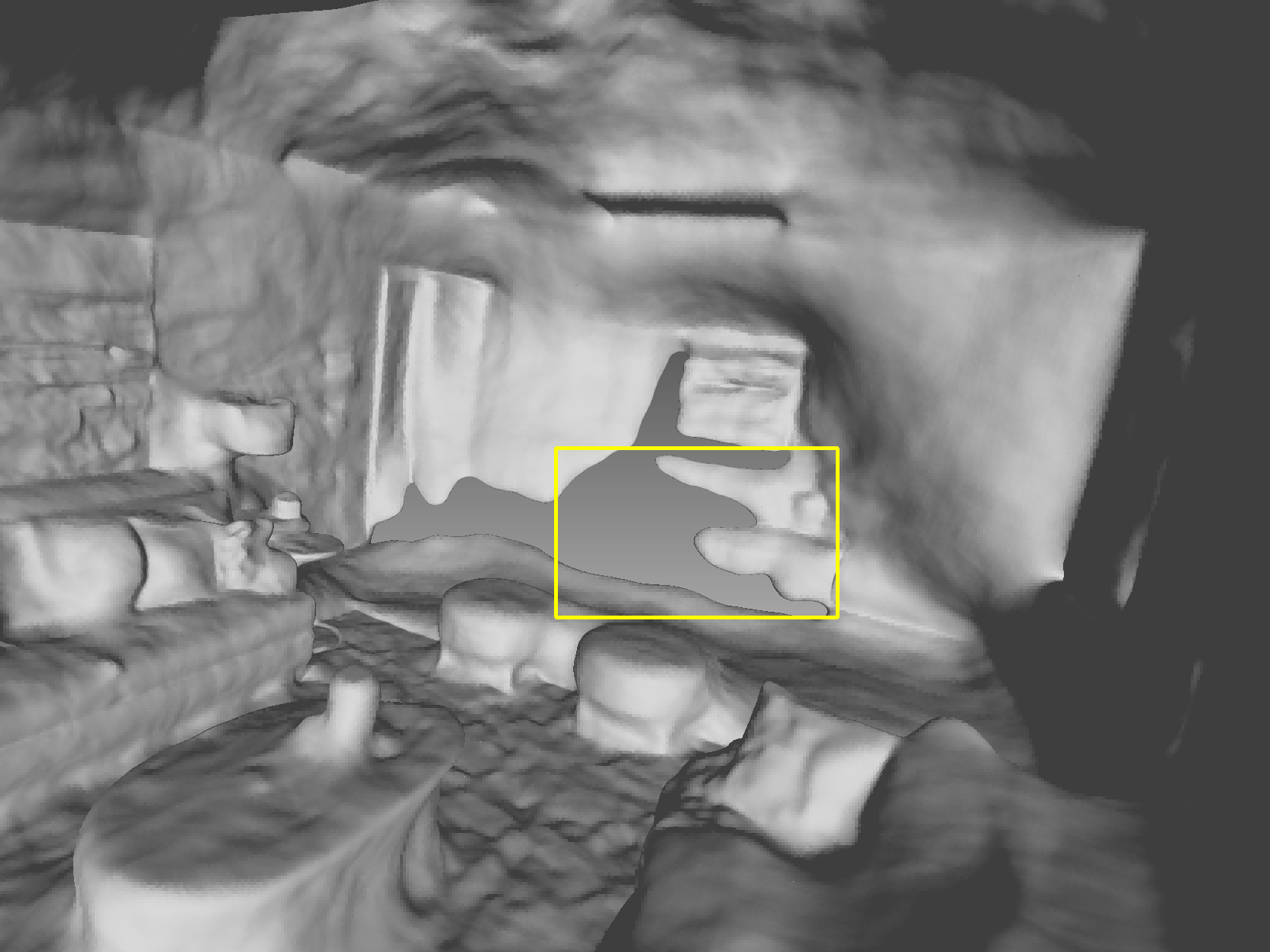}
            \end{minipage}
        \end{subfigure}

	\begin{subfigure}{\linewidth}
            \rotatebox[origin=c]{90}{\normalsize{NeuralRGBD}\hspace{-2.2cm}}
            \begin{minipage}[t]{0.24\linewidth}
                \centering
                \includegraphics[width=1\linewidth]{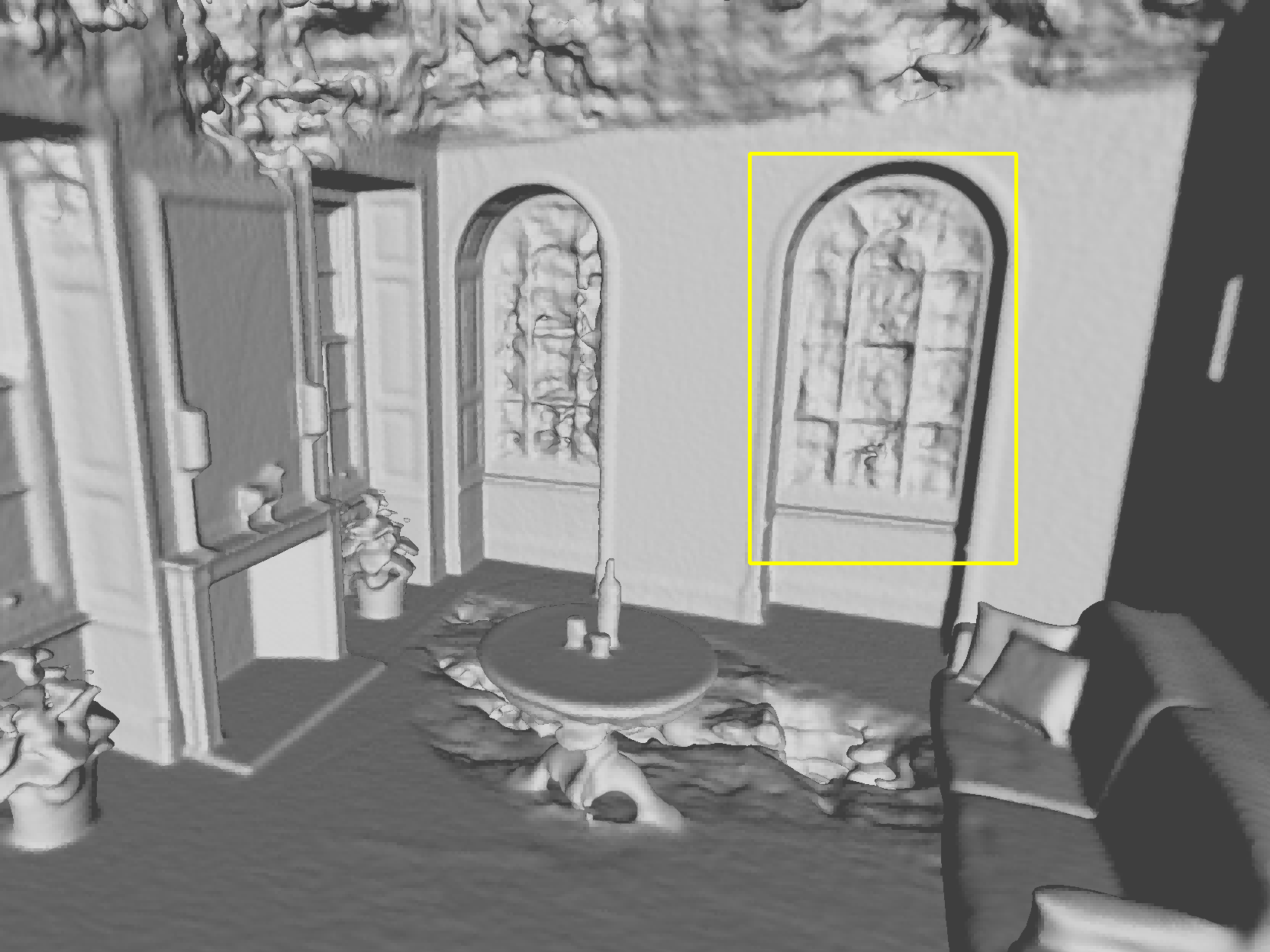}
            \end{minipage}
            \begin{minipage}[t]{0.24\linewidth}
                \centering
                \includegraphics[width=1\linewidth]{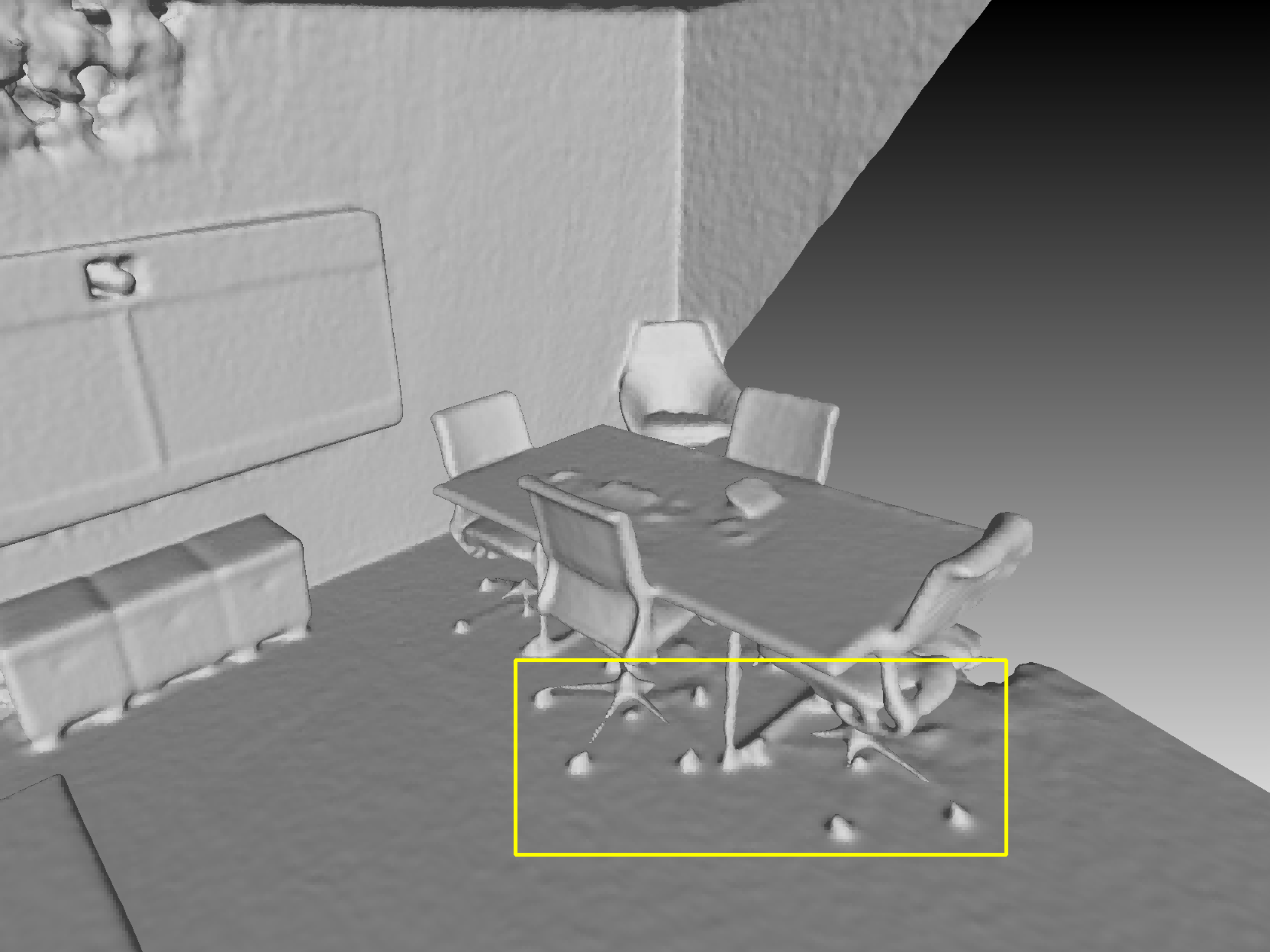}
            \end{minipage}
            \begin{minipage}[t]{0.24\linewidth}
                \centering
                \includegraphics[width=1\linewidth]{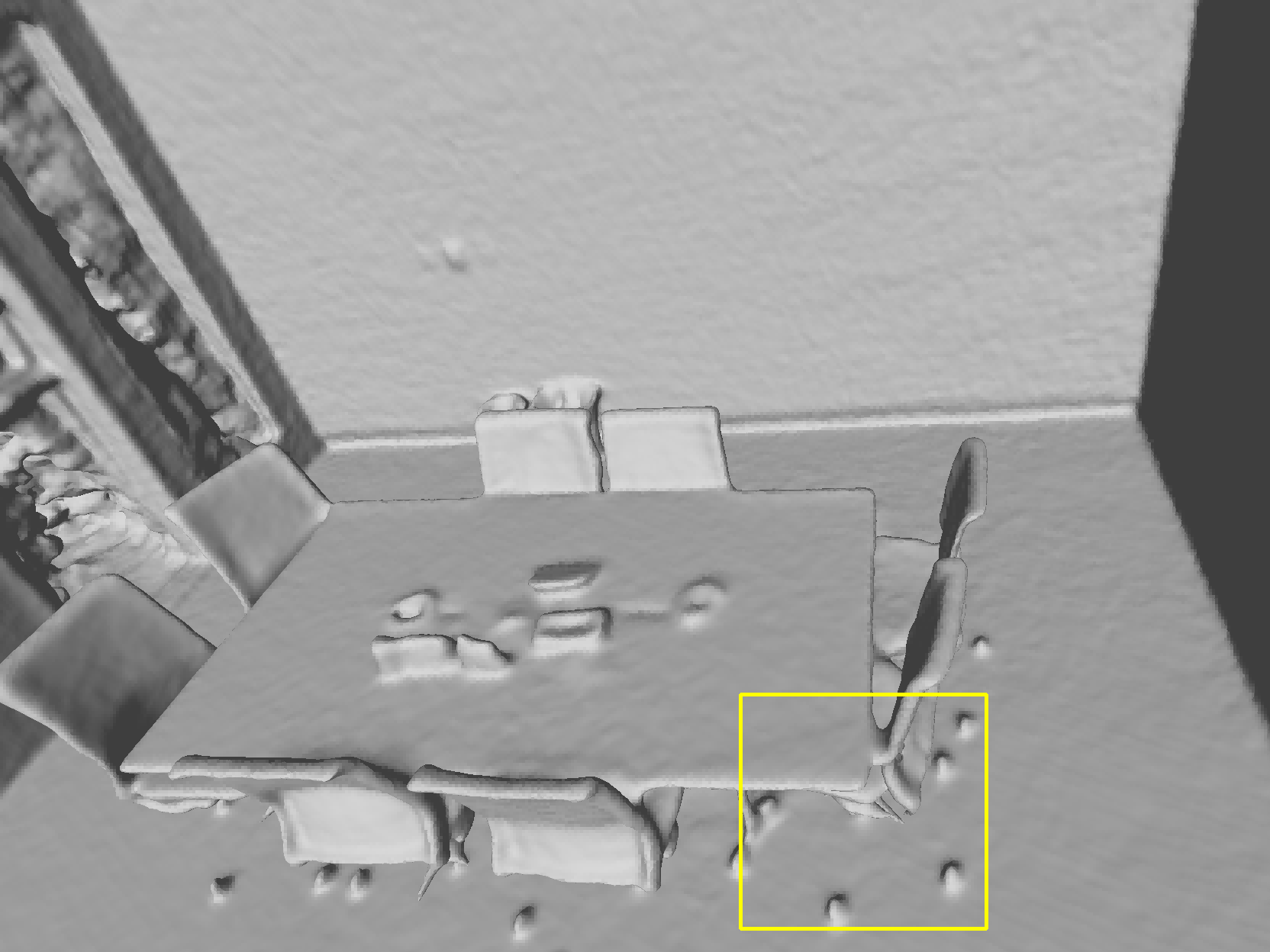}
            \end{minipage}
            \begin{minipage}[t]{0.24\linewidth}
                \centering
                \includegraphics[width=1\linewidth]{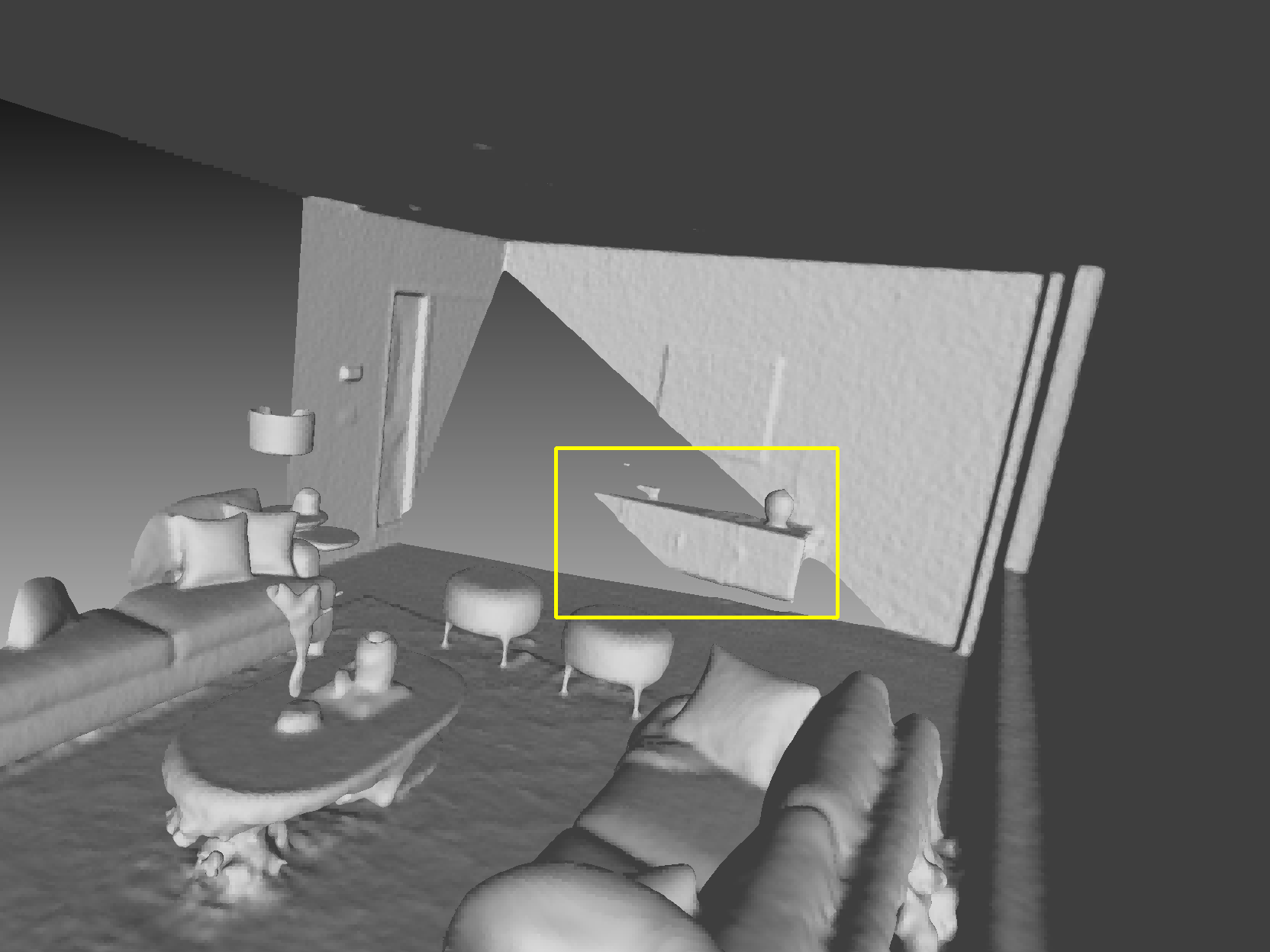}
            \end{minipage}
        \end{subfigure}

	\begin{subfigure}{\linewidth}
            \rotatebox[origin=c]{90}{\normalsize{Go-Surf}\hspace{-2.2cm}}
            \begin{minipage}[t]{0.24\linewidth}
                \centering
                \includegraphics[width=1\linewidth]{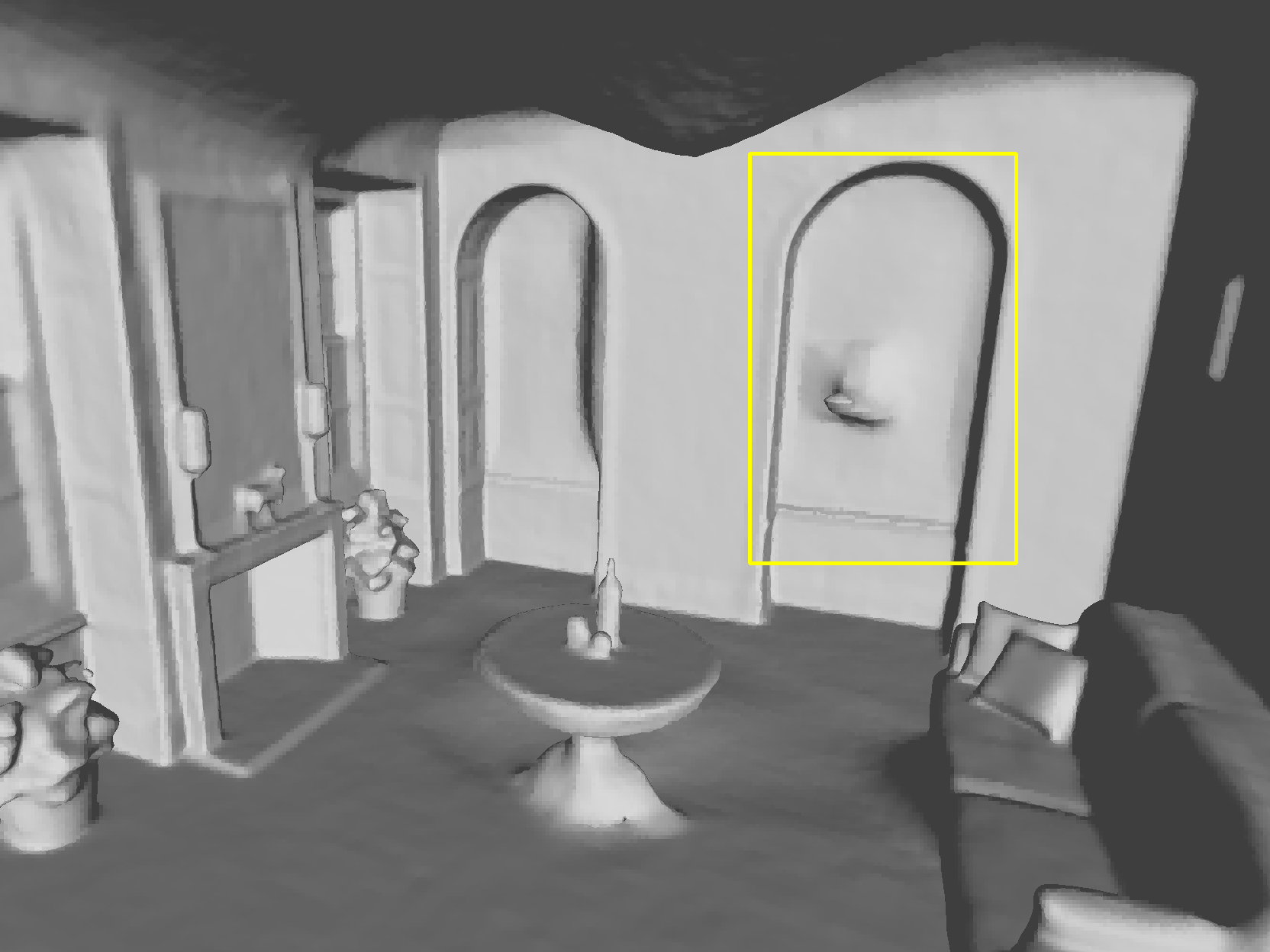}
            \end{minipage}
            \begin{minipage}[t]{0.24\linewidth}
                \centering
                \includegraphics[width=1\linewidth]{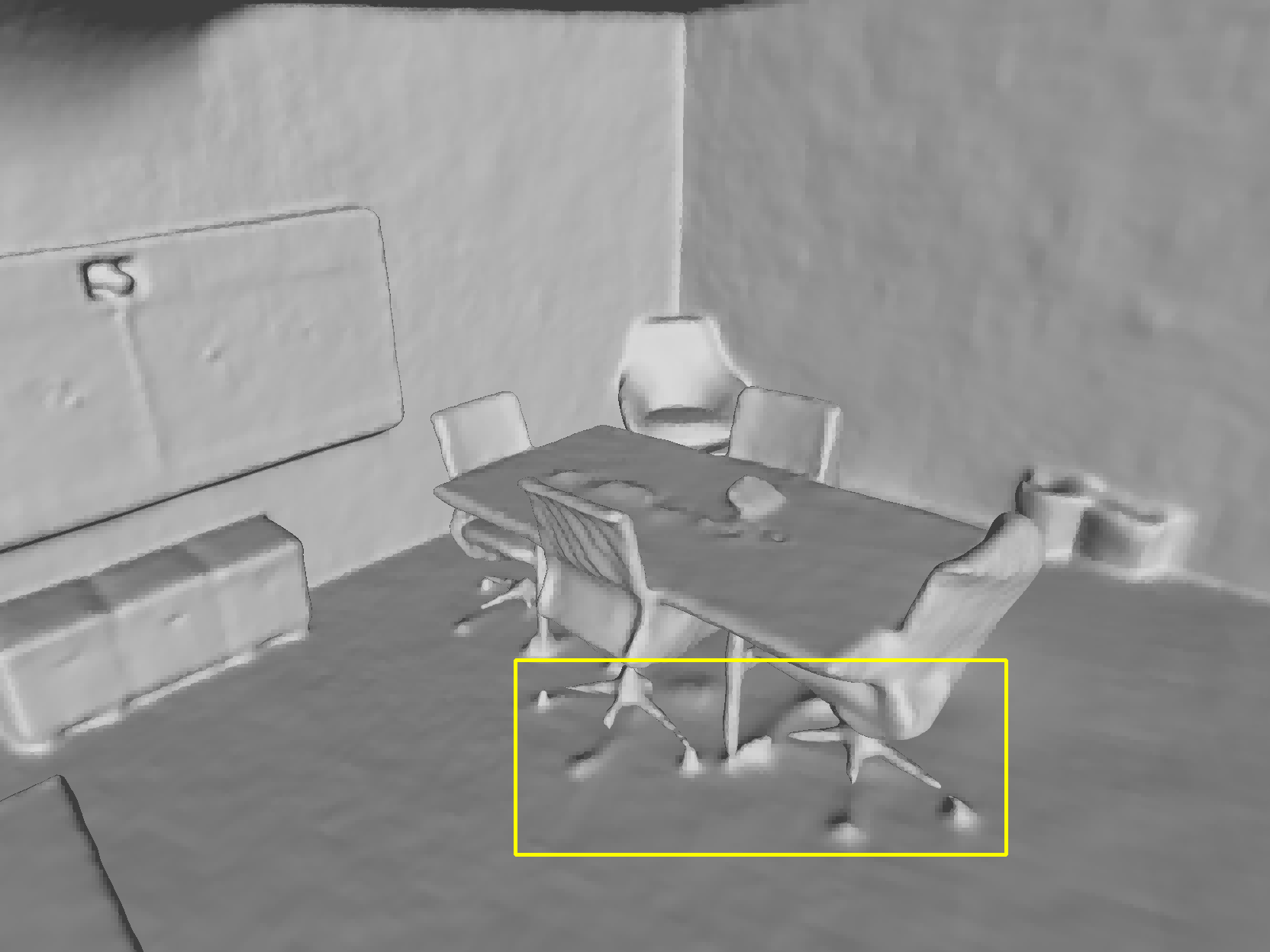}
            \end{minipage}
            \begin{minipage}[t]{0.24\linewidth}
                \centering
                \includegraphics[width=1\linewidth]{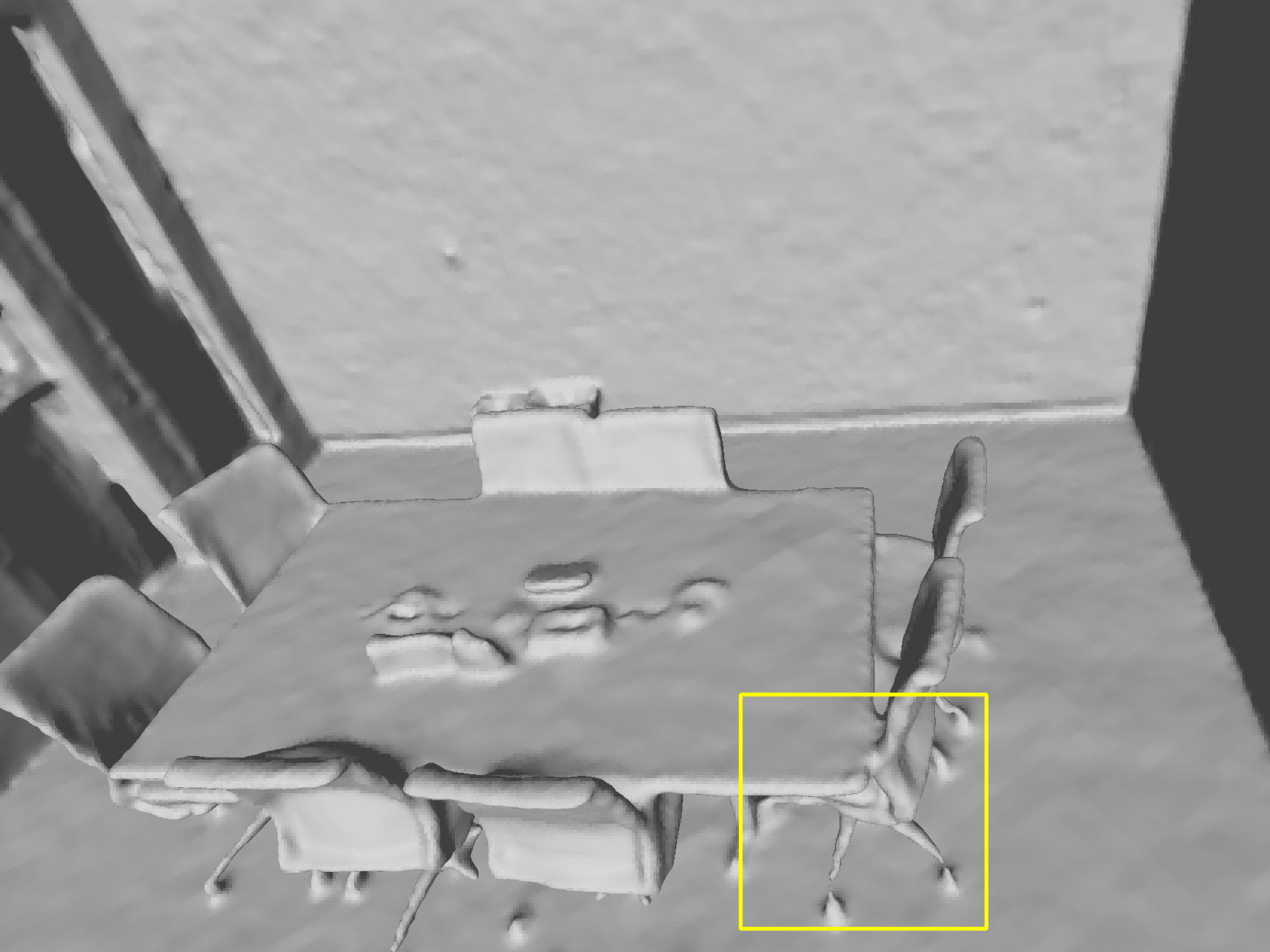}
            \end{minipage}
            \begin{minipage}[t]{0.24\linewidth}
                \centering
                \includegraphics[width=1\linewidth]{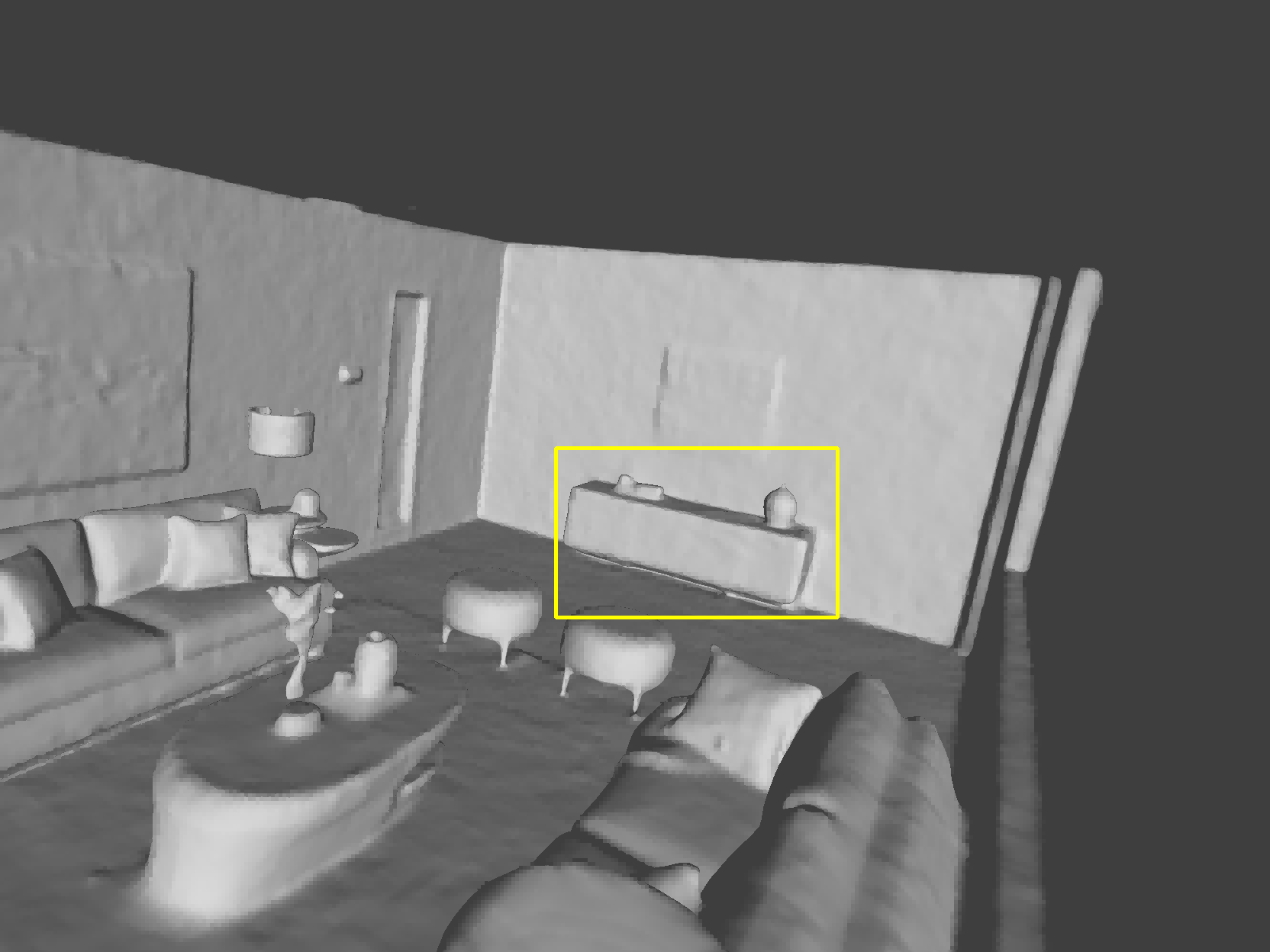}
            \end{minipage}
        \end{subfigure}

	\begin{subfigure}{\linewidth}
            \rotatebox[origin=c]{90}{\normalsize{Ours}\hspace{-2.2cm}}
            \begin{minipage}[t]{0.24\linewidth}
                \centering
                \includegraphics[width=1\linewidth]{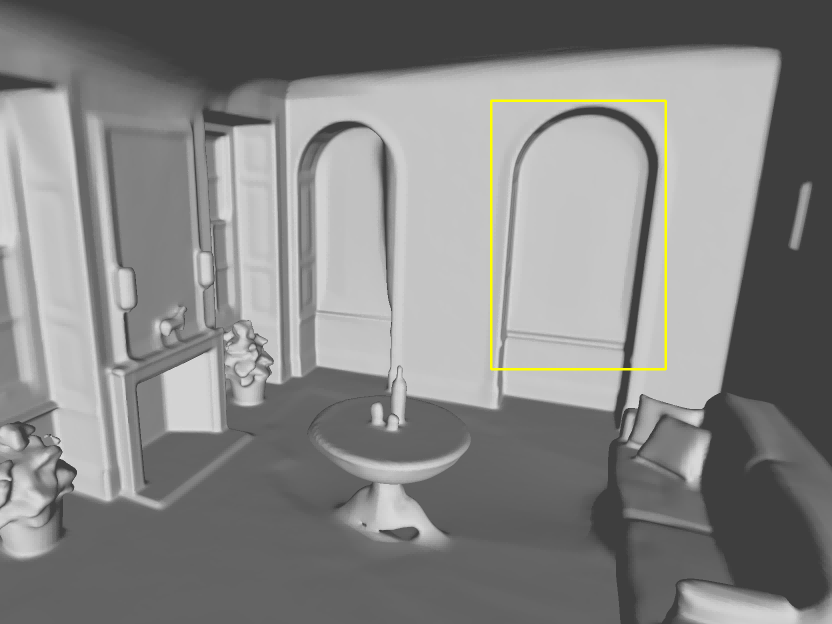}
            \end{minipage}
            \begin{minipage}[t]{0.24\linewidth}
                \centering
                \includegraphics[width=1\linewidth]{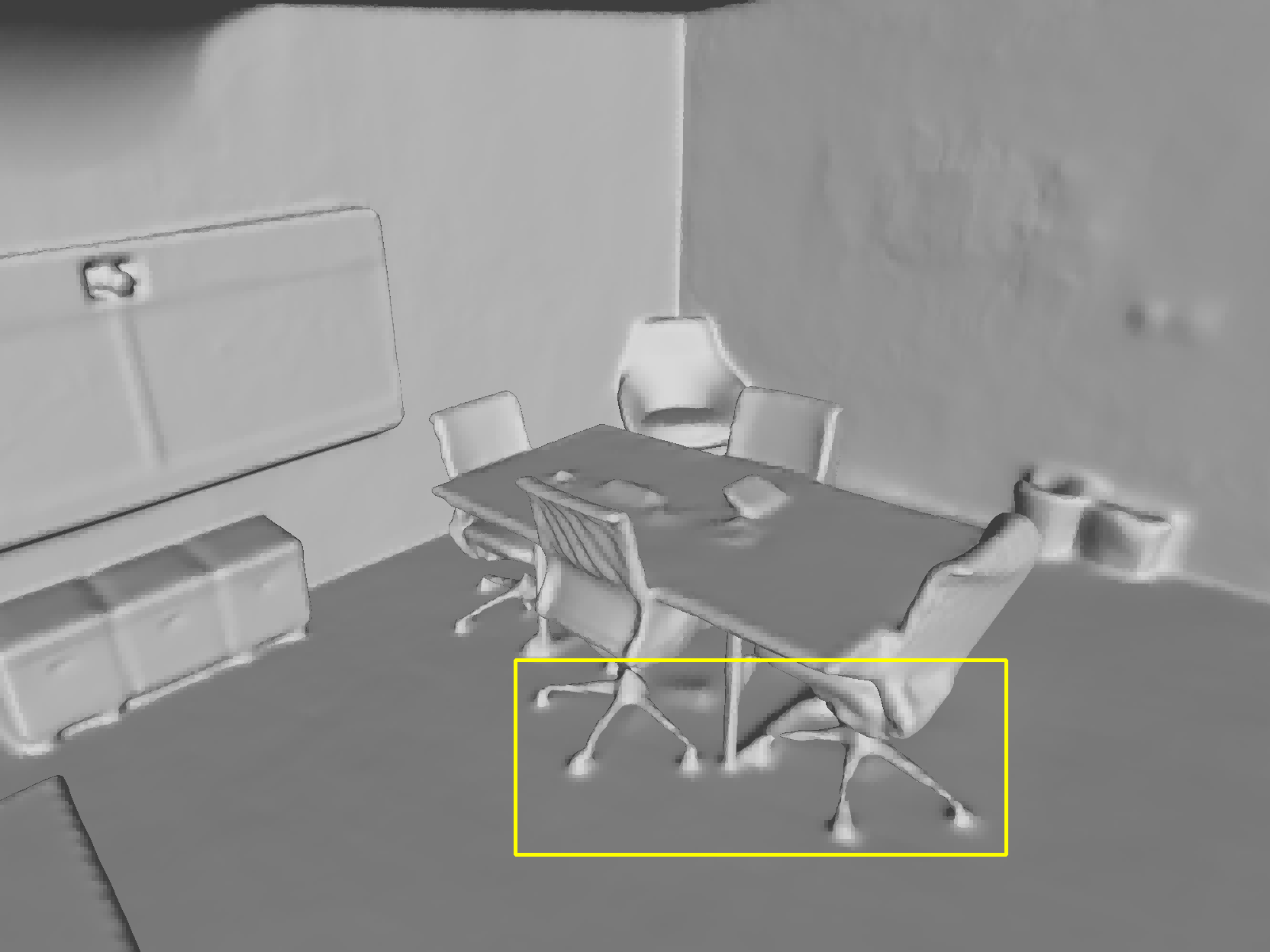}
            \end{minipage}
            \begin{minipage}[t]{0.24\linewidth}
                \centering
                \includegraphics[width=1\linewidth]{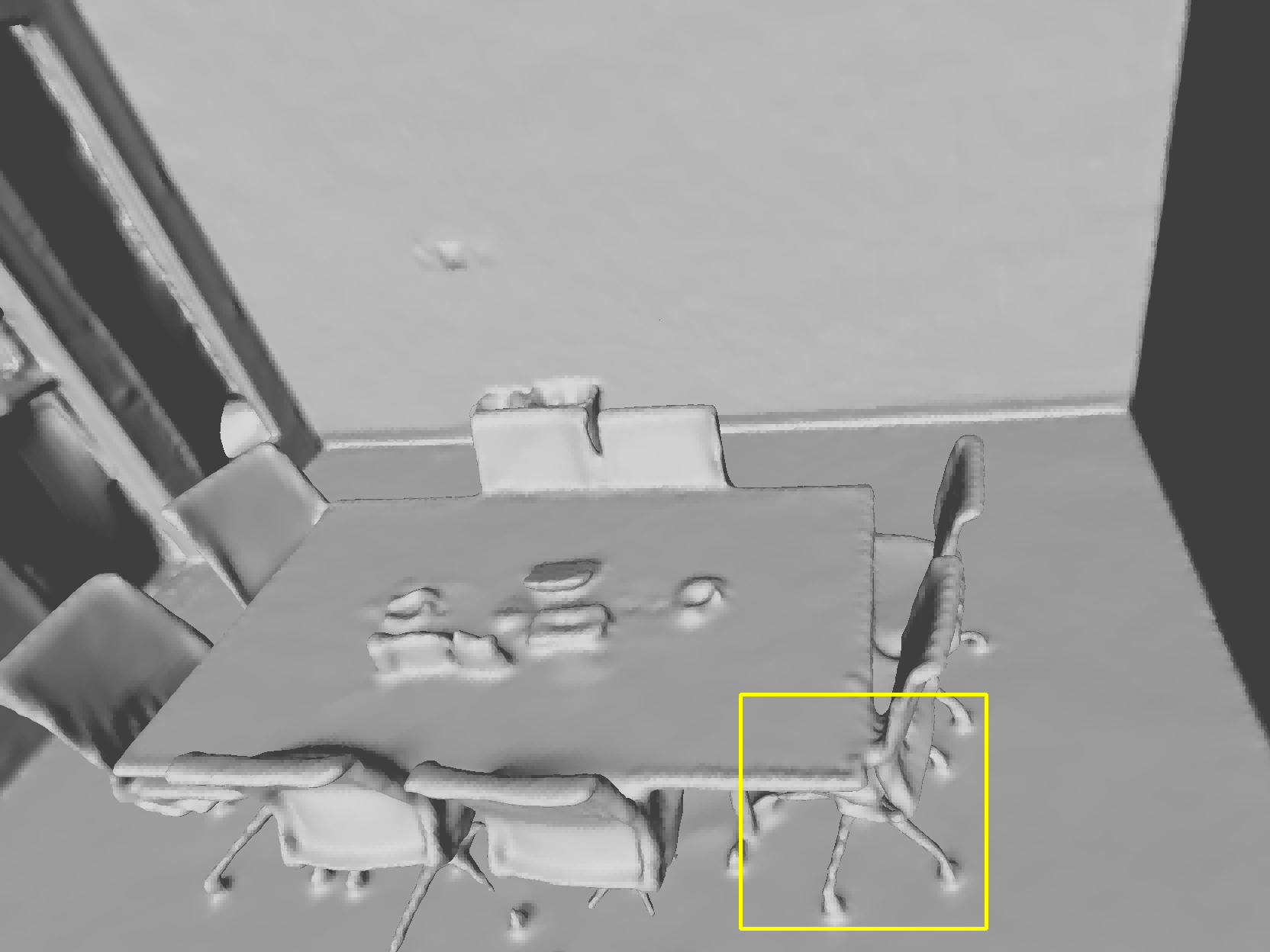}
            \end{minipage}
            \begin{minipage}[t]{0.24\linewidth}
                \centering
                \includegraphics[width=1\linewidth]{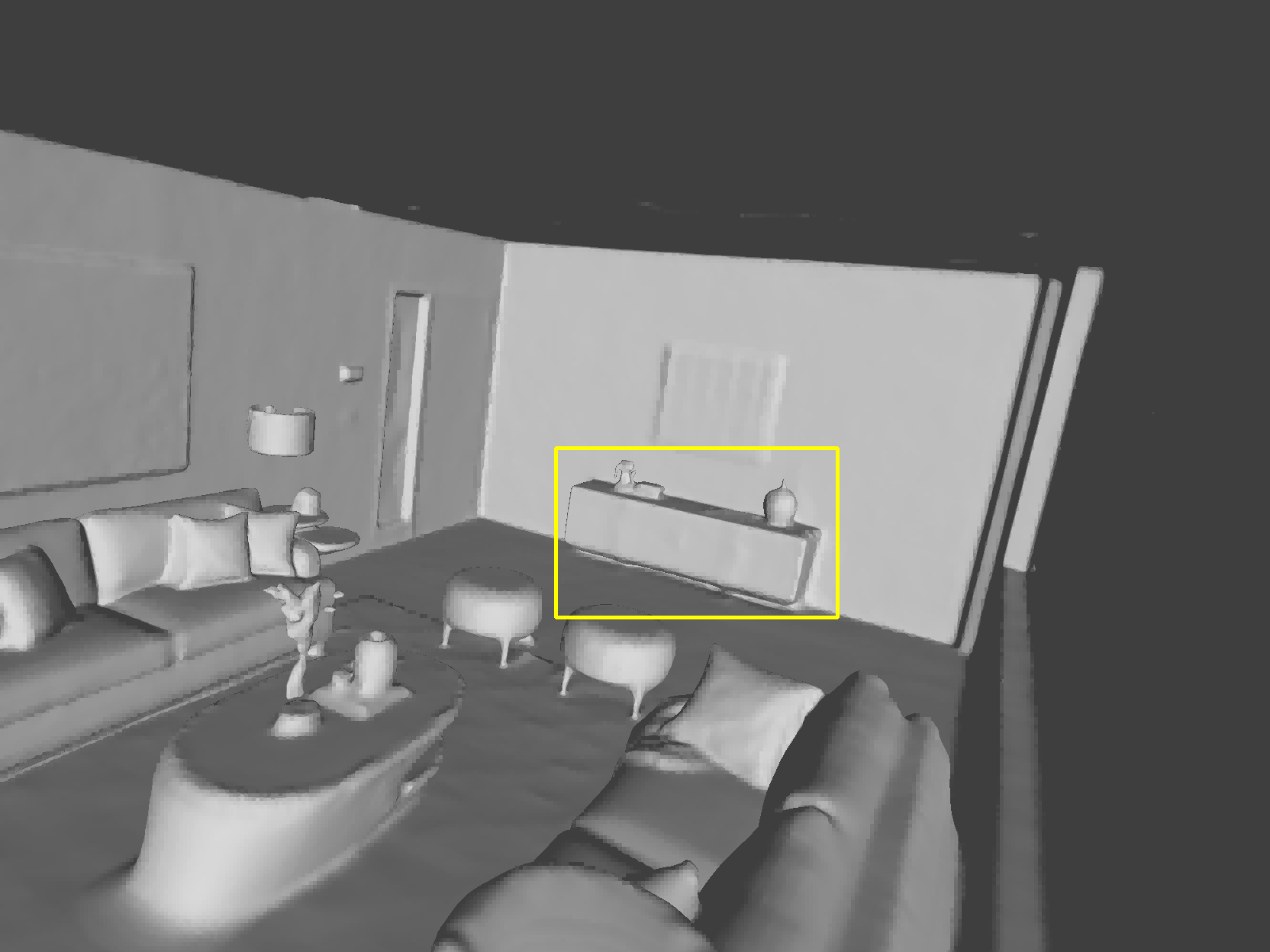}
            \end{minipage}
        \end{subfigure}

	\begin{subfigure}{\linewidth}
            \rotatebox[origin=c]{90}{\normalsize{GT}\hspace{-2.2cm}}
            \begin{minipage}[t]{0.24\linewidth}
                \centering
                \includegraphics[width=1\linewidth]{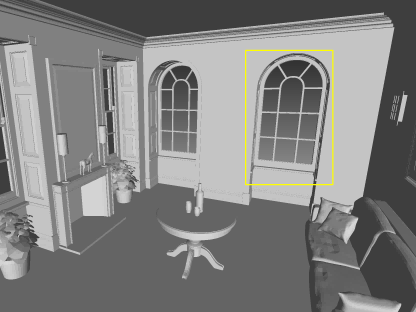}
            \end{minipage}
            \begin{minipage}[t]{0.24\linewidth}
                \centering
                \includegraphics[width=1\linewidth]{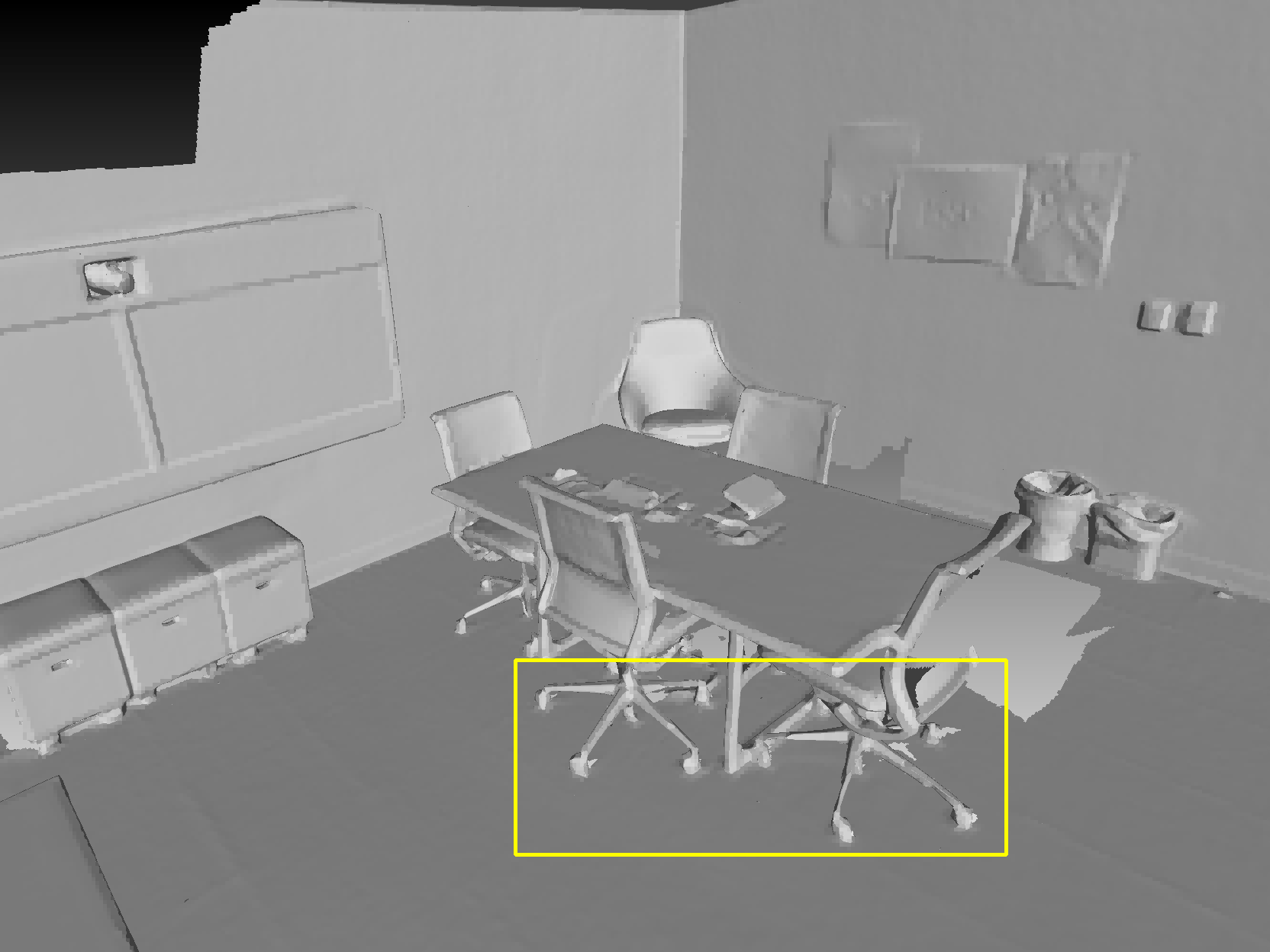}
            \end{minipage}
            \begin{minipage}[t]{0.24\linewidth}
                \centering
                \includegraphics[width=1\linewidth]{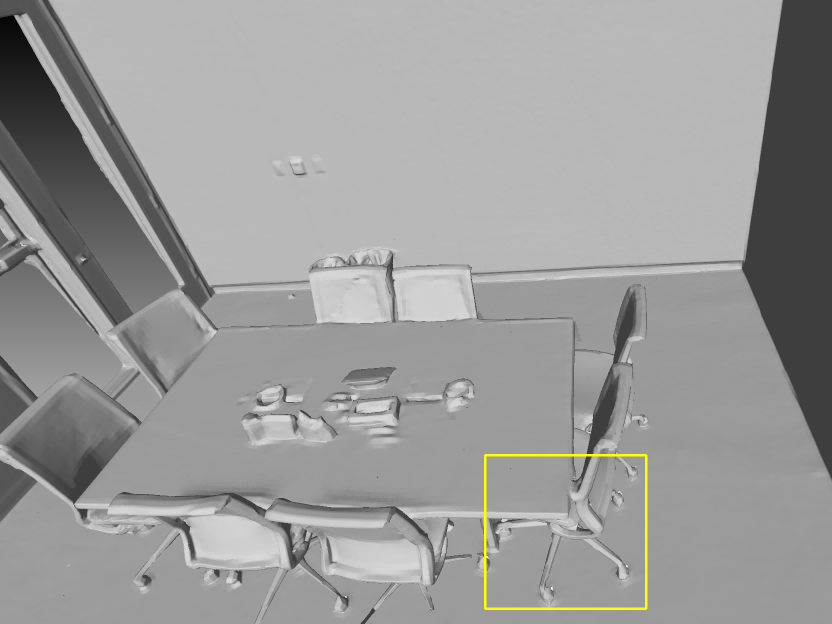}
            \end{minipage}
            \begin{minipage}[t]{0.24\linewidth}
                \centering
                \includegraphics[width=1\linewidth]{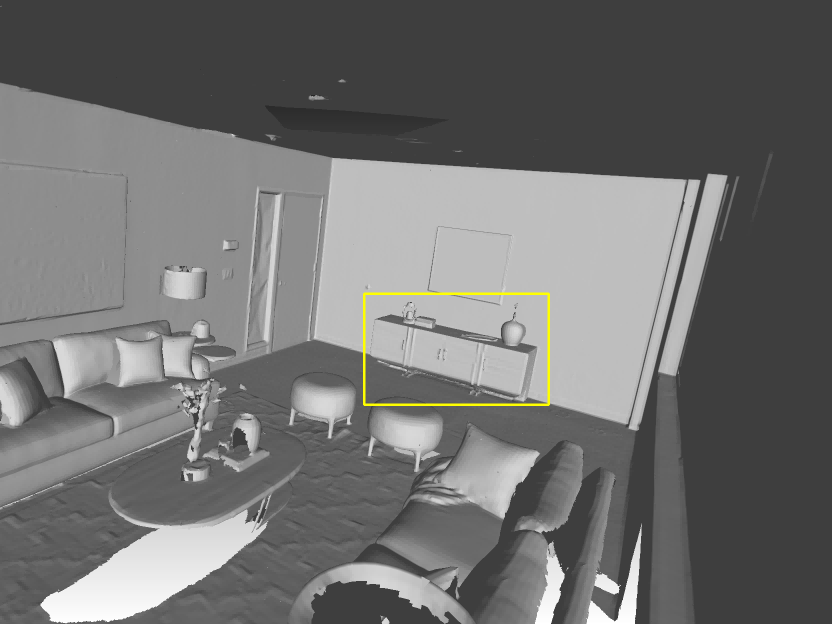}
            \end{minipage}
        \end{subfigure}

	\caption{We show additional reconstruction results on the scene \textit{Grey-white room}, \textit{Office2}, \textit{Office3} and \textit{Room0}. Our approach allows for rich details and smoother planes highlighted in the yellow box.}
	\label{fig:fig7}
\end{figure*}

\begin{figure*}[ht!]
	\centering
 
	\begin{subfigure}{\linewidth}
            \rotatebox[origin=c]{90}{\normalsize{Neus}\hspace{-2.2cm}}
            \begin{minipage}[t]{0.24\linewidth}
                \centering
                \includegraphics[width=1\linewidth]{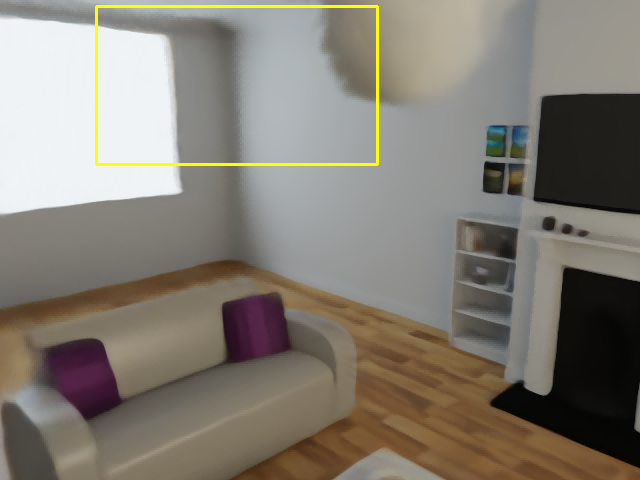}
            \end{minipage}
            \begin{minipage}[t]{0.24\linewidth}
                \centering
                \includegraphics[width=1\linewidth]{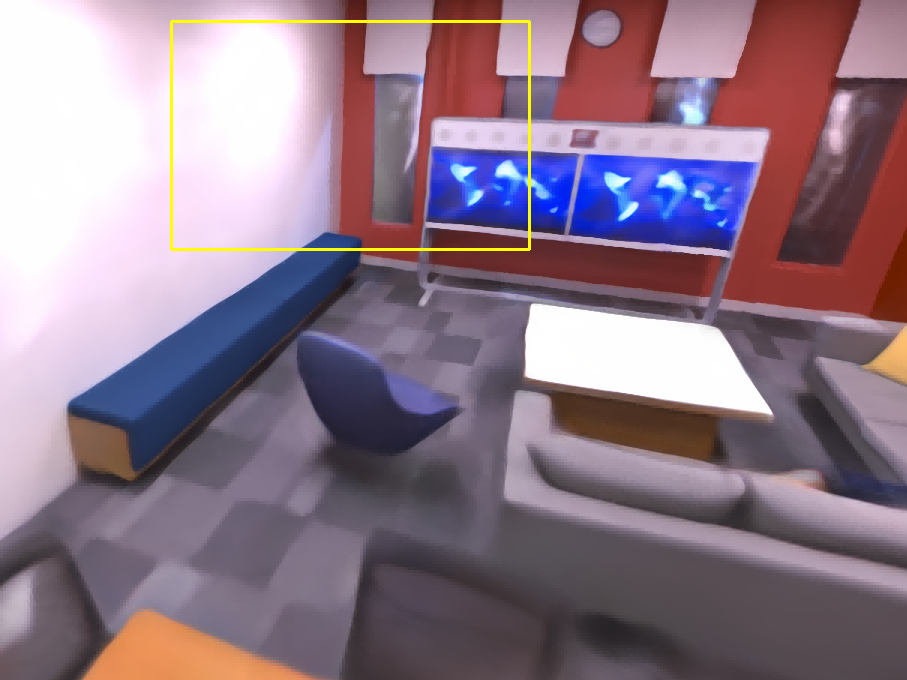}
            \end{minipage}
            \begin{minipage}[t]{0.24\linewidth}
                \centering
                \includegraphics[width=1\linewidth]{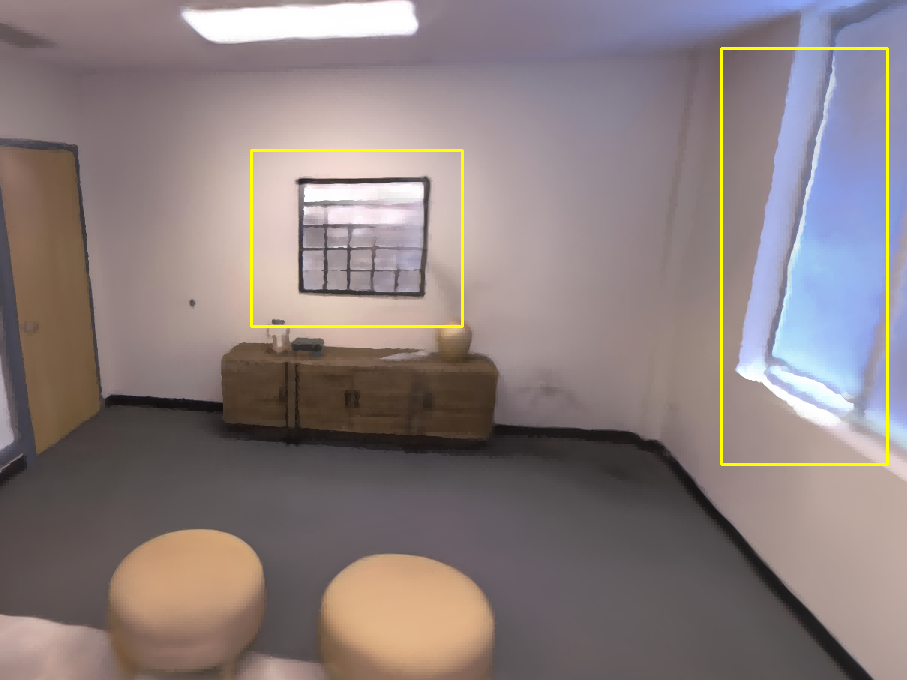}
            \end{minipage}
            \begin{minipage}[t]{0.24\linewidth}
                \centering
                \includegraphics[width=1\linewidth]{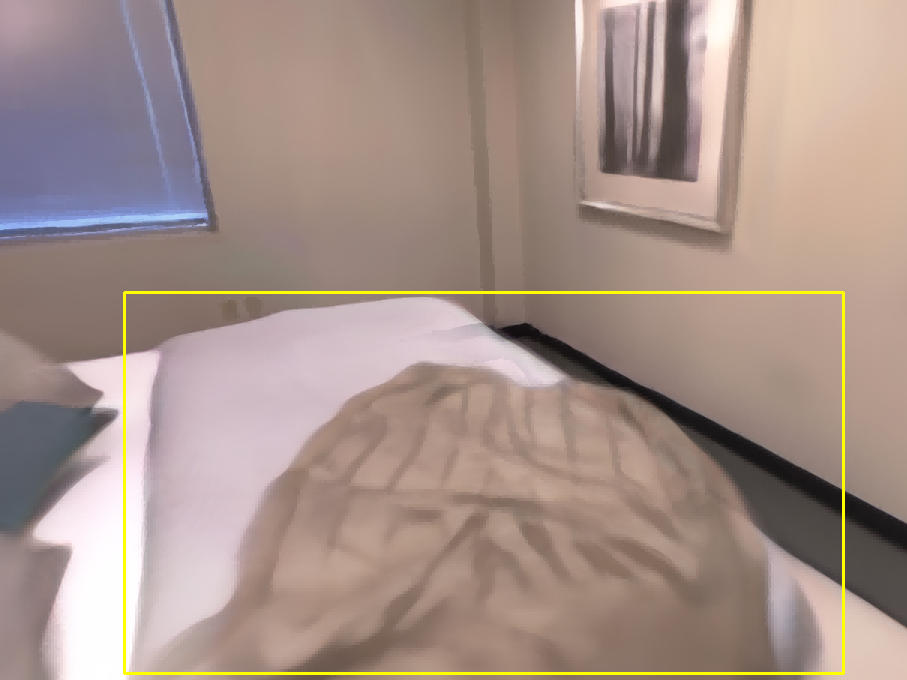}
            \end{minipage}
        \end{subfigure}
        
	% \begin{subfigure}{\linewidth}
 %            \rotatebox[origin=c]{90}{\normalsize{VolSDF}\hspace{-2.2cm}}
 %            \begin{minipage}[t]{0.24\linewidth}
 %                \centering
 %                \includegraphics[width=1\linewidth]{figure_supple/figure2/2-1.png}
 %            \end{minipage}
 %            \begin{minipage}[t]{0.24\linewidth}
 %                \centering
 %                \includegraphics[width=1\linewidth]{figure_supple/figure2/2-2.png}
 %            \end{minipage}
 %            \begin{minipage}[t]{0.24\linewidth}
 %                \centering
 %                \includegraphics[width=1\linewidth]{figure_supple/figure2/2-3.png}
 %            \end{minipage}
 %            \begin{minipage}[t]{0.24\linewidth}
 %                \centering
 %                \includegraphics[width=1\linewidth]{figure_supple/figure2/2-4.png}
 %            \end{minipage}
 %        \end{subfigure}

	\begin{subfigure}{\linewidth}
            \rotatebox[origin=c]{90}{\normalsize{InstantNGP}\hspace{-2.2cm}}
            \begin{minipage}[t]{0.24\linewidth}
                \centering
                \includegraphics[width=1\linewidth]{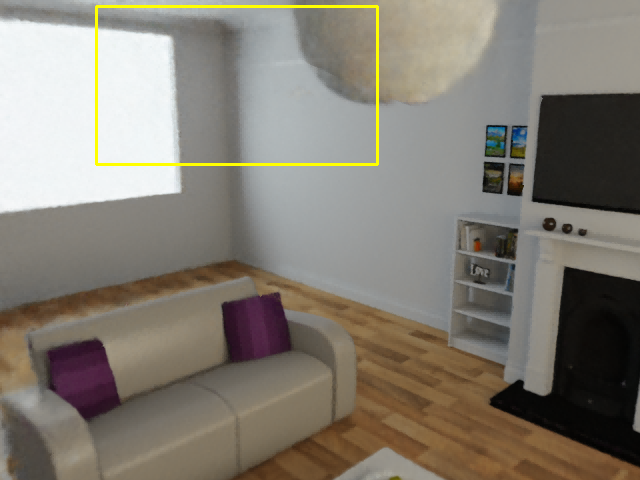}
            \end{minipage}
            \begin{minipage}[t]{0.24\linewidth}
                \centering
                \includegraphics[width=1\linewidth]{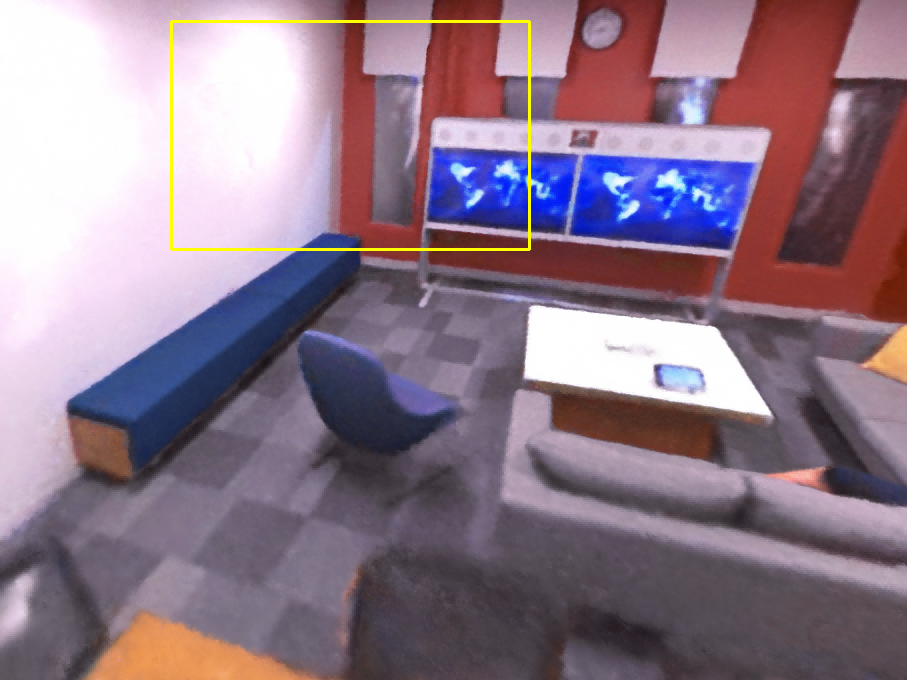}
            \end{minipage}
            \begin{minipage}[t]{0.24\linewidth}
                \centering
                \includegraphics[width=1\linewidth]{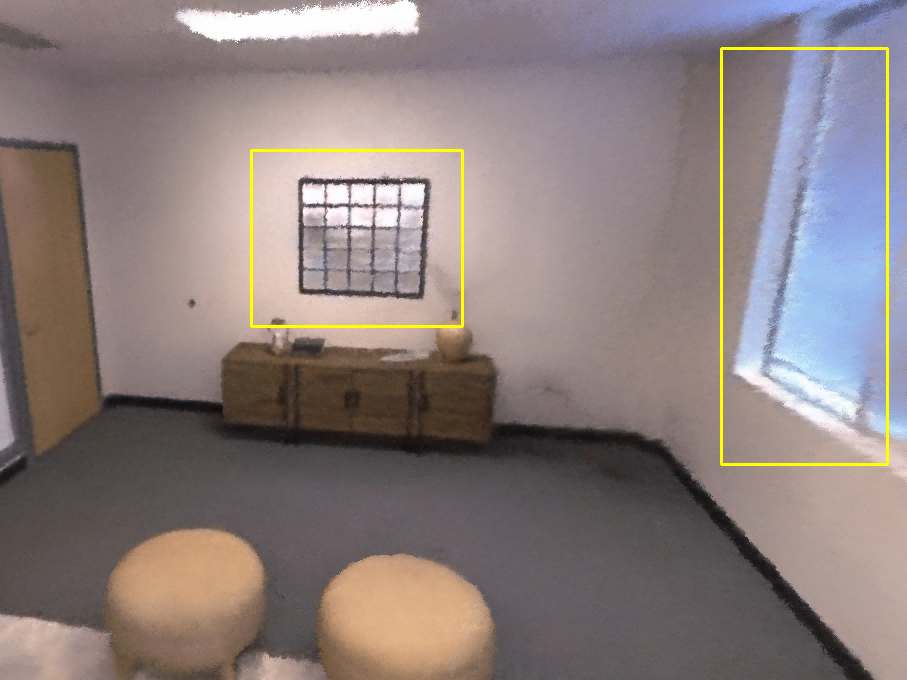}
            \end{minipage}
            \begin{minipage}[t]{0.24\linewidth}
                \centering
                \includegraphics[width=1\linewidth]{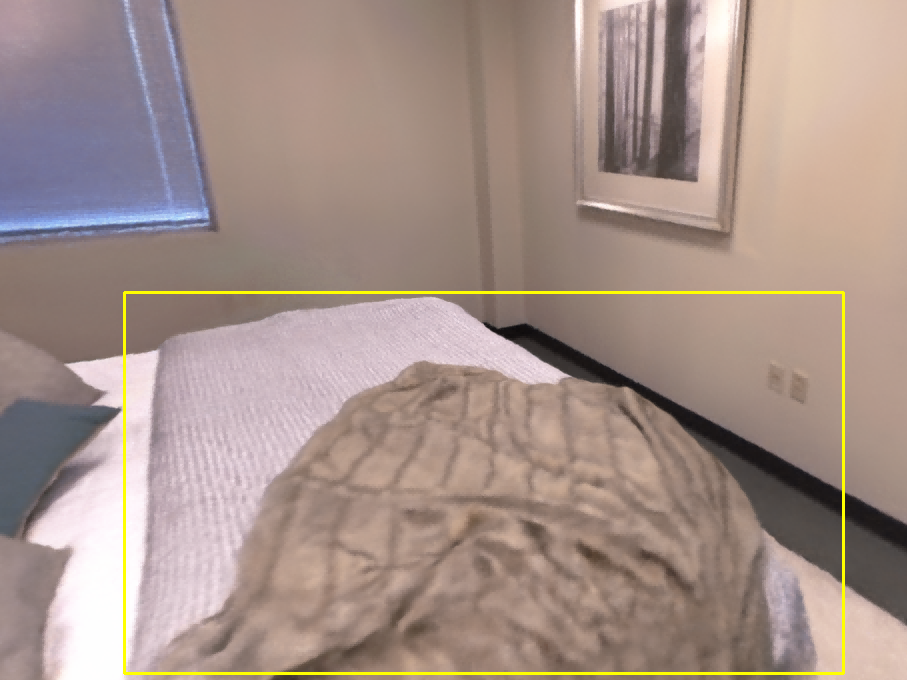}
            \end{minipage}
        \end{subfigure}

	\begin{subfigure}{\linewidth}
            \rotatebox[origin=c]{90}{\normalsize{DVGO}\hspace{-2.2cm}}
            \begin{minipage}[t]{0.24\linewidth}
                \centering
                \includegraphics[width=1\linewidth]{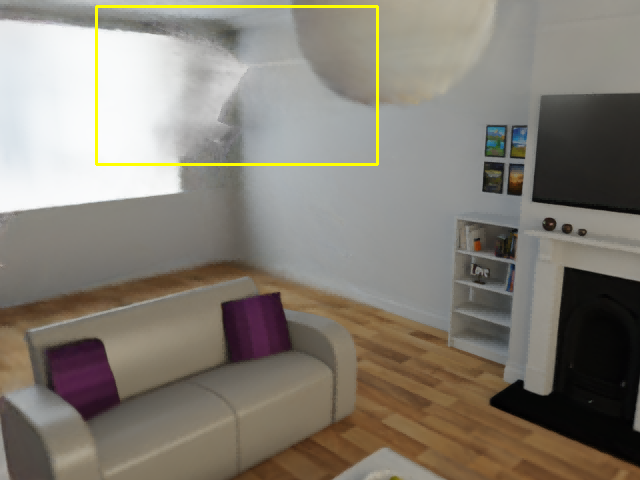}
            \end{minipage}
            \begin{minipage}[t]{0.24\linewidth}
                \centering
                \includegraphics[width=1\linewidth]{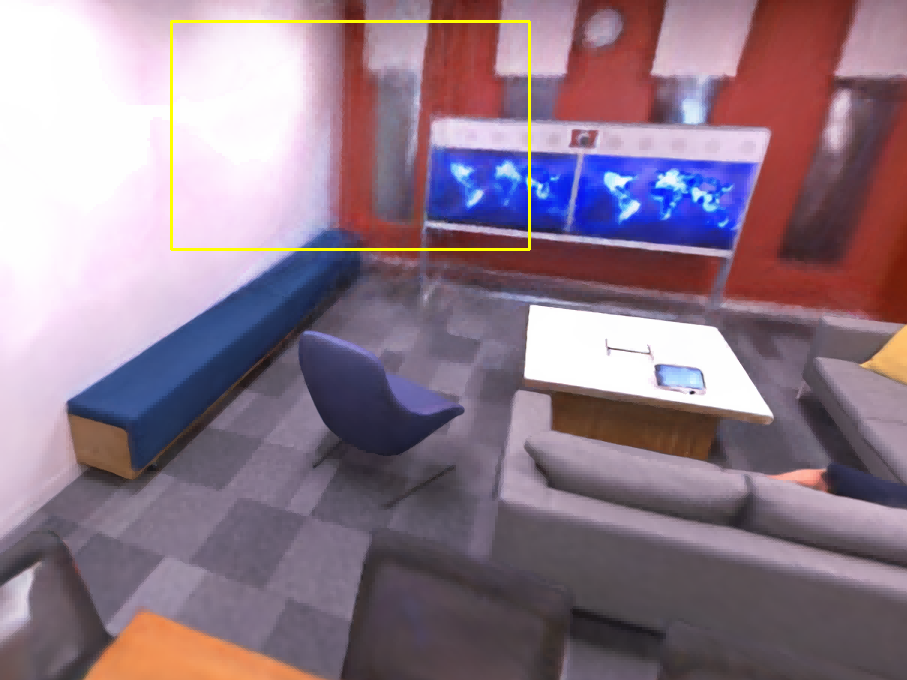}
            \end{minipage}
            \begin{minipage}[t]{0.24\linewidth}
                \centering
                \includegraphics[width=1\linewidth]{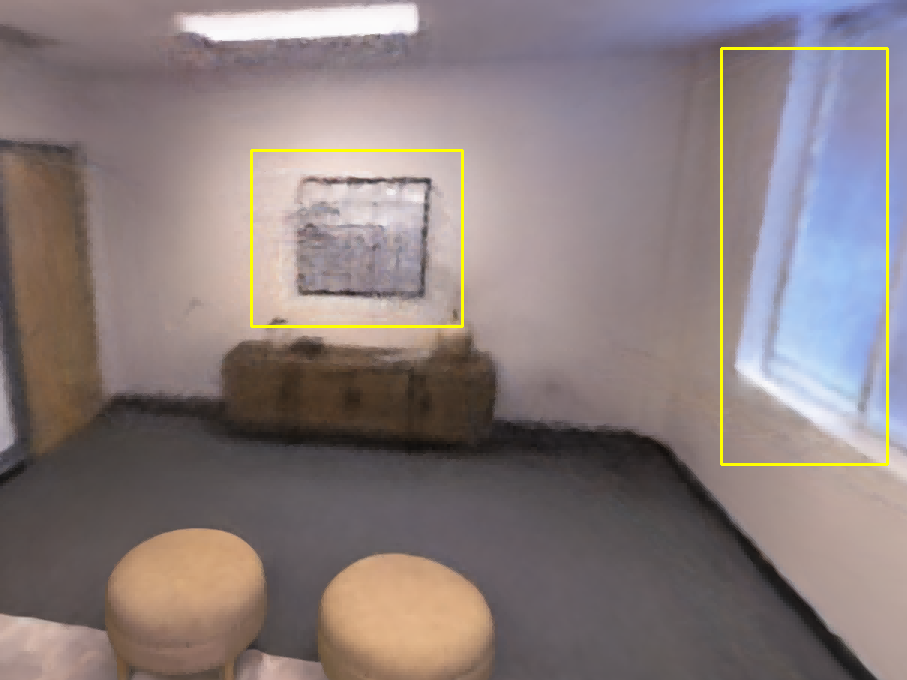}
            \end{minipage}
            \begin{minipage}[t]{0.24\linewidth}
                \centering
                \includegraphics[width=1\linewidth]{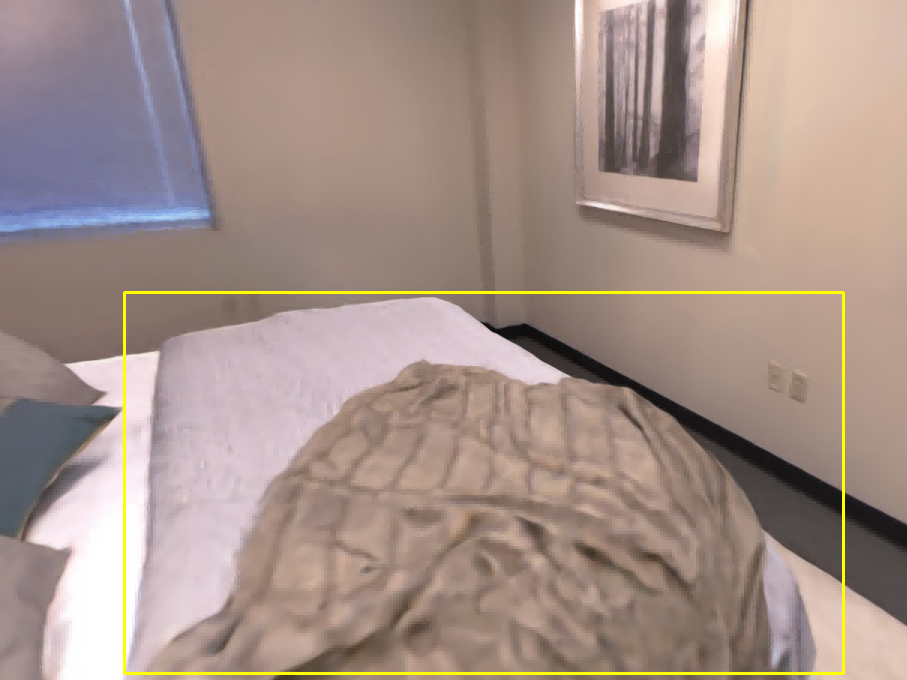}
            \end{minipage}
        \end{subfigure}

	\begin{subfigure}{\linewidth}
            \rotatebox[origin=c]{90}{\normalsize{Go-Surf}\hspace{-2.2cm}}
            \begin{minipage}[t]{0.24\linewidth}
                \centering
                \includegraphics[width=1\linewidth]{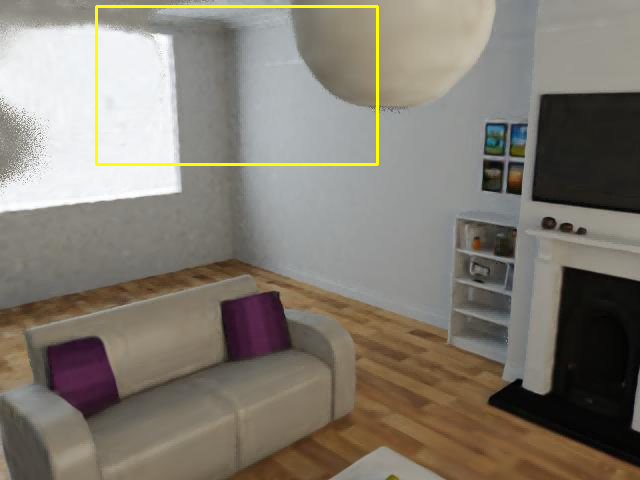}
            \end{minipage}
            \begin{minipage}[t]{0.24\linewidth}
                \centering
                \includegraphics[width=1\linewidth]{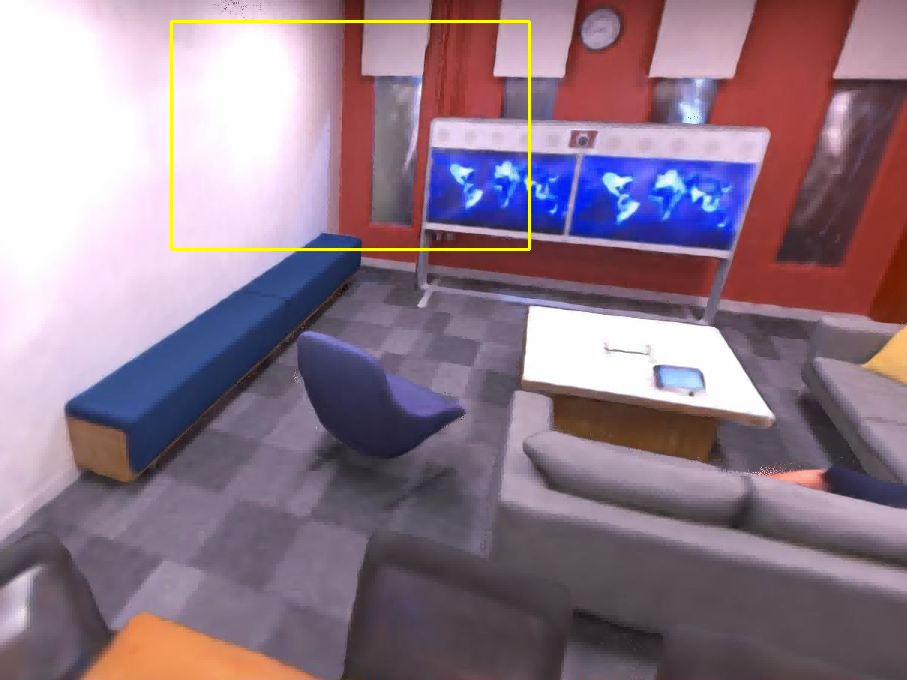}
            \end{minipage}
            \begin{minipage}[t]{0.24\linewidth}
                \centering
                \includegraphics[width=1\linewidth]{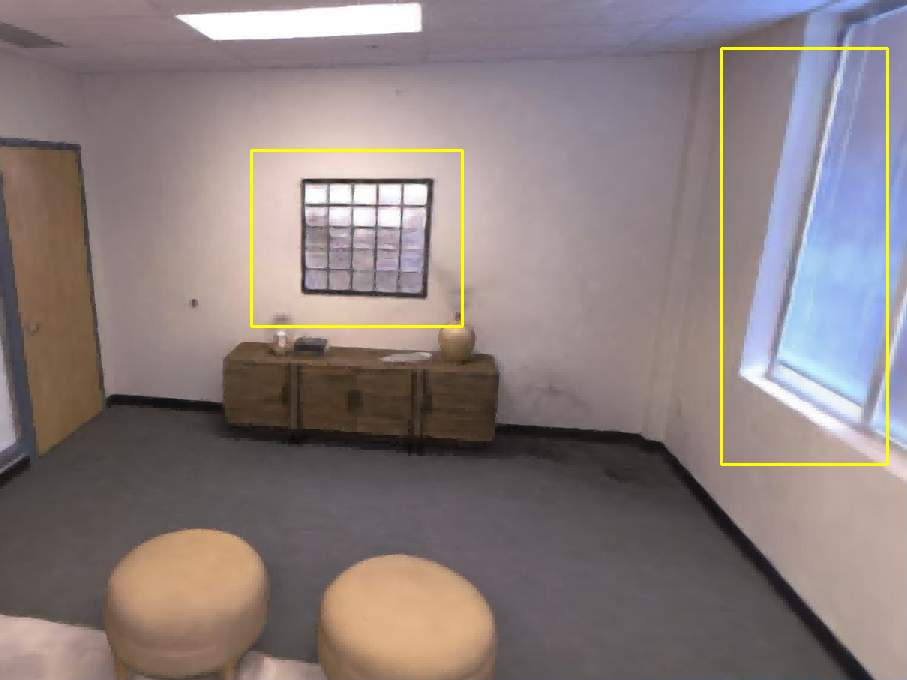}
            \end{minipage}
            \begin{minipage}[t]{0.24\linewidth}
                \centering
                \includegraphics[width=1\linewidth]{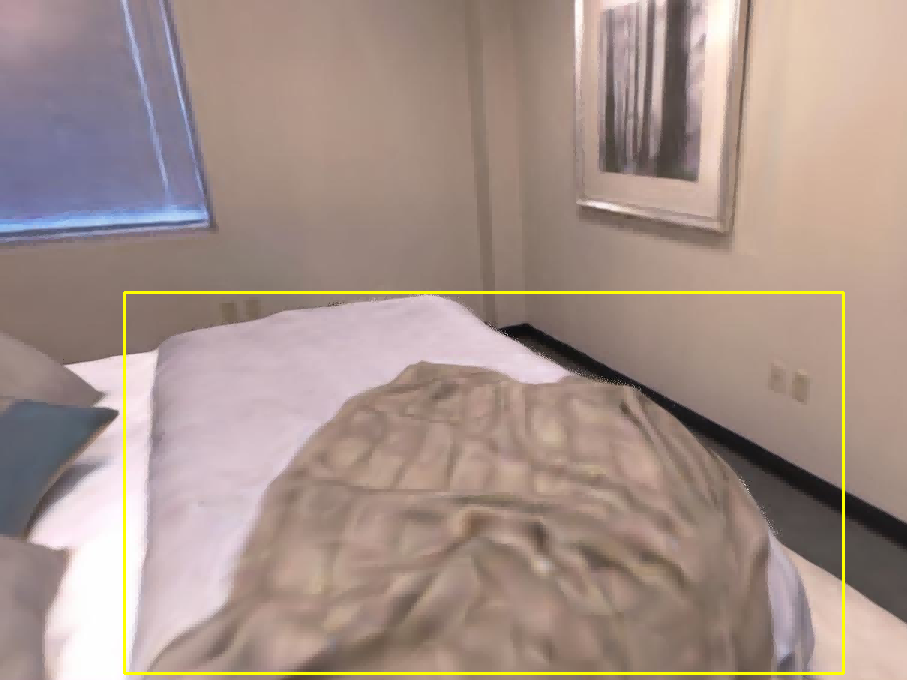}
            \end{minipage}
        \end{subfigure}

	\begin{subfigure}{\linewidth}
            \rotatebox[origin=c]{90}{\normalsize{NeuralRGBD}\hspace{-2.2cm}}
            \begin{minipage}[t]{0.24\linewidth}
                \centering
                \includegraphics[width=1\linewidth]{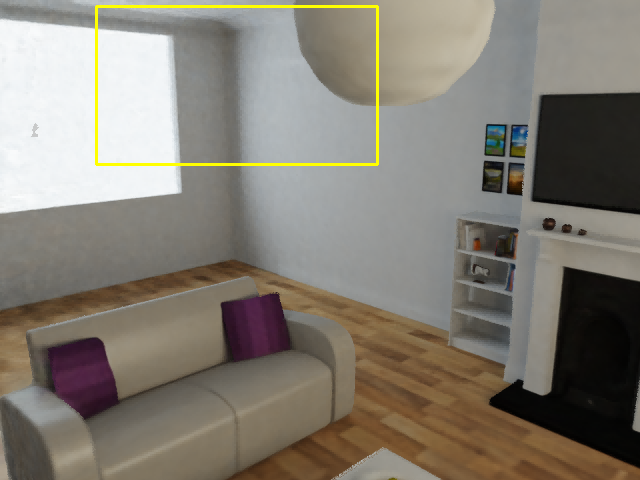}
            \end{minipage}
            \begin{minipage}[t]{0.24\linewidth}
                \centering
                \includegraphics[width=1\linewidth]{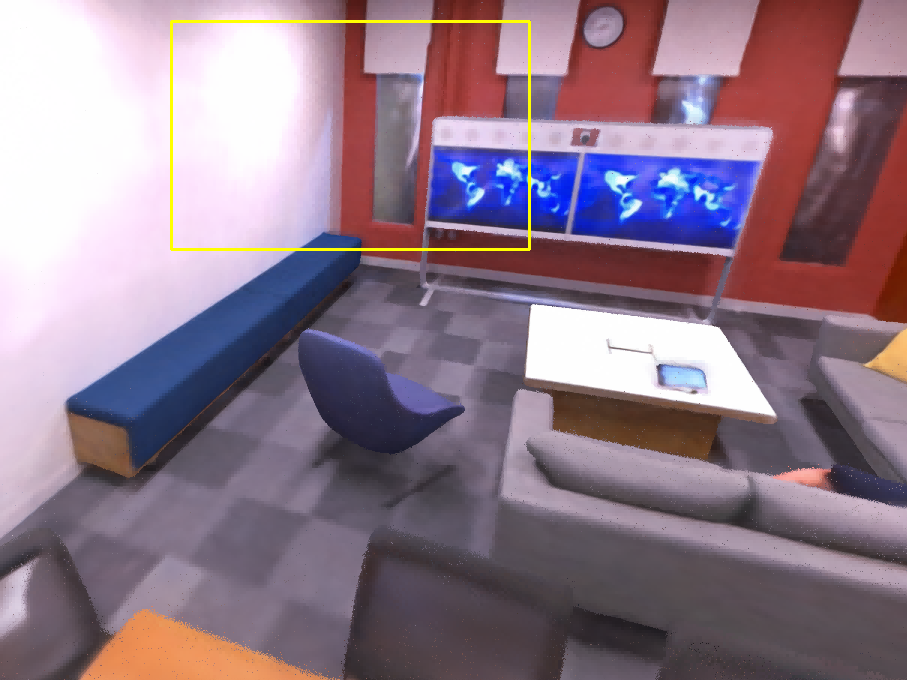}
            \end{minipage}
            \begin{minipage}[t]{0.24\linewidth}
                \centering
                \includegraphics[width=1\linewidth]{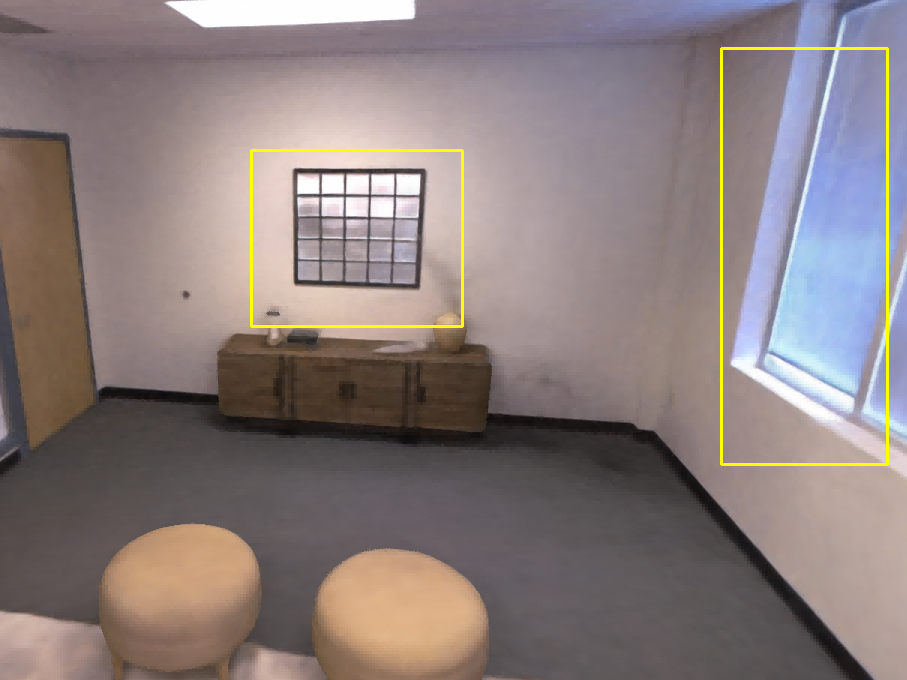}
            \end{minipage}
            \begin{minipage}[t]{0.24\linewidth}
                \centering
                \includegraphics[width=1\linewidth]{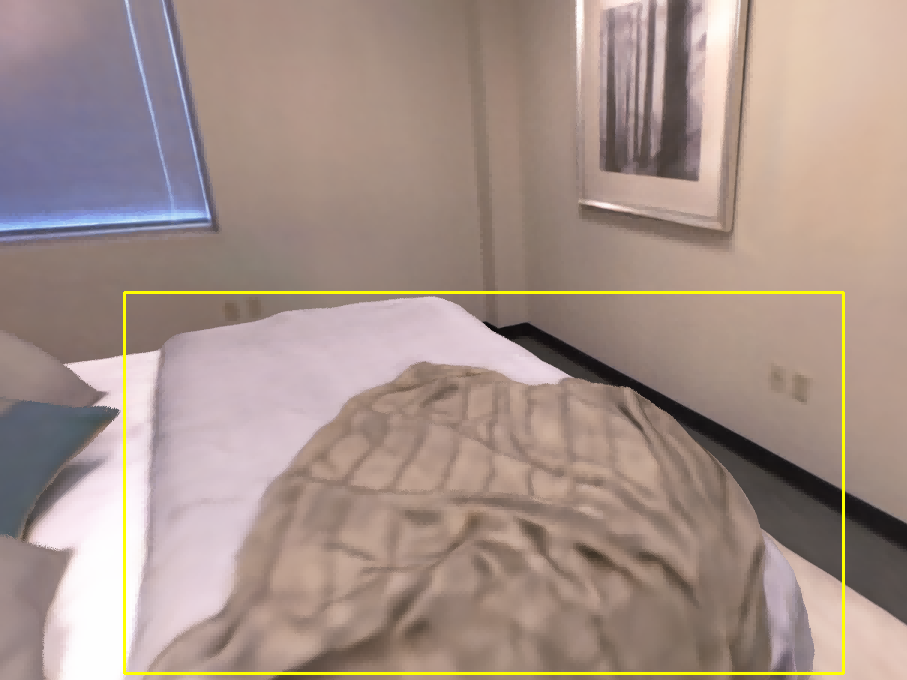}
            \end{minipage}
        \end{subfigure}

	\begin{subfigure}{\linewidth}
            \rotatebox[origin=c]{90}{\normalsize{Ours}\hspace{-2.2cm}}
            \begin{minipage}[t]{0.24\linewidth}
                \centering
                \includegraphics[width=1\linewidth]{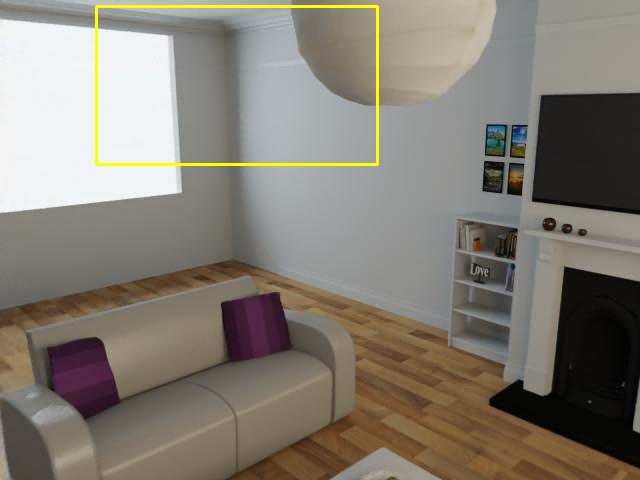}
            \end{minipage}
            \begin{minipage}[t]{0.24\linewidth}
                \centering
                \includegraphics[width=1\linewidth]{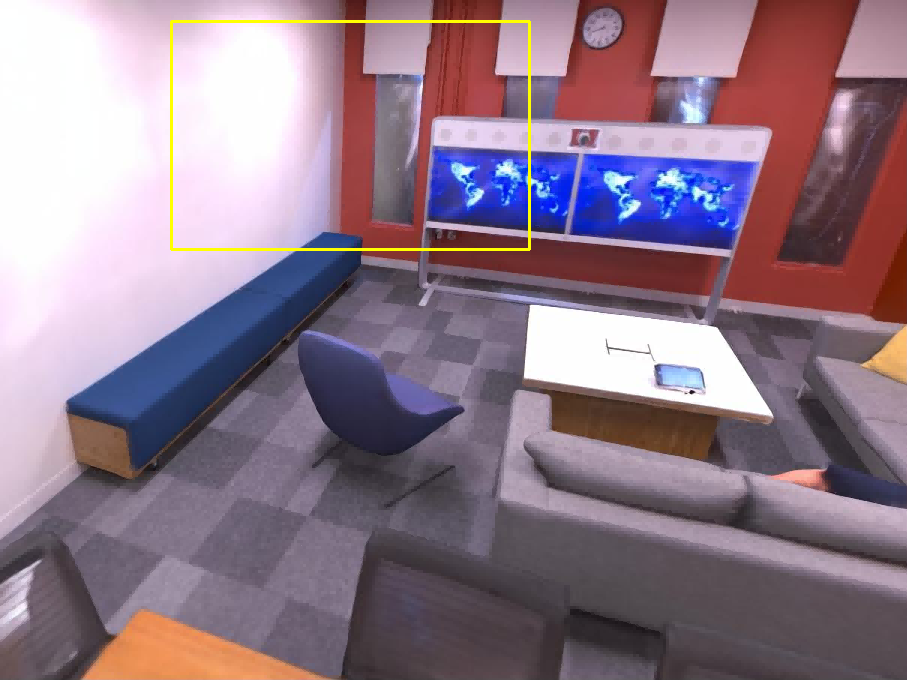}
            \end{minipage}
            \begin{minipage}[t]{0.24\linewidth}
                \centering
                \includegraphics[width=1\linewidth]{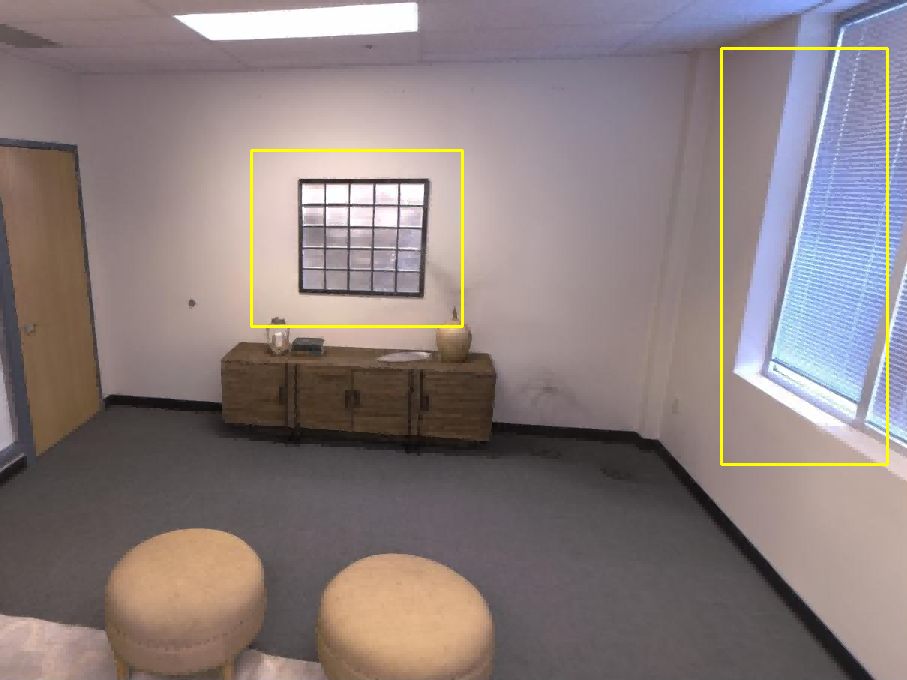}
            \end{minipage}
            \begin{minipage}[t]{0.24\linewidth}
                \centering
                \includegraphics[width=1\linewidth]{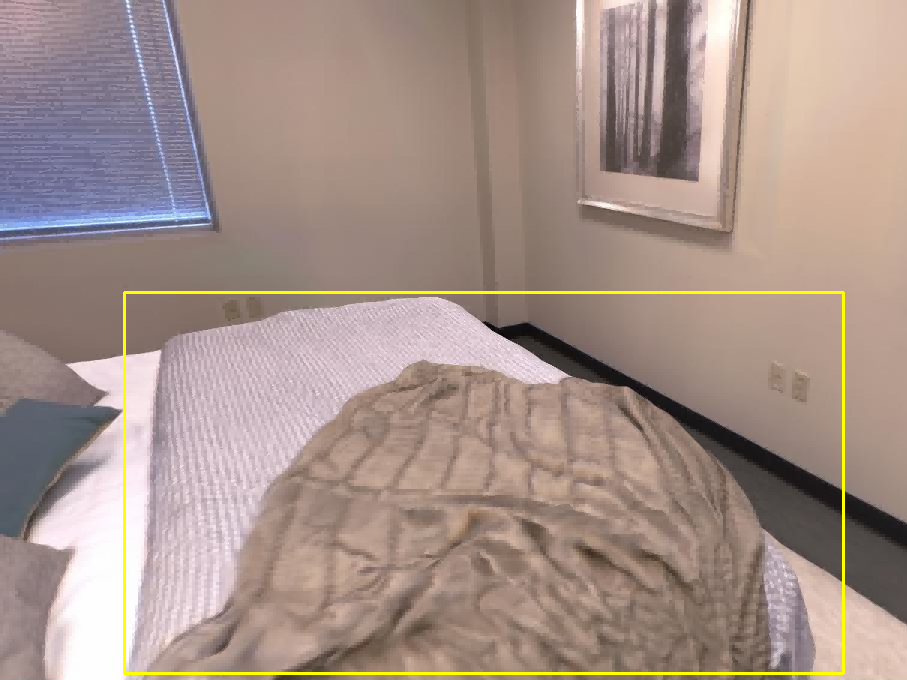}
            \end{minipage}
        \end{subfigure}

	\begin{subfigure}{\linewidth}
            \rotatebox[origin=c]{90}{\normalsize{GT}\hspace{-2.2cm}}
            \begin{minipage}[t]{0.24\linewidth}
                \centering
                \includegraphics[width=1\linewidth]{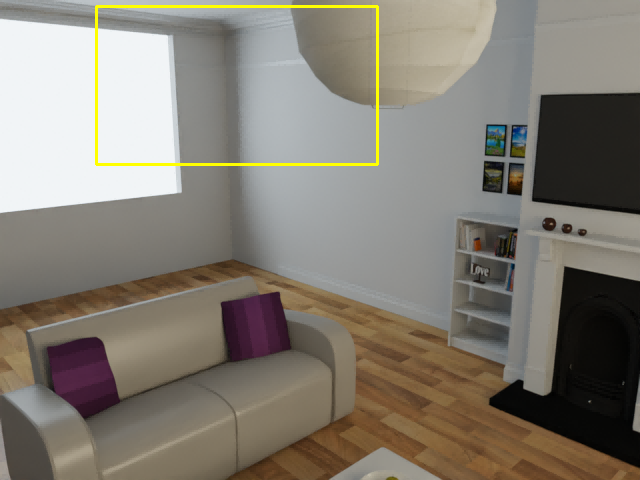}
                % \caption{Whiteroom}
            \end{minipage}
            \begin{minipage}[t]{0.24\linewidth}
                \centering
                \includegraphics[width=1\linewidth]{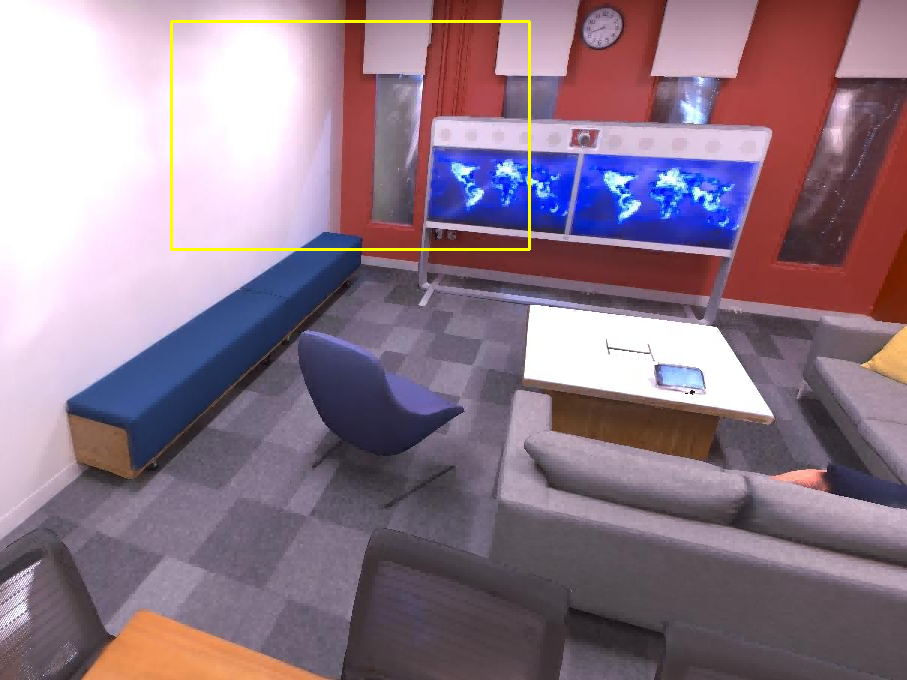}
                % \caption{Office3}
            \end{minipage}
            \begin{minipage}[t]{0.24\linewidth}
                \centering
                \includegraphics[width=1\linewidth]{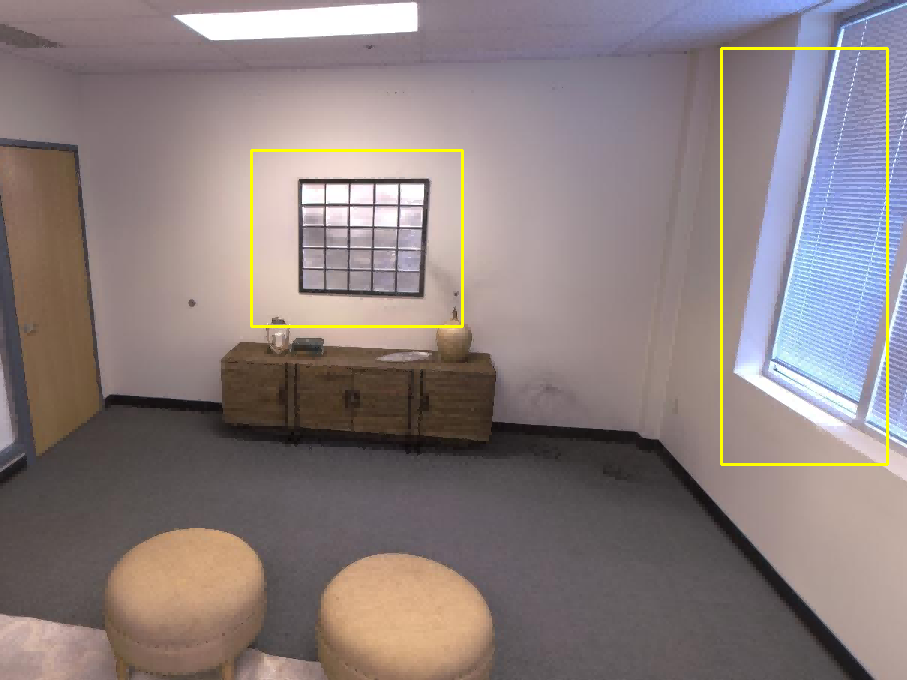}
                % \caption{Room0}
            \end{minipage}
            \begin{minipage}[t]{0.24\linewidth}
                \centering
                \includegraphics[width=1\linewidth]{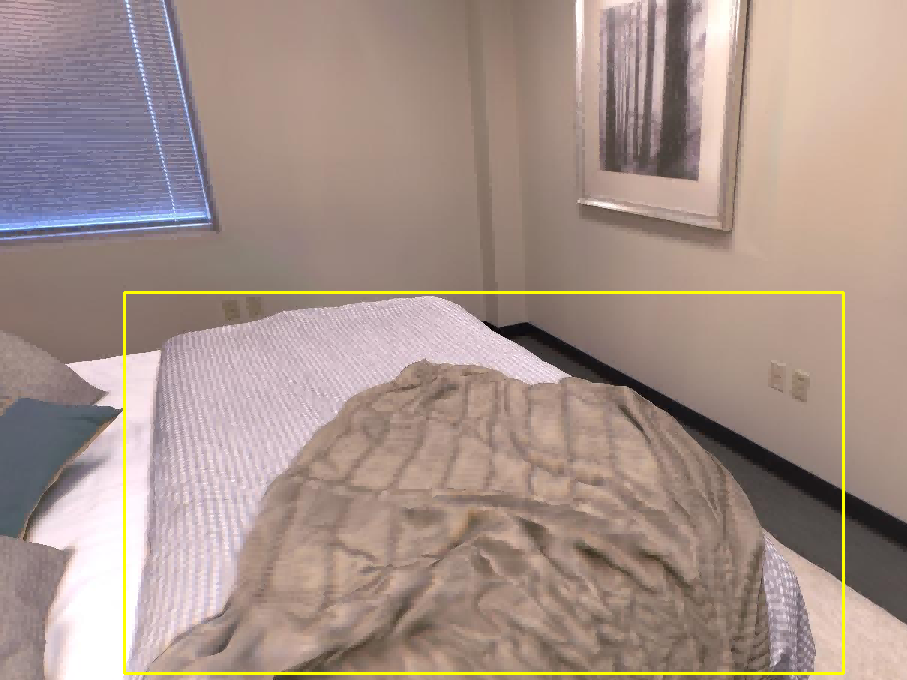}
                % \caption{Room1}
            \end{minipage}
        \end{subfigure}

	\caption{Additional results of view synthesis on scenes \textit{Whiteroom}, \textit{Office3}, \textit{Room0}, \textit{Room1}. The methods for novel view synthesis, such as \textit{InstantNGP}, \textit{DVGO}, fail to render clear results at texture-less regions, and methods focusing on geometry reconstruction, such as \textit{Neural-RGBD}, \textit{Go-Surf}, fail to restore the appearance of the regions with complex texture.}
	\label{fig:fig8}
\end{figure*}

More color decomposition results are shown in \cref{fig:fig10}. We can see that our method successfully decomposes the full color into view-independent color and view-dependent color such as the reflective table, the book and the TV.

\begin{table}
  \caption{Model storage and runtime of Du-NeRF. We list \textit{Scene size} ($\text{m}^3$), \textit{Model size}, number of parameters (\textit{Params.}) and \textit{Runtime}. Our method reaches convergence in at most 1 hour for all scenes.} 
  \label{tab:tab13}
  \centering
  \huge
  \resizebox{0.5\textwidth}{!}{
  \begin{tabular}{@{}lccccc@{}}
    \toprule
    \textbf{Scene}  & \textbf{Scene size} & \textbf{Model size} & \textbf{Params.} & \textbf{Runtime} \\
    
    \midrule
    \textbf{Breakfast room}   & $4.1 \times 3.3 \times 4.7$ & $116$ MB & $28.9$ M & $50$ min\\
    \textbf{Green room}   & $8.0 \times 3.1 \times 4.7$ & $170$ MB & $42.5$ M & $53$ min\\
    \textbf{Grey-white room}   & $5.9 \times 3.1 \times 4.4$ & $127$ MB & $31.7$ M & $50$ min \\
    \textbf{ICL living room }  & $5.3 \times 2.9 \times 5.4$ & $128$ MB & $32.1$ M & $51$ min \\
    \textbf{Complete kitchen}   & $9.3 \times 3.3 \times 10.0$ & $298$ MB & $74.1$ M & $73$ min \\
    \textbf{Kitchen}   & $7.0 \times 3.4 \times 8.7$ & $229$ MB & $57.0$ M & $61$ min \\
    \textbf{Morning apartment}   & $3.5 \times 2.3 \times 4.0$ & $81$ MB & $20.3$ M & $47$ min\\
    \textbf{Staircase}   & $6.8 \times 3.7 \times 6.5$ & $206$ MB & $51.7$ M & $63$ min \\
    \textbf{Thin Geometry}   & $3.4 \times 1.2 \times 3.6$ & $65$ MB & $16.3$ M & $48$ min \\
    \textbf{White room}   & $5.6 \times 3.8 \times 7.8$ & $201$ MB & $50.3$ M & $59$ min  \\
    \midrule
    \textbf{Office0}   & $4.7 \times 5.3 \times 3.3$ & $130$ MB & $32.5$ M & $41$ min \\
    \textbf{Office1}   & $5.2 \times 4.5 \times 3.3$ & $115$ MB & $28.9$ M & $42$ min  \\
    \textbf{Office2}   & $6.8 \times 8.5 \times 3.2$ & $230$ MB & $57.4$ M & $52$ min  \\
    \textbf{Office3}   & $9.0 \times 9.7 \times 3.5$ & $298$ MB & $74.5$ M & $70$ min  \\
    \textbf{Office4}   & $6.9 \times 6.9 \times 3.2$ & $193$ MB & $48.3$ M & $48$ min  \\

    \textbf{Room0}   & $8.2 \times 5.1 \times 3.2$ & $170$ MB & $42.7$ M & $46$ min  \\
    \textbf{Room1}   & $7.1 \times 6.1 \times 3.1$ & $175$ MB & $43.7$ M & $45$ min  \\
    \textbf{Room2}   & $7.2 \times 5.3 \times 4.0$ & $184$ MB & $45.9$ M & $46$ min  \\
    
    \bottomrule
  \end{tabular}
  }
\end{table}

\section{Runtime and Memory Requirements}\label{Runtime and Memory Requirements}
There is a detailed breakdown of the runtime and memory usage in our experiments on all of the datasets, as shown in \cref{tab:tab13}. It can be seen that our method could reach convergence in at most 1 hour on all scenes. Similar to previous methods that trade memory for time, Du-NeRF requires hundreds of MB to store the multi-resolution grid.

\begin{table*}[ht!]
    \begin{tabular*}{\linewidth}{llcccccccc}
    \toprule
\textbf{Scene} & \textbf{Method} & \textbf{Acc $\downarrow$} & \textbf{Com $\downarrow$} & \textbf{C-$l_1 \downarrow$} & \textbf{NC $\uparrow$} & \textbf{F-score $\uparrow$} & \textbf{PSNR $\uparrow$} & \textbf{SSIM $\uparrow$} & \textbf{LPIPS $\downarrow$}  \\ \midrule

\multirow{8}{*}{\textbf{Breakfast Room}}   & BundleFusion  & 0.0134          & 0.2000          & 0.1070          & 0.9170          & 0.8000           & -                & -               & -               \\
                                           & Neus          & 0.6270          & 0.7720          & 0.7000          & 0.6410          & 0.0100           & 27.2880          & 0.8230          & 0.1940          \\
                                           & VolSDF        & 0.0820          & 0.3360          & 0.2090          & 0.6880          & 0.2970           & 28.9610          & 0.8730          & 0.1570          \\
                                           & NeuralRGBD    & 0.0145          & 0.0148          & 0.0147          & 0.9650          & 0.9900           & 32.5340          & 0.9280          & 0.1090          \\
                                           & GO-Surf       & \textbf{0.0141} & 0.0150          & 0.0145          & 0.9630          & 0.9810           & 29.0600          & 0.8830          & 0.1650          \\
                                           & InstantNGP    & 0.2310          & 0.2970          & 0.2640          & 0.5670          & 0.2420           & 31.6560          & 0.8970          & 0.0853          \\
                                           & DVGO          & 0.5310          & 0.9010          & 0.7160          & 0.5100          & 0.0184           & 34.2720          & 0.9530          & 0.0640          \\
                                           & \textbf{Ours} & \textbf{0.0141} & \textbf{0.0135} & \textbf{0.0138} & \textbf{0.9640} & \textbf{0.9860}  & \textbf{38.2610} & \textbf{0.9830} & \textbf{0.0238} \\ \midrule
\multirow{8}{*}{\textbf{Complete Kitchen}} & BundleFusion  & 0.0366          & 1.0780          & 0.5570          & 0.7470          & 0.4450           & -                & -               & -               \\
                                           & Neus          & 0.2950          & 1.5580          & 0.9270          & 0.5610          & 0.0641           & 26.3270          & 0.8410          & 0.2540          \\
                                           & VolSDF        & 0.3260          & 0.9100          & 0.6180          & 0.6800          & 0.1470           & 26.0160          & 0.8570          & 0.2400          \\
                                           & NeuralRGBD    & \textbf{0.0189} & 0.1100          & 0.0647          & 0.8960          & 0.8790           & 32.0310          & 0.9100          & 0.2020          \\
                                           & GO-Surf       & 0.0254          & 0.0295          & 0.0274          & 0.9395          & 0.8930           & 29.6040          & 0.8680          & 0.1760          \\
                                           & InstantNGP    & 0.1950          & 0.8730          & 0.5340          & 0.5740          & 0.2350           & 30.4066          & 0.8850          & 0.1470          \\
                                           & DVGO          & 0.2970          & 1.1580          & 0.7280          & 0.5610          & 0.1990           & 31.0570          & 0.9010          & 0.2180          \\
                                           & \textbf{Ours} & 0.0222          & \textbf{0.0259} & \textbf{0.0240} & \textbf{0.9410} & \textbf{0.9010}  & \textbf{34.5170} & \textbf{0.9520} & \textbf{0.0682} \\ \midrule
\multirow{8}{*}{\textbf{Green Room}}       & BundleFusion  & 0.0140          & 0.1965          & 0.1053          & 0.9090          & 0.8140           & -                & -               & -               \\
                                           & Neus          & 0.2130          & 0.3430          & 0.2780          & 0.7400          & 0.1130           & 29.4000          & 0.8930          & 0.1450          \\
                                           & VolSDF        & 0.1360          & 0.4320          & 0.2840          & 0.6490          & 0.2250           & 29.8840          & 0.9100          & 0.1380          \\
                                           & NeuralRGBD    & \textbf{0.0104} & \textbf{0.0140} & \textbf{0.0122} & \textbf{0.9340} & \textbf{0.9910}  & 34.0800          & 0.9520          & 0.0770          \\
                                           & GO-Surf       & 0.0124          & 0.0156          & 0.0140          & 0.9275          & 0.9825           & 31.0520          & 0.9220          & 0.1170          \\
                                           & InstantNGP    & 0.2440          & 0.9710          & 0.6070          & 0.5370          & 0.1180           & 34.9470          & 0.9460          & 0.0364          \\
                                           & DVGO          & 0.2940          & 0.5190          & 0.4070          & 0.5640          & 0.2490           & 34.9800          & 0.9550          & 0.0760          \\
                                           & \textbf{Ours} & 0.0122          & 0.0150          & 0.0136          & 0.9290          & 0.9850           & \textbf{38.5530} & \textbf{0.9780} & \textbf{0.0247} \\ \midrule
\multirow{8}{*}{\textbf{Grey White Room}}  & BundleFusion  & 0.0202          & 0.2743          & 0.1472          & 0.8230          & 0.7380           & -                & -               & -               \\
                                           & Neus          & 0.2620          & 0.4820          & 0.3720          & 0.6290          & 0.1160           & 28.7820          & 0.8630          & 0.1640          \\
                                           & VolSDF        & 0.1930          & 0.3360          & 0.2650          & 0.7070          & 0.2550           & 30.4400          & 0.8950          & 0.1480          \\
                                           & NeuralRGBD    & \textbf{0.0134} & \textbf{0.0151} & \textbf{0.0143} & \textbf{0.9310} & \textbf{0.9940}  & 35.1630          & 0.9470          & 0.0900          \\
                                           & GO-Surf       & 0.0145          & 0.0167          & 0.0156          & 0.9255          & 0.9875           & 30.8900          & 0.9115          & 0.1490          \\
                                           & InstantNGP    & 0.1390          & 0.9380          & 0.5390          & 0.5010          & 0.1810           & 32.0200          & 0.8790          & 0.1270          \\
                                           & DVGO          & 0.2410          & 0.4430          & 0.3420          & 0.5640          & 0.2990           & 34.7160          & 0.9470          & 0.0930          \\
                                           & \textbf{Ours} & 0.0140          & 0.0155          & 0.0147          & 0.9260          & 0.9900           & \textbf{37.7320} & \textbf{0.9700} & \textbf{0.0406} \\ \midrule
\multirow{8}{*}{\textbf{Icl Living Room}}  & BundleFusion  & 0.0104          & 0.2697          & 0.1400          & 0.9120          & 0.7720           & -                & -               & -               \\
                                           & Neus          & 0.3570          & 0.8040          & 0.5810          & 0.6270          & 0.1000           & 31.9550          & 0.9010          & 0.1090          \\
                                           & VolSDF        & 0.2520          & 0.8130          & 0.5330          & 0.6370          & 0.1450           & 30.6400          & 0.8980          & 0.1400          \\
                                           & NeuralRGBD    & \textbf{0.0089} & 0.0840          & 0.0462          & 0.9070          & 0.9010           & 34.3810          & 0.9300          & 0.1960          \\
                                           & GO-Surf       & 0.0101          & 0.0129          & 0.0115          & 0.9670          & 0.9910           & 31.7410          & 0.9080          & 0.2425          \\
                                           & InstantNGP    & 0.7180          & 1.8900          & 1.3000          & 0.5140          & 0.0100           & 27.1670          & 0.7650          & 0.2760          \\
                                           & DVGO          & 0.3000          & 0.8670          & 0.5830          & 0.5430          & 0.2130           & 33.9930          & 0.9330          & 0.2420          \\
                                           & \textbf{Ours} & 0.0112          & \textbf{0.0140} & \textbf{0.0126} & \textbf{0.9690} & \textbf{0.9920}  & \textbf{36.9860} & \textbf{0.9570} & \textbf{0.0638} \\

    \bottomrule
    
    \end{tabular*}
    \caption{Reconstruction and view synthesis results of NeuralRGBD dataset. The best performances are highlighted in bold.}
    \label{tab:tab9}

\end{table*}

\begin{table*}[ht!]
    \begin{tabular*}{\linewidth}{llcccccccc}
    \toprule
\textbf{Scene} & \textbf{Method} & \textbf{Acc $\downarrow$} & \textbf{Com $\downarrow$} & \textbf{C-$l_1 \downarrow$} & \textbf{NC $\uparrow$} & \textbf{F-score $\uparrow$} & \textbf{PSNR $\uparrow$} & \textbf{SSIM $\uparrow$} & \textbf{LPIPS $\downarrow$}  \\ \midrule

\multirow{8}{*}{\textbf{Kitchen}}                                                      & BundleFusion  & 0.0170          & 0.5960          & 0.3065          & 0.8510          & 0.6390          & -                & -               & -               \\
                                                                                       & Neus          & 0.4680          & 0.8490          & 0.6580          & 0.5740          & 0.0400          & 25.5880          & 0.8240          & 0.2340          \\
                                                                                       & VolSDF        & 0.2130          & 0.3760          & 0.2950          & 0.7000          & 0.2810          & 25.7570          & 0.8520          & 0.2300          \\
                                                                                       & NeuralRGBD    & \textbf{0.0198} & \textbf{0.1450} & 0.0824          & 0.9000          & 0.8630          & 31.6270          & 0.9170          & 0.1480          \\
                                                                                       & GO-Surf       & 0.0204          & 0.0265          & \textbf{0.0235}          & \textbf{0.9340}          & \textbf{0.9430}          & 27.8260          & 0.8735          & 0.2000          \\
                                                                                       & InstantNGP    & 0.1620          & 0.6470          & 0.4050          & 0.5520          & 0.2360          & 29.3850          & 0.8650          & 0.1380          \\
                                                                                       & DVGO          & 0.2690          & 0.5120          & 0.3900          & 0.5640          & 0.3340          & 30.5230          & 0.9360          & 0.1310          \\
                                                                                       & \textbf{Ours} & 0.0207          & 0.0266          & 0.0236 & 0.9330 & 0.9400 & \textbf{35.6100} & \textbf{0.9650} & \textbf{0.0458} \\ \hline
\multirow{8}{*}{\textbf{\begin{tabular}[c]{@{}l@{}}Morning \\ Apartment\end{tabular}}} & BundleFusion  & 0.0093          & 0.0153          & 0.0123          & 0.8880          & 0.9760          & -                & -               & -               \\
                                                                                       & Neus          & 0.2130          & 0.2430          & 0.2280          & 0.6660          & 0.2180          & 27.5640          & 0.8370          & 0.2090          \\
                                                                                       & VolSDF        & 0.0804          & 0.1450          & 0.1130          & 0.7300          & 0.3700          & 29.2440          & 0.8870          & 0.1660          \\
                                                                                       & NeuralRGBD    & \textbf{0.0088} & \textbf{0.0117} & \textbf{0.0103} & \textbf{0.8920} & \textbf{0.9870} & 33.1350          & 0.9270          & 0.1080          \\
                                                                                       & GO-Surf       & 0.0106          & 0.0145          & 0.0125          & 0.8840          & 0.9750          & 28.3880          & 0.8570          & 0.2245          \\
                                                                                       & InstantNGP    & 0.4260          & 0.4920          & 0.4590          & 0.5030          & 0.1000          & 24.2740          & 0.6230          & 0.4640          \\
                                                                                       & DVGO          & 0.1510          & 0.1760          & 0.1630          & 0.5480          & 0.5100          & 33.9590          & 0.9480          & 0.0740          \\
                                                                                       & \textbf{Ours} & 0.0099          & 0.0136          & 0.0118          & 0.8870          & 0.9780          & \textbf{36.6370} & \textbf{0.9660} & \textbf{0.0392} \\ \hline
\multirow{8}{*}{\textbf{Staircase}}                                                    & BundleFusion  & 0.0160          & 1.0020          & 0.5088          & 0.7960          & 0.4310          & -                & -               & -               \\
                                                                                       & Neus          & 0.3800          & 0.8910          & 0.6360          & 0.6190          & 0.1140          & 29.4540          & 0.8430          & 0.2710          \\
                                                                                       & VolSDF        & 0.1070          & 0.7340          & 0.4200          & 0.6720          & 0.2440          & 30.9470          & 0.8490          & 0.2570          \\
                                                                                       & NeuralRGBD    & \textbf{0.0213} & 0.0441          & 0.0327          & 0.9420          & 0.9010          & 34.8610          & 0.9070          & 0.2210          \\
                                                                                       & GO-Surf       & 0.0233          & 0.0285          & 0.0259          & 0.9490          & 0.8855          & 32.1520          & 0.8805          & 0.2560          \\
                                                                                       & InstantNGP    & 0.2880          & 0.3240          & 0.3060          & 0.5820          & 0.3130          & 31.1140          & 0.8250          & 0.2320          \\
                                                                                       & DVGO          & 0.2710          & 0.7460          & 0.5080          & 0.5750          & 0.2580          & 34.5150          & 0.9160          & 0.2200          \\
                                                                                       & \textbf{Ours} & 0.0221          & \textbf{0.0268} & \textbf{0.0244} & \textbf{0.9510} & \textbf{0.9280} & \textbf{36.7910} & \textbf{0.9550} & \textbf{0.0710} \\ \hline
\multirow{8}{*}{\textbf{Thin Geometry}}                                                & BundleFusion  & 0.0227          & 0.0762          & 0.0495          & 0.8640          & 0.7160          & -                & -               & -               \\
                                                                                       & Neus          & 0.1550          & 0.5760          & 0.3650          & 0.5380          & 0.0860          & 19.2300          & 0.7830          & 0.1760          \\
                                                                                       & VolSDF        & 0.1140          & 0.3960          & 0.2550          & 0.5990          & 0.2710          & 24.7540          & 0.8900          & 0.1360          \\
                                                                                       & NeuralRGBD    & 0.0093          & 0.0254          & 0.0173          & \textbf{0.9090} & 0.9420          & 18.9310          & 0.6600          & 0.5580          \\
                                                                                       & GO-Surf       & 0.0107          & 0.0186          & 0.0146          & 0.9005          & 0.9440          & 25.8900          & 0.8820          & 0.1400          \\
                                                                                       & InstantNGP    & 0.0821          & 0.4290          & 0.2560          & 0.6130          & 0.3070          & 32.4840          & 0.9330          & 0.0347          \\
                                                                                       & DVGO          & 0.1040          & 0.0750          & 0.0900          & 0.5720          & 0.4750          & 35.1540          & 0.9680          & 0.0330          \\
                                                                                       & Ours          & \textbf{0.0087} & \textbf{0.0122} & \textbf{0.0105} & 0.9070          & \textbf{0.9790} & \textbf{34.1260} & \textbf{0.9640} & \textbf{0.0286} \\ \hline
\multirow{8}{*}{\textbf{Whiteroom}}                                                    & BundleFusion  & 0.0184          & 0.8690          & 0.4440          & 0.8060          & 0.4700          & -                & -               & -               \\
                                                                                       & Neus          & 0.2040          & 0.3930          & 0.2980          & 0.6820          & 0.1710          & 29.0620          & 0.8770          & 0.1610          \\
                                                                                       & VolSDF        & 0.1240          & 0.3370          & 0.2300          & 0.7450          & 0.3870          & 30.5300          & 0.9050          & 0.1230          \\
                                                                                       & NeuralRGBD    & \textbf{0.0202} & 0.0437          & 0.0320          & 0.9200          & 0.9098          & 33.1980          & 0.9330          & 0.1240          \\
                                                                                       & GO-Surf       & 0.0225          & 0.0359          & 0.0292          & 0.9285          & 0.9070          & 29.2535          & 0.8985          & 0.1640          \\
                                                                                       & InstantNGP    & 0.1560          & 0.4570          & 0.3060          & 0.6060          & 0.3370          & 31.7830          & 0.9090          & 0.0956          \\
                                                                                       & DVGO          & 0.2230          & 0.5060          & 0.3650          & 0.5880          & 0.3130          & 33.1580          & 0.9400          & 0.0970          \\
                                                                                       & Ours          & 0.0214          & \textbf{0.0340} & \textbf{0.0277} & \textbf{0.9270} & \textbf{0.9210} & \textbf{36.2280} & \textbf{0.9690} & \textbf{0.0409} \\

    \bottomrule
    
    \end{tabular*}
    \caption{Reconstruction and view synthesis results of NeuralRGBD dataset. The best performances are highlighted in bold.}
    \label{tab:tab10}

\end{table*}

\begin{table*}[ht!]
    \begin{tabular*}{\linewidth}{lc|cccccccc|c}
    \toprule
\textbf{Methods}                  & \textbf{Evaluation} & \textbf{office0} & \textbf{office1} & \textbf{office2} & \textbf{office3} & \textbf{office4} & \textbf{room0} & \textbf{room1} & \textbf{room2} & \textbf{Mean} \\ \hline
\multirow{4}{*}{BundleFusion}     & C-$l_1 \downarrow$          & 0.0109           & 0.0110           & 0.0310           & 0.0660           & 0.0165           & 0.0587         & 0.0110         & 0.0339         & 0.0299        \\
                                  & F-score $\uparrow$             & 0.9880           & 0.9910           & 0.9190           & 0.8410           & 0.9690           & 0.8820         & 0.9900         & 0.9100         & 0.9360        \\
                                  & PSNR $\uparrow$                & -                & -                & -                & -                & -                & -              & -              & -              & -             \\
                                  & LPIPS $\downarrow$               & -       & -       & -       & -       & -       & -     & -     & -     & -    \\ \hline
\multirow{4}{*}{Neus}             & C-$l_1 \downarrow$          & 0.0154           & 0.2280           & 0.2390           & 0.2000           & 0.3440           & 0.2140         & 0.1570         & 0.2940         & 0.2288        \\
                                  & F-score $\uparrow$             & 0.2340           & 0.1800           & 0.1730           & 0.3050           & 0.0850           & 0.2300         & 0.2210         & 0.1240         & 0.1940        \\
                                  & PSNR $\uparrow$                 & 33.274           & 33.874           & 26.372           & 26.893           & 28.546           & 25.913         & 28.068         & 28.570         & 28.939       \\
                                  & LPIPS $\downarrow$                & 0.1570           & 0.1240           & 0.1790           & 0.1640           & 0.1660           & 0.2660         & 0.2120         & 0.1790         & 0.1810        \\ \hline
\multirow{4}{*}{VolSDF}           & C-$l_1 \downarrow$           & 0.1270           & 0.2050           & 0.1820           & 0.2830           & 0.2980           & 0.2750         & 0.1580         & 0.2160         & 0.2180        \\
                                  & F-score $\uparrow$             & 0.4200           & 0.2440           & 0.4150           & 0.3250           & 0.2980           & 0.3280         & 0.3810         & 0.3010         & 0.3390        \\
                                  & PSNR $\uparrow$                & 34.327           & 35.733           & 28.996           & 27.943           & 30.595           & 26.374         & 28.807         & 30.227         & 30.375       \\
                                  & LPIPS $\downarrow$               & 0.1570           & 0.1120           & 0.1650           & 0.1730           & 0.1590           & 0.2660         & 0.2140         & 0.1560         & 0.1750        \\ \hline
\multirow{4}{*}{NeuralRGBD}       & C-$l_1 \downarrow$          & 0.0635           & \textbf{0.0088}           & 0.0544           & 0.0177           & 0.3340           & 0.0892         & 0.1010         & 0.3480         & 0.1271        \\
                                  & F-score $\uparrow$             & 0.8550           & 0.9890           & 0.9380           & 0.9790           & 0.7040           & 0.8740         & 0.8420         & 0.5960         & 0.8470        \\
                                  & PSNR $\uparrow$                 & 36.563           & 38.148           & 31.559           & 30.509           & 33.934           & 28.311         & 30.866         & 31.452         & 32.668       \\
                                  & LPIPS $\downarrow$               & 0.1640           & 0.1590           & 0.2000           & 0.1840           & 0.1600           & 0.2750         & 0.2410         & 0.2000         & 0.1980        \\ \hline
\multirow{4}{*}{GO-Surf}          & C-$l_1 \downarrow$          & 0.0099           & 0.0104           & 0.0125           & 0.0156           & 0.0133           & 0.0131         & 0.0103         & 0.0117         & 0.0121        \\
                                  & F-score $\uparrow$             & 0.9880           & 0.9925           & 0.9875           & 0.9825           & 0.9890           & 0.9935         & 0.9960         & \textbf{0.9880}         & 0.9896        \\
                                  & PSNR $\uparrow$                & 35.105           & 35.655           & 29.059           & 29.138           & 31.784           & 27.378         & 29.522         & 30.096         & 30.967       \\
                                  & LPIPS $\downarrow$               & 0.1620           & 0.1740           & 0.2355           & 0.1940           & 0.1835           & 0.2975         & 0.2650         & 0.2270         & 0.2170        \\ \hline
\multirow{4}{*}{InstantNGP}       & C-$l_1 \downarrow$          & 0.2560           & 1.2320           & 0.4940           & 0.4940           & 1.1500           & 0.7870         & 0.6270         & 0.7900         & 0.7288        \\
                                  & F-score $\uparrow$             & 0.2010           & 0.1020           & 0.2660           & 0.2660           & 0.1030           & 0.1050         & 0.1510         & 0.0720         & 0.1583        \\
                                  & PSNR $\uparrow$                & 36.667           & 37.679           & 30.793           & 30.793           & 32.772           & 26.912         & 30.786         & 32.411         & 32.352       \\
                                  & LPIPS $\downarrow$               & 0.1710           & 0.0754           & 0.1240           & 0.1240           & 0.2140           & 0.2620         & 0.1350         & 0.0976         & 0.1500        \\ \hline
\multirow{4}{*}{DVGO}             & C-$l_1 \downarrow$          & 0.2160           & 0.2790           & 0.2010           & 0.3270           & 0.3660           & 0.4160         & 0.2470         & 0.3120         & 0.2955        \\
                                  & F-score $\uparrow$             & 0.3170           & 0.2040           & 0.3010           & 0.2470           & 0.1330           & 0.2360         & 0.2630         & 0.2110         & 0.2390        \\
                                  & PSNR $\uparrow$               & 36.859           & 37.878           & 31.596           & 28.349           & 32.583           & 26.985         & 30.250         & 31.192         & 31.962       \\
                                  & LPIPS $\downarrow$               & 0.1670           & 0.1640           & 0.2100           & 0.2260           & 0.2100           & 0.3500         & 0.2510         & 0.2090         & 0.2230        \\ \hline
\multirow{4}{*}{\textbf{Du-NeRF}} & C-$l_1 \downarrow$          & \textbf{0.0096}  & 0.0102           & \textbf{0.0116}  & \textbf{0.0141}  & \textbf{0.0118}  & \textbf{0.0119} & \textbf{0.0093} & \textbf{0.0108} & \textbf{0.0112}        \\
                                  & F-score $\uparrow$            & \textbf{0.9890}  & \textbf{0.9930}  & \textbf{0.9900}  & \textbf{0.9860}  & \textbf{0.9910}  & \textbf{0.9950} & \textbf{0.9970} & \textbf{0.9880} & \textbf{0.9911}        \\
                                  & PSNR $\uparrow$                & \textbf{41.365}  & \textbf{41.922}  & \textbf{34.913}  & \textbf{34.622}  & \textbf{37.776}  & \textbf{34.184} & \textbf{36.123} & \textbf{35.928} & \textbf{37.104}       \\
                                  & LPIPS $\downarrow$               & \textbf{0.0546}  & \textbf{0.0630}  & \textbf{0.0933}  & \textbf{0.0790}  & \textbf{0.0775}  & \textbf{0.0824} & \textbf{0.0673} & \textbf{0.0790} & \textbf{0.0740} \\ 
    \bottomrule
    
    \end{tabular*}
    \caption{Reconstruction and view synthesis results of Replica dataset. The best performances are highlighted in bold.}
    \label{tab:tab11}

\end{table*}

